\documentclass[a4paper,10pt]{article}
\usepackage[utf8]{inputenc}

\usepackage{amsmath,amsthm}
\usepackage{graphicx, subfigure, caption}
\usepackage{color}
\usepackage{longtable}
\usepackage[table]{xcolor}

\usepackage[%
 centering,
 paperwidth=21cm, paperheight=29.7cm,
 textwidth=15cm, textheight=23cm,
 headheight=13.6pt,
 tmargin=80pt,
 hoffset=0mm, voffset=10mm,
 hmarginratio=1:1, vmarginratio=1:1,
 nomarginpar,
]{geometry}
\usepackage[vlined]{algorithm2e}
\usepackage{authblk}
\usepackage{bbold}
\usepackage{multirow}
\usepackage{comment}
\usepackage{hyperref}

\def\R{{\cal R}}
\theoremstyle{definition}
\newtheorem{defn}{Definition}
\theoremstyle{plain}
\newtheorem{lemma}{Lemma}

\title{A Computational Model for Amodal Completion}
\author[1]{Maria Oliver}
\author[1]{Gloria Haro}
\author[2]{Mariella Dimiccoli}
\author[3]{Baptiste Mazin}
\author[1]{Coloma Ballester}
\affil[1]{Universitat Pompeu Fabra (UPF)}
\affil[2]{Centre de Visió per Computador (CVC)}
\affil[3]{DxO Labs}
\date{}

\begin{document}
\maketitle
\begin{abstract}
This paper presents a computational model to recover the most likely interpretation of the 3D scene structure from a planar image, where some objects may occlude others. 
The estimated scene interpretation is obtained by integrating some global and local cues and provides both the complete  disoccluded objects that form the scene and their ordering according to depth. Our method first computes several distal scenes which are compatible with the proximal planar image. To compute these different hypothesized scenes, we propose a perceptually inspired  object disocclusion method, which works by minimizing the Euler's elastica as well as by incorporating the relatability of partially occluded contours and the convexity of the disoccluded objects. Then, to estimate the preferred scene we rely on a Bayesian model and define probabilities taking into account the global complexity of the objects in the hypothesized scenes as well as the effort of bringing these objects in their relative position in the planar image, which is also measured by an Euler's elastica-based quantity. The model is illustrated with numerical experiments on, both, synthetic and real images showing the ability of our model to reconstruct the occluded objects and the preferred perceptual order among them. We also present results on images of the  Berkeley dataset with provided figure-ground ground-truth labeling.
\end{abstract}

\section{Introduction}
\label{sec:intro}
\begin{figure}[h]
    \centering
    \subfigure[Image source~\cite{MartinFTM01}]{\label{subfig:exReal}\includegraphics[height=2.85cm]{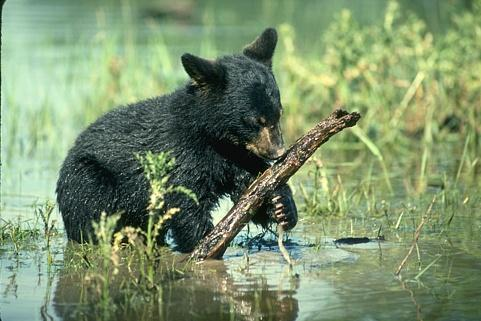}} \hspace{1.2cm}
    \subfigure[Image source~\cite{Kanizsa1991}]{\label{subfig:exSintetic}\includegraphics[height=2.85cm]{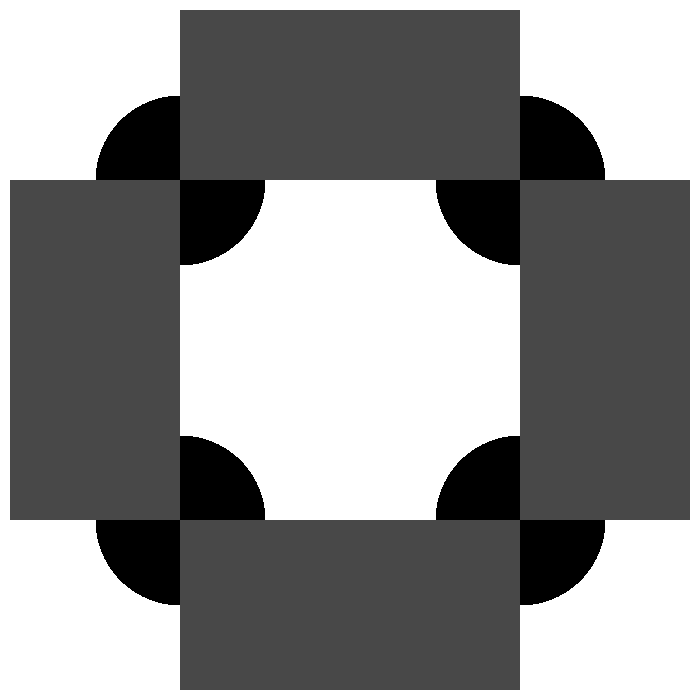}}
    \caption{Some examples where our brain experiences visual completion.}
    \label{fig:examples}
\end{figure}

Visual completion is a pervasive process in our daily life that works by hallucinating contours and surfaces in the scene when there is not a physical magnitude for them. Whenever we look at an image, our brain  unconsciously reconstructs the 3D scene by completing partially occluded objects while inferring their relative depth order into the scene (see Figure~\ref{fig:examples}). In Figure~\ref{subfig:exSintetic}, for instance, our brain prefers to interpret the scene as four disks partially occluded by four rectangles instead of, e.g., the more straightforward  description of eight quarters of a disk and four rectangles fitting together.

\begin{figure}[h]
    \centering
    \subfigure[Relatable contours]{ \label{subfig:relatable} \includegraphics[height = 2.4cm]{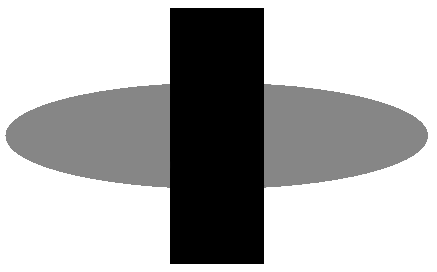}}\hspace{0.5cm}
    \subfigure[Non-relatable contours]{\label{subfig:non-relatable}\includegraphics[height=2.4cm]{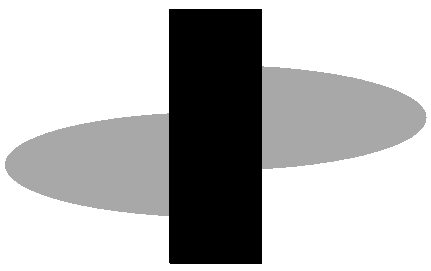}}\hspace{0.8cm}
    \caption{How many perceptual objects? (a) Because of the presence of two relatable contours, which are made of two pairs of relatable end-points (in the upper and lower  part of the gray shape, respectively), we perceive an ellipse occluded by a rectangle. (b) There is no pair of relatable contours so we perceive three different shapes.
    }
    \label{fig:relatability}
\end{figure}

In this paper, we are interested in computationally modeling this perceptual phenomenon, recovering what the brain infers about the structure and the relative depth of the objects composing the scene from a planar image. To simplify the analysis of our approach, we focus on scenes where objects appear at two different depths, ones occluding the others. The current approach can handle scenes with both partially occluded and fully visible objects. Our contribution is twofold: firstly, we propose a computational method relying on  perceptual findings related to amodal visual completion to compute the disoccluded objects that form the possible 3D interpretations or  configurations that arise  from a planar image; and secondly, we propose a Bayesian probabilistic model which chooses between these possible  interpretations of a planar image the most plausible one, justifying the visual completion human experience. The disocclusion method works by minimizing the Euler's elastica and incorporates the concepts of relatability of the occluded contours and  convexity of the disoccluded objects. Roughly speaking, two contours are relatable if they can be connected with a smooth contour without inflection points~\cite{KellmanShipley} (see Figure~\ref{subfig:relatable}--\subref{subfig:non-relatable})) . An equivalent and more precise definition  is given in Section~\ref{sec:theModel} (see Def.~\ref{defRel}) but let us now notice that the relatability property implies that two contours can be relatable no matter how far away their corresponding ending points are.
Once the objects conforming the scene are disoccluded, we follow a Bayesian approach and give definitions for the prior probability and the likelihood, measured, respectively, by the object complexities and  an elastica-based quantity. As a consequence, our probability model takes into account the shape of the objects in the hypothesized scenes as well as the effort of bringing these objects in their relative positions in the visual image. 

The structure of the paper is as follows. In Section~\ref{sec:relatedwork} we review the  related work and the fundamentals of visual completion. Section \ref{sec:theModel} is devoted to present the proposed approach: in particular, Subsection \ref{sec:binaryinpainting} presents the  object disocclusion method, while Subsection~\ref{sec:approach} details the probabilistic model. Section \ref{sec:algorithm} explains the numerical algorithm while Section~\ref{sec:experimentalresults} provides experimental results. Finally, we present our conclusions in Section \ref{sec:conclusions}. 

\section{{Related research}}\label{sec:relatedwork}

\begin{figure}[h]
    \centering
    \subfigure[Global-local convergent completion process]{ \label{subfig:convergent} \includegraphics[width=7cm]{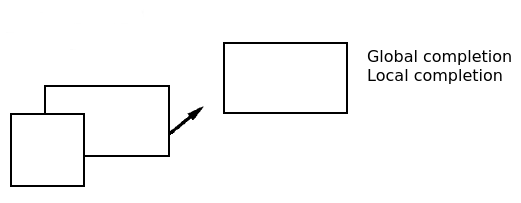}} \hspace{0.5cm}
    \subfigure[Global-local divergent completion process]{\label{subfig:divergent} \includegraphics[width=7cm]{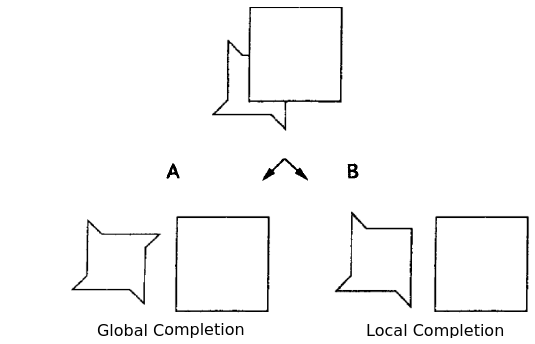}}
    \caption{Examples of global-local completion processes. Images adapted from \cite{van1995competing}.}
    \label{fig:convergentDivergent}
\end{figure}

\begin{figure}[h]
    \centering
    \includegraphics[height = 2cm]{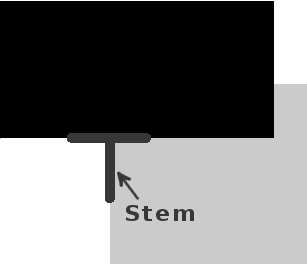}
    \caption{Example of a T-junction arising in presence of partial occlusion.}
    \label{fig:t-junction}
\end{figure}

The visual completion phenomenon has been intensively investigated during the past fifty years and it is still an active area of research \cite{kanizsa1972amodal,kanizsa1985seeing,michotte1991amodal,gerbino1987effect,ringach1996spatial,murray2004setting,froyen2015modal,carrigan2015differentiating}. Nowadays, it is well acknowledged that occlusion patterns evoke both local and global completion processes, but how the final perception outcome is conveyed is still not well understood. 
Local completion has been related to the good continuation of visible contours \cite{Wertheimer1923}, while global completion is driven by the simplicity principle \cite{koffka1935,Kanizsa1979}:  The assumption that the visual system  favors  interpretations characterized by phenomenal simplicity, such as symmetry, repetition, familiarity or context properties \cite{Sekuler}, and regularity, 
and typically leads toward the simplest completed shape, even though the good continuation principle may be violated. Figure~\ref{subfig:divergent} shows an example where two different amodal completions occur depending on whether a global cue as symmetry is incorporated or only more local cues, while in Figure~\ref{subfig:convergent}, both interpretations coincide. Some authors (e.g.,~\cite{moravec86} and references therein) have noticed that features favoring completion through good continuation are read out more quickly (in the very first second) than are features favoring completion through symmetry (which are incorporated in the following 9 seconds). The incorporation of different cues was also studied by Rubin~\cite{rubin2001role} who experimentally proved that local and global occlusion cues affect the perception of amodal completion at different stages of visual processing. As for global cues, the author focused in relatability and surface similarity, being cues that seem to be instantaneously used at first stages of occlusion perception. 
Relatability (see Figure~\ref{fig:relatability}) was first introduced by  Kellman and Shipley~\cite{KellmanShipley}, who noticed that it is a necessary global condition for completion to occur. 
Then, for the perception of amodal completion, Rubin proposed that the detection of local cues such as T-junctions (see Figure~\ref{fig:t-junction})
generates a local pattern of activation which launches a process of propagation of the contour which is either enhanced or stopped depending on whether or not other global cues such as relatability or surface similarity hold.

The evidence that the visual system generates multiple interpretations and that visual completion is the result of a competition between them,  was discussed by van  Lier  et  al. \cite{van1995competing,van1995multiple}. 
In  Figure\ref{subfig:divergent} is shown that sometimes global and local processes may diverge, and since it is not always the same process which is prevalent, a theory based on either local or global principles alone cannot hold. Later,  van  Lier  et  al.~\cite{VanDerHelm1}  proposed  an  integrative model of global and local aspects of occlusion.
In the present work, the global cues that we use are relatability and convexity (which are instantaneously considered at first stages of occlusion perception). As discussed in the following Section, we leave the incorporation of other global properties such as global self-similarities for a future work.

In computer vision, the computational translation of the visual completion phenomenon is commonly referred to as disocclusion or inpainting. A pioneering contribution to the recovery of image plane geometry was given by Nitzberg, Mumford and Shiota~\cite{NitzbergMumfordShiota}.
The authors proposed a variational model for segmenting the image into objects which should be ordered according to their depth in the scene, providing the so-called image 2.1 sketch. The minimization of the functional should be able to find the occluding and the occluded objects, while finding the occluded boundaries. 
The energy functional is defined in terms of region simplification and completion of occluded contours. Contours completion is achieved by linking signatures of occlusion, the T-junctions (see Figure~\ref{fig:t-junction}), with the Euler's elastica, so that the completion tends to respect the principle of good continuation~\cite{Kanizsa1979,Kanizsa1991}. 
Despite its theoretical importance, the complexity of minimizing this energy makes the approach far from practical applications. On the other hand, it was one of the main sources of inspiration for the first inpainting algorithms. A first formulation was proposed by Masnou and  Morel~\cite{masnoumorel1998}, who tried to interpolate the data into the uncomplete region by minimizing an energy functional based on the Elastica. A similar approach was followed in~\cite{BBCSV-01} and in the work of Chan and Shen~\cite{ChanShen}. Those methods belong to the so-called geometry-oriented methods where the images are modeled as functions with some degree of smoothness, expressed, for instance, in terms of the curvature of the level lines \cite{masnoumorel1998,BBCSV-01,ChanShen,masnou2002disocclusion,CittiSarti2006,inpaintingCao2011}
or the total variation of the image~\cite{inpaintingTVchan2001}. 
Binary inpainting tools for images are also used to disocclude shapes and thus can be considered as geometry-oriented methods. Binary inpainting can be based on diffusion processes followed by thresholding, named as thresholding dynamics (e.g.~\cite{MBO}). Thresholding dynamic interpolations also usually minimize a geometric functional,  based either on the length, area, or curvature of the shape contours \cite{MBO,ERT-05,GH-03}.

This work is focused on a computational model that, given an image of a 3D scene, it automatically outputs the preferred -- according to human perception -- interpretation of the  scene in terms of depth configuration of the scene objects together with their completion in case of occlusions.
A related and inspiring work in the literature is the proposal of van Lier et al.~\cite{VanDerHelm1}.
 They proposed to choose the preferred scene interpretation based on the minimum complexity or description code, taking into account local and global aspects of occlusion. 
 Their model assumes that the most likely interpretation  is the one that minimizes the sum of the complexity of three components of the visual pattern: (i) The internal structure, related to each of the visible shapes separately, (ii) the external structure, related to the positional relation between these shapes, and (iii) the virtual structure, related to the occluded parts of the shapes. 
The perceptual complexity of each of these three components is expressed in terms of structural information theory (SIT)~\cite{SIT}, a formal coding model that encodes complexity in terms of descriptive parameters.
However, van Lier et al.~do not automatically complete the occluded objects and the complexities are manually estimated from line drawings; thus their approach can not be directly applied to images in a computer vision task. The same authors noticed in \cite{van2011bayesian} that the global minimum principle can be settled in a Bayesian framework (see \cite{knill1996} and references therein) by properly defining prior and conditional probabilities. In this paper, based on a Bayesian framework, we propose a fully automatic method that can be applied to any image decomposed in shapes. 

\section{The model} \label{sec:theModel}

\begin{figure}[h]
 \centering
o \subfigure[Image source: Edoardo Accenti, 1967]{\includegraphics[height=0.35\textwidth]{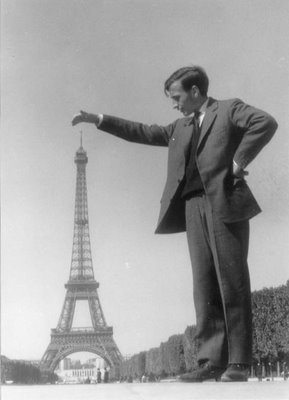}} \hspace{0.5cm}
 \subfigure[Image source: Laurent Laveder]{\includegraphics[height=0.35\textwidth]{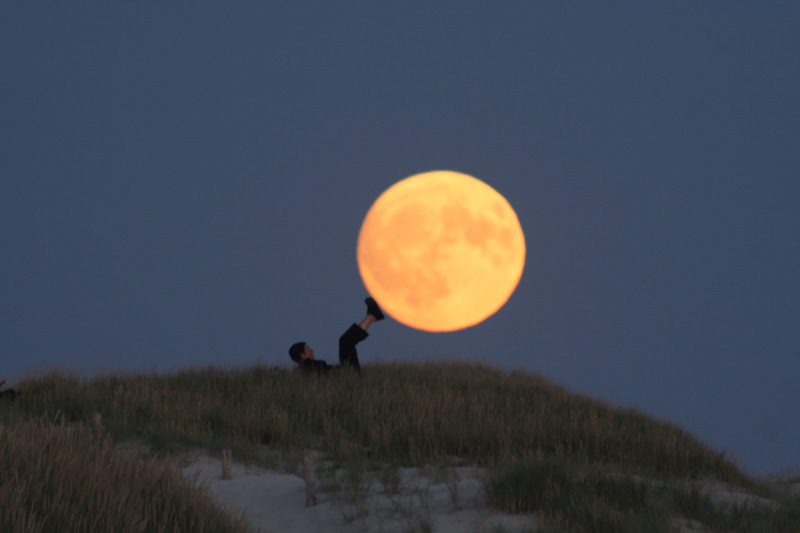}}
\caption{Two well-known optical illusions where the actual depth order of the scene objects is ambiguous or undetermined.}
\label{fig:illusions}
\end{figure}

\begin{figure}[h]
 \centering
  \subfigure[Three possible hypothesis.]{\label{subfig:3hyp}\includegraphics[width=0.35\textwidth]{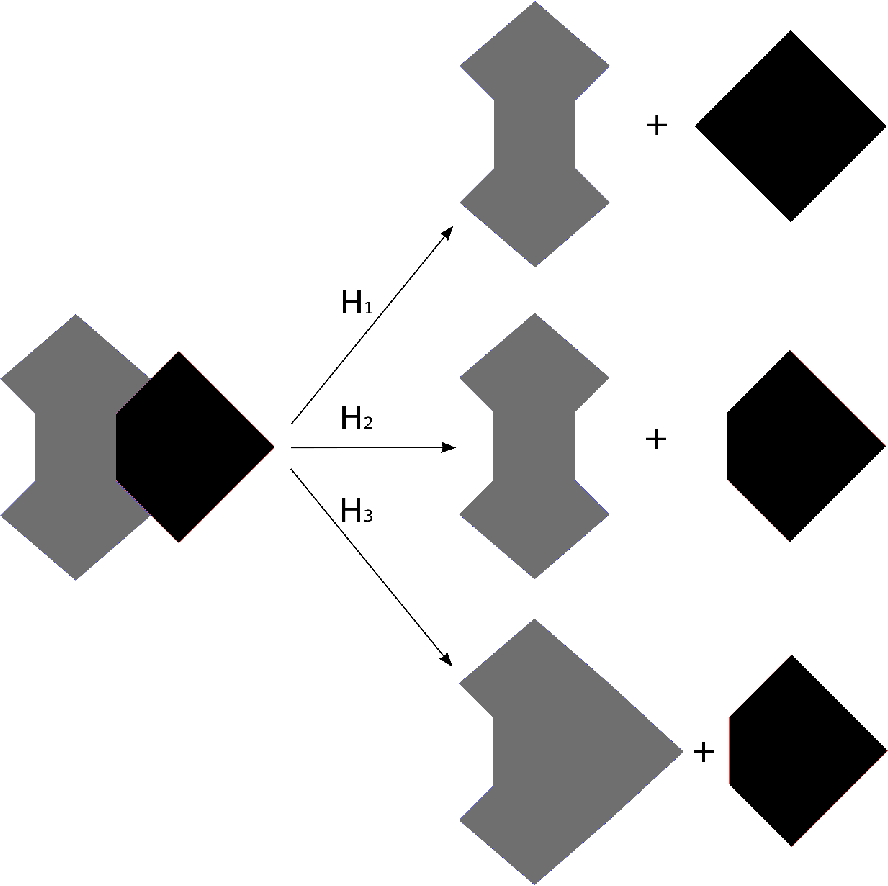}}\hspace{1.0cm}
  \subfigure[Two possible hypothesis.]{\label{subfig:2hyp}\includegraphics[width=0.35\textwidth]{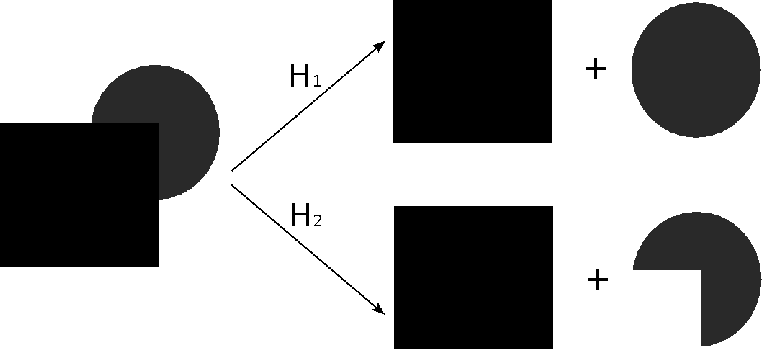}}
\caption{
Two examples with its corresponding hypothesis, $H_i$, for describing the 3D scene structure that gives rise to the observed image (the left-one in each example). In Figure~\ref{subfig:2hyp}, where $H_3=H_2$, the depth of the three-quarters-of-a-disk shape and the square is not the same: in $H_3$ both objects appear at the same depth and the real relative size is (exactly) the one reflected in the proximal image, while in $H_2$ the three-quarters-of-a-disk shape is closer to the observer and the square could be of real bigger size but farther away}
\label{fig:nHypothesis}
\end{figure}

Our model is grounded in two elements: a disocclusion method that computes the different objects conforming different potential scenes that are compatible with the given planar image, and a probabilistic model that quantitatively justifies which scene configuration is the preferred. As we are considering two-depth images, the possible interpretations or hypothesis of the real 3D scene are three, namely: object A occluding object B, B occluding A, or A and B fitting together forming a mosaic (see Figure~\ref{fig:nHypothesis}). We will denote them by $H_1$, $H_2$, and $H_3$. Let us remark that sometimes the objects in the third interpretation $H_3$ (i.e., A and B fitting together) coincide with one of the others, both perceptually and using our algorithm (an example is shown in Figure~\ref{subfig:2hyp}), or even the objects in all three hypothesis coincide (an example is shown in Table \ref{tab:alltogether3}). Even if the objects forming the scene coincide in different hypothesis, the depth ordering is not the same in each hypothesis. In Section~\ref{sec:experimentalresults} we will provide some experiments analyzing this phenomenon which is related to the well-known optical illusion of relative depth perception of the objects (see Figure~\ref{fig:illusions}). In this work, when $H_3$ coincides with one of the other hypothesis, we assume that both objects appear at the same depth (see Figure~\ref{subfig:2hyp}) and then the associated probabilities will decide which hypothesis has the highest likelihood. In the Perception community, the observed image is often called the \textit{proximal stimulus} (e.g., the left image in Figure~\ref{subfig:3hyp} and \subref{subfig:2hyp}), and each of the hypothesized interpretations $H_i$ is called the \textit{distal stimulus}.

Our method first computes several distal interpretations of the scene which are compatible with the proximal planar image. 
Rubin~\cite{rubin2001role} studied the role of T-junctions, relatability and surface-similarity in the amodal completion phenomenon and illusory contour perception. The author proposed that T-junctions, being a local cue for
occlusion, are used to launch the completion process when contours are relatable~\cite{KellmanShipley}.
Then, the Gestalt law of good continuation plays an important role. 
This motivates us to use the Euler's elastica in order to smoothly continue the contours. 
Let us recall that, given two T-junctions at points $x_1$ and $x_2$, with
tangents $\tau_{x_1}$ and $\tau_{x_2}$ to the respective terminating stems (also called T-stems, see Figure~\ref{fig:t-junction}), 
Euler solved in 1744~\cite{Mumford} the problem of joining them with
a smooth continuation curve minimizing
\begin{equation}\label{eq:Elastica}
\int_{\mathcal{\gamma}} (\kappa^2(s) + \beta) \mathrm{d}s, 
\end{equation}
where $\beta>0$ and the minimum is taken among all the curves $\mathcal{\gamma}$ joining
$x_1$ and $x_2$ with tangents $\tau_{x_1}$ and $\tau_{x_2}$, respectively, $\kappa(s)$ denotes the curvature of $\mathcal{\mathcal{\gamma}}$, $\mathrm{d}s$ its arc length, and $\beta$ is a positive constant. 
The parameter $\beta$ plays a geometric role by settling the expected underlying a priory regularity. In this sense, with a larger $\beta$, the energy favors the completion with straight lines (minimal length).
Otherwise, smooth curves of low curvature are favored even if their length is increased.  Figure~\ref{fig:beta-parameter} illustrates the effect of the parameter $\beta$; as $\beta$ decreases the disoccluded shape converges to a disk; in the limit case $\beta=0$, the energy to be minimized is the Willmore energy (with the boundary constraints on the tangents).
The elastica energy  is not lower semicontinuous~\cite{BelletiniEtAl} and some relaxed versions have been proposed~\cite{BelletiniEtAl,masnoumorel1998,BBCSV-01}, which are compatible with Kanisza's amodal completion theory~\cite{Kanizsa1991}. It 
has been frequently used to solve different computer vision problems (e.g.~\cite{NitzbergMumfordShiota,Mumford,masnoumorel1998,ThornberWilliams,LeungMalik} among others). In a recent work that proposes a computational method for modal completion \cite{kang2014illusory}, the elastica is a key ingredient to obtain illusory contours. A method for both modal and amodal completion which uses geodesics in the group of rotations and translations was proposed in \cite{CittiSarti2006}. In this work, the elastica is used in two ways. We propose in Section~\ref{sec:binaryinpainting} an elastica-based object disocclusion method which incorporates the relatability of partially occluded contours and the convexity of the disoccluded objects. On the other hand, the elastica is also used in Section~\ref{sec:approach} to select the most probable disoccluded scene.

\subsection{Elastica-based object disocclusion}\label{sec:binaryinpainting}

For disoccluding the objects, we focus on the completion that takes place in the first time instants of observation \cite{moravec86} or when the local processes dominate due to limited regularities in the object or low saliency of the symmetry cues versus the good continuation \cite{Sekuler}. In the perception community, this completion is usually called local completion.
 The disocclusion method we propose integrates global and local cues:
Global cues such as relatability and convexity are incorporated in the initial step of our algorithm, followed by local ones such as smooth continuation. 

We propose to disocclude partially occluded objects by a binary inpainting algorithm that simulates the minimization of the elastica \eqref{eq:Elastica}. Disocclusion, also known as image completion or inpainting, is the recovery of missing or corrupted parts of an image in a given region so that the reconstructed image looks natural. Most available methods for inpainting can be divided into two groups: geometry~\cite{masnoumorel1998,BBCSV-01,ChanShen,inpaintingTVchan2001,masnou2002disocclusion,ERT-05,inpaintingCao2011} and texture-oriented methods~\cite{inpainting_demanet_2003,inpainting_criminisi_2004,inpainting_wexler_irani_2007,inpainting_kawai_2009,inpainting_aujol_2010,inpainting_arias_2011,mansfield_2011}. The synthesis of methods of these two types is still an open question~\cite{inpaintingCao2011}. Since we are interested in recovering objects or shapes, we will focus on geometry-oriented methods, where images are usually modeled as functions with some degree of smoothness, expressed, for instance, in terms of the curvature of the level lines or the total variation of the image. Taking advantage of this structure, these methods interpolate the inpainting domain by continuing the geometric structure of the image (its level lines or edges), usually as the solution of a (geometric) variational problem or by means of a partial differential equation.

\begin{figure}[h]
    \centering
    \subfigure[Input]{\includegraphics[width = 2.5cm]{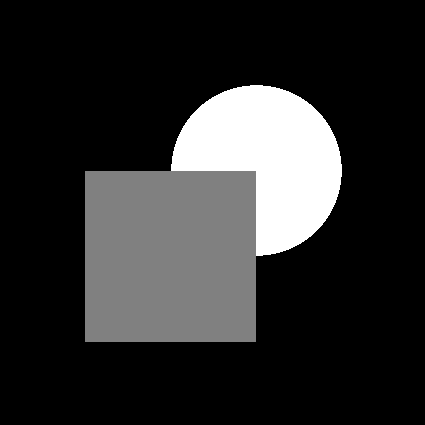}}
    \subfigure[Initialization]{\includegraphics[width = 2.5cm]{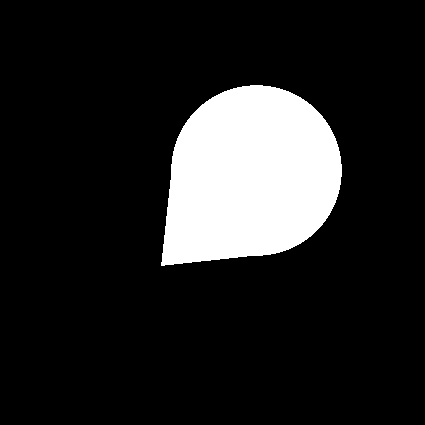}}\\
    \subfigure[$\beta=0.3$\newline  $\tilde{p}=0.48$]{\includegraphics[width = 2.5cm]{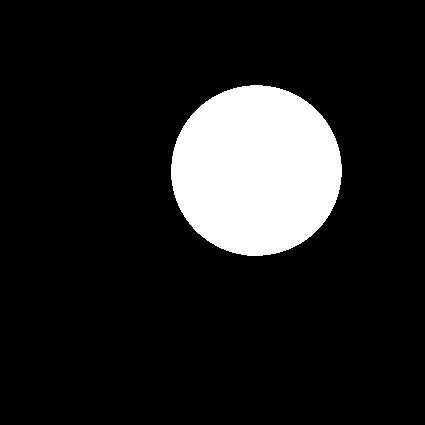}}    
    \subfigure[$\beta=0.6$\newline $\tilde{p}=0.48$]{\includegraphics[width = 2.5cm]{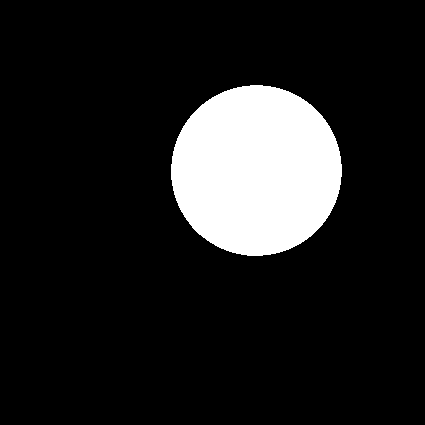}}    
    \subfigure[$\beta=0.9$\newline $\tilde{p}=0.43$]{\includegraphics[width = 2.5cm]{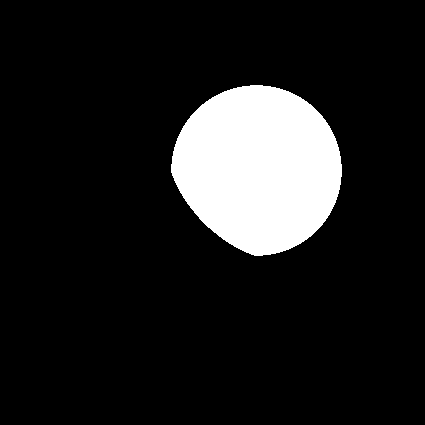}}
    \subfigure[$\beta=1.2$\newline $\tilde{p}=0.41$]{\includegraphics[width = 2.5cm]{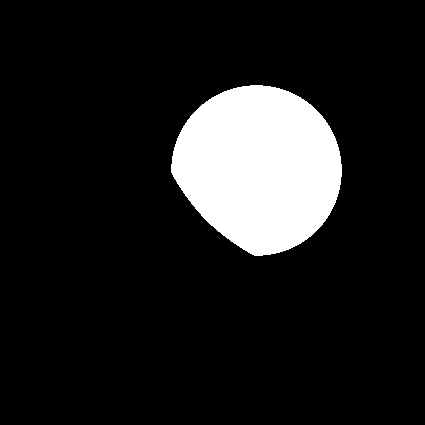}}
    \subfigure[$\beta=1.5$\newline $\tilde{p}=0.40$]{\includegraphics[width = 2.5cm]{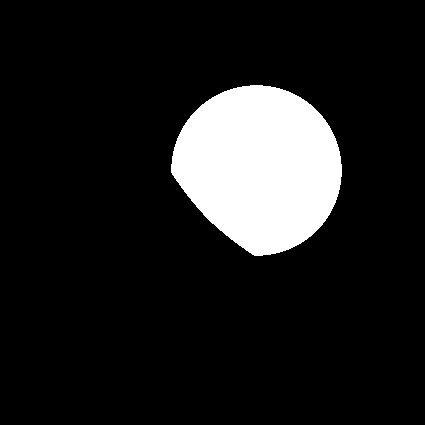}}
    \caption{Disocclusion results for different values of $\beta$ and its associated certainty $\tilde{p}$.  When $\beta$ grows, more weight is given to the length of the disoccluded boundary $\gamma$ and then disoccluded straight lines are favored no matter if curvature singularities appear at the endpoints of  $\gamma$. On the other hand, when $\beta>0$ is small, more weight is given to the square of the curvature and thus the algorithm avoids singularities of the curvature no matter if it produces a bigger length (as $\beta$ decresases the disoccluded shape converges to a disk; in the limit case $\beta=0$, the energy to be minimized is the Willmore energy).}
    \label{fig:beta-parameter}
\end{figure}

\begin{figure}[h]
    \centering
    \subfigure[$\beta=0.253$]{\includegraphics[width = 2.5cm]{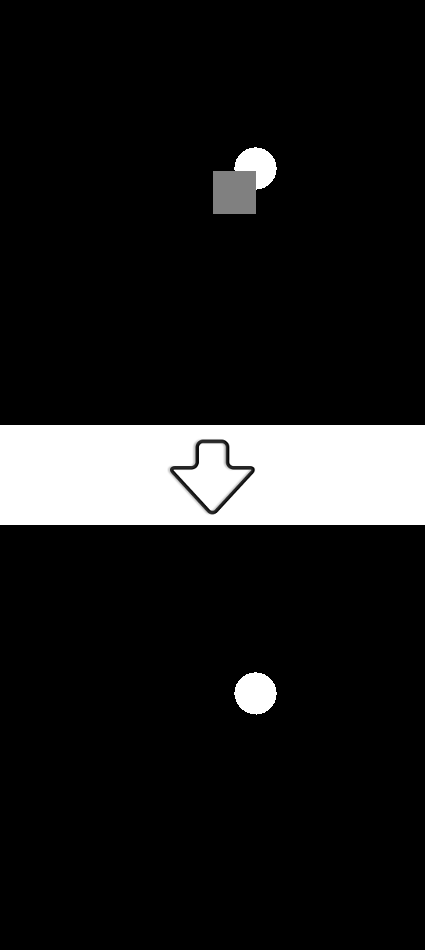}}
    \subfigure[$\beta=0.375$]{\includegraphics[width = 2.5cm]{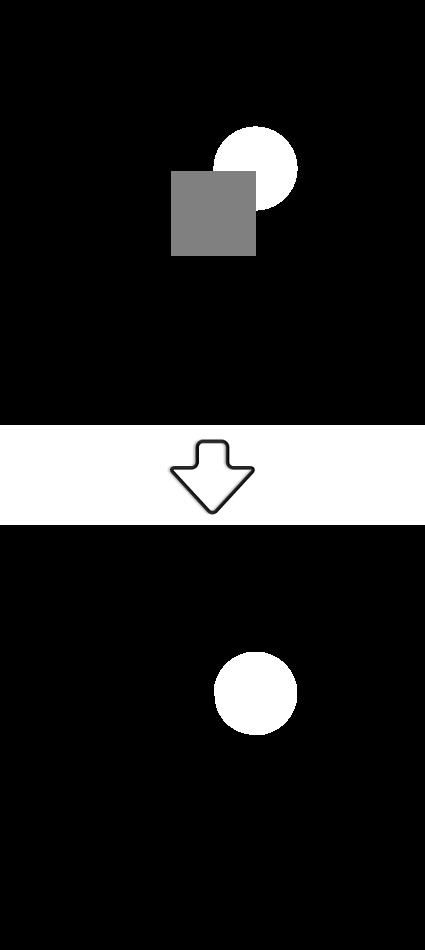}}
    \subfigure[$\beta=0.53$]{\includegraphics[width = 2.5cm]{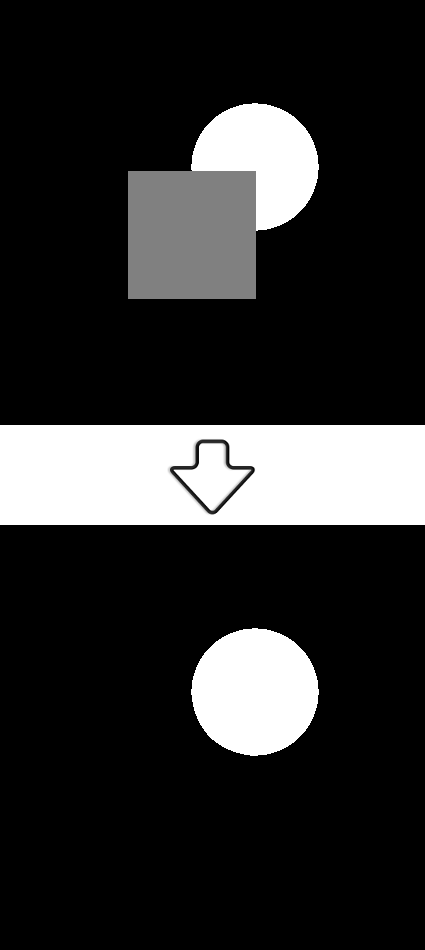}}
    \subfigure[$\beta=0.6$]{\includegraphics[width = 2.5cm]{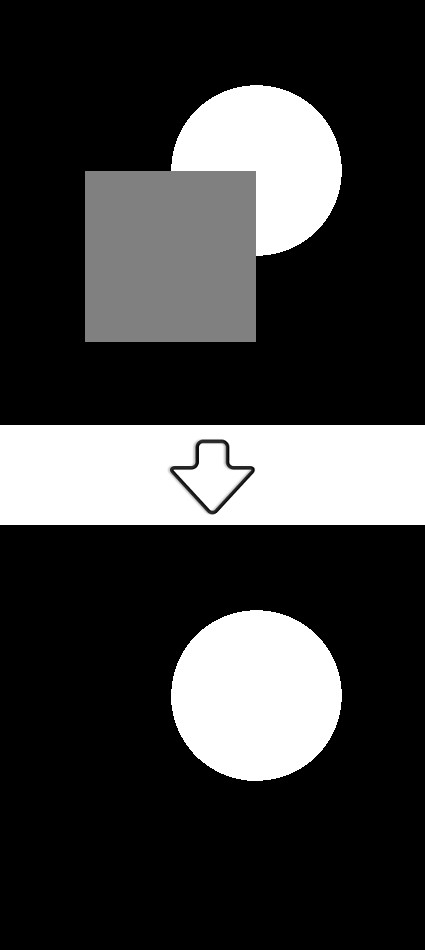}}\\
    \caption{Depending on the resolution of the proximal stimulus, the parameter $\beta$ needs to be adapted to obtain the same underlying shape regularity (in this case, a constant curvature of the completed shape). For smaller circles, a smaller value of $\beta$ is needed to obtain a perfect disk (more details are given in the text).}
    \label{fig:b_vs_r}
\end{figure}

In this paper, we are concerned solely by the shape of the objects. Thus, we work with segmented objects and we perform a geometric inpainting of the binary images that represent these objects. More precisely,
we disocclude each object in each hypothesis by  separately considering the hypothesized occluding object as the inpainting mask. The object is automatically completed in such a way that its boundary minimizes a relaxed version of the elastica \eqref{eq:Elastica}. For that, the object to be completed is represented in a binary image (given by the object segmentation) and its completion is performed through a threshold dynamics algorithm 
which mainly consists in a diffusion process followed by a thresholding. In our case, the minimization algorithm iteratively alternates one step of the Grzibovskis-Heintz scheme~\cite{GH-03} that decreases $\int_{\mathcal{\gamma}} \kappa^2 \mathrm{d}s$, one step of the standard Merriman-Bence-Osher scheme~\cite{MBO} that decreases $\beta \int_{\mathcal{\gamma}}  \mathrm{d}s$, and a thresholding step, as proposed by Esedoglu et al. in~\cite{ERT-05}. We present the pseudo-code and more details in Algorithm~\ref{thr.dyn.alg}. Figure~\ref{fig:beta-parameter} shows an example illustrating how the parameter $\beta$ affects the disoccluded shape. When $\beta$ is big, more weight is given to the length of the curve and then straight lines are favored. When $\beta$ decreases the disoccluded shape converges to a disk avoiding singularities of the curvature no matter if it produces a bigger length. On the other hand, depending on the resolution of the proximal stimulus, which translates into a smaller or bigger curvature of curvy boundaries, the parameter $\beta$ needs to be adapted to obtain the same underlying shape regularity. An example is shown in Figure~\ref{fig:b_vs_r}: circles with larger radius need a larger value of $\beta$ in order to obtain the same regularity of the disoccluded shape.  The reason is the following: the curvature of smooth plane curves is defined as the inverse of the radius of the osculating circle (the unique circle which most closely approximates the curve near the point). Therefore, there is a relationship between the numerical curvature of the disoccluded objects and the a priori regularity imposed through the parameter $\beta$: the larger the $\beta$, the larger the expected radius of the osculating circle.

In Section~\ref{sec:approach} we define prior probabilities and likelihoods which take into account global and local properties of the shape. As a consequence, the Bayesian approach is able to choose the more likely amodal completion not only between the different hypothesis on the scene configuration for a fixed disocclusion parameter $\beta$   (as shown in Sect.~\ref{sec:approach}), but also between several disocclusions associated to different parameters $\beta$, and therefore to integrate some global completion properties such as symmetry or repetitions. For instance, Figure~\ref{fig:beta-parameter} presents the different hypothesis certainty,  denoted by $\tilde{p}$ in Sect~\ref{sec:approach}, for different values of $\beta$.  However, for the experimental results in Section~\ref{sec:experimentalresults}, where the non-occluded part of the shapes have no constant curvature, the parameter $\beta$ has been fixed; an efficient method to compute the best $\beta$ goes out of the scope of the present work. 

\subsubsection{Initialization of the inpainting mask}\label{initialization}
\begin{figure}[h]
    \centering
    \subfigure[]{\includegraphics[width = 2.5cm]{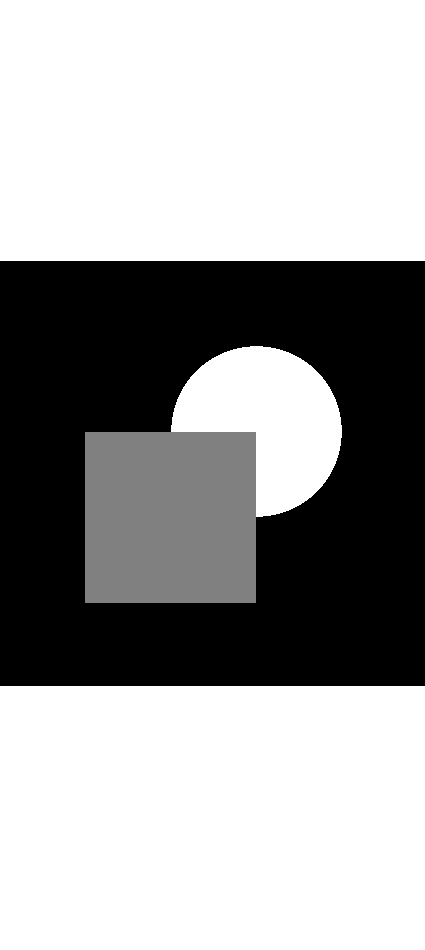}}
    \subfigure[]{\includegraphics[width = 2.5cm]{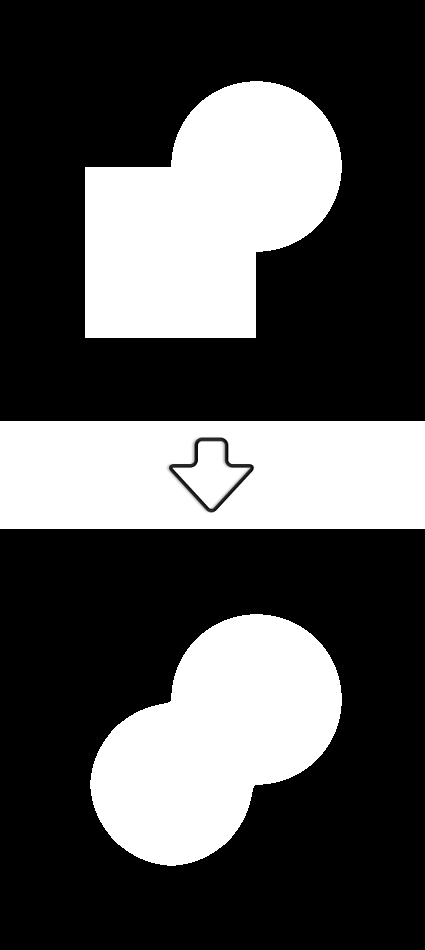}}
    \subfigure[]{\includegraphics[width = 2.5cm]{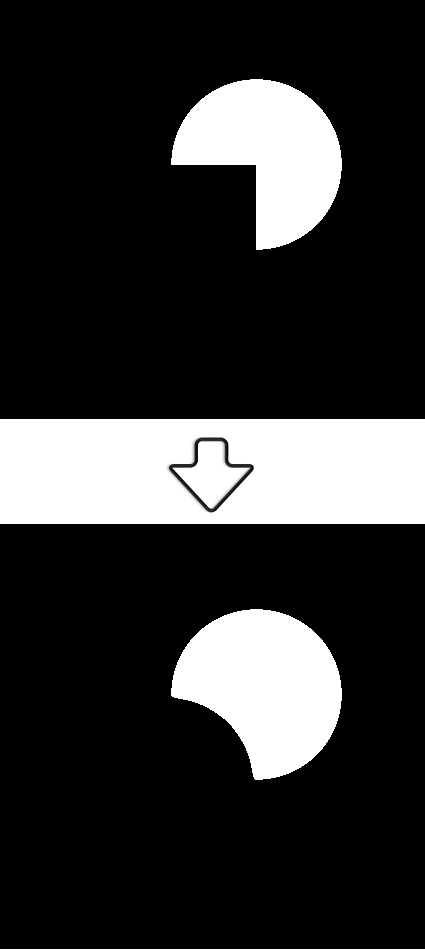}}
    \subfigure[]{\includegraphics[width = 2.5cm]{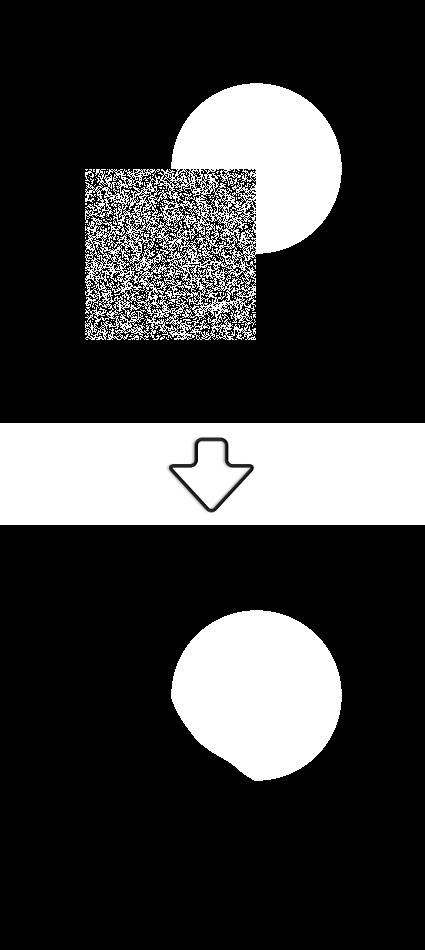}}
    \subfigure[]{\includegraphics[width = 2.5cm]{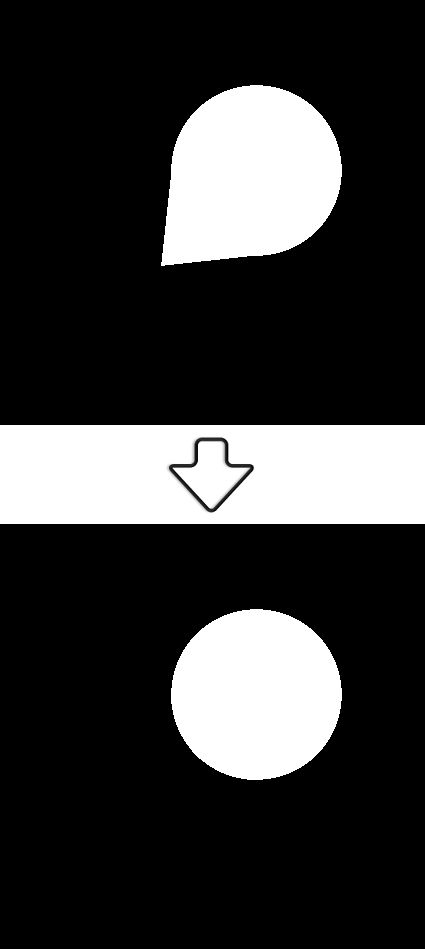}}\\
    \caption{Each column contains an inpainting experiment with the same input image (a) where the inpainting mask is given by the square. The first row shows several initializations, namely, initializating the mask with white (b), black (c), random (d) or with our proposal (e). The second row shows the inpainted result of the corresponding image above. Notice how the proposed initialization gives a better result (according to the Gestalt laws of perception) and produces a completion that maintains the tangents at the endpoints of the disoccluded boundary.}
    \label{fig:initializations}
\end{figure}

Since the elastica energy \eqref{eq:Elastica} is not convex, the inpainting result depends on the  initial condition inside the inpainting mask. Let us illustrate it with a simple example. In Figure~\ref{fig:initializations} we show the inpainting results (shown in the second row) obtained by minimizing the elastica with different initializations, namely, initializing the mask with white, black, random (black and white  chosen randomly from a uniform distribution) or with our proposal, which is explained in the remainder of this section. Notice how the proposed initialization gives a better result (according to the Gestalt laws of perception) and produces a completion that maintains the tangents at the endpoints of the disoccluded boundary.

In order to automatically compute an initialization of the inpainting problem sufficiently close to what humans perceive as disoccluded objects by amodal completion, we  incorporate perceptual cues such as relatability of object contours~\cite{KellmanShipley} and convexity of the disoccluded objects.\par

The notion of relatability (see Figure~\ref{fig:relatability}) was introduced by Kellman and Shipley~\cite{KellmanShipley} in the attempt of defining under which conditions visual completion occurs. Let us recall the definition of relatability. 
\begin{defn}[Relatability~\cite{KellmanShipley,SinghHoffman}]
Let $E_1$ and $E_2$ be two (non-closed) edges and let $x_1$ and $x_2$ be one of their respective end-points. Let $\tau_{x_1}$ and $\tau_{x_2}$ be the tangents at these points, pointing to the direction along which the interpolated contour must continue. Consider the semi-lines $s_1=\{ x_1+\lambda\tau_{x_1},\; \lambda\geq 0\}$ and $s_2=\{ x_2+\lambda\tau_{x_2},\; \lambda\geq 0\}$. Then, $E_1$ and $E_2$ are relatable if: (a)  $s_1$ and $s_2$ intersect, and (b) their outer angle of intersection (i.e. from $\tau_{x_1}$ to $-\tau_{x_2}$) is acute or $90^{\circ}$. \label{defRel}
\end{defn}
In~\cite{SinghHoffman}, the authors showed that this definition is equivalent to the existence of a  smooth contour without inflection  points connecting $x_1$ and $x_2$, and that the interpolating curve does not turn through a total angle of more than $\frac{\pi}{2}$.

Since (non-occluded) objects in the world tend to be convex \cite{burge2010natural}, we favor the convexity of the disoccluded object by taking advantage of the following  well-known property of convex sets.

\begin{lemma}
    Every closed convex set in $R^n$ is the intersection of the closed half-spaces that contain it.
    \label{lemma}
\end{lemma}

\begin{figure}[h]
    \centering
    \subfigure[]{\label{subfig:ps1} \includegraphics[width = 2.5cm]{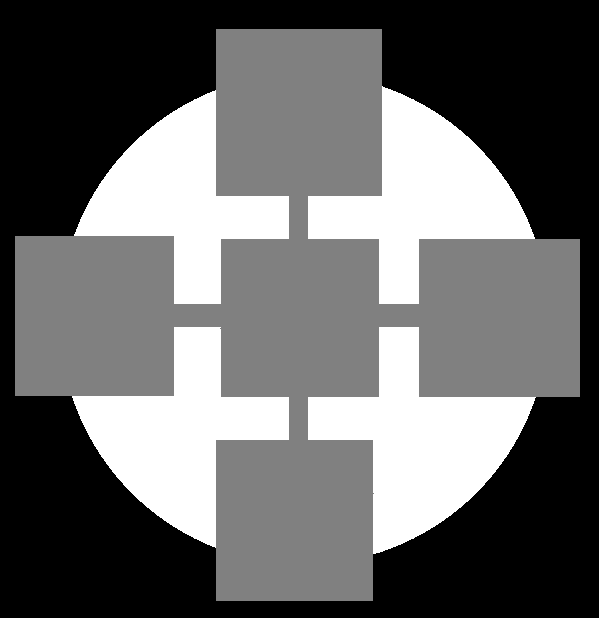}}
    \subfigure[]{\label{subfig:tangents1}\includegraphics[width = 2.5cm]{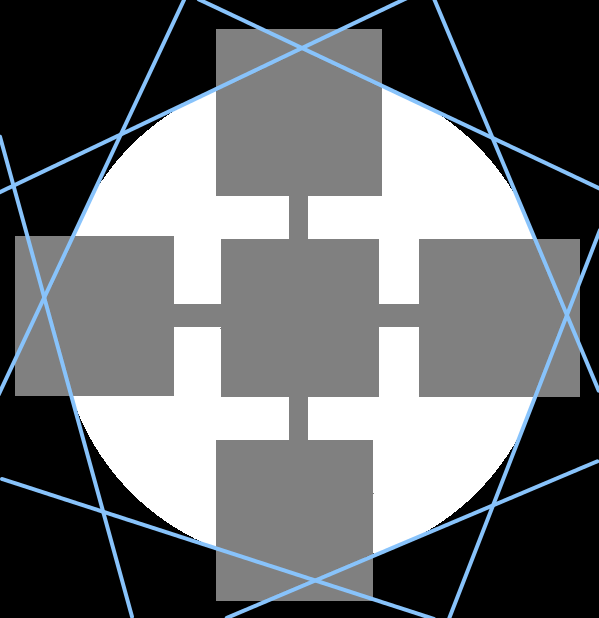}}
    \subfigure[]{\label{subfig:rel1}\includegraphics[width = 2.5cm]{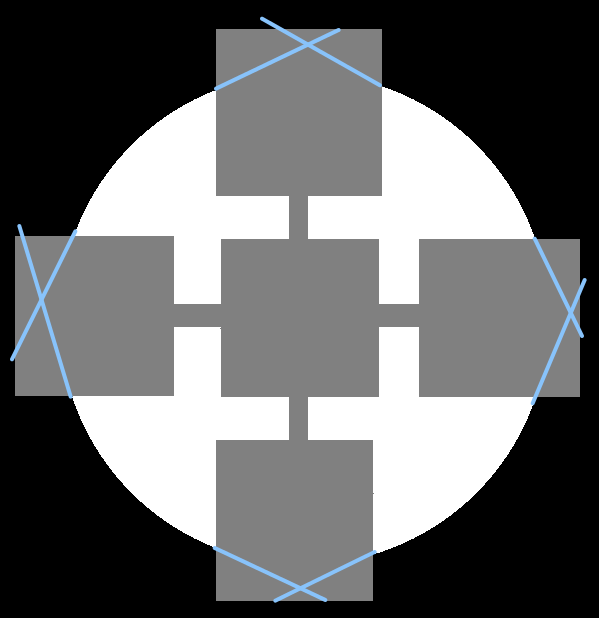}}
    \subfigure[]{\label{subfig:lema1}\includegraphics[width = 2.5cm]{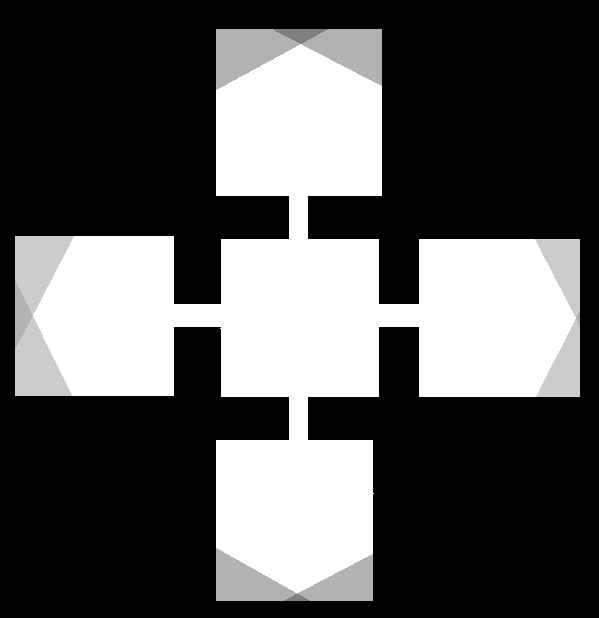}}
    \subfigure[]{\label{subfig:iniMask1}\includegraphics[width = 2.5cm]{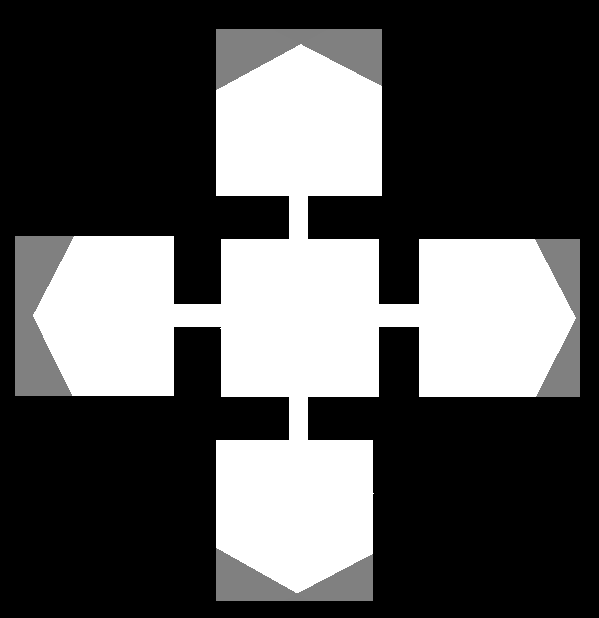}}\\
    \subfigure[]{\label{subfig:ps2} \includegraphics[width = 2.5cm]{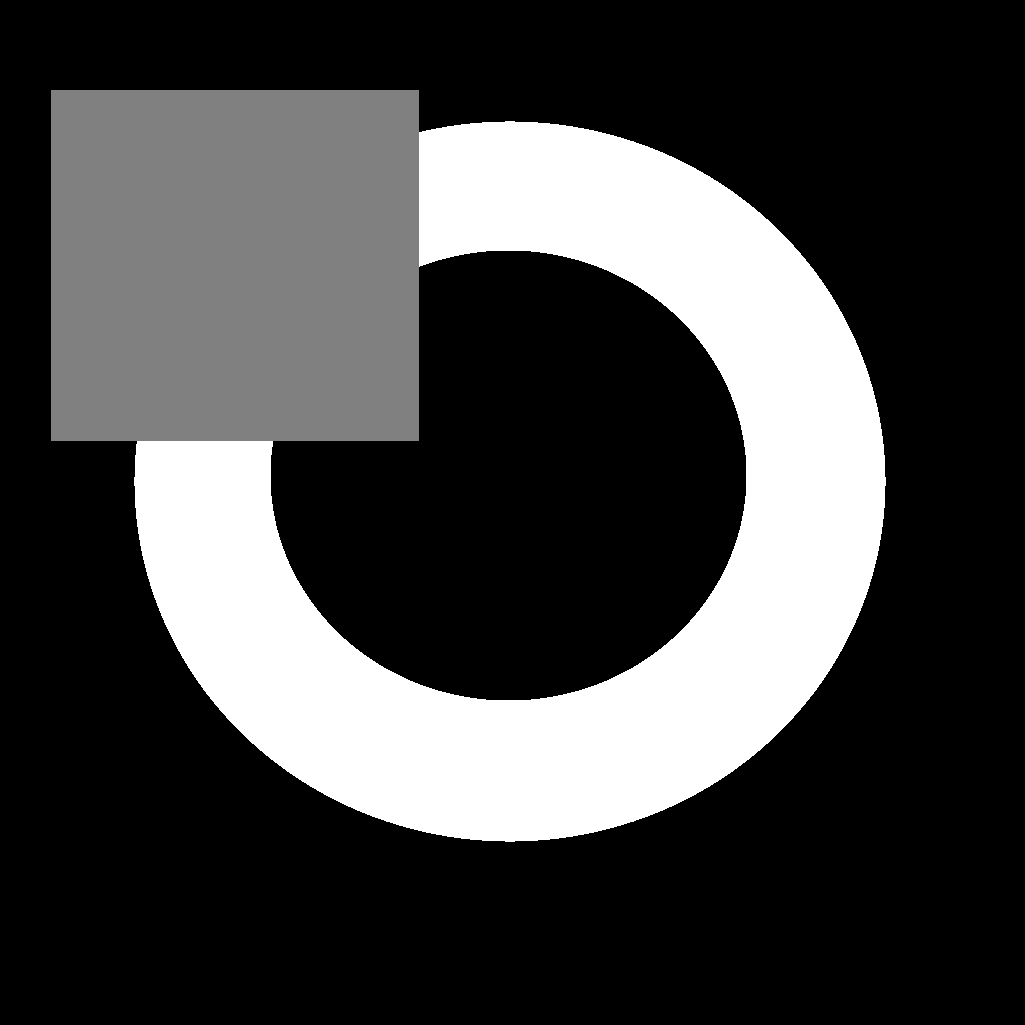}}
    \subfigure[]{\label{subfig:tangents2}\includegraphics[width = 2.5cm]{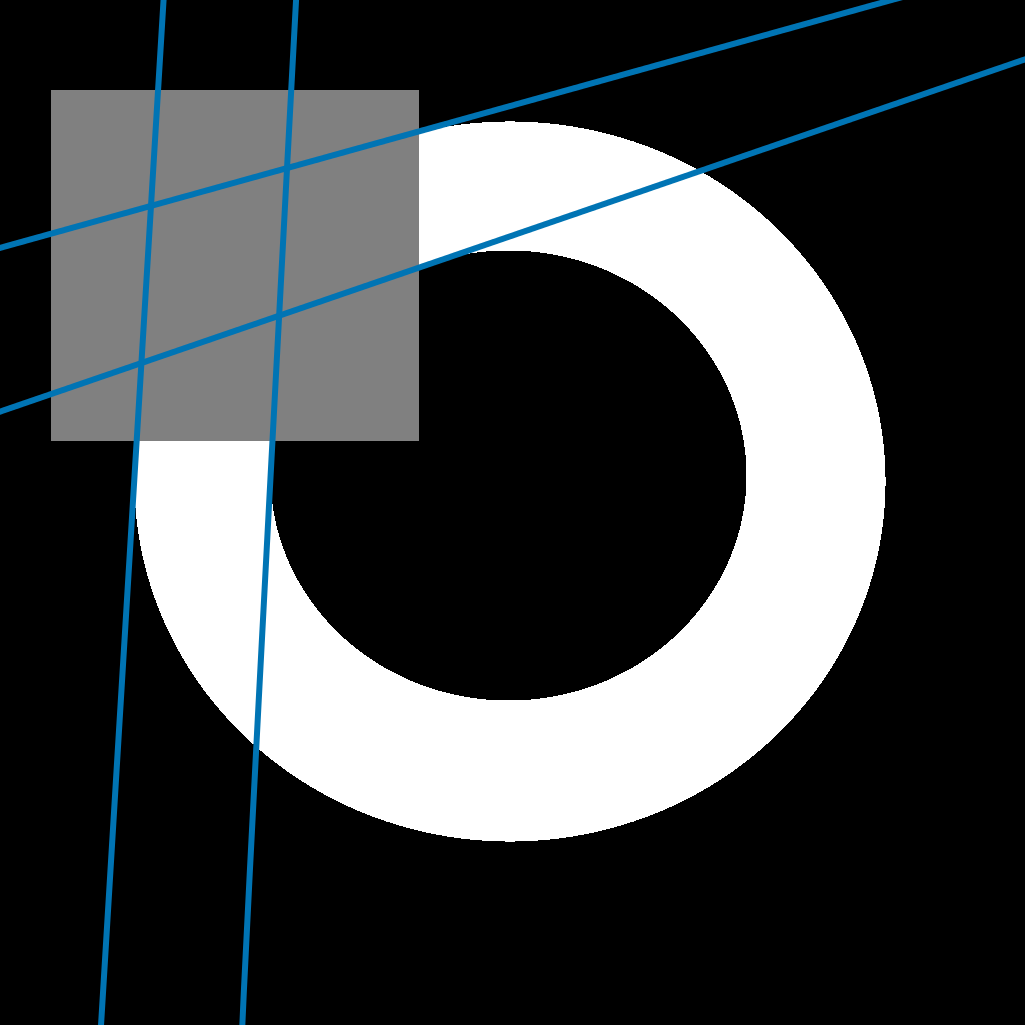}}
    \subfigure[]{\label{subfig:rel2}\includegraphics[width = 2.5cm]{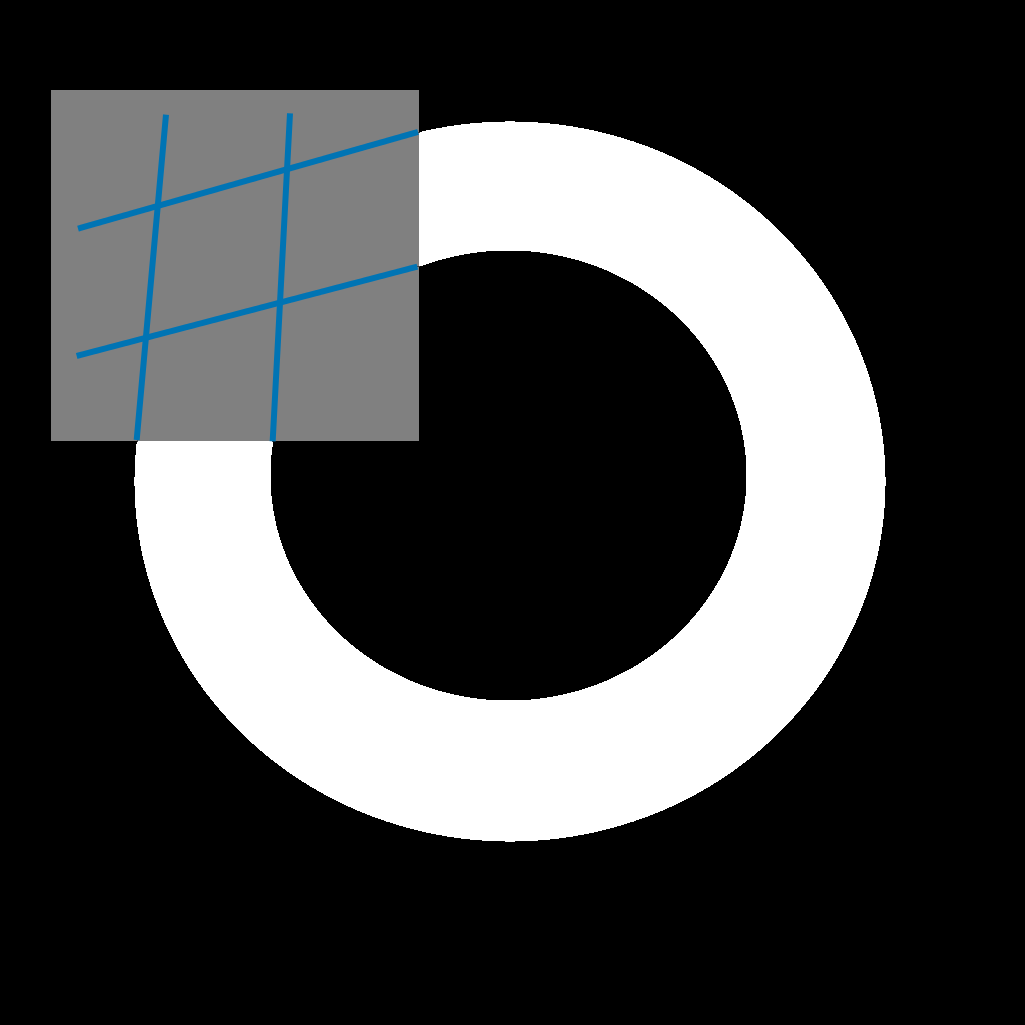}}
    \subfigure[]{\label{subfig:lema2}\includegraphics[width = 2.5cm]{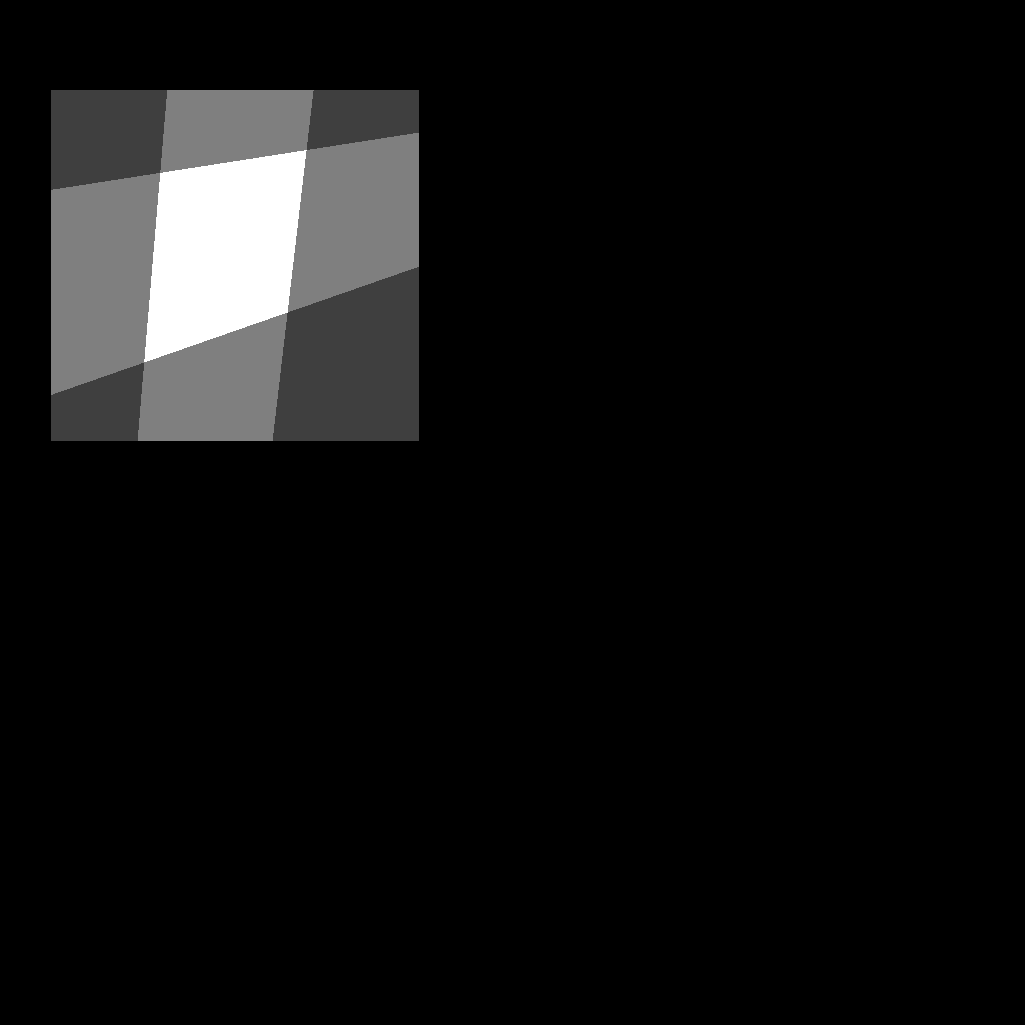}}
    \subfigure[]{\label{subfig:iniMask2}\includegraphics[width = 2.5cm]{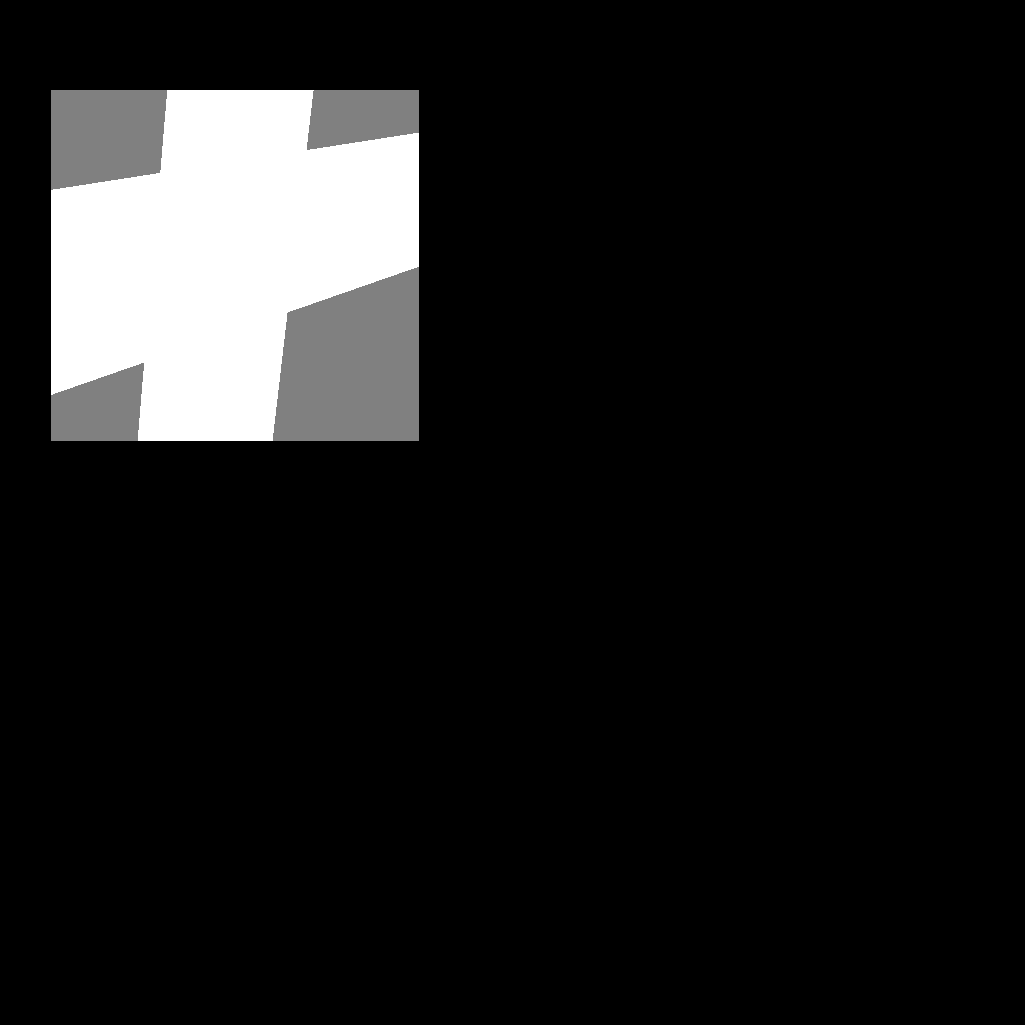}}    
    \caption{Two examples of relatability- and convexity-based initialization of the inpainting mask. (a), (f): shapes to disocclude (in white) and inpainting mask (in gray); (b), (g): extension by semi-lines (in blue, see Def.~\ref{defRel}) of the tangents of the level-lines arriving to the inpainting mask; (c), (h): relatable countours (in blue); (d), (i): votes on the different half-spaces defined by any pair of relatable contours (darker colors mean less votes); (e), (j): final initialization of the mask, after thresholding.}
    \label{fig:iniMask1}
\end{figure}

The automatic initialization of the binary image inside the inpainting mask is illustrated in Figure~\ref{fig:iniMask1}. In practice, our algorithm considers all the end-points of the object contours (given in this case by the level lines) arriving to the inpainting mask together with their tangents (illustrated in Figure~\ref{subfig:tangents1} by a line passing through them), and computes all the possible pairs of relatable contours (shown in Figures~\ref{subfig:rel1} and~\ref{subfig:rel2}). In order to compute these tangents we use the Line Segment Detector~\cite{lsd}. Then, for each pair of relatable contours, for the end-point $x_i$ and tangent $\tau_{x_i}$ we consider the half-space $\{x\in\R^2: \langle\tau_{x_i}^{\bot},x\rangle - \langle \tau_{x_i}^{\bot},x_i\rangle\geq 0\}$ (or $\leq 0$, depending on which half-space the object is), and we assign a vote to  the half-space on which the known object is. Figure~\ref{subfig:lema1} and Figure~\ref{subfig:lema2} displays the image gathering these votes in the inpainting mask, for the shapes shown in Figure~\ref{subfig:ps1} and Figure~\ref{subfig:ps2}, respectively (the shape to disocclude is shown in white and the inpainting mask is shown in gray, respectively). Let us remark that, in order to better illustrate our perceptually inspired initialization, in Figure~\ref{subfig:lema1}  and Figure~\ref{subfig:iniMask1} (respectively, in Figure~\ref{subfig:lema2}  and Figure~\ref{subfig:iniMask2}) we only show the computed values inside the inpainting mask. Finally, we binarize the image containing the votes with a threshold based on a rank order  filter of these votes. We order the votes in increasing order and start with a threshold with the value ranked at percentile $75$th. If no new connected components appear in the initialization with this threshold we keep it. Otherwise we decrease the threshold (taking the preceding ordered value) and repeat the process until no new connected components appear. Two different examples of this binary image are shown in Figure~\ref{subfig:iniMask1} and \ref{subfig:iniMask2}, they are the  initialization of the binary inpainting algorithm. Figure~\ref{subfig:iniMask1} shows an example where the threshold on the votes correspond to the $75$th percentile while in Figure~\ref{subfig:iniMask2} the threshold was automatically decreased to the $65$th percentile in order to obtain an initialization with a single connected component.

\subsection{Elastica-based probabilistic model }\label{sec:approach}
In this section, we follow a Bayesian approach \cite{knill1996} in order to choose among
all possible interpretations of the scene, the most plausible one. We propose definitions for the prior and the conditional probabilities which take into account the global complexity of the objects in the hypothesized scenes as well as the effort of bringing these objects in their relative positions in the visual image. As a consequence, the result of this probability model indicates that the most simple interpretation is the one that more likely results from the amodal completion process, which was also suggested by \cite{van2011bayesian}.

Inspired by  the work of van Lier et al.~\cite{VanDerHelm1}, our probabilistic model takes into account
the complexity of the objects. Each hypothesized scene is formed both by completely visible objects and disoccluded objects
(computed using the method described in previous Section~\ref{sec:binaryinpainting}); their respective global complexities are taken into account to define the prior probability of the hypothesized scene under analysis.  
The likelihood, i.e.~the conditional probability of the given image (proximal stimulus) given a certain hypothesis (distal stimulus) is defined through an Euler's elastica-based quantity  that measures two attributes: the effort of bringing these objects in their relative positions given in the image and the smoothness of the disoccluded boundaries. Our probabilistic model provides a formalization allowing to computationally verify directly on images the proposal of van der Helm \cite{VanDerHelm1,van2011bayesian} (based on  manually estimating complexity from line drawings) giving a probabilistic interpretation of the visual completion process. 

We propose to predict and justify the preferred interpretation by maximizing the responsibility or a posterior probability, given by the Bayes' rule as 
\begin{equation}\label{Bayes}
 p(H_i/I)= \frac{p(I/H_i) p(H_i)}{p(I)}
\end{equation}
over the hypothesized interpretations $H_i$, where $I$ is the proximal stimulus or given image. As the quotient $p(I)$ remains the same for all hypothesis $H_i$ in the maximization process, we propose to select the preferred hypothesis $H_{\mathrm{P}}$ by 
\begin{equation}\label{max}
H_{\mathrm{P}}=\mathrm{arg}\max_i{p}(H_i/I)=\mathrm{arg}\max_i \tilde{p}(I/H_i)\,\tilde{p}(H_i).
\end{equation}

\begin{figure}[h]
  \centering
  \subfigure[$H_1$]{\label{subfig:b_dc}\includegraphics[width=2.8cm]{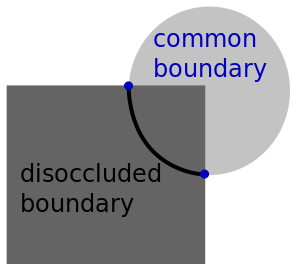}}\hspace{1.5cm}
  \subfigure[$H_2$]{\label{subfig:b_c}\includegraphics[width=2.8cm]{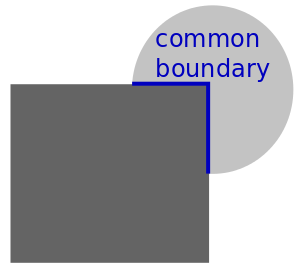}}
\caption{Boundaries $\mathcal{B}^c_i$ and $\mathcal{B}^d_i$ ($i=1,2$) for hypothesis $H_1$ and $H_2$ from Figure~\ref{subfig:2hyp}. (a) $H_1$: square in front of circle; when the circle is disoccluded the common boundary $\mathcal{B}^c_1$ among both objects is formed by the 2 T-junction points (in blue), while the disoccluded boundary $\mathcal{B}_1^d$ is made of all the boundary of the circle that was behind the square (in black) (b) $H_2$:  square at the same depth than the circle-part; the common boundary $\mathcal{B}^c_2$ is shown in blue while the disoccluded boundary coincides with it, $\mathcal{B}_1^d=\mathcal{B}_1^c$, due to the fact that we consider closed objects (see details en the text).}
\label{fig:boundaries}
\end{figure}

Given the underlying hypothesis $H_i$ and the proximal image $I$, we define the conditional probability $p(I/H_i)$ as 
\begin{equation}\label{condPr}
p(I/H_i) \propto \tilde{p}(I/H_i) = e^{-\omega_1\int_{\mathcal{B}^c_i} (\kappa^2 + \beta) \mathrm{d}s } e^{-\omega_1\int_{\mathcal{B}^d_i} (\kappa^2 + \beta)  \mathrm{d}s}, 
\end{equation}
where $\mathcal{B}^c_i$ and $\mathcal{B}^d_i$ stand for common and disoccluded boundaries, respectively (see Figure~\ref{fig:boundaries} for an example with two hypothesis) and $\omega_1$ is a normalization constant. Formula \eqref{condPr} measures the responsibility 
that hypothesis $H_i$ takes for explaining the proximal stimulus $I$ as well as the deviation of $I$ from $H_i$.

\begin{figure}[h]
  \centering
  \subfigure[]{\label{subfig:facil}\includegraphics[width=2.8cm]{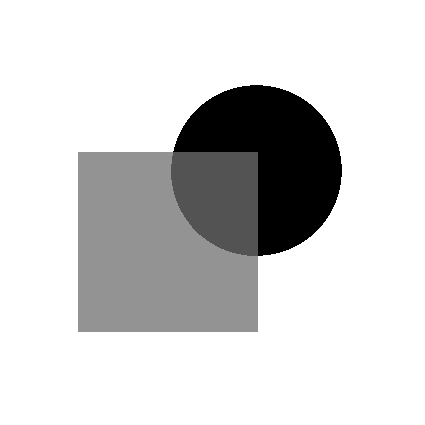}}
  \subfigure[]{\label{subfig:dificil}\includegraphics[width=2.8cm]{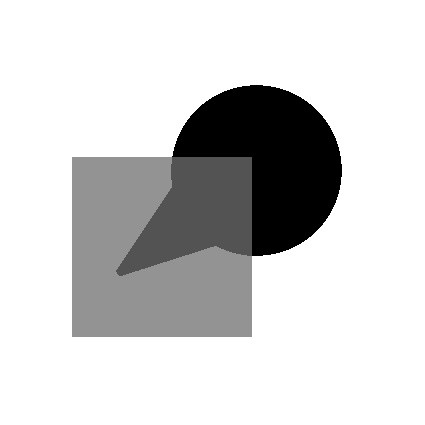}}
\caption{Example of two different disocclusions and how they affect the likelihood. In the example (a), where the reconstructed object is a disk, we can move the disk at many different positions behind the square to obtain the same kind of image (stimulus) we are observing. While in example (b) there is no so much freedom to move the position of the occluded object without changing the appearance of the observed image.}
\label{fig:probs}
\end{figure}

With the first integral in \eqref{condPr} we compute the difficulty of bringing the two objects together in order to get the perceived image taking into account only the known boundary of the objects; for example, it is easier to obtain configuration \ref{subfig:b_dc} than \ref{subfig:b_c} as in the first case only two points need to coincide, independently of the two coinciding points we will perceive the same image, and in the other case, $H_2$, a larger boundary needs to coincide in order to perceive exactly that configuration. The second integral takes into account the regularity of the occluded boundary of the shape to define the probability of obtaining a particular stimulus; for example in Figure~\ref{subfig:facil} we can move the disk at many different positions behind the square to obtain the same image we are observing, but in Figure~\ref{subfig:dificil} the movements we can do are more limited, as the perceived image will change drastically. Let us remark that due to the way we disocclude the objects the resulting disoccluded boundaries are always smooth; if we had different models of disocclusion this term would help to distinguish among them (in addition to the prior term). For instance, with our disocclusion model based on the elastica we are not able to recover the occluded object in Figure~\ref{fig:probs}(b) or the objects A in Figure~\ref{subfig:divergent}.
The probability distribution in \eqref{condPr} also appeared in Mumford~\cite{Mumford} and Williams and Jacobs~\cite{WilliamsJacobs}, who characterized the probability distribution of the shape of boundary completions based on the paths followed by a particle undergoing a stochastic motion, a directional random walk. It turns out that 
the elastica has the interpretation of being the mode of the probability distribution underlying this stochastic process restricted to curves with prescribed boundary behavior (the maximum
likelihood curve with which to reconstruct hidden contours).

The prior probabilities are defined as
\begin{equation}\label{Pr}
p(H_i) \propto \tilde{p}(H_i) = e^{-\omega_2\  \mathrm{compl}(O^1_i)} e^{-\omega_2\ \mathrm{compl}(O^2_i)},
\end{equation}
where $\omega_2$ is a normalizing constant, and $O^1_i$ and $O^2_i$ are the (disoccluded) objects in the 
hypothesized interpretation $H_i$. 
The factor $\mathrm{compl}(O^j_i)$
denotes the complexity of the object or shape $O^j_i$ at depth $j$. In the case that the object at one depth is formed by more than one connected component the complexity is computed separately for each connected component and their sum constitutes the complexity of  $O^j_i$. We use the definition of complexity of a shape defined by Chen and Sundaram in~\cite{complexity}, 
\begin{equation}
    C = \left(1 + R\right)\left(0.6\cdot\min\left(C_{\text{dist}}, C_{\text{angle}}\right) + 0.07\max\left(C_{\text{dist}}, C_{\text{angle}}\right) + 0.33P\right), 
    \label{compl.form}
\end{equation}
which takes into account global properties of the shape such the global distance entropy ($C_{\text{dist}}$), the local angle entropy  ($C_{\text{angle}}$), the perceptual smoothness ($P$), and a measure of shape randomness ($R$). The global distance is defined in \cite{complexity} as the distance of boundary points to the centroid of the shape. The local angle is the angle formed by the two segments joining three consecutive boundary points. The perceptual smoothness is computed using the local angle (as closer to $\pi$ the angle, the smoother the shape). Finally, the shape randomness is the maximum difference between two random traces obtained from the two more distant points of the boundary. Therefore, the  prior probability considers global properties such as shape contour symmetries and repetitions. 

Let us notice that with these definitions our whole model for amodal completion is able to choose, not only between the different hypothesis for a fixed disocclusion parameter $\beta$ but also between several disocclusions associated to different parameters $\beta$, and therefore to take into account global completion properties such as symmetry or repetitions. In Figure~\ref{fig:beta-parameter} there is an example illustrating this computational ability, where the different probabilities associated to the different disocclusion results depending on $\beta$ are given. 

Both normalization constants, $\omega_1$ and $\omega_2$, are defined, respectively, by the inverse of the maximum value, over all the hypothesis $H_i$, of the elastica and the object complexity.

\begin{figure}[h]
  \centering
    \subfigure[]{\includegraphics[width=3cm]{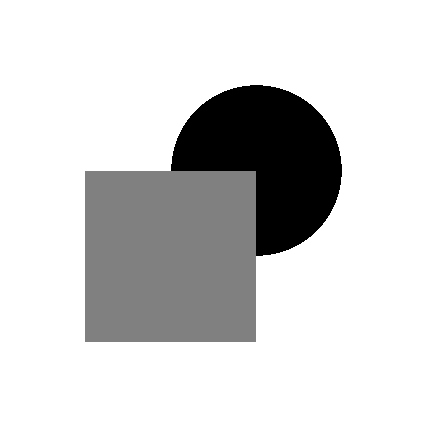}}
    \subfigure[]{\includegraphics[width=3cm]{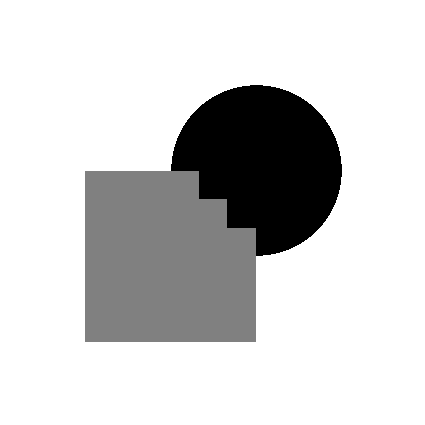}}
    \subfigure[]{\includegraphics[width=3cm]{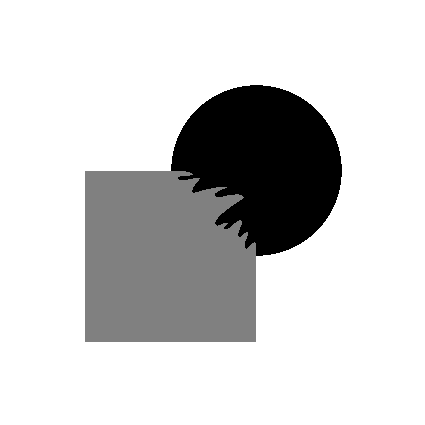}}
\caption{Three different proximal stimulus or images. From left to right, visual completion will become more and more evident than the interpretation of two pieces fitting together, both perceptually and quantitatively, with probability~\eqref{condPr}.}
\label{fig:3discandsquares}
\end{figure}

Let us comment on the term $e^{-\omega_1\int_{\mathcal{B}^c}(\kappa^2 + \beta)\mathrm{d}s}$ in our definition \eqref{condPr}. When visual completion occurs while propagating the stem, (e.g., hypothesis $H_1$ in Figure~\ref{subfig:2hyp}; also, hypothesis $H_1$ in Figure~\ref{subfig:b_dc}), the common boundaries $\mathcal{B}^c$ between the objects are reduced to the T-junctions. In this case: $\int_{\mathcal{B}^c}(\kappa^2 + \beta)\mathrm{d}s=0$ and thus $e^{-\omega_1\int_{\mathcal{B}^c}(\kappa^2 + \beta)\mathrm{d}s}=1$.
Let us notice that in the distal stimulus, since we are considering closed objects, $\mathcal{B}^c$ belongs to both objects. Therefore, in the hypothesis where the objects are interpreted as being fit-together (e.g., hypothesis $H_2$ in Figure~\ref{subfig:2hyp}; also, hypothesis $H_2$ in Figure~\ref{subfig:b_c}), a disoccluded boundary $\mathcal{B}^d$ appears which coincides with $\mathcal{B}^c$ (i.e.,  $\mathcal{B}^d = \mathcal{B}^c$). 
Let us also comment on the effect of the regularity of $\mathcal{B}^c$. Figure~\ref{fig:3discandsquares} presents three different proximal stimuli or images. The numerical computation of the term $e^{-\omega_1\int_{\mathcal{B}^c}(\kappa^2 + \beta)\mathrm{d}s} $ associated to each of the three images will decrease from left to right in the fit-together (or mosaic) interpretation, the same behavior applies to the complexity-related terms $ e^{-\omega_2\mathrm{compl}(O^i_1)} $ and $ e^{-\omega_2\mathrm{compl}(O^2_i)} $. 
Therefore, the visual completion will become more and more evident and the interpretation of two complex pieces fitting together will become perceptually less favorable.

Let us finally remark that we are not considering \emph{all} possibles configurations~\cite{von2009toward} but only the ones favored by relatability, convexity, and good continuation. On the other hand, even if global cues such as symmetry or repetitions are taken into account in our probability model, we do not incorporate them in the disocclusion algorithm. In the future, we plan to integrate it with other disocclusion strategies (such as, e.g., exemplar-based methods~\cite{inpainting_aujol_2010,inpainting_arias_2011} or~\cite{HayashiSasaki}) allowing to model these global properties and obtain, e.g., the objects A in Figure~\ref{subfig:divergent}.

\section{Algorithm and implementation details}\label{sec:algorithm}
Algorithm~\ref{algoSummary} shows the steps of the whole numerical algorithm.  Let us detail it:
Our algorithm needs a decomposition of the given image into objects and object parts
which are interpreted as projections of real 3D objects on the image plane.
This decomposition can be given either from the classical decomposition
in level sets, in bi-level sets or segmenting the image from a criterion.
In this paper, for the synthetic images, we use the decomposition in bi-level sets, 
which are defined as usual by 
$X^{(\lambda_n,\lambda_{n+1})}I=\{ x\in\Omega : \lambda_n\leq I(x)<\lambda_{n+1}\}$, 
where $\Omega$ is the image domain and $\{\lambda_n\}\subset\R$ is a finite strictly increasing sequence; and for the real images, we use the segmented shapes from the Berkeley segmentation dataset~\cite{MartinFTM01}.
In this way we obtain the objects that appear in the image; these objects will be denoted by $X_1$ and $X_2$. From $X_1$ and $X_2$ the three hypothesis will be considered by the algorithm: $X_1$ occluding the distal object $D_2$ (corresponding to the proximal $X_2$), $X_2$ occluding the distal object $D_1$ (corresponding to the proximal $X_1$), and $X_1$ and $X_2$ fit together. Now, by applying the disocclusion method of Section \ref{sec:binaryinpainting} where $X_1$ and $X_2$ are, respectively, the inpainting mask, we compute the \emph{complete} hypothesis $H_1=X_1\cup D_2$ and $H_2=X_2\cup D_1$, respectively. Then, to this two hypothesis, we always add the additional hypothesis $H_3=X_1\cup X_2$ of the \emph{mosaic} interpretation (which is obtained when we do not apply the disocclusion  algorithm). For each $H_i$ we compute the probabilities $\tilde{p}(I/H_i)$ and $\tilde{p}(H_i)$ from the definitions in Sect.~\ref{sec:approach}. Finally, we compute the perceptually preferred hypothesis $H_P$ by (\ref{max}).
\begin{algorithm}[H]
\SetAlgoLined
 \SetKwInOut{Input}{Input}\SetKwInOut{Output}{Output}
 \BlankLine
 \BlankLine
 \BlankLine
 \Input{An image I (the proximal stimulus) with objects $X_1$ and $X_2$ at two depths.}
 \Output{The set of distal hypothesis $H_1$, $H_2$, $H_3$, each one made of complete objects at two depths, and the preferred one $H_{\mathrm{P}}$ (with $\mathrm{P}\in\{1,2,3\}$).}
 \BlankLine
 \For{$i\in\{1,2,3\}$}{
 \eIf{$i \not= 3$}{
 $\bullet$ Consider $X_i$ as inpainting mask and initialize the inpainting mask using the perceptual method described in Sect.~\ref{initialization} \\
 $\bullet$ Disocclude object $X_j$, with $j\not= i$, using Algorithm \ref{thr.dyn.alg} implementing the elastica-based method of Sect.~\ref{sec:binaryinpainting}. From it, we obtain the disoccluded object $D_j$ and the completed hypothesis $H_i = X_i \cup D_j$.
 }
 {
 $\bullet$ Set $H_3 = X_1\cup X_2$
 }
 $\bullet$ Compute the probabilities $\tilde{p}(I/H_i)$ with Algorithm~\ref{elasticaBoundaries} and $\tilde{p}(H_i)$ (equation \eqref{compl.form}) from the definitions in Sect.~\ref{sec:approach}. \\
 }
Set $H_{\mathrm{P}}=\mathrm{arg}\max\limits_i \tilde{p}(I/H_i)\,\tilde{p}(H_i).$
 \BlankLine
 \BlankLine
\caption{Pseudo-code summarizing the proposal.}
\label{algoSummary}
\end{algorithm}

In Algorithm~\ref{thr.dyn.alg} we describe the threshold dynamics method we use for disocclusion, and in Algorithm~\ref{elasticaBoundaries} we present the algorithm for computing the conditional probability $\tilde{p}(I/H_i), i=1,2,3$.

\begin{algorithm}[H]
\SetAlgoLined
 \SetKwInOut{Input}{Input}\SetKwInOut{Output}{Output}
 \BlankLine
 \BlankLine
 \BlankLine
 \Input{A binary image $I$ containing a region without information (the inpainting region $\tilde{M}$), the inpainting region $\tilde{M} \subset \Omega$, and a elastica parameter $\beta>0$.} 
 \Output{Disoccluded object $D$ (given by an inpainted binary image $\bar{I}$)}
 \BlankLine
 $\bullet$ Set $\alpha = 0.99$ and $\delta t = 12$.\\
 $\bullet$ Set the initial shape $\Sigma_0 = \left\{x: I(x)=1\right\}$. \\
 $\bullet$ Set $n=0$ and $\Sigma_1=\Omega$. \\
 \While{$||\Sigma_{n+1} - \Sigma_n|| > 10^{-3}$}{
 \begin{enumerate}
     \item A step of Grzibovskis-Heintz algorithm. Set:
     \[
     \Gamma_1=\left\{x: 2\alpha G_{\sqrt{\delta t}}\ast\mathbb{1}_{\Sigma_n}(x) - 2G_{\alpha^2\sqrt{\delta t}}\ast\mathbb{1}_{\Sigma_n}(x) \leq \alpha - 1\right\} .
     \]
     \item A step of standard  Merriman-Bence-Osher algorithm. Set:
     \[
     \Gamma_2 = \left\{x:G_{\beta\delta t}\ast\mathbb{1}_{\Gamma_1}(x)\geq\frac{1}{2}\right\} .
     \]
     \item Fidelity step. Set
     \[
     \Sigma_{n+1} = \left(\Gamma_2\cap\tilde{M}\right)\cup\left(\Omega\backslash\tilde{M}\right) .
     \]
          $n=n+1$.
 \end{enumerate}}
  $\bullet$ $\bar{I} = \mathbb{1}_{\Sigma_{n}} $  and $D=\Sigma_{n}$.
 \BlankLine
\caption{Pseudo-code of the disocclusion algorithm. }
\label{thr.dyn.alg}
\end{algorithm}

Let us add some details regarding Algorithm~\ref{thr.dyn.alg}. The Gaussian convolution has been computed using the Lindeberg's   discrete scale-space method and its implementation described in~\cite{otero2016computing}, that is, we use that the Gaussian convolution $v:t\mapsto G_{\sqrt{\delta t}}\ast u$ is the solution of the heat equation $\frac{\partial v}{\partial t} = \Delta v$ for a diffusion time $\delta t$ (set to $12$ in our experiments, to guarantee the prescribed upper and lower bounds depending on the curvature of the visible shape~\cite{MBO}) so we only need to discretize partial derivatives. We refer to \cite{otero2016computing} for more details on the discretization. Parameter $\alpha$ needs to be close but less than $1$~\cite{ERT-05,esedoglu2008threshold}.

\begin{algorithm}[H]
\SetAlgoLined
 \SetKwInOut{Input}{Input}\SetKwInOut{Output}{Output}
 \BlankLine
 \BlankLine
 \BlankLine
 \Input{Inpainting masks $X_1, X_2$, disoccluded objects $D_1,D_2$, elastica parameter $\beta$.} 
 \Output{Conditional probability of each hypothesis $H_1,H_2,H_3$.}
 \BlankLine
 $\bullet$ Compute the boundaries $\partial{X_1}$, $\partial{X_2}$, $\partial{D_1}$, $\partial{D}_2$ of $X_1$, $X_2$, $D_1$, $D_2$, respectively.\\
 $\bullet$ Set $\mathcal{B}^c_i = \partial{X}_1\cap\partial{X}_2,\quad i=1,2,3$.\\
 \For{$i=1,2$}{
 $\mathcal{B}^d_i = \partial{D}_i\backslash \partial{X}_i$}
 $\bullet$ Set $\mathcal{B}^d_3 = \mathcal{B}^c_3$ \\
 \For{$\{i=1,2,3\}$}{
 $E_{B_i} = \sum\limits_{x \in \mathcal{B}^c_i}(\kappa^2(x) + \beta) + \sum\limits_{x \in \mathcal{B}^d_i}(\kappa^2(x) + \beta)$\\
 }
 $\bullet$ Set $\omega_1 = \max\{E_{B_1},E_{B_2},E_{B_3}\}$\\
 \For{$i=1,2,3$} {
 $\tilde{p}(I/H_i) = \exp\{-\omega_1 E_{B_i}\}$ 
 }
 \BlankLine
\caption{Pseudo-code of the algorithm computing $\tilde{p}(I/H_i)$ for $i=1,2,3$.}
\label{elasticaBoundaries}
\end{algorithm}

Regarding Algorithm~\ref{elasticaBoundaries}, the discrete boundaries of each shape are computed as external boundaries and using 4-connectivity. On the other hand, in order to compute the curvature of a discrete curve (or boundary) $\gamma$, we use the method of~\cite{sethian1985curvature} and compute
\begin{equation}
    \kappa(x) = \operatorname{div}\left(\frac{\nabla u(x)}{|\nabla u(x)|}\right),
\end{equation}
where $u$ is the signed distance function to the boundary $\gamma$. We use forward derivatives to compute the gradient, backward derivatives for the divergence. The discrete signed distance function $u$ is computed using the algorithm explained in~\cite{meijster2002general}.

Finally, the prior probability is computed using~\eqref{Pr} with the complexity measure given by~\eqref{compl.form}. We consider as boundary points for computing \eqref{compl.form} all the pixels that form the boundary of an object. In case of an object formed by more than one connected component we compute the complexity \eqref{compl.form} of every connected component and the final complexity measure is the addition of the individual complexities. For details about how to compute $R, C_{\text{dist}}, C_{\text{angle}}$ and $P$ we refer to Section~\ref{sec:approach} and to~\cite{complexity}.

\section{Experimental results}\label{sec:experimentalresults}

The proposed method has been tested with both synthetic and real images. Parameter $\beta$ , which sets the underlying a priori regularity (see comments on its role in Section~\ref{sec:binaryinpainting}) has been fixed to $0.6$ for all the experiments in order to have an algorithm as general as possible. There are two exceptions, namely, Proximal 2 of Table~\ref{tab:alltogether1} and  Example 4 of Table~\ref{tab:alltogether2_fail}, where $\beta$ is fixed to 1.2 and 1.7, respectively, due to the biggest size of the circular shapes. As explained in Section~\ref{sec:binaryinpainting}, there is a relationship between the numerical curvature of the disoccluded objects and the a priori regularity imposed through the parameter $\beta$: the larger the $\beta$, the larger the expected radius of the osculating circle locally approximating the curve. 

The experiments of this Section are organized as follows: In Sections~\ref{sec:synthetic},~\ref{sec:real} and~\ref{subsec:fig-ground} we introduce the experiments which agree with our perception, while in Section~\ref{failures} we show and discuss the experiments that failed. The synthetic experiments, which are described in the following Section~\ref{sec:synthetic},  are shown in Tables \ref{tab:alltogether1}, \ref{tab:alltogether2} and \ref{tab:alltogether3} while the experiments on real images (described in Section~\ref{sec:real}) are shown in Tables \ref{2segs}, \ref{tab:alltogether7}, \ref{tab:alltogether5} and \ref{tab:alltogether6}. Table \ref{tab:alltogether7} shows our results on images  of the Berkeley dataset with provided figure-ground ground-truth labeling~\cite{figuregroundBerkeley}. Table~\ref{tab:alltogether8} in Section~\ref{subsec:fig-ground} shows the ability of our method to also decide on (perceptually) fully visible objects over a background. Finally, Tables~\ref{tab:alltogether1_fail},~\ref{tab:alltogether2_fail} and~\ref{tab:alltogether5_fail} present the synthetic and real results where our method did not agree with human perception. For each row in each table we show a complete experiment. 

Let us recall that our method assumes the proximal stimulus to be decomposed into objects and object parts (which can be interpreted as projections of real 3D objects on the image plane). As in the synthetic experiments, the images are formed by objects  with a single and unique color, this already gives a segmentation and we apply our algorithm directly. For the real experiments, we use a segmentation of the image. In particular,  we have taken segmented images from the Berkeley segmentation dataset~\cite{MartinFTM01} and from \cite{SemanticBenchmark}. 

\subsection{Synthetic images}\label{sec:synthetic}

Tables \ref{tab:alltogether1},  \ref{tab:alltogether2} and \ref{tab:alltogether3} show some experiments on synthetic images. For each row in each table, a complete experiment is shown. We first present the proximal image (piecewise constant) $I$ on the left, followed by the three hypothesis $H_i$ (each one separated by a gray box), together with the values $\tilde{p}(I/H_i)$ and $\tilde{p}(H_i)$ proportional to the conditional probability and the prior probability, respectively, and the probability value $p(H_i|I)$. 
Let us remark that we have normalized the probabilities in such a way that $p(H_1/I)+p(H_2/I)+p(H_3/I)=1$. 
The probability value of the preferred hypothesis $H_P$ is highlighted in boldface. For the first two hypothesis, $H_1$ and $H_2$, we display the objects at depth 1 on the left, and the disoccluded objects (at depth 2) on the right. Let us recall that the objects at depth 1 are considered the inpainting mask for disoccluding the objects at depth 2. Finally the last column is the hypothesis $H_3$ where the two objects are fitting together at the same depth. 

Let us comment on the results in Tables \ref{tab:alltogether2} and \ref{tab:alltogether3}. In Table \ref{tab:alltogether2}, the third hypothesis is not shown because it coincides with $H_2$ due to the fact that the disocclusion algorithm does not change the objects being disoccluded. Besides, in Table \ref{tab:alltogether3} there is shown a synthetic experiment where the three hypothesis coincide on account of the obtained disoccluded objects: The disocclusion algorithm applied in the first two hypothesis does not change the objects and thus $H_1 = H_2 = H_3$, and the posterior probability is the same for all three hypothesis. Let us remark that, even if the objects forming the scene coincide in different hypothesis, the depth ordering is not the same in each hypothesis. Let us singularize and explain one example in Figure~\ref{subfig:2hyp}, where our method produces $H_3=H_2$. However, as for depth order, $H_3$ is interpreted as two objects at the same depth (and having the real relative size which is observed in the proximal image) while $H_2$ can be interpreted as three quarters of a disk which is closer to the observer, plus a square which can be of bigger size but farther away from the three-quarters-of-a-disk shape and whose boundary partially coincides with part of the boundary of the three-quarters-of-a-disk shape. Notice that this situation is related to the well-known ambiguity in depth of some proximal stimulus, sometimes causing optical illusion of relative depth perception as those in the images displayed in Figure~\ref{fig:illusions}. 

These experiments show that our method agrees with human perception. The perception literature acknowledges that, in a T-junction, the occluder is the surface on the T-head side while the surfaces on the T-stem side continue behind the occluder~\cite{rubin2001role}. This phenomenon is validated by our method: The  preferred hypothesis, highlighted in boldface, is the one that is obtained by continuation of the T-stems. Let us  comment on the results corresponding to Proximal 6 and 7 of Table~\ref{tab:alltogether1}, which include quite similar shapes with equal occlusion signatures but different common boundaries among the shapes. In Proximal 7 and 8 (and also in Proximal 9 of Table~\ref{tab:alltogether2}), the local perception cue at the T-junctions indicates that there is an occluded disk which continues behind an incomplete square (the occluder). Our method is able to choose the corresponding preferred hypothesis as is shown by the probability values $p(H_1/I)$. In Proximal 12 on Table~\ref{tab:alltogether2}, according to the T-junctions we should prefer $H_1$ (as the method chooses), but for symmetries most of us perceptually prefer $H_2$. In this case, according to the prior $H_2$ should be preferred (as symmetries are valued positively), but as the disoccluded and common boundary are so large in $H_2$, the value $\tilde{p}(I/H_1)$ is much bigger, and $H_1$ is selected as preferred. Finally, let us comment on Proximal 4 which is a well-known and perceptually controversial example. The preferred hypothesis for Proximal 4 is the one that agrees with the T-junction cues and the one reported in \cite{VanDerHelm1} to be the most preferred by the subjects participating in their psychophysics experiments. However, in this case the posterior probabilities for $H_2$ and $H_1$ are quite close and, according to our personal experience (e.g., by incorporating our knowledge about the world and objects of similar shapes),  some people prefer the interpretation according to $H_2$. 
Both experiments, Proximal 4 and 12, are examples where hypothesis $H_1$ and  $H_2$ correspond, respectively, to a local and global completion of the occluded object. In both cases, our algorithm favors local completion, that is, a completion that produces good continuation instead of the global one which produces a more symmetric object (notice that the local completion in both cases produces a symmetric object with respect to one axis). As commented in Section \ref{sec:relatedwork}, both kind of completions interplay and the prevalence of one of them depends on the observation time \cite{moravec86} and on the saliency of the good-continuation versus symmetry in the completion \cite{Sekuler}.

\setlength{\tabcolsep}{3pt}
\begin{longtable}{llllll}
 {\includegraphics[width = 1.75cm]{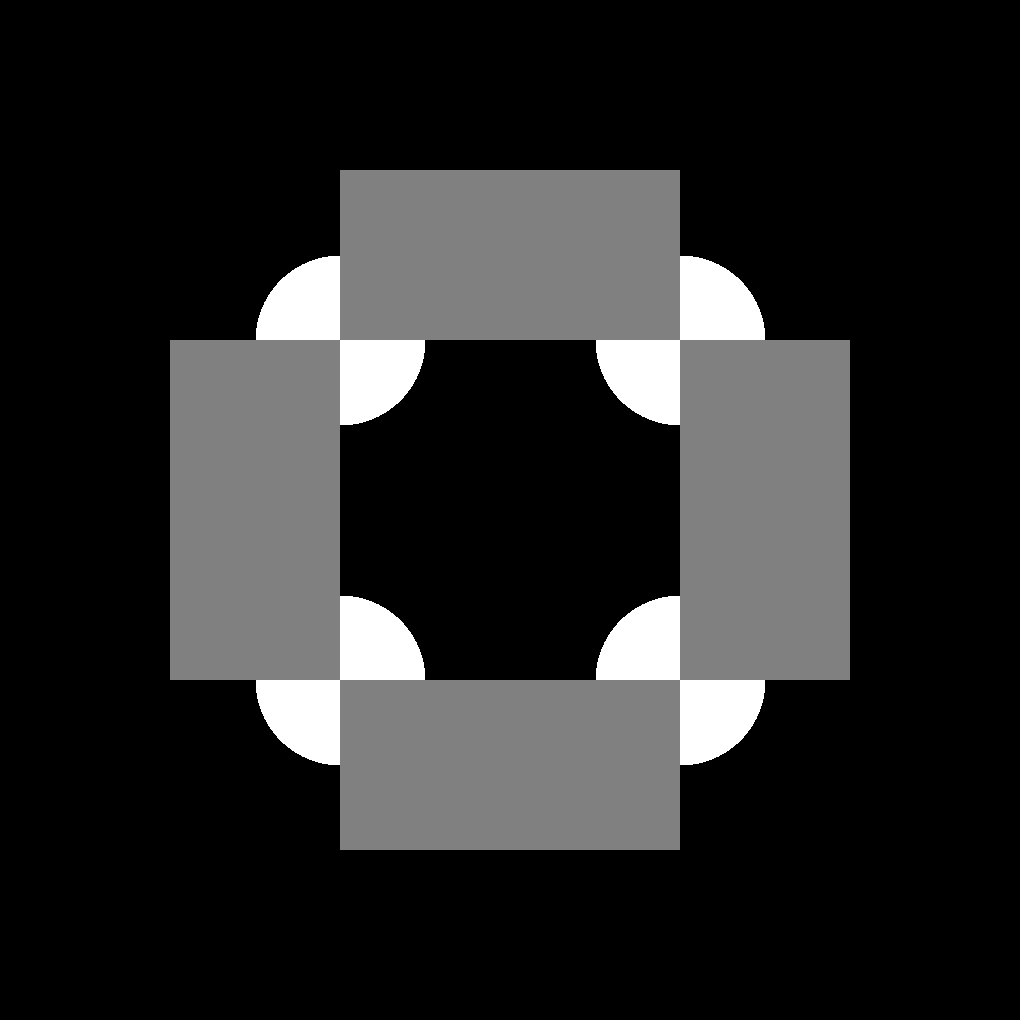}} & 
 \cellcolor{gray!25} \includegraphics[width = 1.75cm]{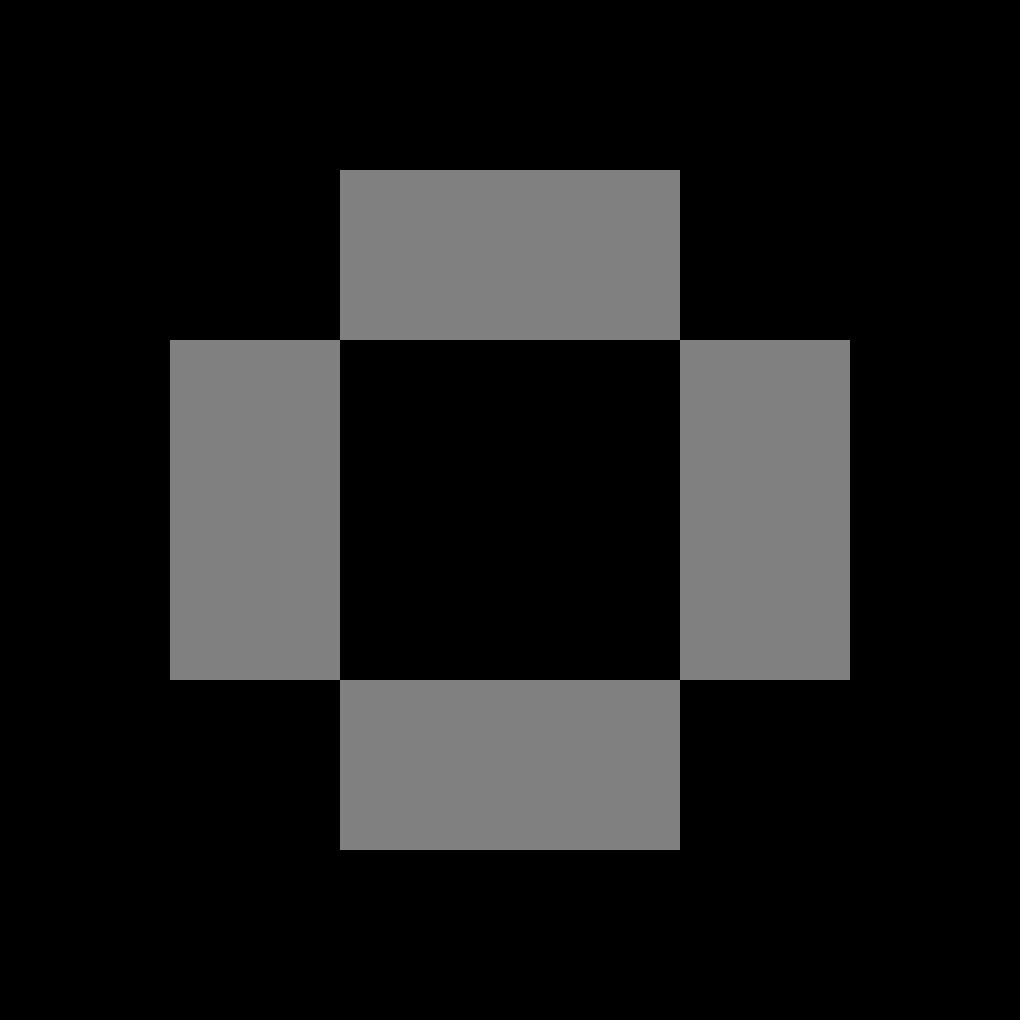} 
 \includegraphics[width = 1.75cm]{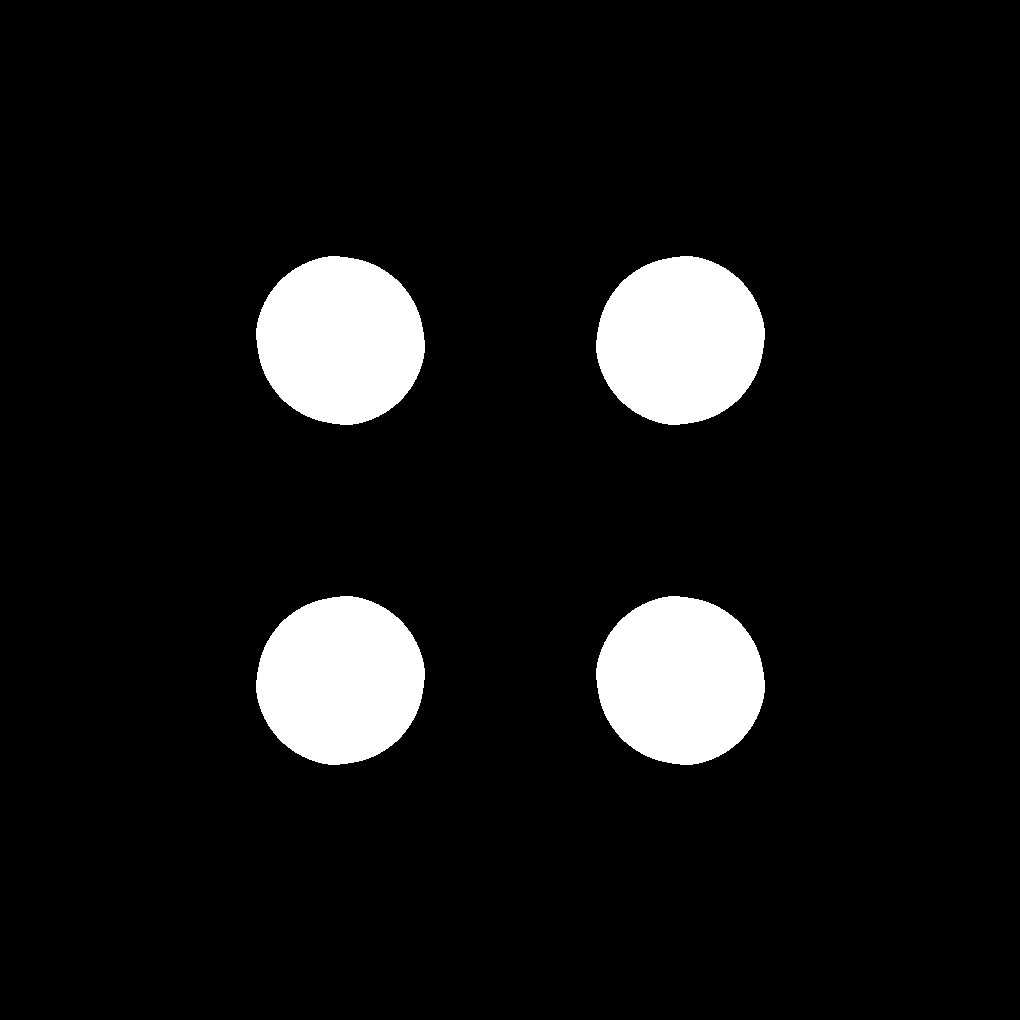} & &
 \cellcolor{gray!25} \includegraphics[width = 1.75cm]{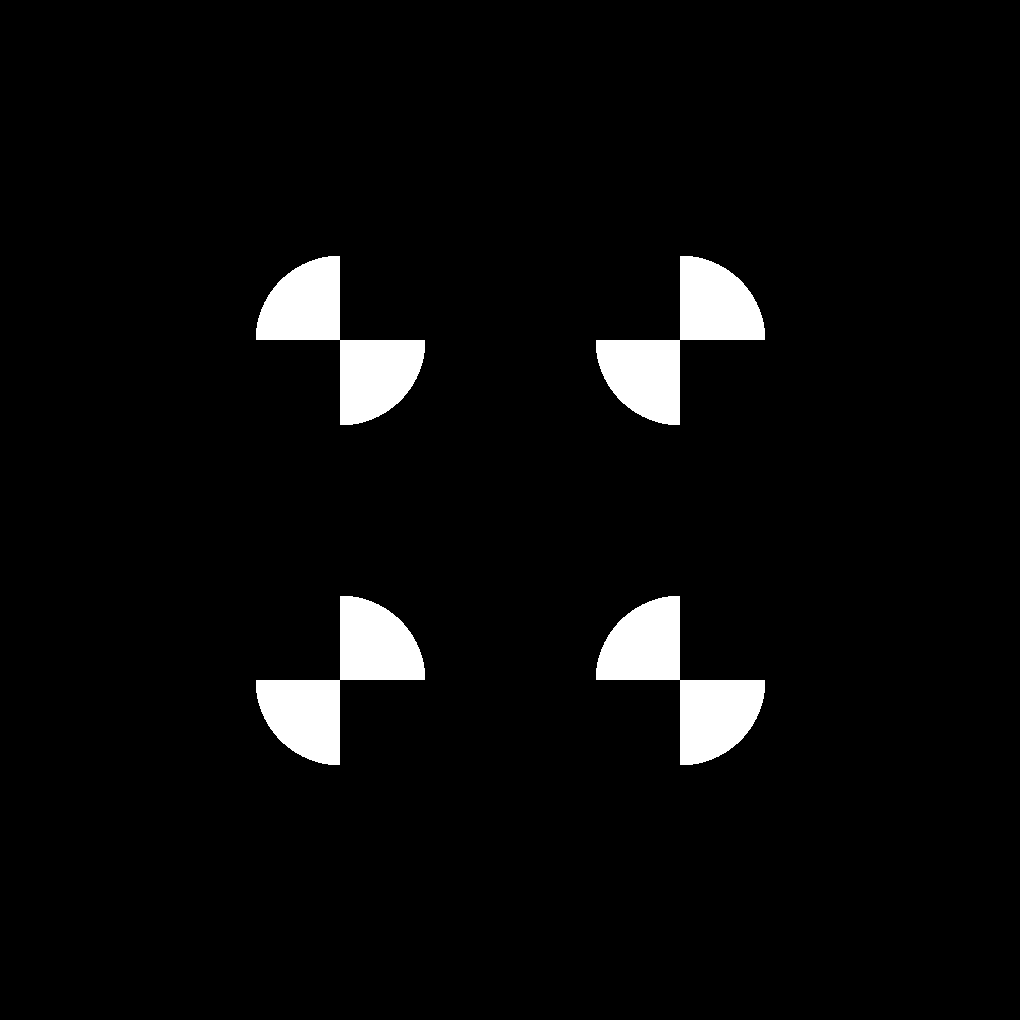} 
 \includegraphics[width = 1.75cm]{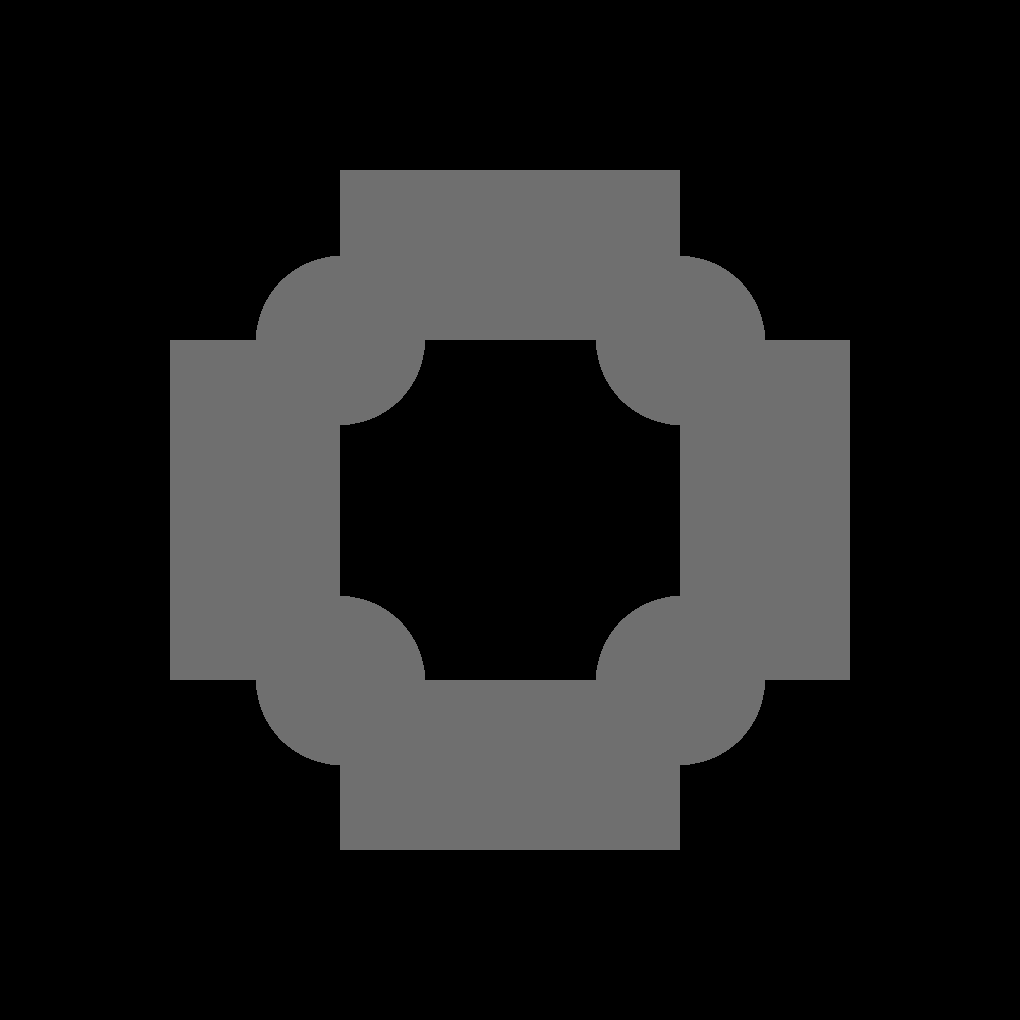} & &
 \cellcolor{gray!25} \includegraphics[width = 1.75cm]{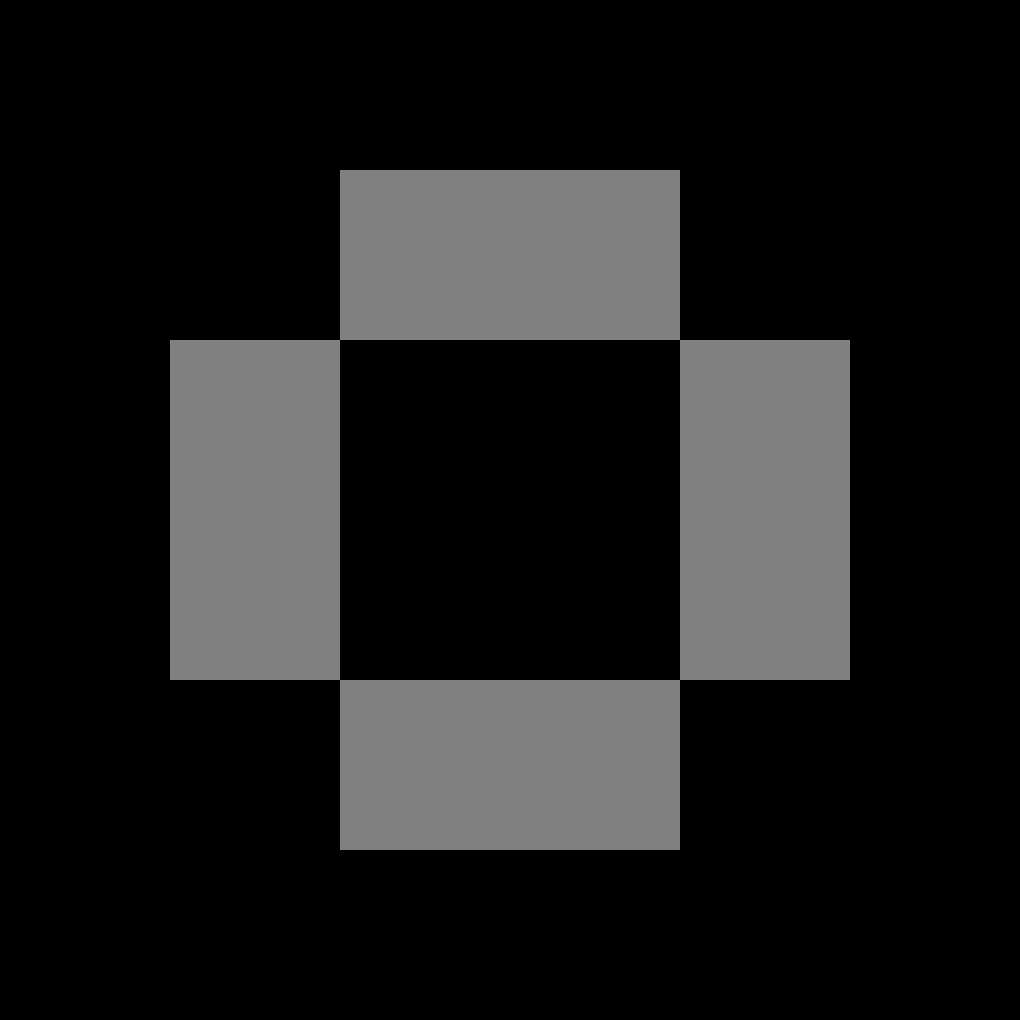} 
 \includegraphics[width = 1.75cm]{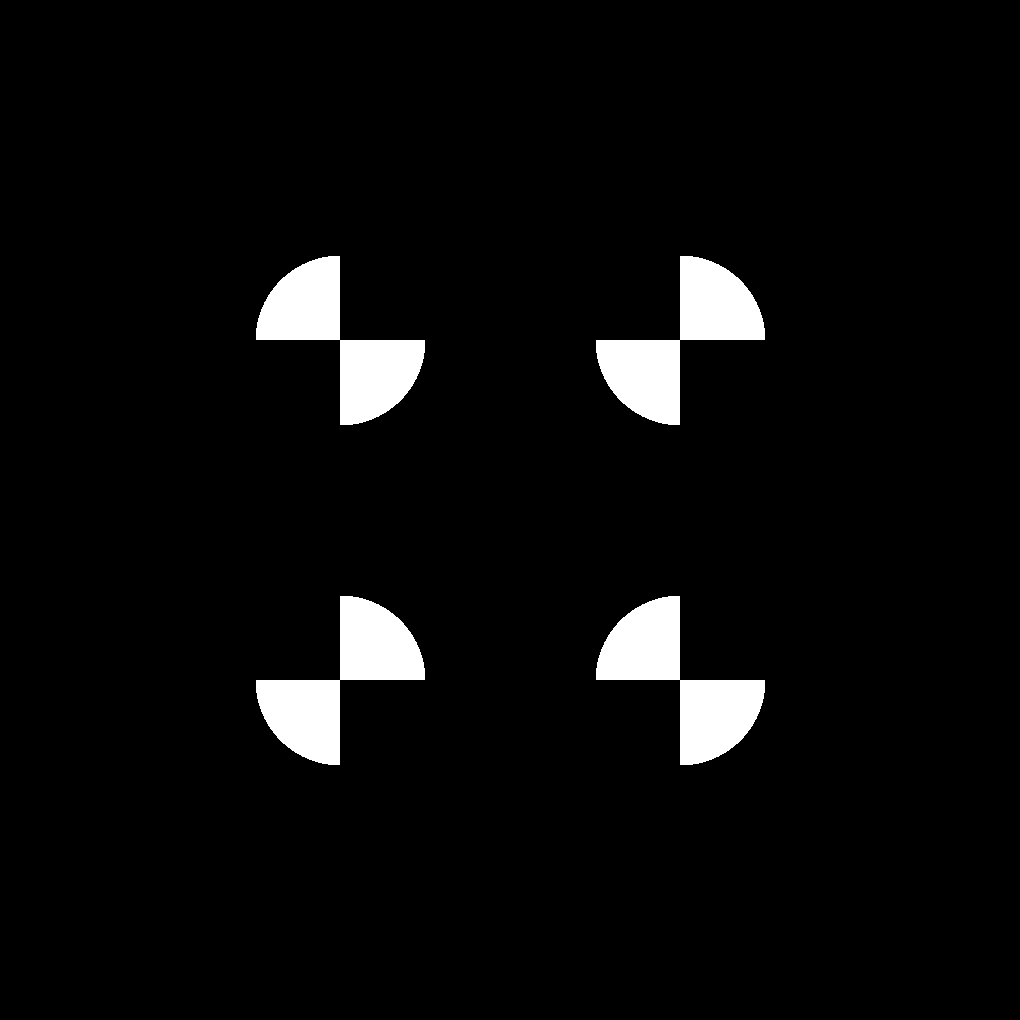} \\
 & \cellcolor{gray!25}$\tilde{p}(I/H_1)=0.7505$ & & \cellcolor{gray!25}$\tilde{p}(I/H_2)=0.5059$ & & \cellcolor{gray!25}$\tilde{p}(I/H_3)=0.4230$\\
 {\scriptsize{Proximal 1}} & \cellcolor{gray!25}$\tilde{p}(H_1)=0.4660$ & &
 \cellcolor{gray!25}$\tilde{p}(H_2)=0.3877$ & & \cellcolor{gray!25}$\tilde{p}(H_3)=0.3679$\\
 & \cellcolor{gray!25}$\mathbf{p(H_1/I)=0.5398}$ & & \cellcolor{gray!25}$p(H_2/I)=0.2201$ & & \cellcolor{gray!25}$p(H_3/I)=0.2402$\\ \\
 {\includegraphics[width = 1.75cm]{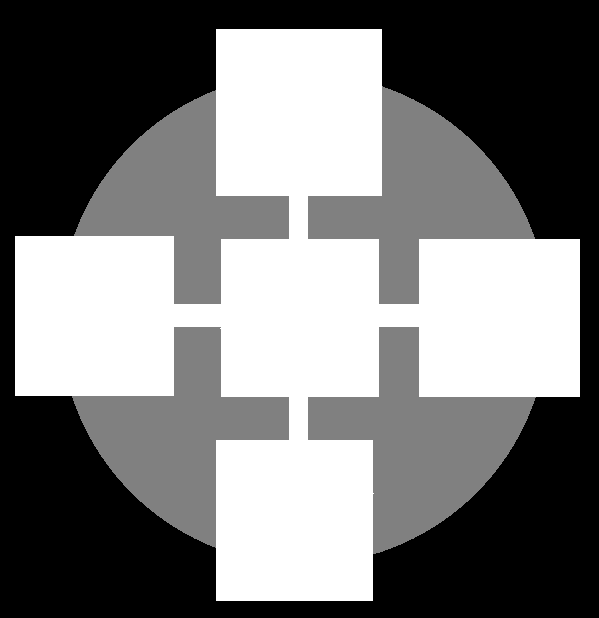}} & 
 \cellcolor{gray!25} \includegraphics[width = 1.75cm]{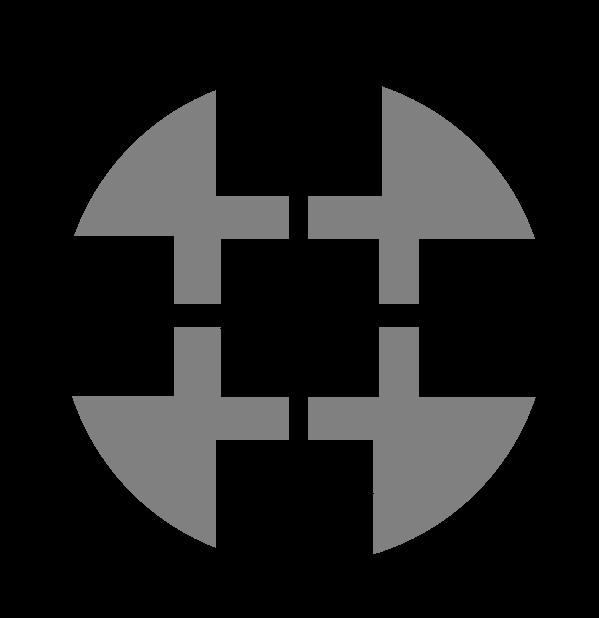} 
 \includegraphics[width = 1.75cm]{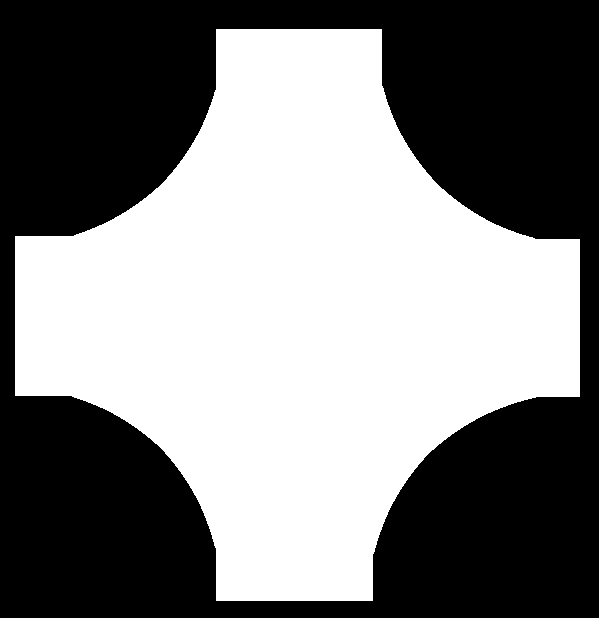} & &
 \cellcolor{gray!25} \includegraphics[width = 1.75cm]{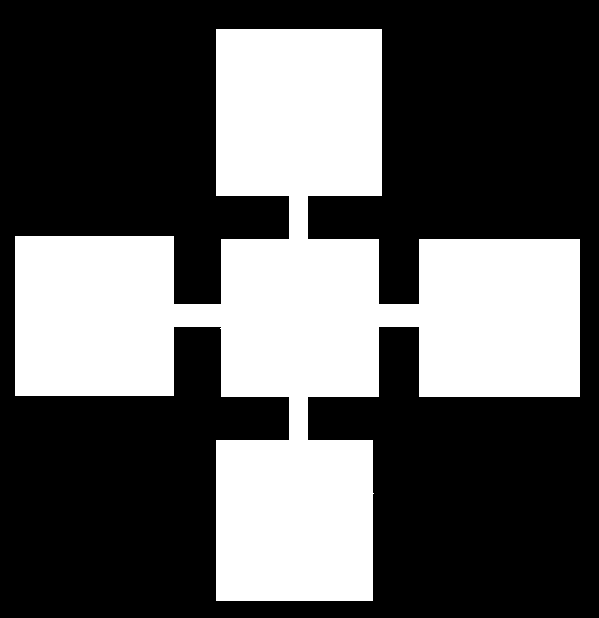} 
 \includegraphics[width = 1.75cm]{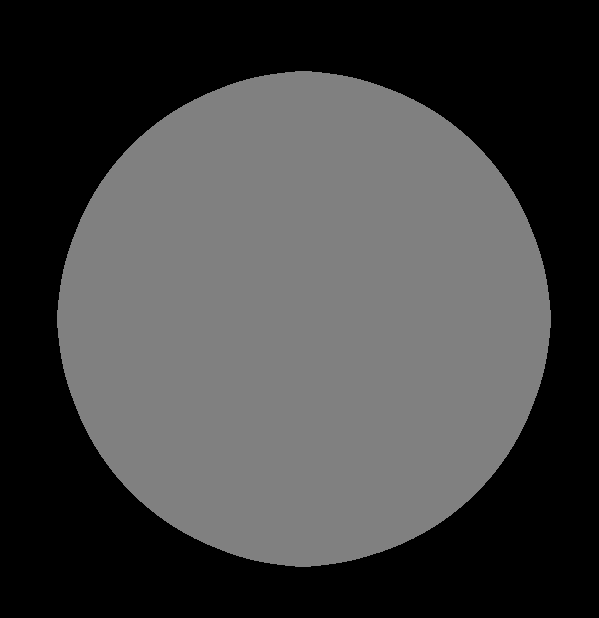} & &
 \cellcolor{gray!25} \includegraphics[width = 1.75cm]{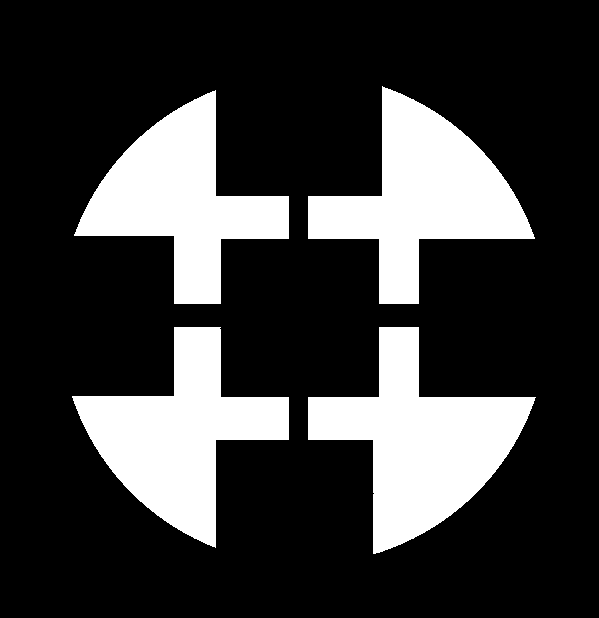} 
 \includegraphics[width = 1.75cm]{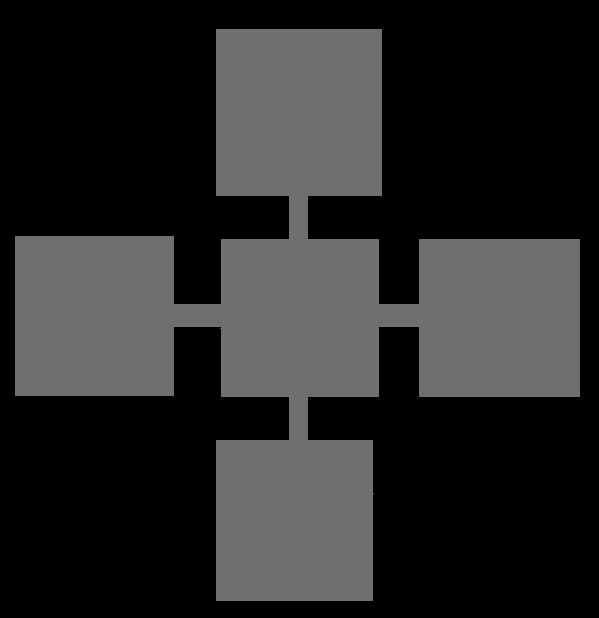} \\
 & \cellcolor{gray!25}$\tilde{p}(I/H_1)=0.7557$ & & \cellcolor{gray!25}$\tilde{p}(I/H_2)=0.8388$ & & \cellcolor{gray!25}$\tilde{p}(I/H_3)=0.3679$\\
 {\scriptsize{Proximal 2}} & \cellcolor{gray!25}$\tilde{p}(H_1)=0.3679$ & & \cellcolor{gray!25}$\tilde{p}(H_2)=0.8793$ & & \cellcolor{gray!25}$\tilde{p}(H_3)=0.4327$\\
 & \cellcolor{gray!25}$p(H_1/I)=0.2366$ & & \cellcolor{gray!25}$\mathbf{p(H_2/I)=0.6279}$ & & \cellcolor{gray!25}$p(H_3/I)=0.1355$\\ \\
 {\includegraphics[width = 1.75cm]{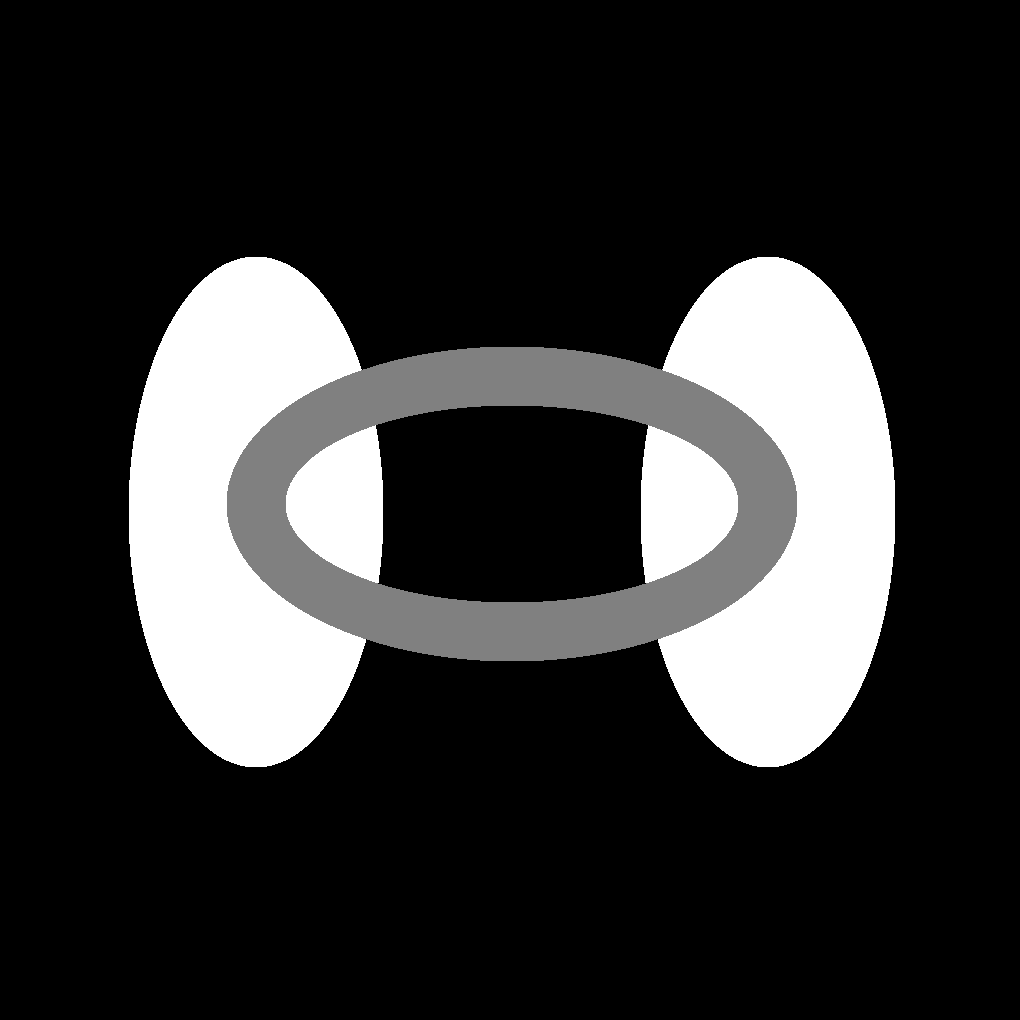}} & 
 \cellcolor{gray!25} \includegraphics[width = 1.75cm]{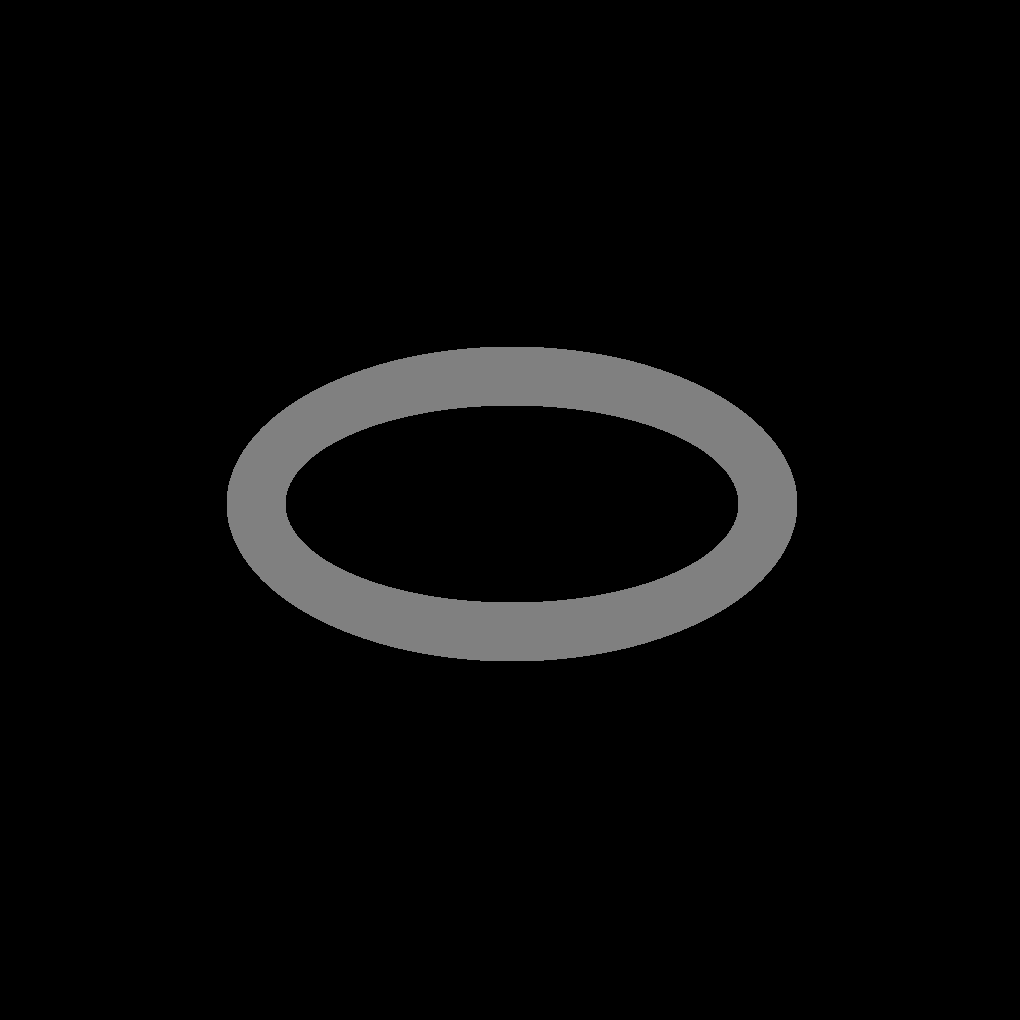} 
 \includegraphics[width = 1.75cm]{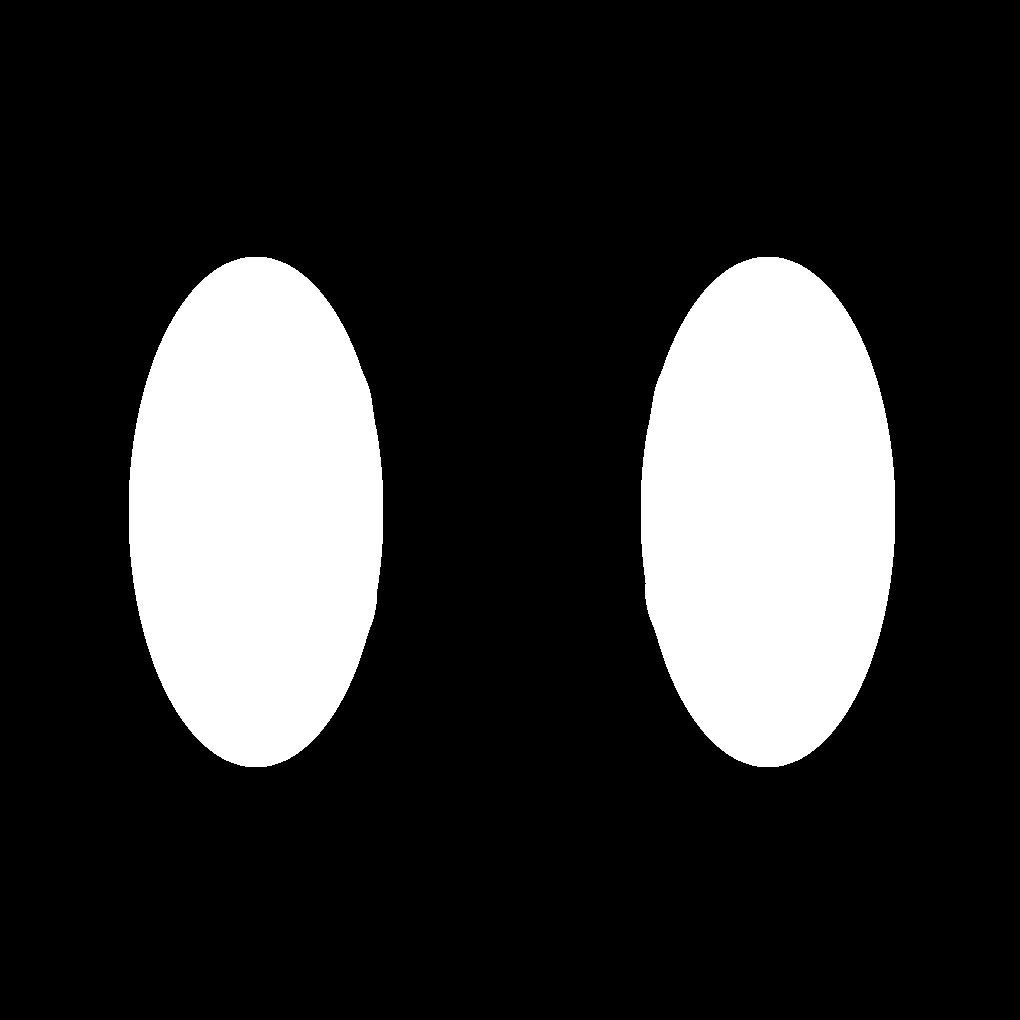} & &
 \cellcolor{gray!25} \includegraphics[width = 1.75cm]{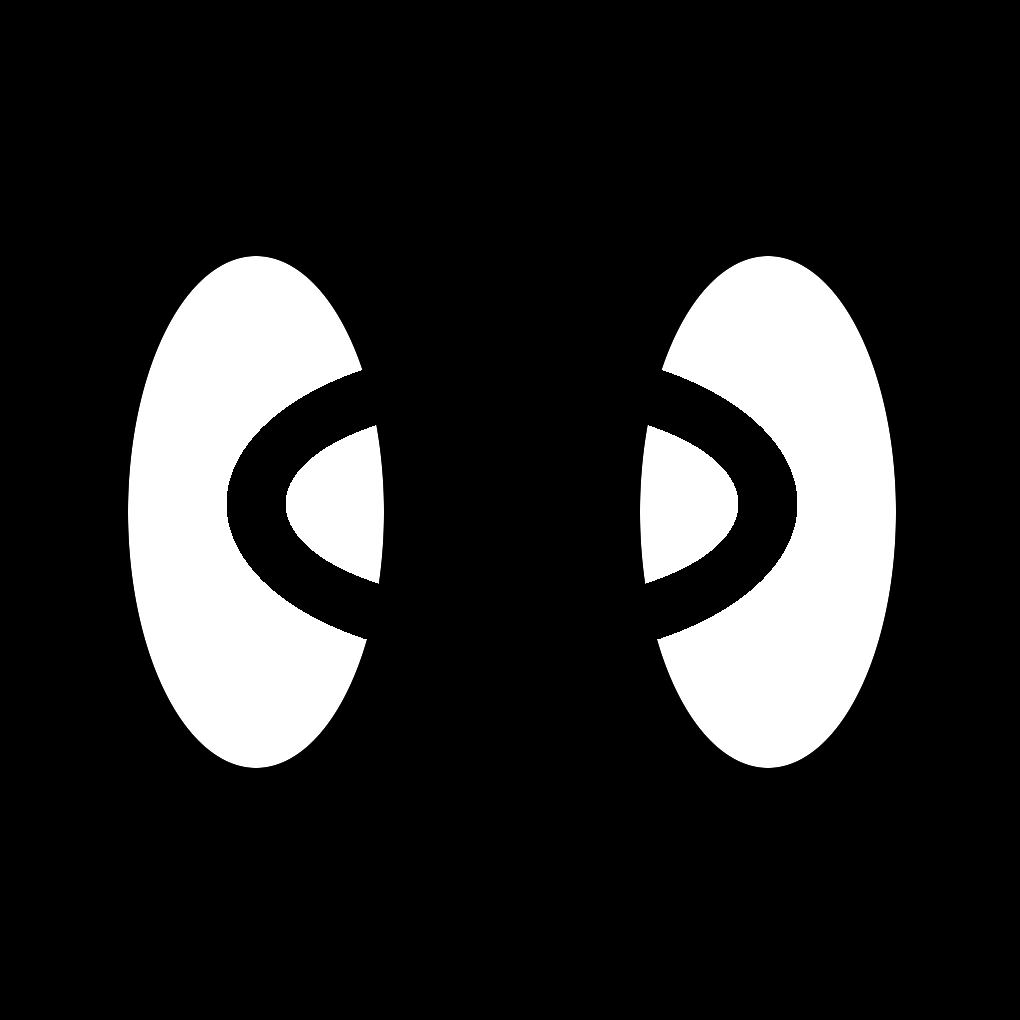} 
 \includegraphics[width = 1.75cm]{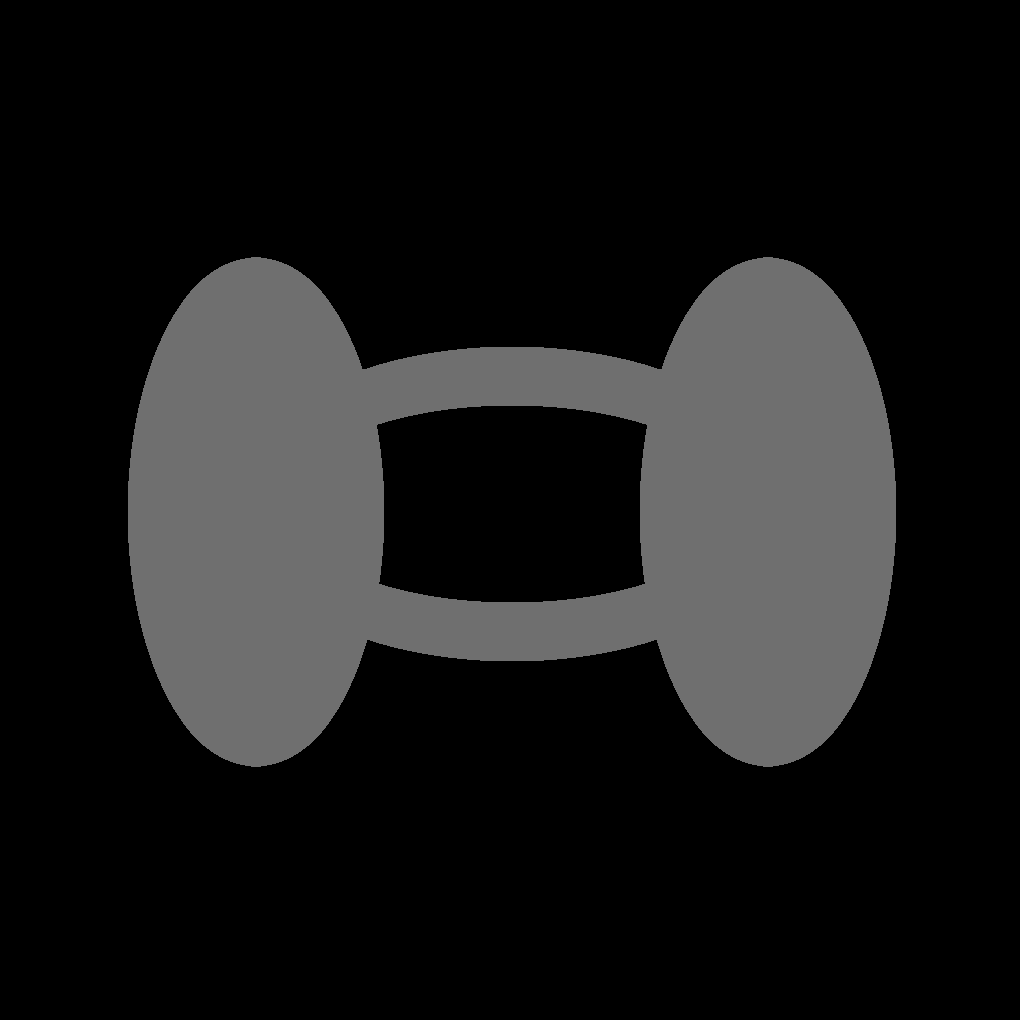} & &
 \cellcolor{gray!25} \includegraphics[width = 1.75cm]{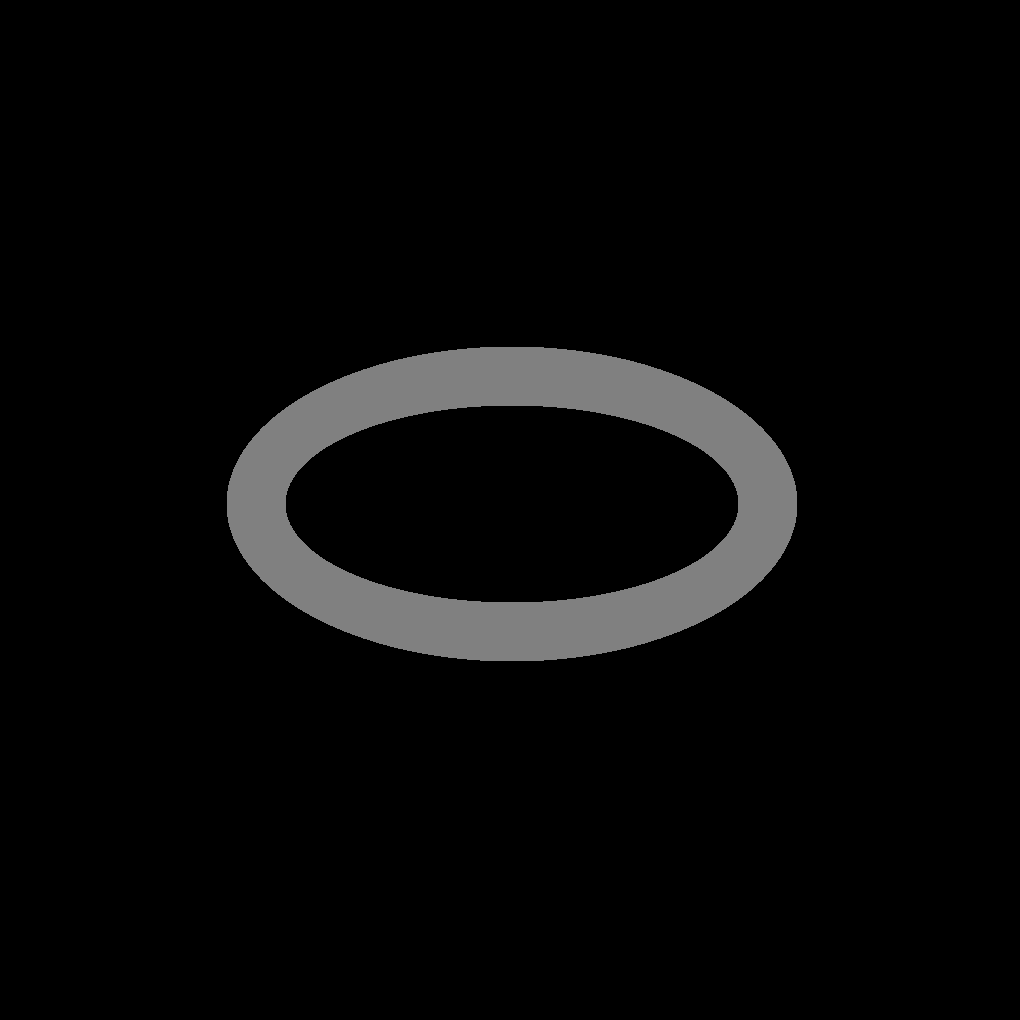} 
 \includegraphics[width = 1.75cm]{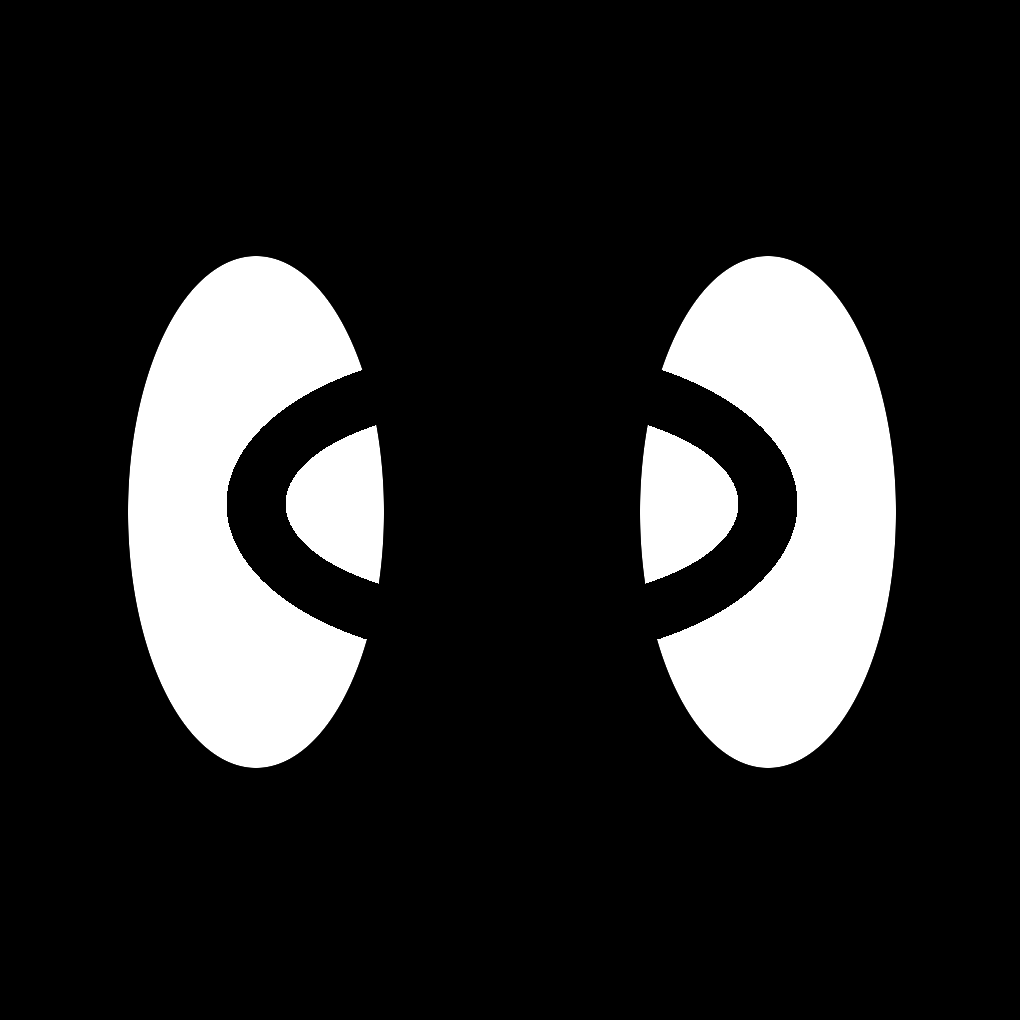} \\
 & \cellcolor{gray!25}$\tilde{p}(I/H_1)=0.9343$ & & \cellcolor{gray!25}$\tilde{p}(I/H_2)=0.3679$ & & \cellcolor{gray!25}$\tilde{p}(I/H_3)=0.4214$\\
  {\scriptsize{Proximal 3}}& \cellcolor{gray!25}$\tilde{p}(H_1)=0.5544$ & & \cellcolor{gray!25}$\tilde{p}(H_2)=0.3738$ & & \cellcolor{gray!25}$\tilde{p}(H_3)=0.3679$\\
 & \cellcolor{gray!25}$\mathbf{p(H_1/I)=0.6391}$ & & \cellcolor{gray!25}$p(H_2/I)=0.1696$ & & \cellcolor{gray!25}$p(H_3/I)=0.1813$\\ \\
 {\includegraphics[width = 1.75cm]{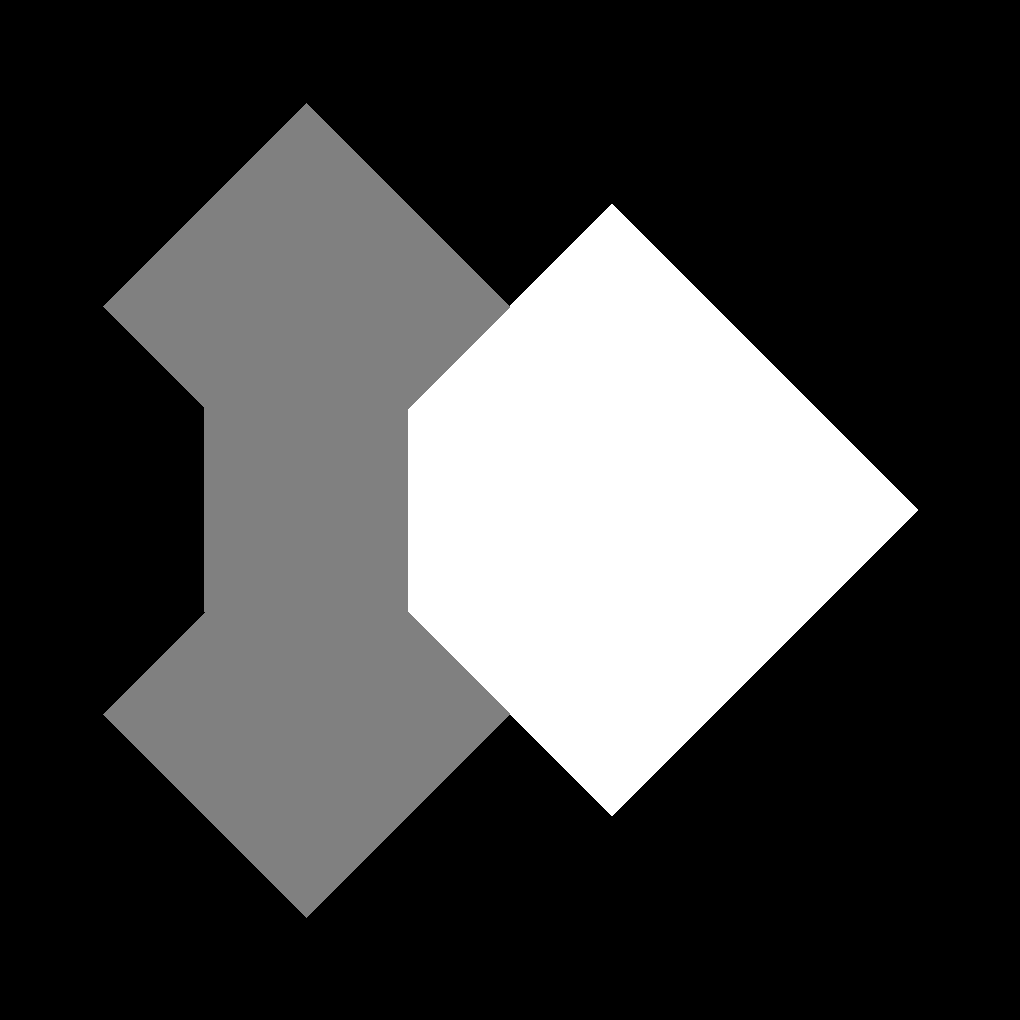}} & 
 \cellcolor{gray!25} \includegraphics[width = 1.75cm]{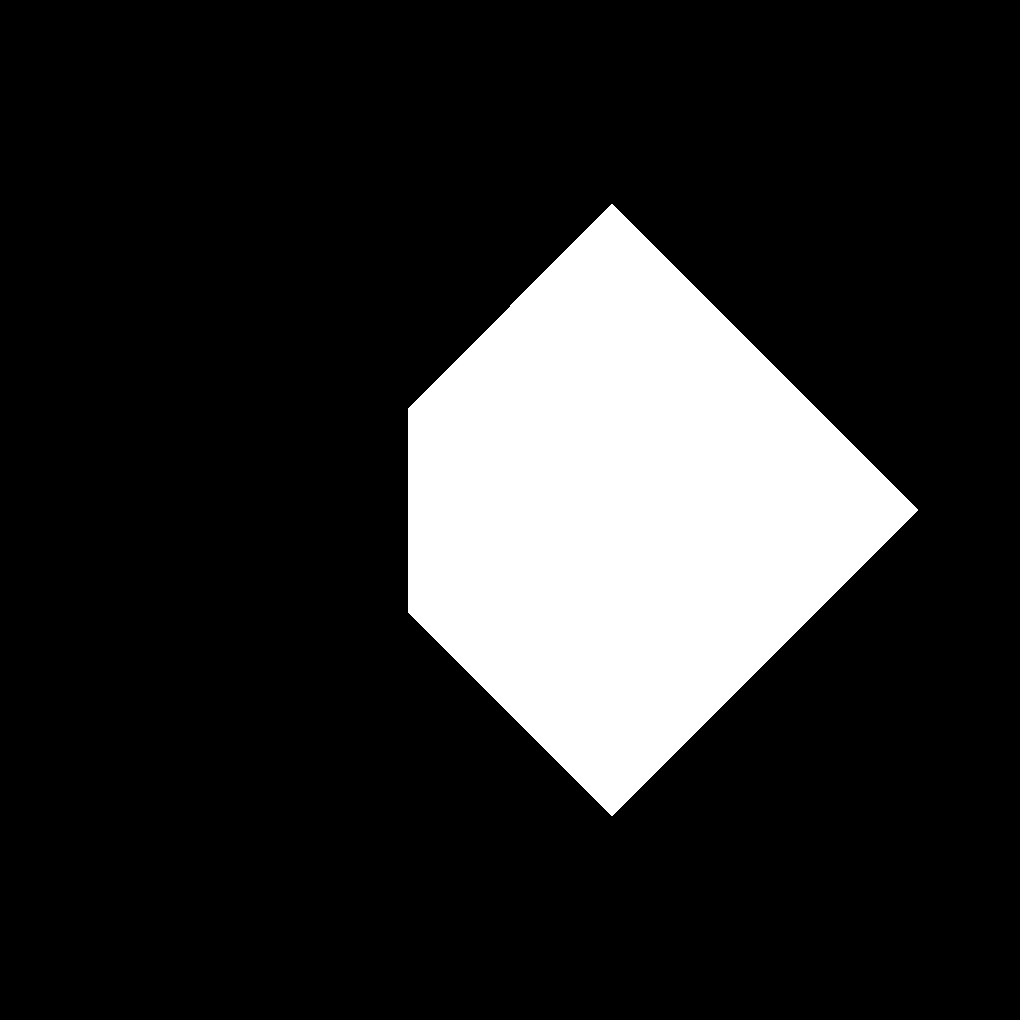} 
 \includegraphics[width = 1.75cm]{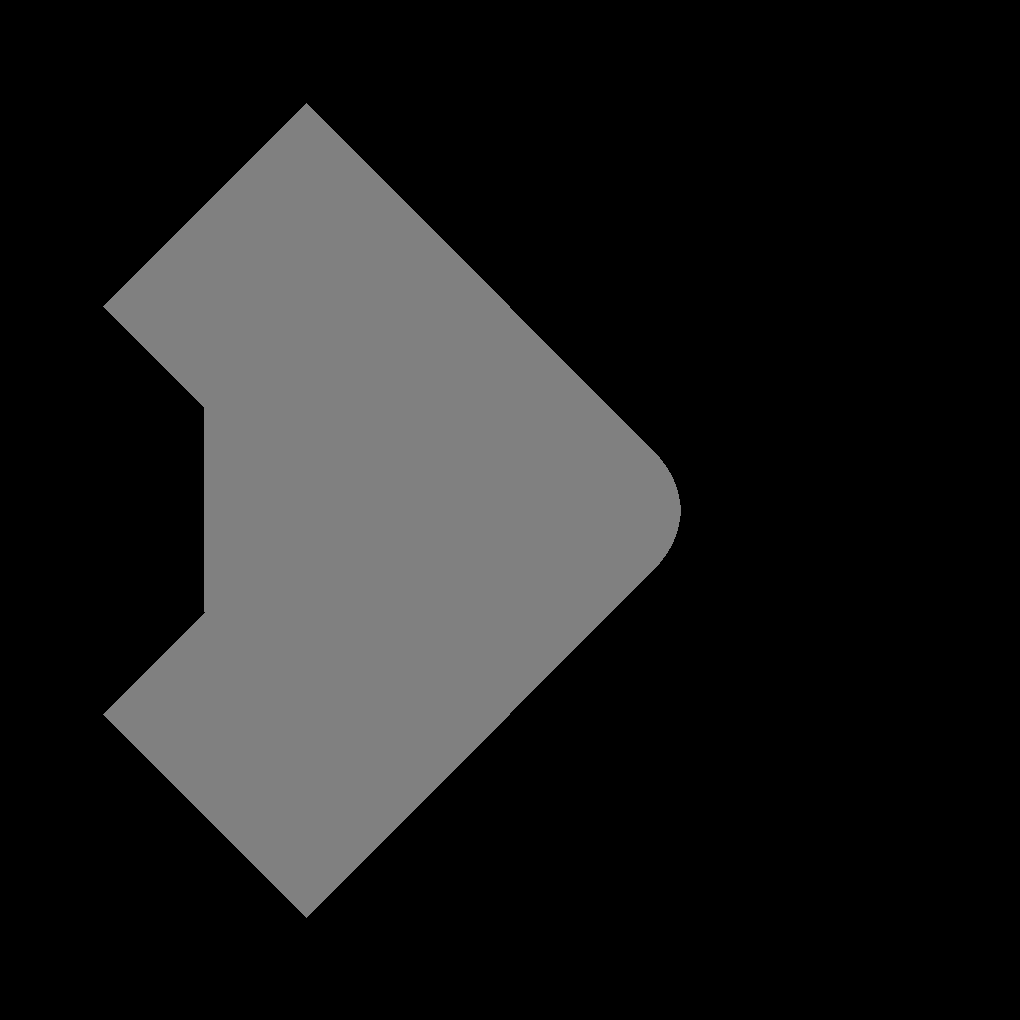} & &
 \cellcolor{gray!25} \includegraphics[width = 1.75cm]{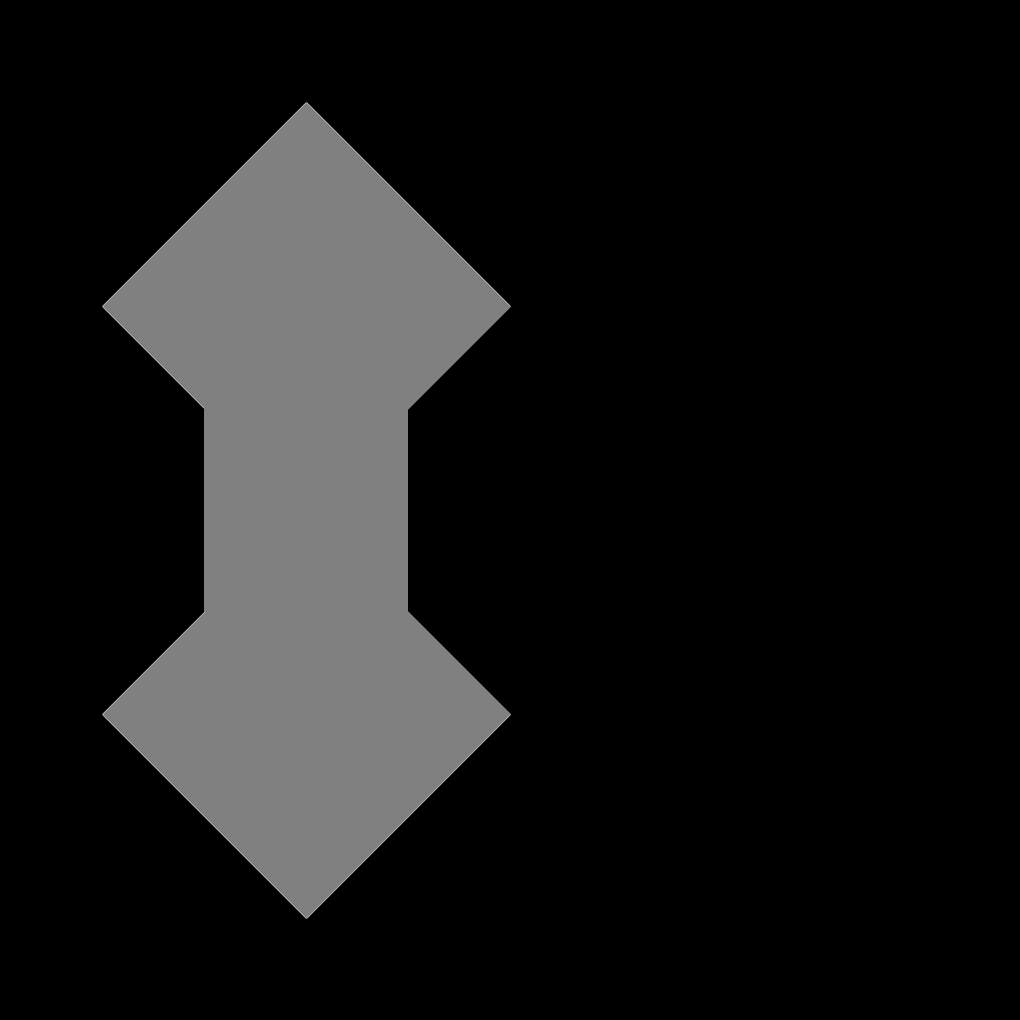} 
 \includegraphics[width = 1.75cm]{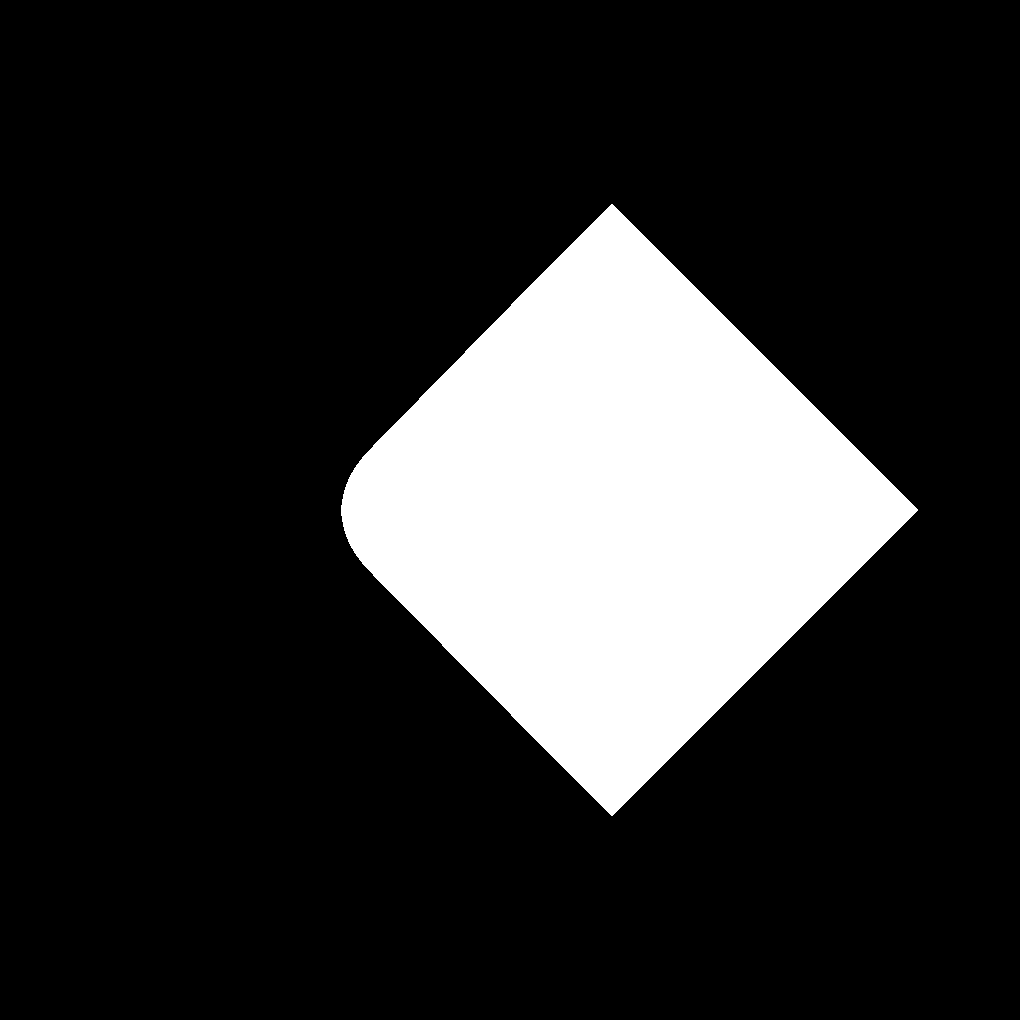} & &
 \cellcolor{gray!25} \includegraphics[width = 1.75cm]{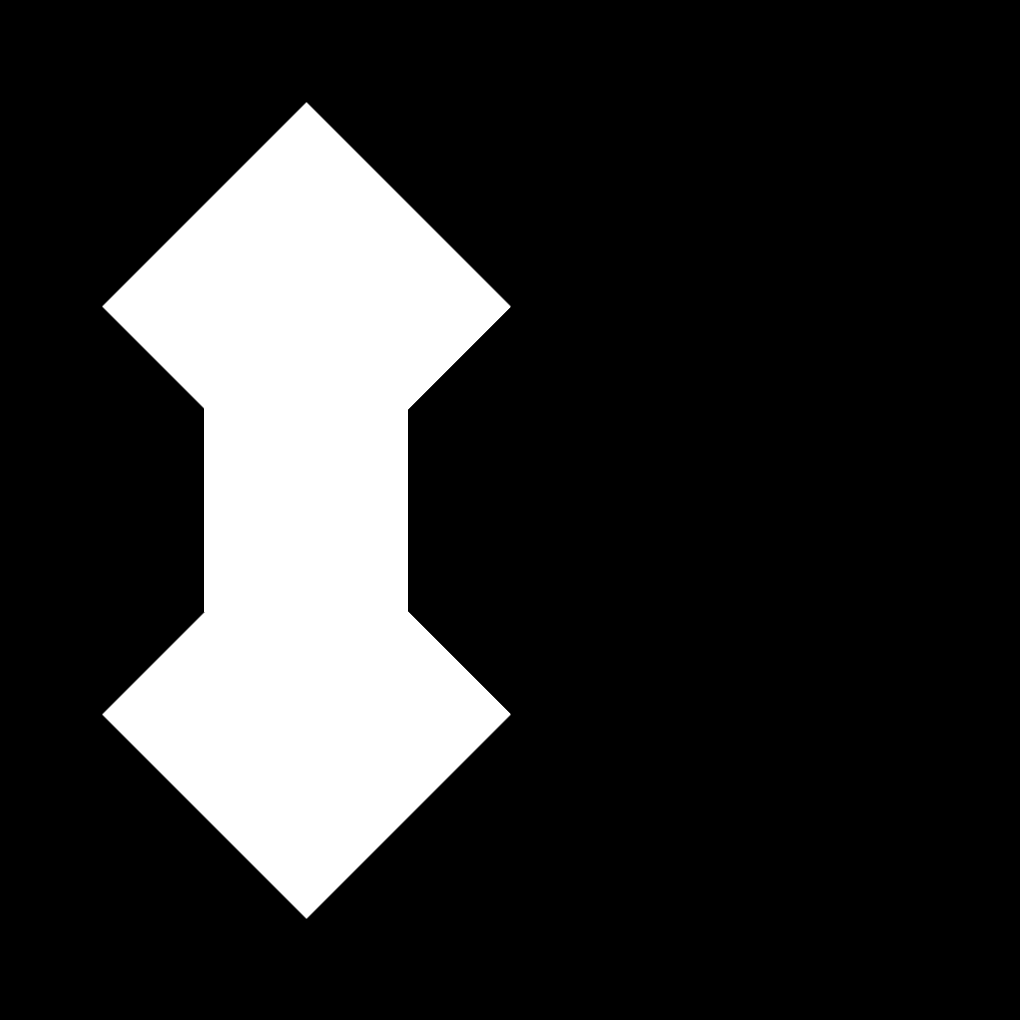} 
 \includegraphics[width = 1.75cm]{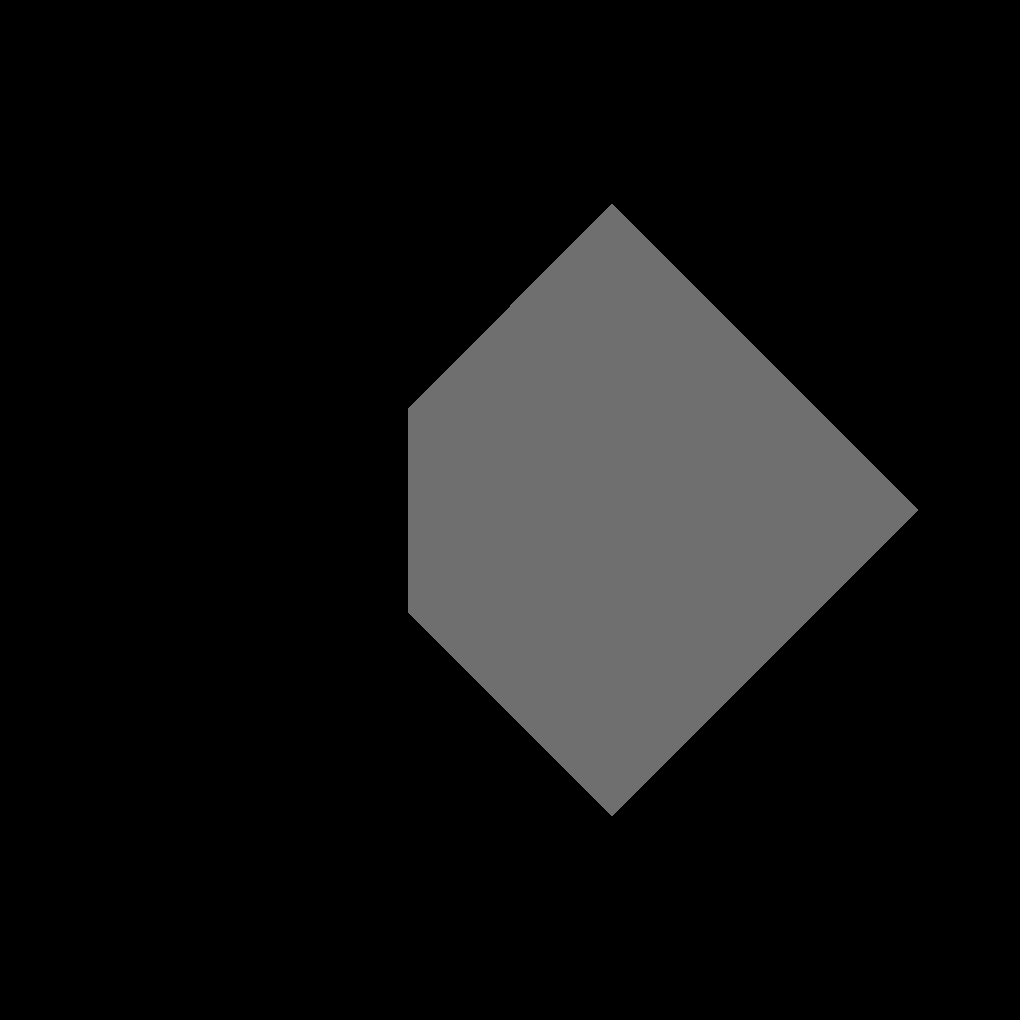} \\
 & \cellcolor{gray!25}$\tilde{p}(I/H_1)=0.7466$ & & \cellcolor{gray!25}$\tilde{p}(I/H_2)=0.7445$ & & \cellcolor{gray!25}$\tilde{p}(I/H_3)=0.3679$\\
  {\scriptsize{Proximal 4}}& \cellcolor{gray!25}$\tilde{p}(H_1)=0.3910$ & & \cellcolor{gray!25}$\tilde{p}(H_2)=0.3679$ & & \cellcolor{gray!25}$\tilde{p}(H_3)=0.4743$\\
 & \cellcolor{gray!25}$\mathbf{p(H_1/I)=0.3943}$ & & \cellcolor{gray!25}$p(H_2/I)=0.3700$ & & \cellcolor{gray!25}$p(H_3/I)=0.2357$\\ \\
 {\includegraphics[width = 1.75cm]{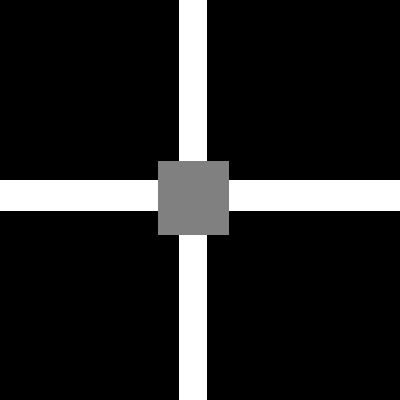}} & 
 \cellcolor{gray!25} \includegraphics[width = 1.75cm]{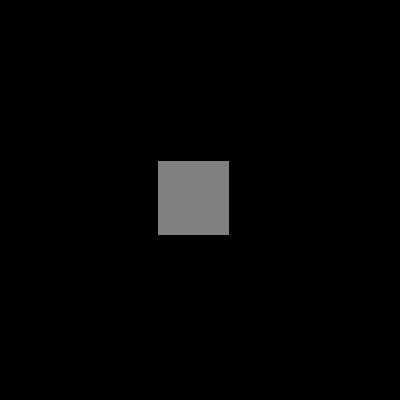} 
 \includegraphics[width = 1.75cm]{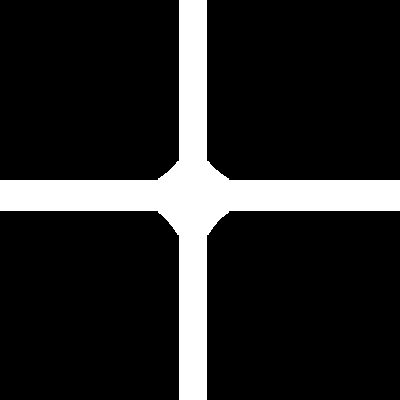} & &
 \cellcolor{gray!25} \includegraphics[width = 1.75cm]{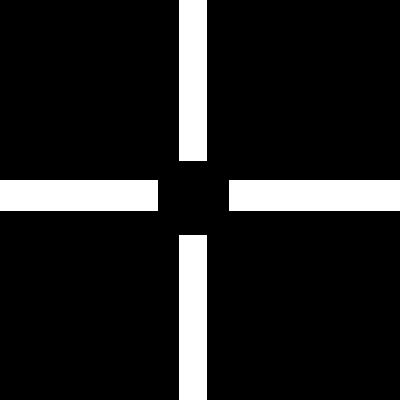} 
 \includegraphics[width = 1.75cm]{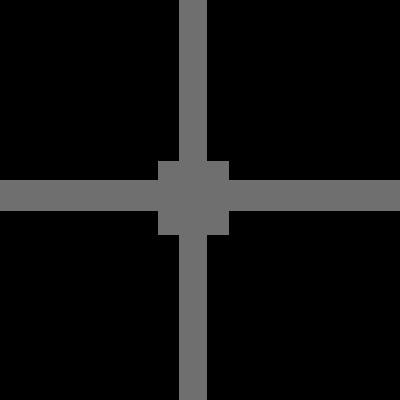} & &
 \cellcolor{gray!25} \includegraphics[width = 1.75cm]{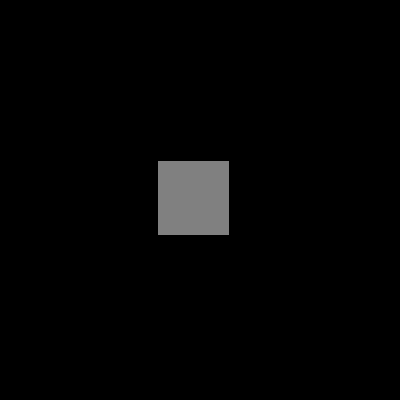} 
 \includegraphics[width = 1.75cm]{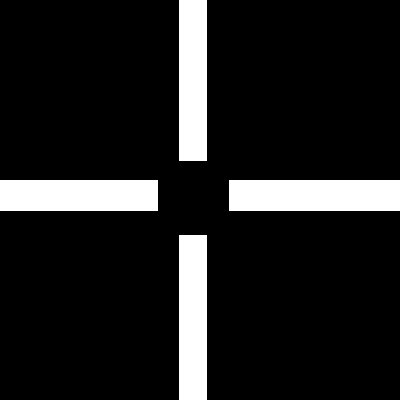} \\
 & \cellcolor{gray!25}$\tilde{p}(I/H_1)=8823$ & & \cellcolor{gray!25}$\tilde{p}(I/H_2)=0.3679$ & & \cellcolor{gray!25}$\tilde{p}(I/H_3)=0.9108$\\
  {\scriptsize{Proximal 5}}& \cellcolor{gray!25}$\tilde{p}(H_1)=0.6020$ & & \cellcolor{gray!25}$\tilde{p}(H_2)=0.3879$ & & \cellcolor{gray!25}$\tilde{p}(H_3)=0.3679$\\
 & \cellcolor{gray!25}$\mathbf{p(H_1/I)=0.5265}$ & & \cellcolor{gray!25}$p(H_2/I)=0.1414$ & & \cellcolor{gray!25}$p(H_3/I)=0.3321$\\ \\
 {\includegraphics[width = 1.75cm]{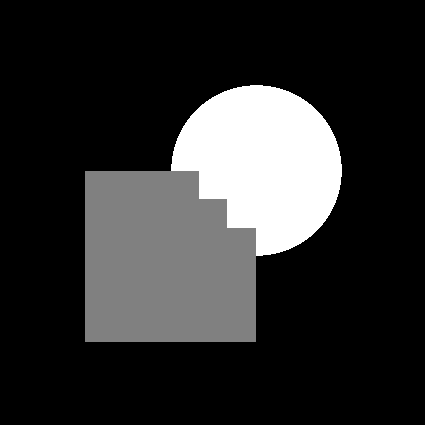}} & 
 \cellcolor{gray!25} \includegraphics[width = 1.75cm]{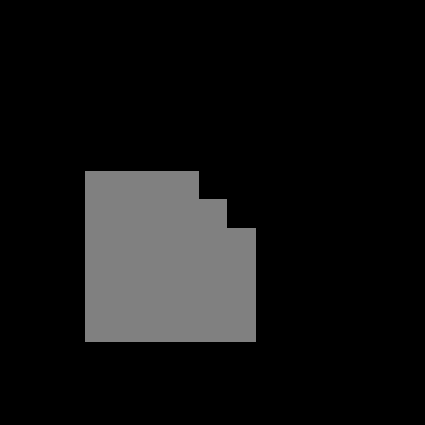} 
 \includegraphics[width = 1.75cm]{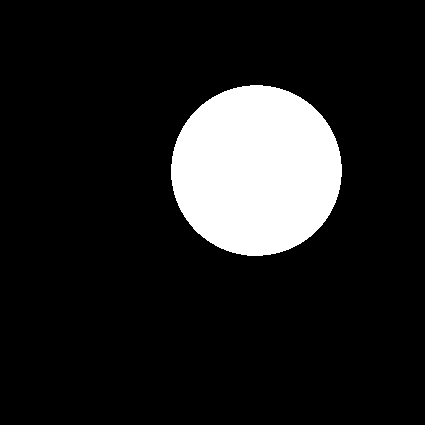} & &
 \cellcolor{gray!25} \includegraphics[width = 1.75cm]{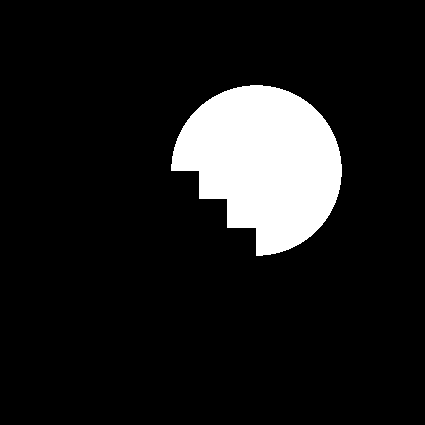} 
 \includegraphics[width = 1.75cm]{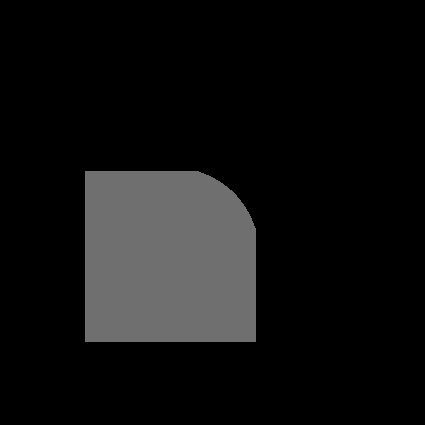} & &
 \cellcolor{gray!25} \includegraphics[width = 1.75cm]{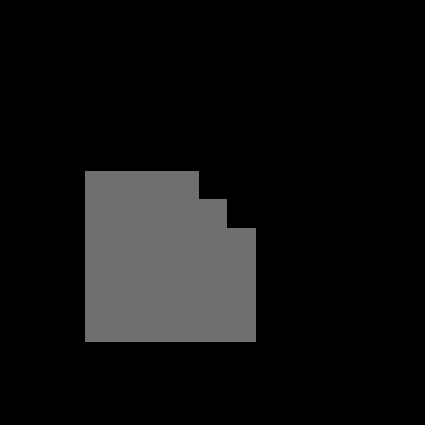} 
 \includegraphics[width = 1.75cm]{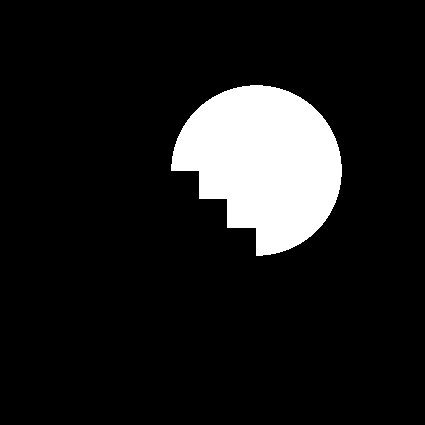} \\
 & \cellcolor{gray!25}$\tilde{p}(I/H_1)=0.4409$ & & \cellcolor{gray!25}$\tilde{p}(I/H_2)=0.5562$ & & \cellcolor{gray!25}$\tilde{p}(I/H_3)=0.3679$\\
  {\scriptsize{Proximal 6}}& \cellcolor{gray!25}$\tilde{p}(H_1)=0.7056$ & & \cellcolor{gray!25}$\tilde{p}(H_2)=0.3679$ & & \cellcolor{gray!25}$\tilde{p}(H_3)=0.4295$\\
 & \cellcolor{gray!25}$\mathbf{p(H_1/I)=0.4618}$ & & \cellcolor{gray!25}$p(H_2/I)=0.3037$ & & \cellcolor{gray!25}$p(H_3/I)=0.2345$\\ \\
 {\includegraphics[width = 1.75cm]{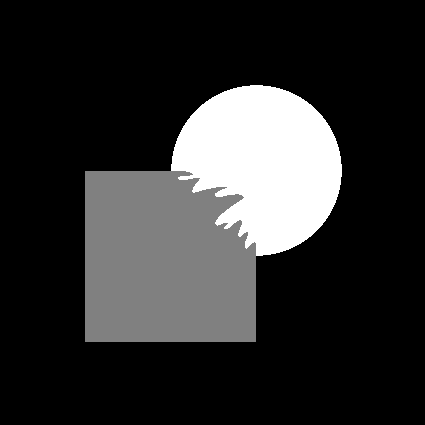}} & 
 \cellcolor{gray!25} \includegraphics[width = 1.75cm]{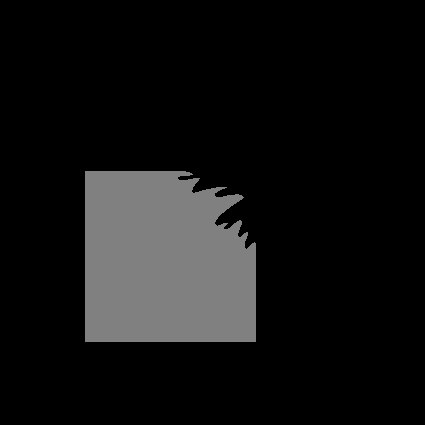} 
 \includegraphics[width = 1.75cm]{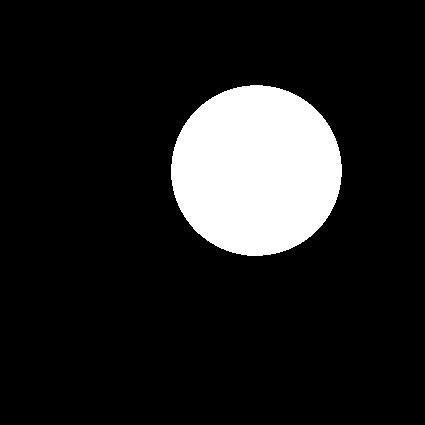} & &
 \cellcolor{gray!25} \includegraphics[width = 1.75cm]{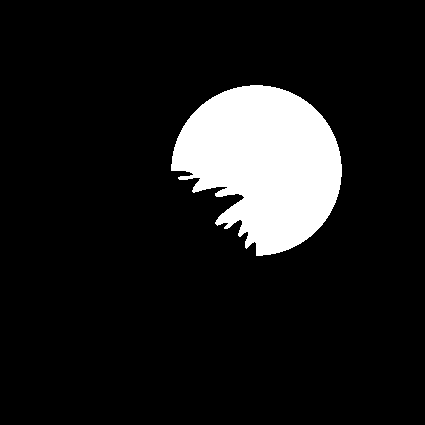} 
 \includegraphics[width = 1.75cm]{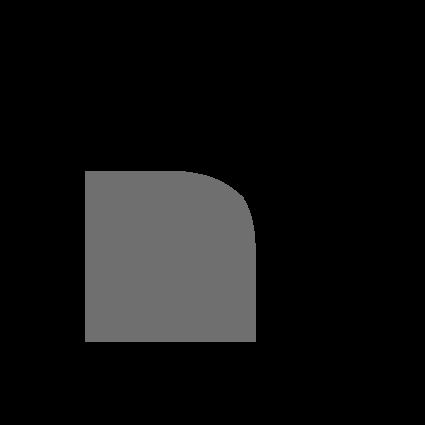} & &
 \cellcolor{gray!25} \includegraphics[width = 1.75cm]{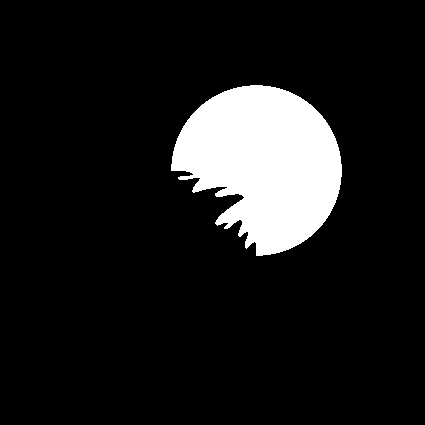} 
 \includegraphics[width = 1.75cm]{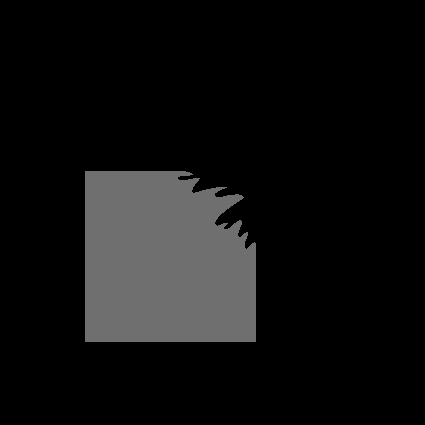} \\
 & \cellcolor{gray!25}$\tilde{p}(I/H_1)=0.7343$ & & \cellcolor{gray!25}$\tilde{p}(I/H_2)=0.7823$ & & \cellcolor{gray!25}$\tilde{p}(I/H_3)=0.3679$\\
  {\scriptsize{Proximal 7}}& \cellcolor{gray!25}$\tilde{p}(H_1)=0.5995$ & & \cellcolor{gray!25}$\tilde{p}(H_2)=0.4087$ & & \cellcolor{gray!25}$\tilde{p}(H_3)=0.3679$\\
 & \cellcolor{gray!25}$\mathbf{p(H_1/I)=0.4917}$ & & \cellcolor{gray!25}$p(H_2/I)=0.3572$ & & \cellcolor{gray!25}$p(H_3/I)=0.1512$\\ \\
 {\includegraphics[width = 1.75cm]{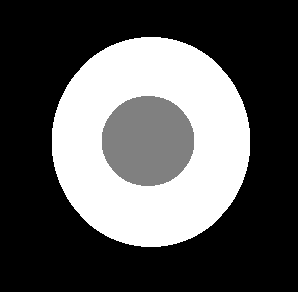}} & 
 \cellcolor{gray!25} \includegraphics[width = 1.75cm]{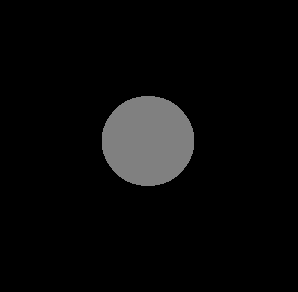} 
 \includegraphics[width = 1.75cm]{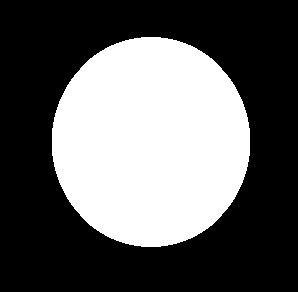} & &
 \cellcolor{gray!25} \includegraphics[width = 1.75cm]{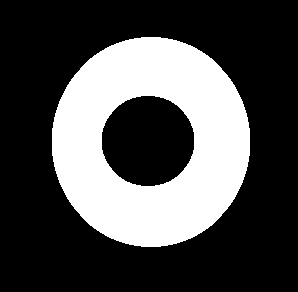} 
 \includegraphics[width = 1.75cm]{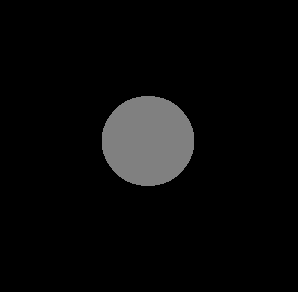} & &
 \cellcolor{gray!25} \includegraphics[width = 1.75cm]{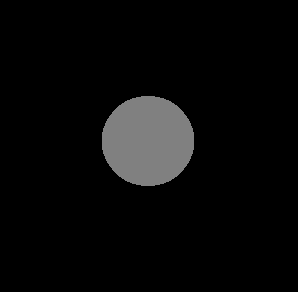} 
 \includegraphics[width = 1.75cm]{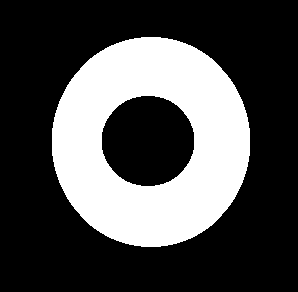} \\
 & \cellcolor{gray!25}$\tilde{p}(I/H_1)=1$ & & \cellcolor{gray!25}$\tilde{p}(I/H_2)=0.3729$ & & \cellcolor{gray!25}$\tilde{p}(I/H_3)=0.3679$\\
  {\scriptsize{Proximal 8}}& \cellcolor{gray!25}$\tilde{p}(H_1)=0.7087$ & & \cellcolor{gray!25}$\tilde{p}(H_2)=0.3679$ & & \cellcolor{gray!25}$\tilde{p}(H_3)=0.3679$\\
 & \cellcolor{gray!25}$\mathbf{p(H_1/I)=0.7223}$ & & \cellcolor{gray!25}$p(H_2/I)=0.1398$ & & \cellcolor{gray!25}$p(H_3/I)=0.1379$\\ \\
  \caption{Synthetic experiments. Each row shows a different experiment: the original image (proximal stimulus) is shown on the left and it is followed by the three different hypothesis (each one separated by a gray box). For the first two hypothesis, $H_i$ for $i=1,2$, we show: the object at depth 1 (left) and the disoccluded object at depth 2 (right). Notice that the object at depth 1 acts as a mask for disoccluding the object at depth 2. In the case of the third hypothesis, $H_3$, both objects are considered to be at the same depth and completely visible in the original image (no disocclusion is applied). 
  In the lower part of each hypothesis we show the values $\tilde{p}(I/H_i)$, $\tilde{p}(H_2)$ (proportional, respectively, to the likelihood and prior probabilities), and the posterior probability $p(H_i/I)$. The probability value of the preferred hypothesis $H_P$ is highlighted in boldface.}
\label{tab:alltogether1}
\end{longtable}

\begin{longtable}{llll}
 {\includegraphics[width = 1.75cm]{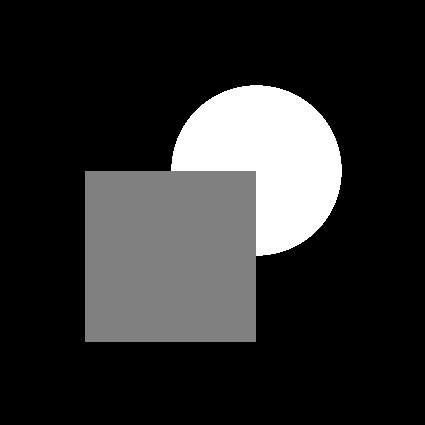}} & 
 \cellcolor{gray!25} \includegraphics[width = 1.75cm]{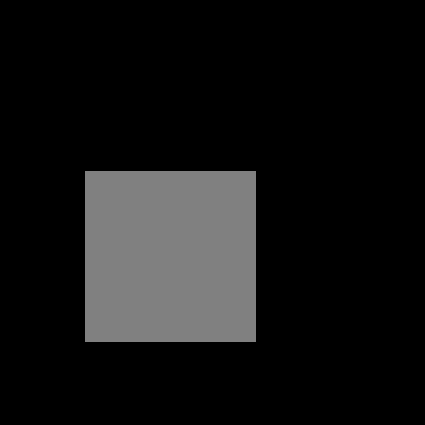} 
 \includegraphics[width = 1.75cm]{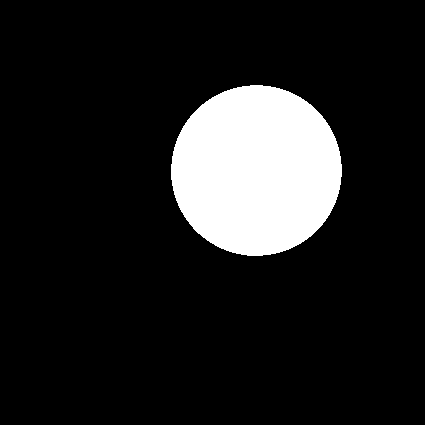} & &
 \cellcolor{gray!25} \includegraphics[width = 1.75cm]{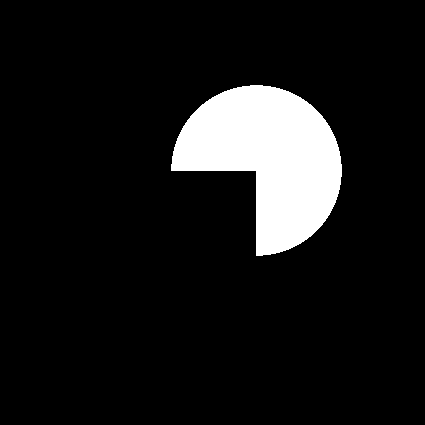} 
 \includegraphics[width = 1.75cm]{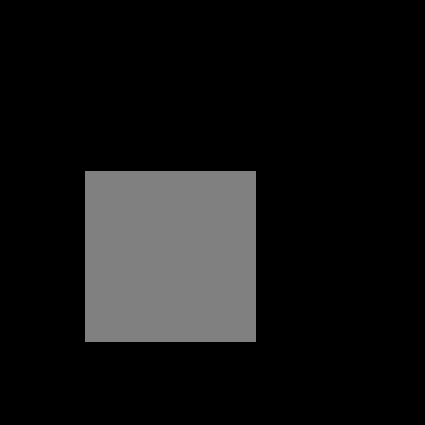} \\
 & \cellcolor{gray!25}$\tilde{p}(I/H_1)=0.6848$ & & \cellcolor{gray!25}$\tilde{p}(I/H_2)=0.3679$\\
  {\scriptsize{Proximal 9}}& \cellcolor{gray!25}$\tilde{p}(H_1)=0.6487$ & & \cellcolor{gray!25}$\tilde{p}(H_2)=0.3679$\\
 & \cellcolor{gray!25}$p(H_1/I)=\mathbf{0.6214}$ & & \cellcolor{gray!25}$p(H_2/I)=0.1893$\\ \\
 {\includegraphics[width = 1.75cm]{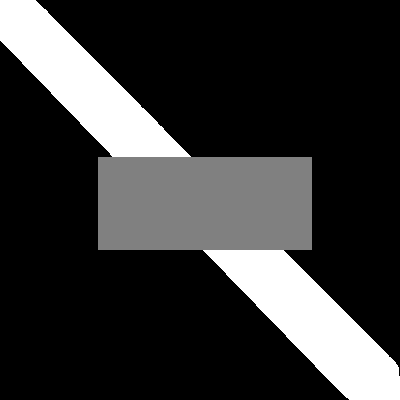}} & 
 \cellcolor{gray!25} \includegraphics[width = 1.75cm]{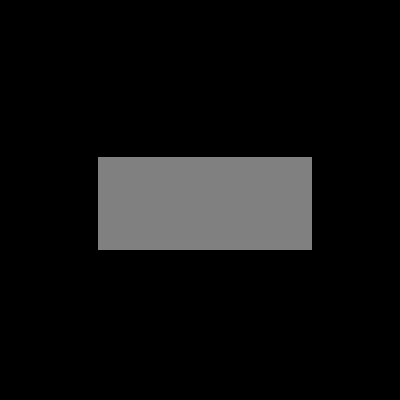} 
 \includegraphics[width = 1.75cm]{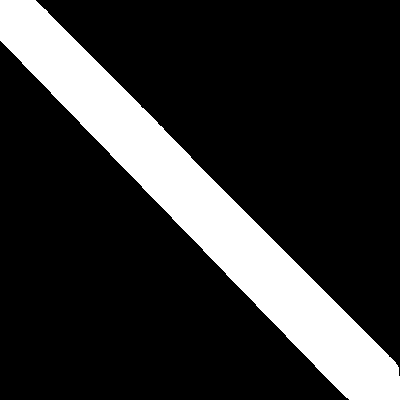} & &
 \cellcolor{gray!25} \includegraphics[width = 1.75cm]{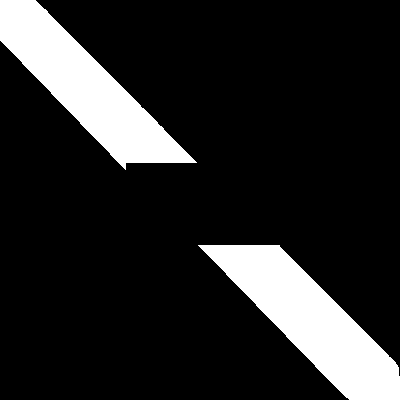} 
 \includegraphics[width = 1.75cm]{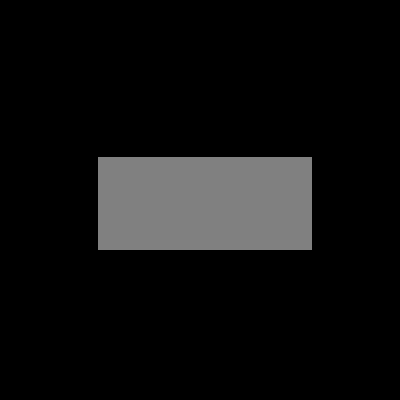} \\
 & \cellcolor{gray!25}$\tilde{p}(I/H_1)=0.3877$ & & \cellcolor{gray!25}$\tilde{p}(I/H_2)=0.3679$\\
  {\scriptsize{Proximal 10}}& \cellcolor{gray!25}$\tilde{p}(H_1)=0.3804$ & & \cellcolor{gray!25}$\tilde{p}(H_2)=0.3679$\\
 & \cellcolor{gray!25}$\mathbf{p(H_1/I)=0.3527}$ & & \cellcolor{gray!25}$p(H_2/I)=0.3237$\\ \\
 {\includegraphics[width = 1.75cm]{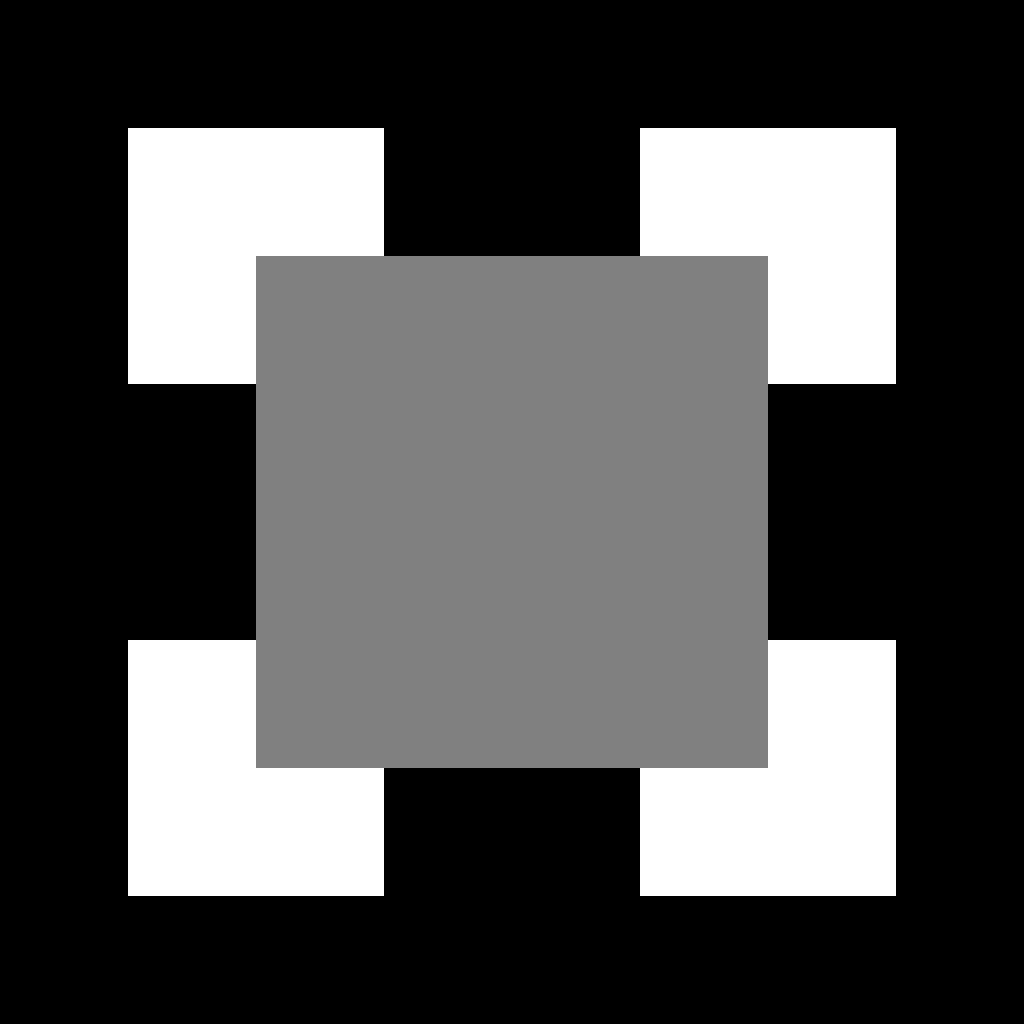}} & 
 \cellcolor{gray!25} \includegraphics[width = 1.75cm]{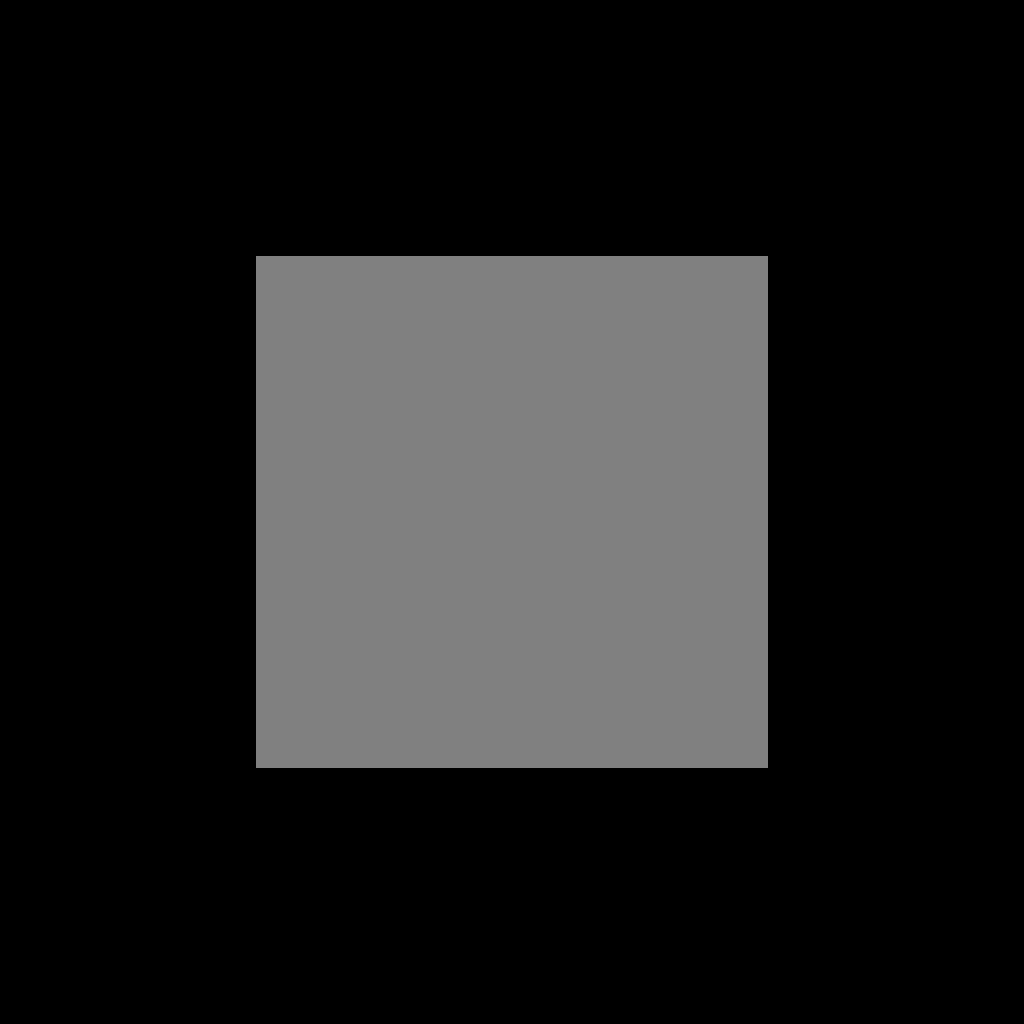} 
 \includegraphics[width = 1.75cm]{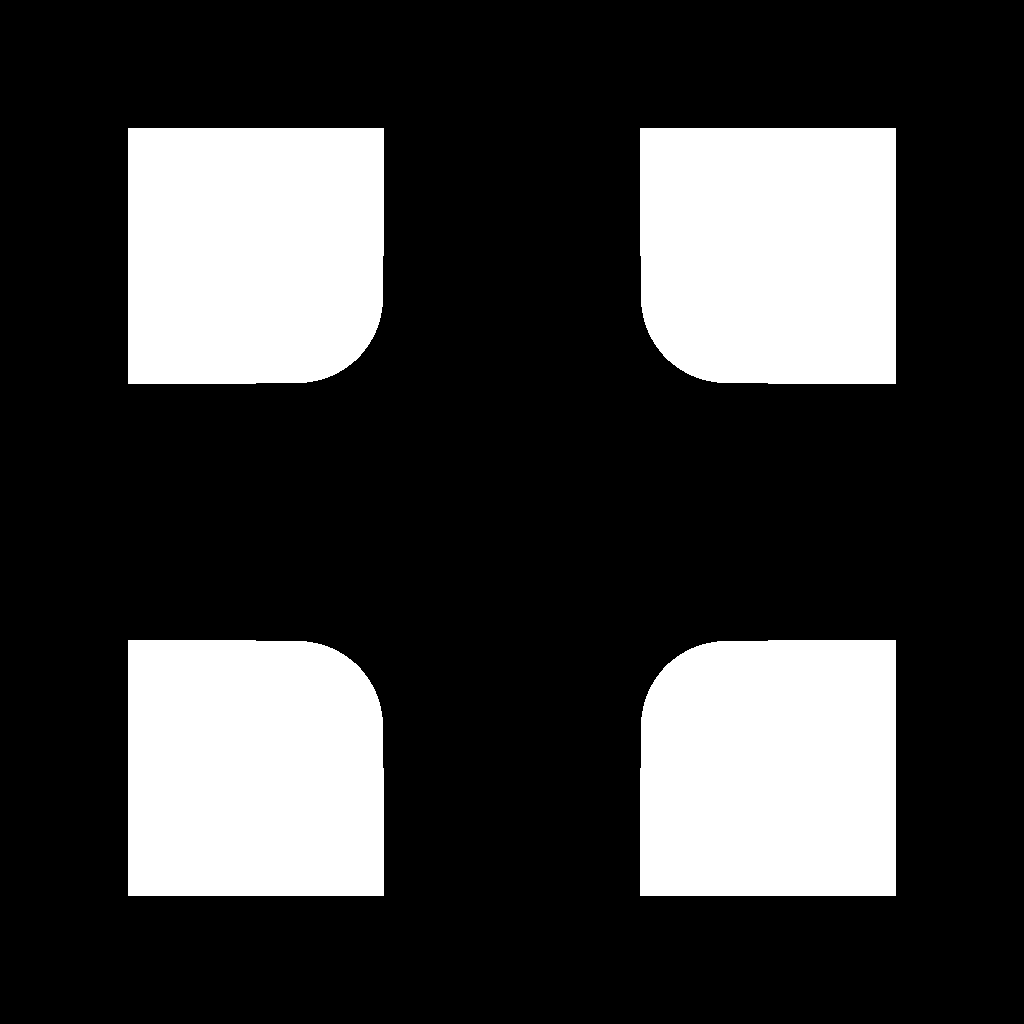} & &
 \cellcolor{gray!25} \includegraphics[width = 1.75cm]{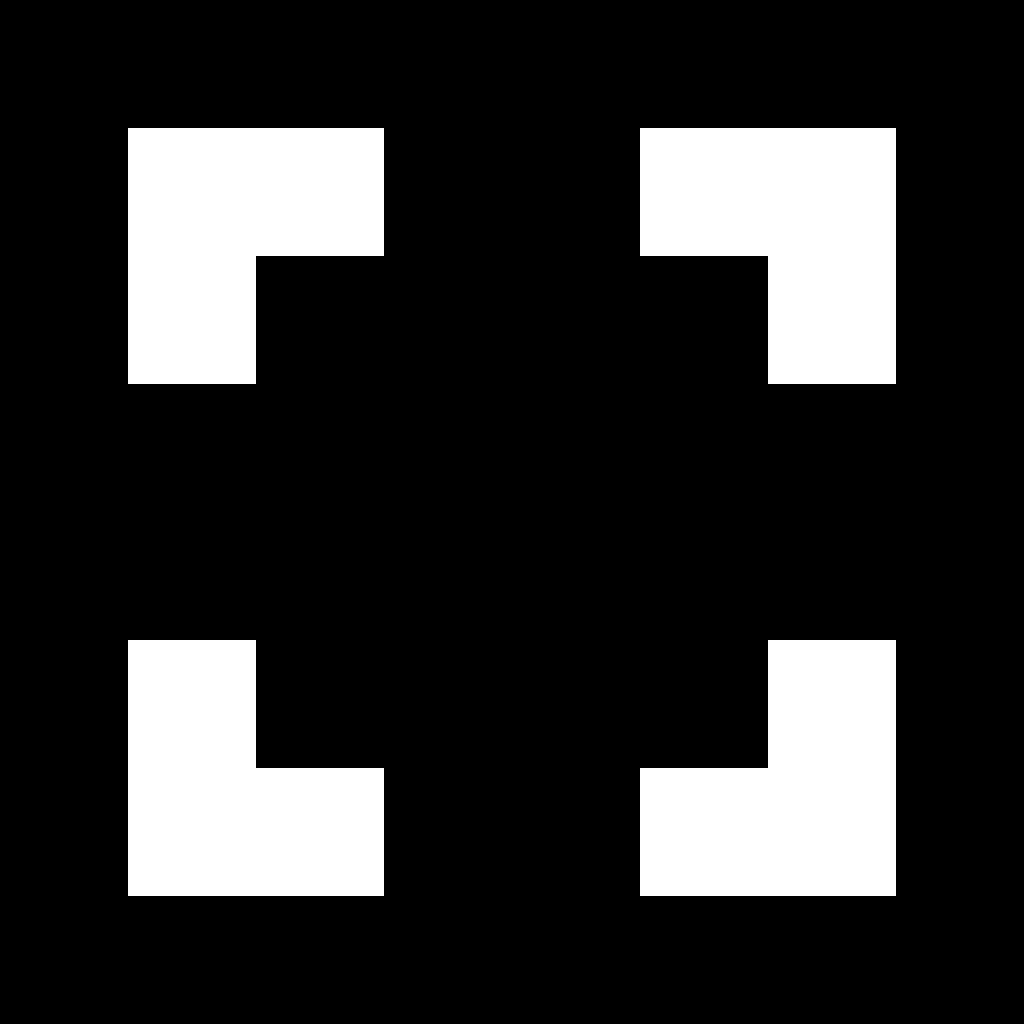} 
 \includegraphics[width = 1.75cm]{figures/s18/Mask1} \\
 & \cellcolor{gray!25}$\tilde{p}(I/H_1)=0.5436$ & & \cellcolor{gray!25}$\tilde{p}(I/H_2)=0.3679$\\
  {\scriptsize{Proximal 11}}& \cellcolor{gray!25}$\tilde{p}(H_1)=0.4051$ & & \cellcolor{gray!25}$\tilde{p}(H_2)=0.3679$\\
 & \cellcolor{gray!25}$\mathbf{p(H_1/I)=0.4486}$ & & \cellcolor{gray!25}$p(H_2/I)=0.2757$\\ \\
 {\includegraphics[width = 1.75cm]{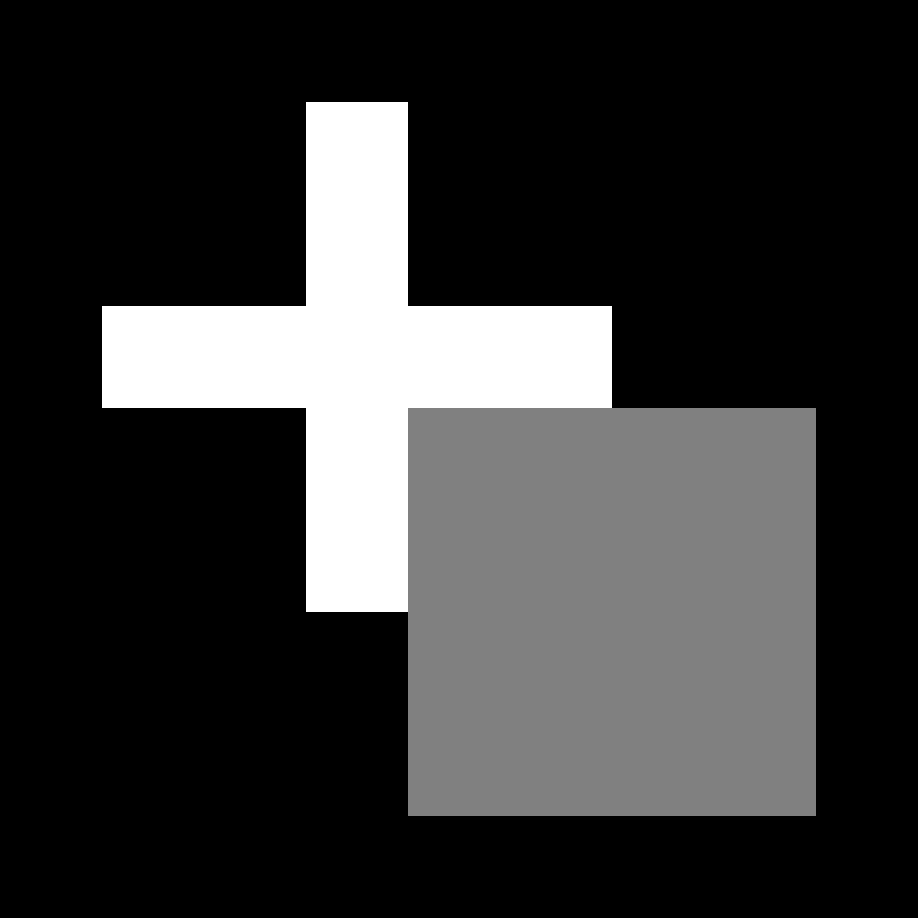}} & 
 \cellcolor{gray!25} \includegraphics[width = 1.75cm]{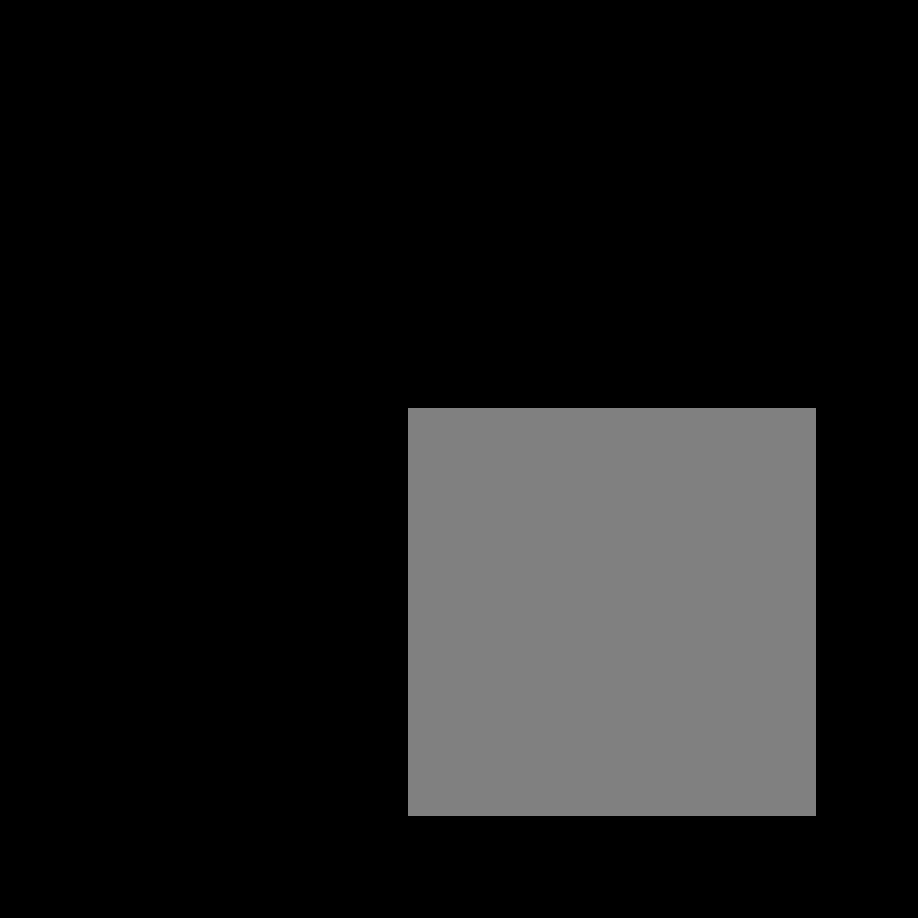} 
 \includegraphics[width = 1.75cm]{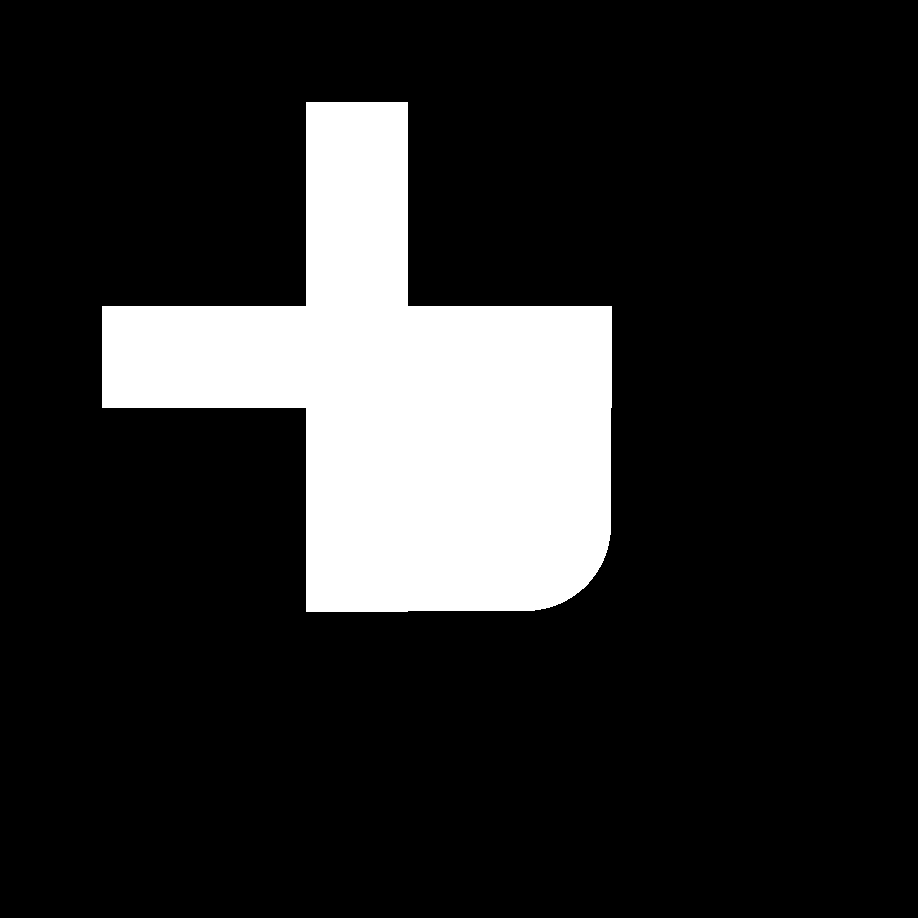} & &
 \cellcolor{gray!25} \includegraphics[width = 1.75cm]{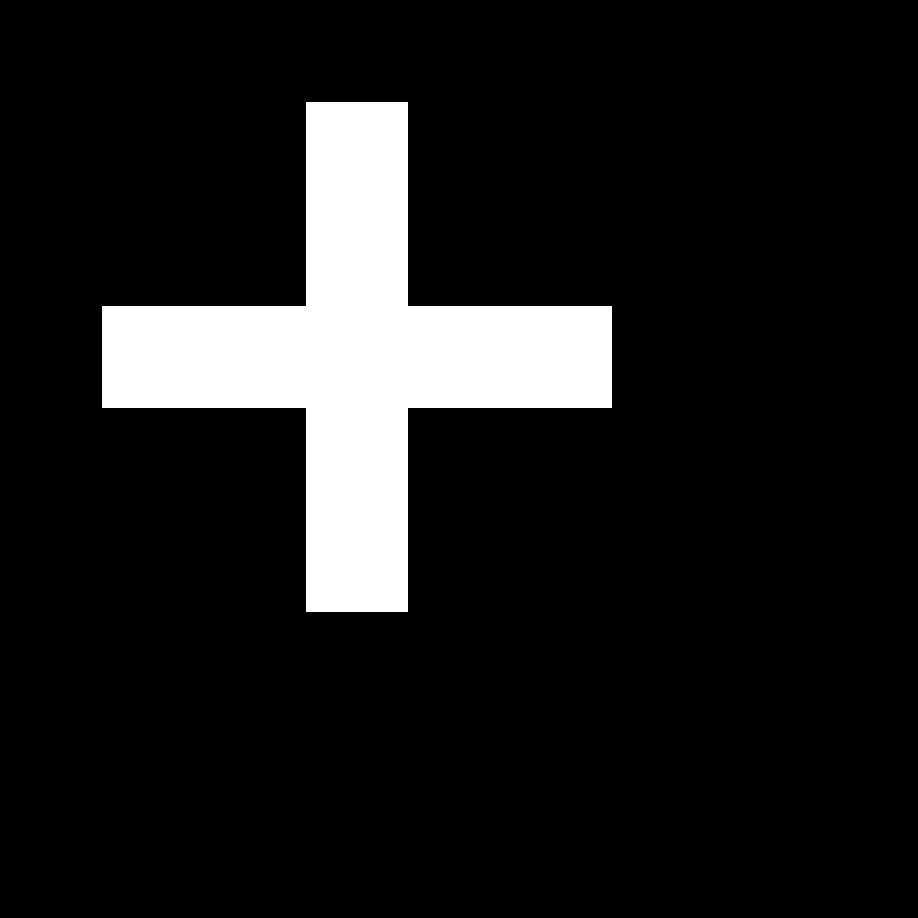} 
 \includegraphics[width = 1.75cm]{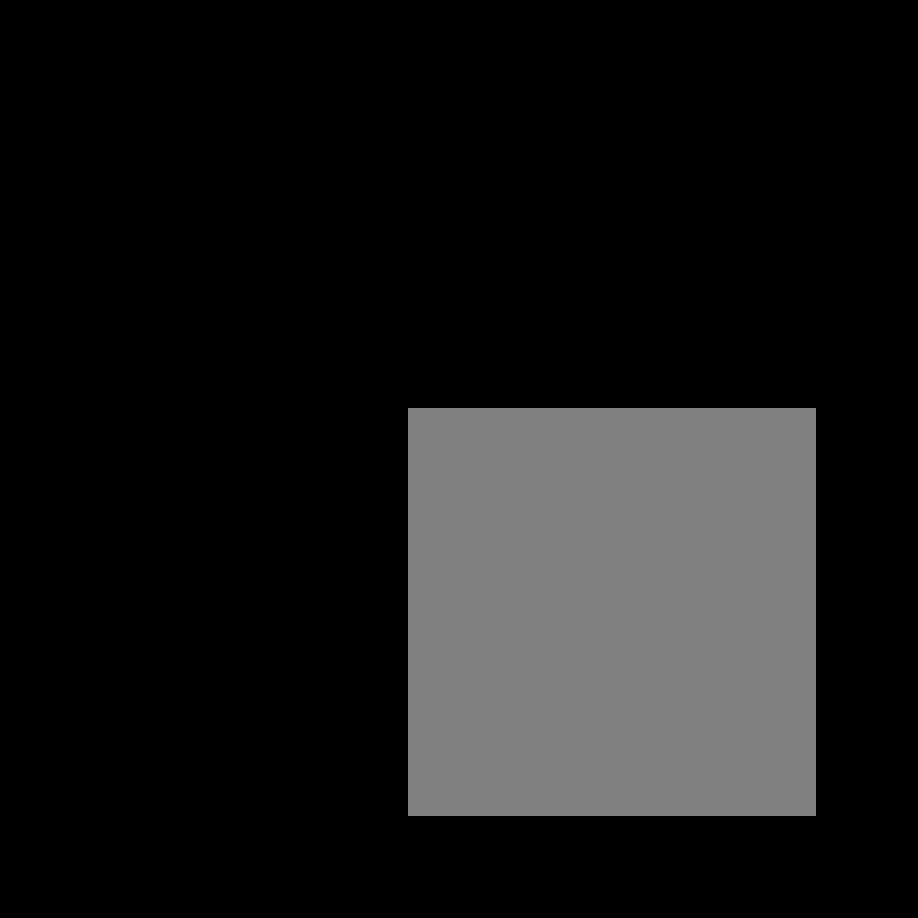} \\
 & \cellcolor{gray!25}$\tilde{p}(I/H_1)=0.6172$ & & \cellcolor{gray!25}$\tilde{p}(I/H_2)=0.3679$\\
  {\scriptsize{Proximal 12}}& \cellcolor{gray!25}$\tilde{p}(H_1)=0.3679$ & & \cellcolor{gray!25}$\tilde{p}(H_2)=0.4324$\\
 & \cellcolor{gray!25}$p(H_1/I)=\mathbf{0.4165}$ & & \cellcolor{gray!25}$p(H_2/I)=0.2918$\\ \\
  \caption{Synthetic experiments. Each row shows a different experiment: the original image (proximal stimulus) is shown on the left and it is followed by two different hypothesis (each one separated by a gray box). For each hypothesis $H_i$ we show: the object at depth 1 (left) and the disoccluded object at depth 2 (right). Notice that the object at depth 1 acts as a mask for disoccluding the object at depth 2. 
  In the lower part of each hypothesis we show the values $\tilde{p}(I/H_i)$, $\tilde{p}(H_2)$ (proportional, respectively, to the likelihood and prior probabilities), and the posterior probability $p(H_i/I)$. The probability value of the preferred hypothesis $H_P$ is highlighted in boldface. The third hypothesis, $H_3$, where both objects are considered to be at the same depth and completely visible in the original image (no disocclusion is applied) is not shown here because it coincides with $H_2$ (due to the fact that the disocclusion algorithm does not change the objects being disoccluded in $H_2$). More details are given in the text.}
\label{tab:alltogether2}
\end{longtable}

\begin{longtable}{ll}
 {\includegraphics[width = 1.75cm]{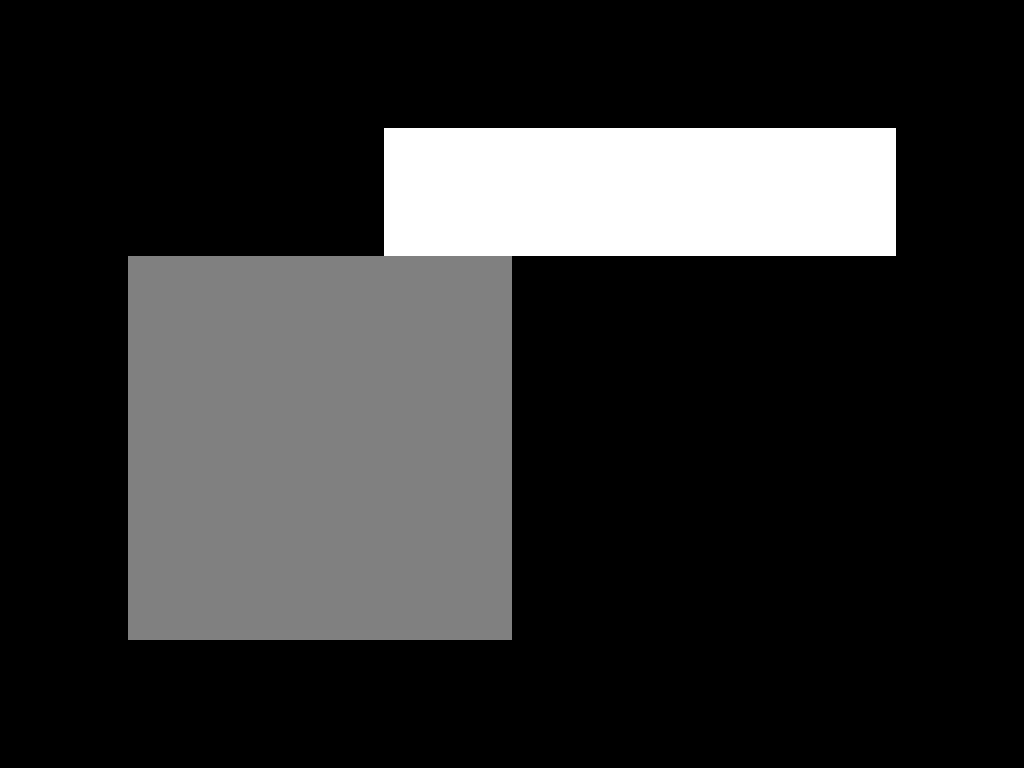}} & 
 \cellcolor{gray!25} \includegraphics[width = 1.75cm]{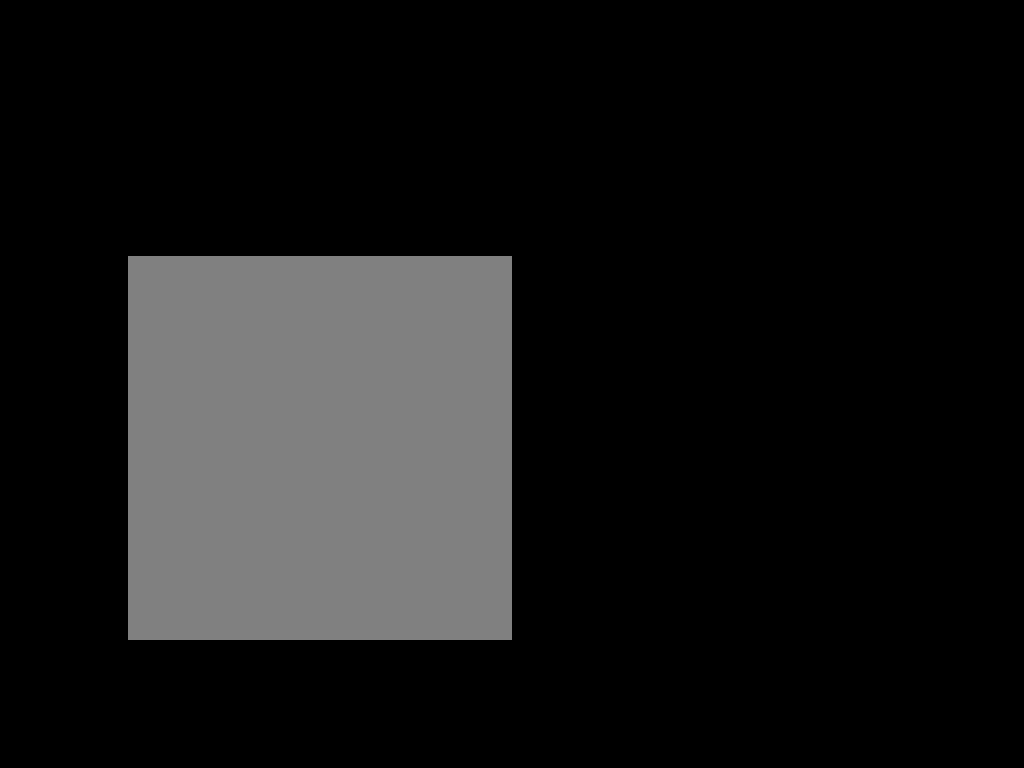} 
 \includegraphics[width = 1.75cm]{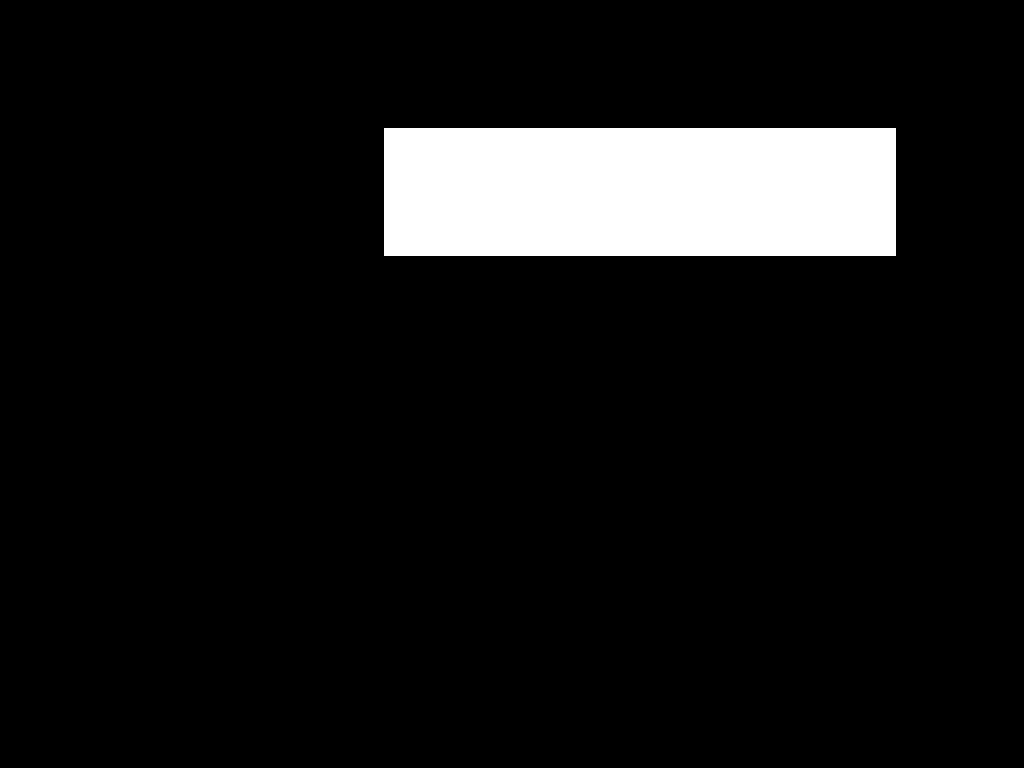}\\
 & \cellcolor{gray!25}$\tilde{p}(I/H)=0.3679$ \\
  {\scriptsize{Proximal 13}}& \cellcolor{gray!25}$\tilde{p}(H)=0.3679$  \\
 & \cellcolor{gray!25}$p(H/I)=0.3333$ \\
 \caption{Synthetic experiment where the three hypothesis coincide (since the disocclusion applied in the first two hypothesis does not change the objects). Thus, we have $H = H_1 = H_2 = H_3$, and the posterior probability is the same for all three hypothesis. As for depth order, $H_3$ is interpreted as two objects at the same depth (and having the real relative size which is observed in the proximal image) while $H_1$ can be interpreted as a gray square which is closer to the observer, plus a white rectangle which can be of bigger size but farther away from the square and whose boundary partially coincides with part of the boundary of the square. Finally, $H_2$ can be interpreted as a white rectangle which is closer to the observer, plus a gray square which can be of bigger size but farther away from the rectangle and whose boundary partially coincides with part of the boundary of the rectangle.}
\label{tab:alltogether3}
\end{longtable}

\subsection{Real images}\label{sec:real}

In this section we show some results on real images from the Berkeley dataset  \cite{MartinFTM01} and the dataset provided in \cite{SemanticBenchmark}. 
For all experiments, we present the real image, the proximal stimulus $I$ which is a segmentation of the real image (one of the segmentations provided in the databases), followed by the three hypothesis $H_i$ (each one separated by a gray box), together with the values $\tilde{p}(I/H_i)$ and $\tilde{p}(H_i)$ proportional to the conditional probability and the prior probability, respectively, and the probability value $p(H_i|I)$. The probability value of the preferred hypothesis $H_P$ is highlighted in boldface. For the first two hypothesis, $H_1$ and $H_2$, we display the objects at depth 1 on the left, and the disoccluded objects (at depth 2) on the right. Let us recall that the objects at depth 1 are considered the inpainting mask for disoccluding the objects at depth 2. Finally the last column is the hypothesis $H_3$ where the two objects are fitting together at the same depth. 

We start illustrating that our method is robust to different segmentations of the same image. Table \ref{2segs} shows a real image with a bear holding a branch and two different segmentations (representing the proximal stimuli). Both segmentations are from the ground truth available in \cite{MartinFTM01}. Segmentation 1 reflects that some flowers are partially occluding the bear and increasing the complexity of the bear shape; 
the flowers do not appear in segmentation 2 and thus the bear shape has a lower complexity (its complexity is $0.53$, while in the previous case, Segmentation 1, was $1.34$). Notice that the values $\tilde{p}$ are not comparable among the two experiments (only among different hypothesis within the same experiment) because they use a different normalizing constant $\omega_2$ (see Section \ref{sec:approach} for further details). 
Finally, the most preferred interpretation of the image coincides using the two different segmentations, i.e., it is a branch partially occluding a bear for both stimulus.

\setlength{\tabcolsep}{3pt}
\begin{longtable}{llllll}
 \multicolumn{6}{c}{\includegraphics[width = 3cm]{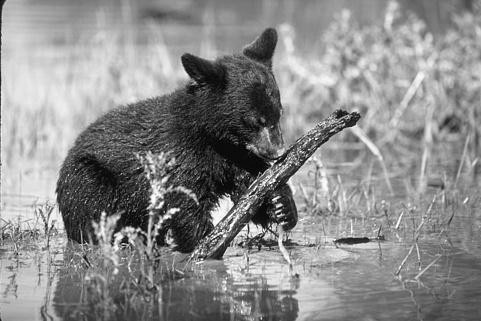}}\\ 
 {\includegraphics[width = 1.75cm]{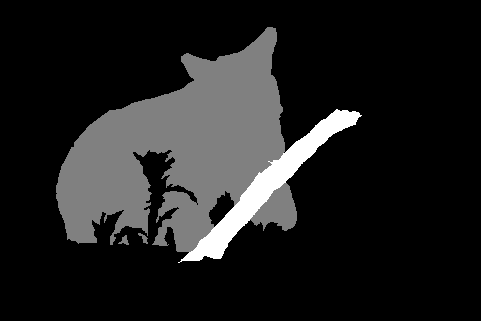}} & 
 \cellcolor{gray!25} \includegraphics[width = 1.75cm]{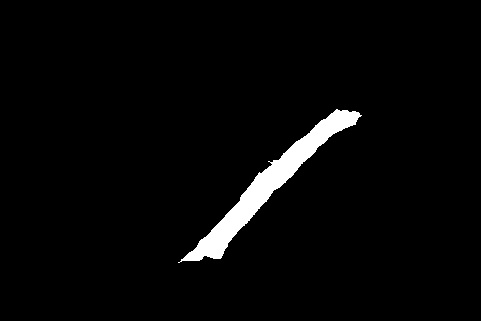} 
 \includegraphics[width = 1.75cm]{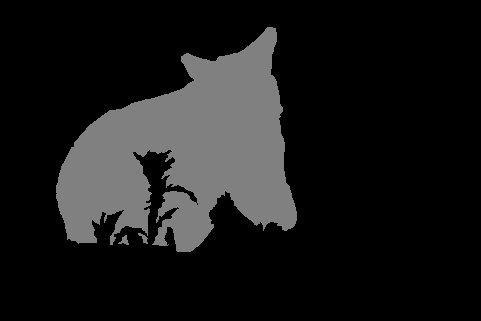} & &
 \cellcolor{gray!25} \includegraphics[width = 1.75cm]{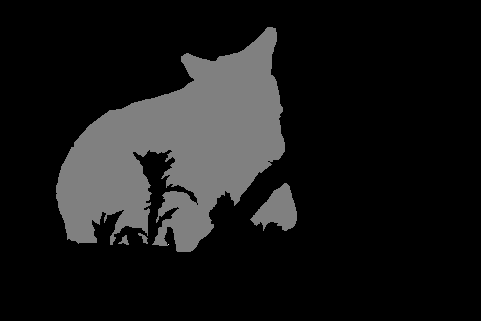} 
 \includegraphics[width = 1.75cm]{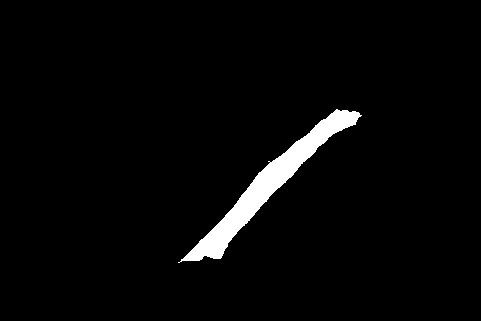} & &
 \cellcolor{gray!25} \includegraphics[width = 1.75cm]{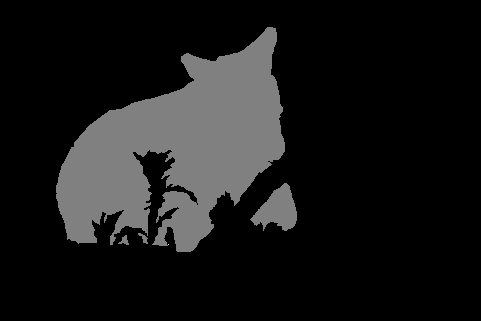} 
 \includegraphics[width = 1.75cm]{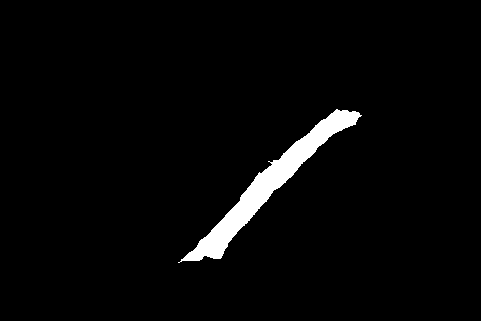} \\
 & \cellcolor{gray!25}$\tilde{p}(I/H_1)=0.7818$ & & \cellcolor{gray!25}$\tilde{p}(I/H_2)=0.7590$ & & \cellcolor{gray!25}$\tilde{p}(I/H_3)=0.3679$\\
 {\scriptsize{Segmentation 1}} 
 & \cellcolor{gray!25}$\tilde{p}(H_1)=0.6996$ & & \cellcolor{gray!25}$\tilde{p}(H_2)=0.3717$ & & \cellcolor{gray!25}$\tilde{p}(H_3)=0.3679$\\
 & \cellcolor{gray!25}$\mathbf{p(H_1/I)=0.4184}$ & & \cellcolor{gray!25}$p(H_2/I)=0.3966$ & & \cellcolor{gray!25}$p(H_3/I)=0.1850$\\ \\
 {\includegraphics[width = 1.75cm]{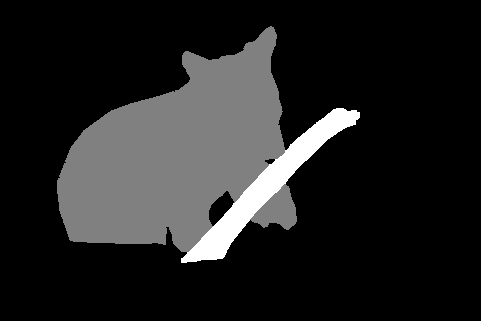}} & 
 \cellcolor{gray!25} \includegraphics[width = 1.75cm]{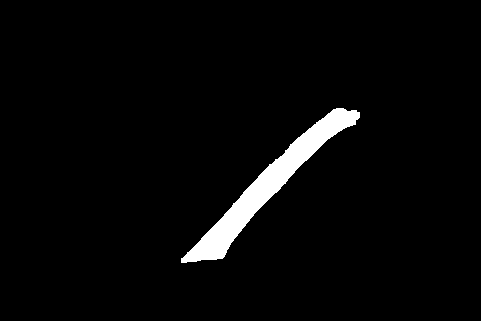} 
 \includegraphics[width = 1.75cm]{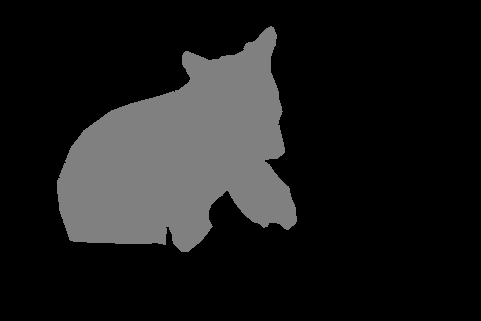} & &
 \cellcolor{gray!25} \includegraphics[width = 1.75cm]{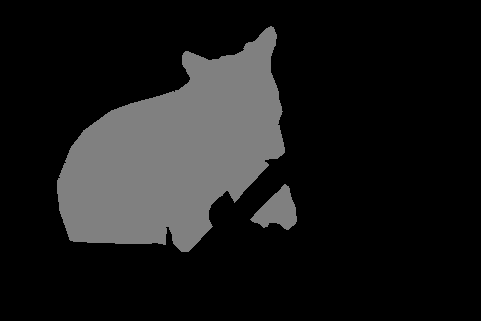} 
 \includegraphics[width = 1.75cm]{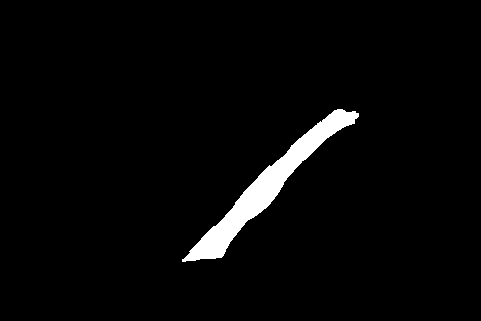} & &
 \cellcolor{gray!25} \includegraphics[width = 1.75cm]{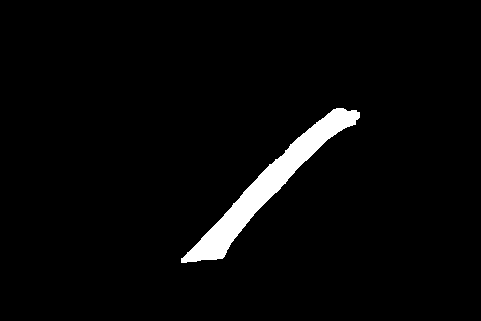} 
 \includegraphics[width = 1.75cm]{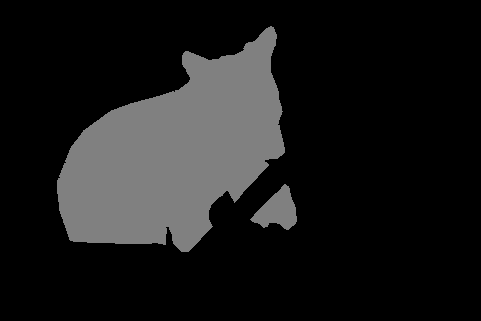} \\
 & \cellcolor{gray!25}$\tilde{p}(I/H_1)=0.6739$ & & \cellcolor{gray!25}$\tilde{p}(I/H_2)=0.6676$ & & \cellcolor{gray!25}$\tilde{p}(I/H_3)=0.3679$\\
 {\scriptsize{Segmentation 2}} 
 & \cellcolor{gray!25}$\tilde{p}(H_1)=0.5265$ & & \cellcolor{gray!25}$\tilde{p}(H_2)=0.3679$ & & \cellcolor{gray!25}$\tilde{p}(H_3)=0.3751$\\
 & \cellcolor{gray!25}$\mathbf{p(H_1/I)=0.4805}$ & & \cellcolor{gray!25}$p(H_2/I)=0.3326$ & & \cellcolor{gray!25}$p(H_3/I)=0.1869$\\ \\
\caption{Experiments with real images from \cite{MartinFTM01}. 
The left-most image is a segmentation of the original image (shown in the first row). 
Both experiments correspond to the same image but considering different segmentations (both extracted from the ground truth segmentation available in \cite{MartinFTM01}). The most preferred interpretation of the image coincides in both experiments, i.e.~a branch partially occluding a bear. Notice that the values $\tilde{p}$ are not comparable among the two experiments (only among different hypothesis within the same experiment) because they use a different normalizing constant $\omega_2$ (see Section \ref{sec:approach} for further details).}
\label{2segs}
\end{longtable}

In Table~\ref{tab:alltogether7} we present results on images of the Berkeley dataset with provided figure-ground ground-truth labeled by humans. Then, Table~\ref{tab:alltogether5} shows experimental results on real images from \cite{SemanticBenchmark} and Table~\ref{tab:alltogether6} shows results on images from the Berkeley Segmentation database \cite{MartinFTM01}. Each row shows a different experiment: the two left-most images are, respectively, the original image and a segmentation of it, they are followed  by the three different hypothesis (each one separated by a gray box).
For the images in Table~\ref{tab:alltogether7}, superimposed on the original image, we display the provided figure-ground ground-truth~\cite{figuregroundBerkeley} as a boundary in two colors, namely, black and white. The black side of the border indicates the object that is behind, while the white region indicates the frontal object.

\setlength{\tabcolsep}{3pt}
\begin{longtable}{lllllll}
 {\includegraphics[width = 1.6cm]{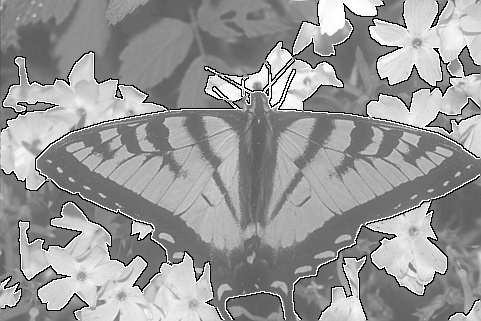}} & 
 {\includegraphics[width = 1.6cm]{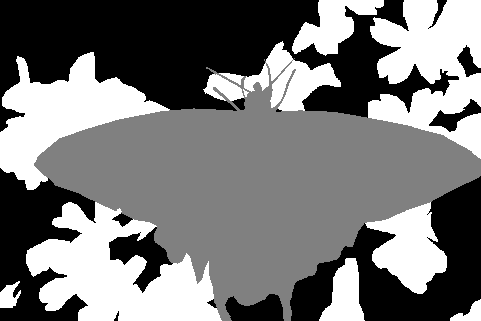}} & 
 \cellcolor{gray!25} \includegraphics[width = 1.6cm]{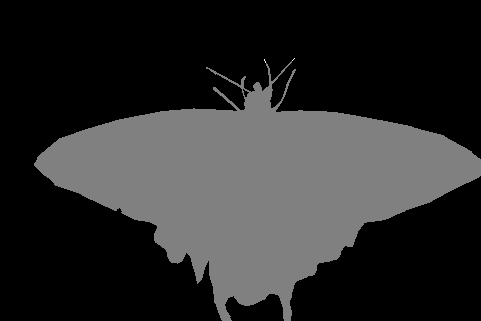} 
 \includegraphics[width = 1.6cm]{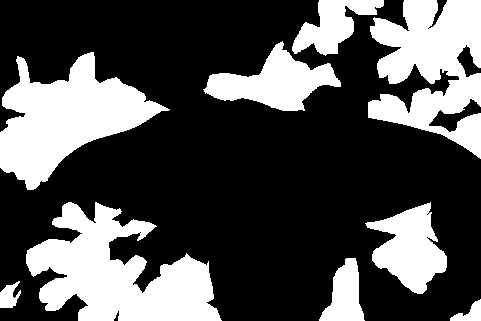} & &
 \cellcolor{gray!25} \includegraphics[width = 1.6cm]{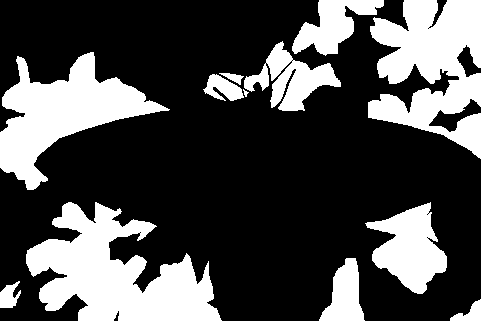} 
 \includegraphics[width = 1.6cm]{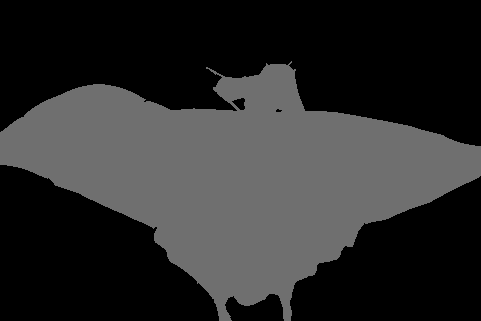} & &
 \cellcolor{gray!25} \includegraphics[width = 1.6cm]{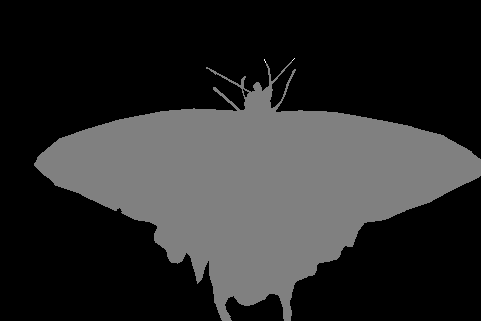} 
 \includegraphics[width = 1.6cm]{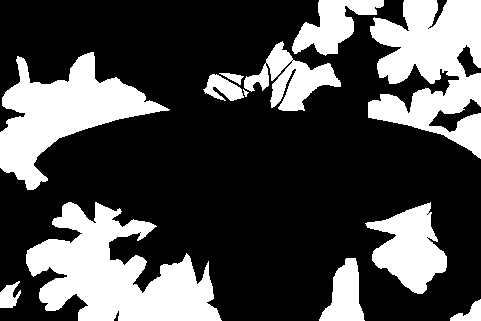} \\
 & & \cellcolor{gray!25}$\tilde{p}(I/H_1)=0.8665$ & & \cellcolor{gray!25}$\tilde{p}(I/H_2)=0.6820$ & & \cellcolor{gray!25}$\tilde{p}(I/H_3)=0.3679$\\
 \multicolumn{2}{c}{\scriptsize{Image 1}} & \cellcolor{gray!25}$\tilde{p}(H_1)=0.5212$ & & \cellcolor{gray!25}$\tilde{p}(H_2)=0.3701$ & & \cellcolor{gray!25}$\tilde{p}(H_3)=0.3679$\\
 & & \cellcolor{gray!25}$\mathbf{p(H_1/I)=0.5380}$ & & \cellcolor{gray!25}$p(H_2/I)=0.3007$ & & \cellcolor{gray!25}$p(H_3/I)=0.1612$\\ \\
 {\includegraphics[width = 1.6cm]{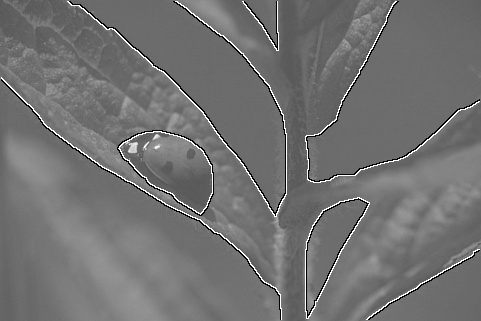}} & 
 {\includegraphics[width = 1.6cm]{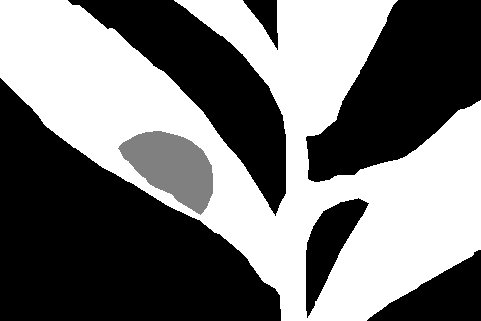}} & 
 \cellcolor{gray!25} \includegraphics[width = 1.6cm]{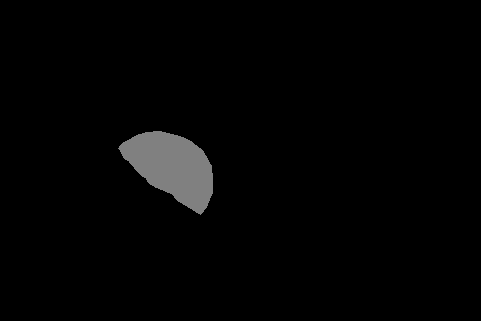} 
 \includegraphics[width = 1.6cm]{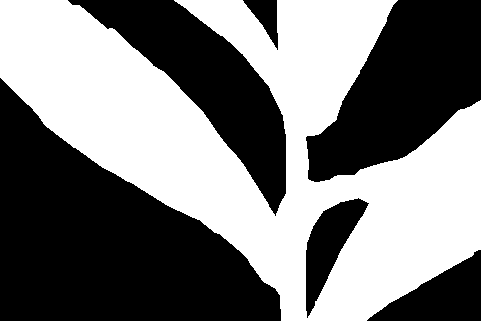} & &
 \cellcolor{gray!25} \includegraphics[width = 1.6cm]{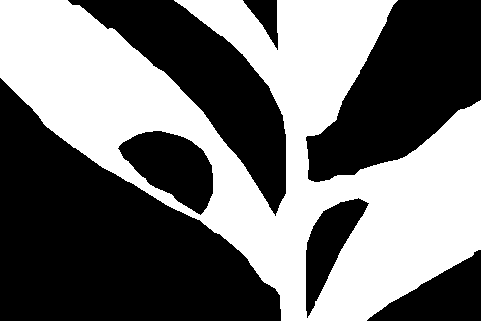} 
 \includegraphics[width = 1.6cm]{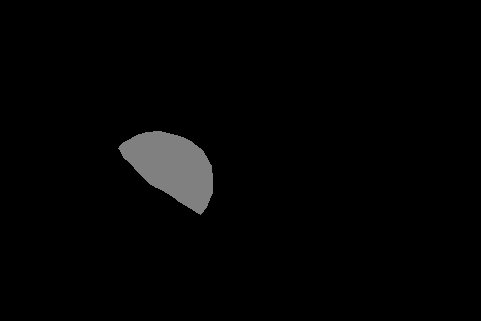} & &
 \cellcolor{gray!25} \includegraphics[width = 1.6cm]{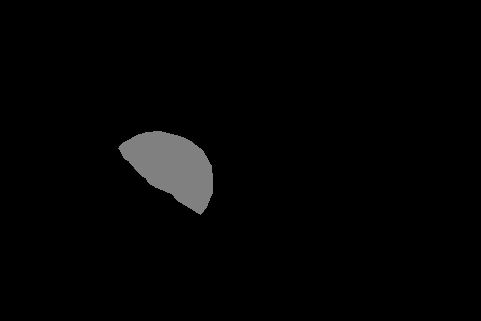} 
 \includegraphics[width = 1.6cm]{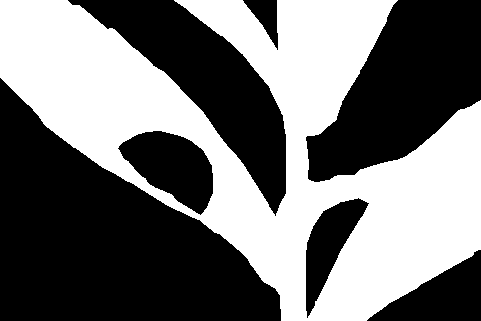} \\
 & & \cellcolor{gray!25}$\tilde{p}(I/H_1)=1$ & & \cellcolor{gray!25}$\tilde{p}(I/H_2)=0.8952$ & & \cellcolor{gray!25}$\tilde{p}(I/H_3)=0.3679$\\
 \multicolumn{2}{c}{\scriptsize{Image 2}} & \cellcolor{gray!25}$\tilde{p}(H_1)=0.3688$ & & \cellcolor{gray!25}$\tilde{p}(H_2)=0.3724$ & & \cellcolor{gray!25}$\tilde{p}(H_3)=0.3692$\\
 & & \cellcolor{gray!25}$\mathbf{p(H_1/I)=0.4410}$ & & \cellcolor{gray!25}$p(H_2/I)=0.3972$ & & \cellcolor{gray!25}$p(H_3/I)=0.1618$\\ \\
 {\includegraphics[width = 1.6cm]{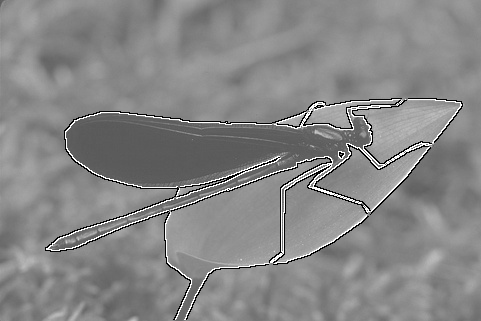}} & 
 {\includegraphics[width = 1.6cm]{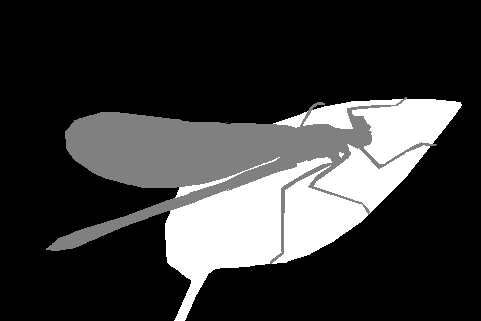}} & 
 \cellcolor{gray!25} \includegraphics[width = 1.6cm]{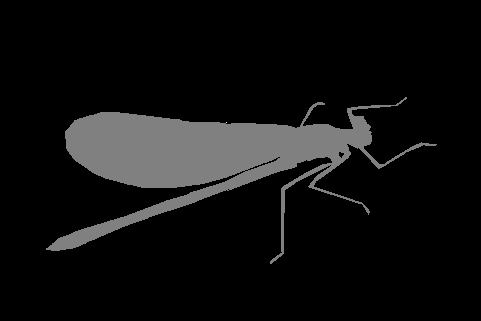} 
 \includegraphics[width = 1.6cm]{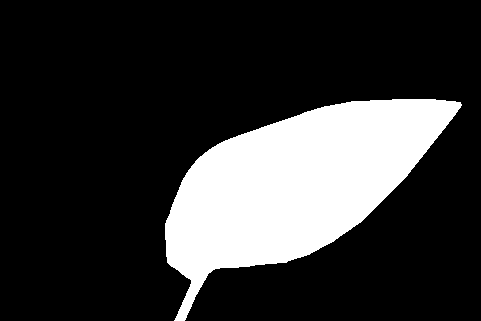} & &
 \cellcolor{gray!25} \includegraphics[width = 1.6cm]{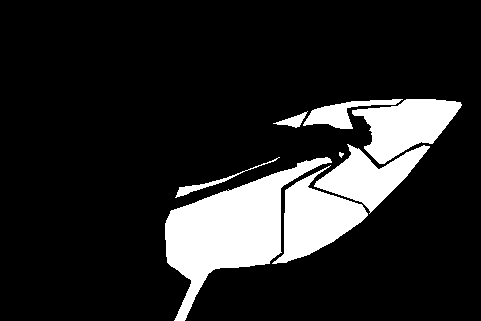} 
 \includegraphics[width = 1.6cm]{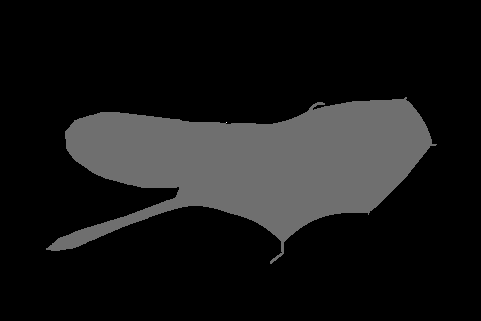} & &
 \cellcolor{gray!25} \includegraphics[width = 1.6cm]{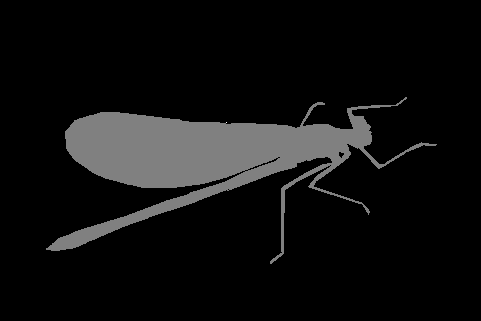} 
 \includegraphics[width = 1.6cm]{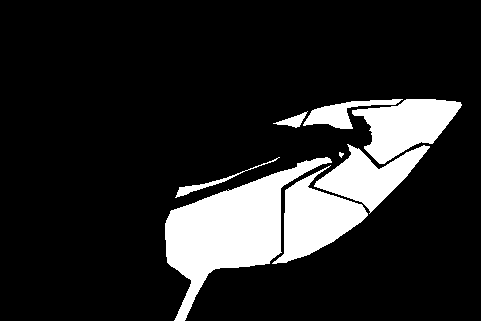} \\
 & & \cellcolor{gray!25}$\tilde{p}(I/H_1)=0.9520$ & & \cellcolor{gray!25}$\tilde{p}(I/H_2)=0.7729$ & & \cellcolor{gray!25}$\tilde{p}(I/H_3)=0.3679$\\
 \multicolumn{2}{c}{\scriptsize{Image 3}} & \cellcolor{gray!25}$\tilde{p}(H_1)=0.3828$ & & \cellcolor{gray!25}$\tilde{p}(H_2)=0.3864$ & & \cellcolor{gray!25}$\tilde{p}(H_3)=0.3679$\\
 & & \cellcolor{gray!25}$\mathbf{p(H_1/I)=0.4564}$ & & \cellcolor{gray!25}$p(H_2/I)=0.3741$ & & \cellcolor{gray!25}$p(H_3/I)=0.1695$\\ \\
 {\includegraphics[width = 1.6cm]{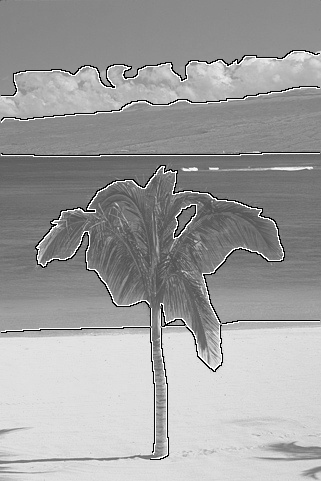}} & 
 {\includegraphics[width = 1.6cm]{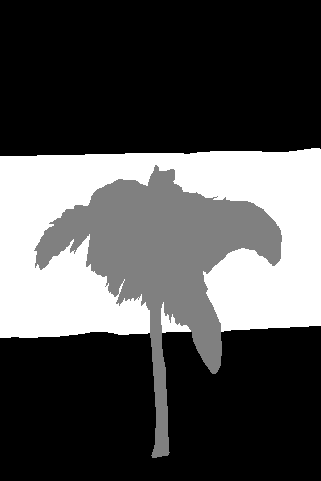}} & 
 \cellcolor{gray!25} \includegraphics[width = 1.6cm]{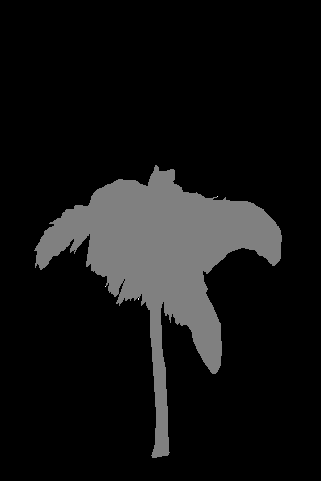} 
 \includegraphics[width = 1.6cm]{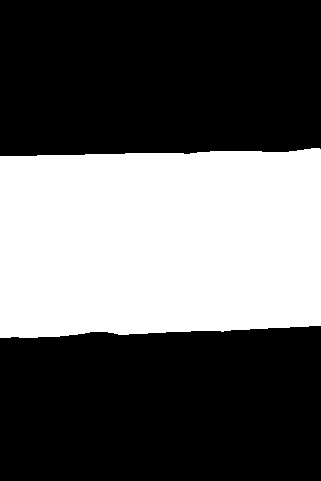} & &
 \cellcolor{gray!25} \includegraphics[width = 1.6cm]{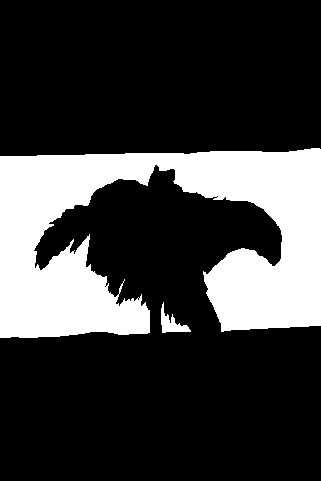} 
 \includegraphics[width = 1.6cm]{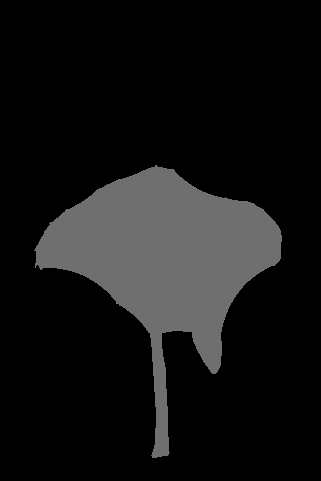} & &
 \cellcolor{gray!25} \includegraphics[width = 1.6cm]{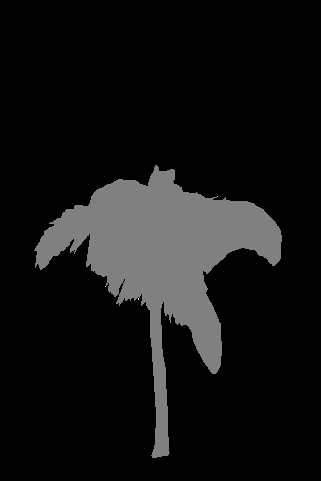} 
 \includegraphics[width = 1.6cm]{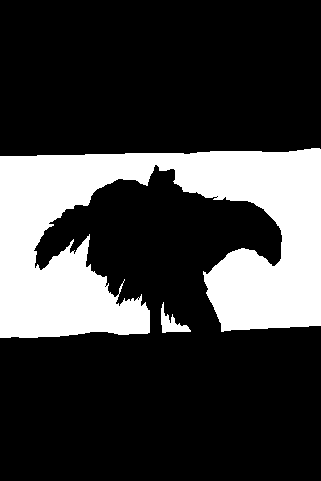} \\
 & & \cellcolor{gray!25}$\tilde{p}(I/H_1)=0.9899$ & & \cellcolor{gray!25}$\tilde{p}(I/H_2)=0.7811$ & & \cellcolor{gray!25}$\tilde{p}(I/H_3)=0.3679$\\
 \multicolumn{2}{c}{\scriptsize{Image 4}} & \cellcolor{gray!25}$\tilde{p}(H_1)=0.4554$ & & \cellcolor{gray!25}$\tilde{p}(H_2)=0.4124$ & & \cellcolor{gray!25}$\tilde{p}(H_3)=0.3924$\\
 & & \cellcolor{gray!25}$\mathbf{p(H_1/I)=0.5064}$ & & \cellcolor{gray!25}$p(H_2/I)=0.3416$ & & \cellcolor{gray!25}$p(H_3/I)=0.1520$\\ \\
 {\includegraphics[width = 1.6cm]{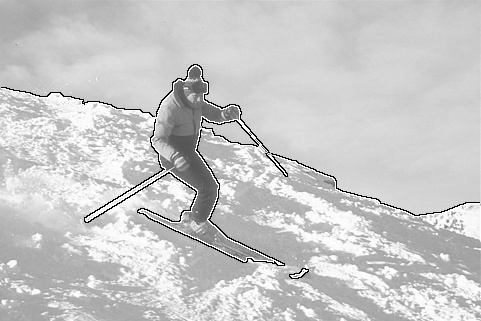}} & 
 {\includegraphics[width = 1.6cm]{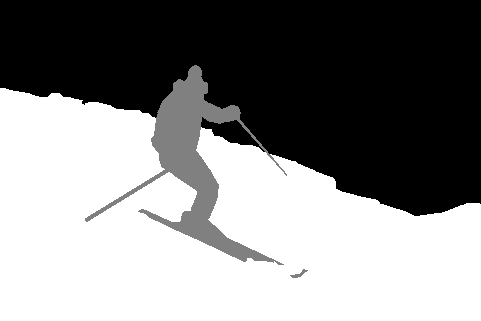}} & 
 \cellcolor{gray!25} \includegraphics[width = 1.6cm]{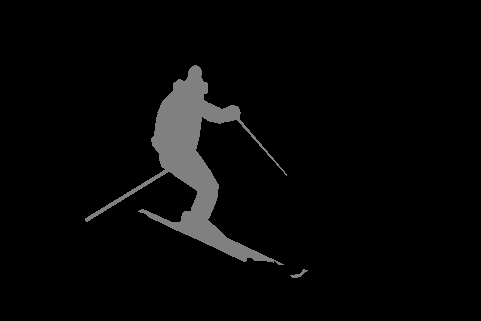} 
 \includegraphics[width = 1.6cm]{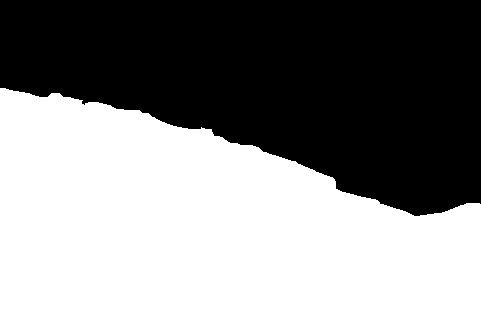} & &
 \cellcolor{gray!25} \includegraphics[width = 1.6cm]{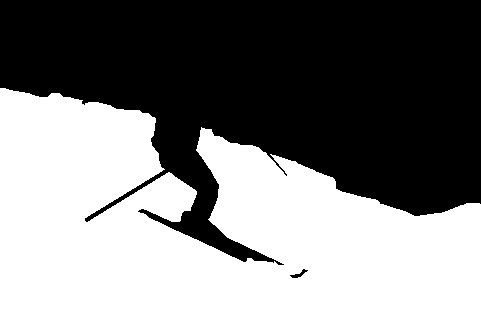} 
 \includegraphics[width = 1.6cm]{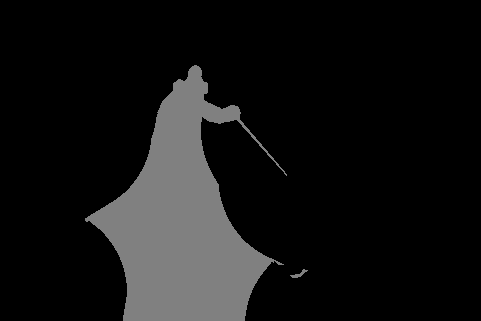} & &
 \cellcolor{gray!25} \includegraphics[width = 1.6cm]{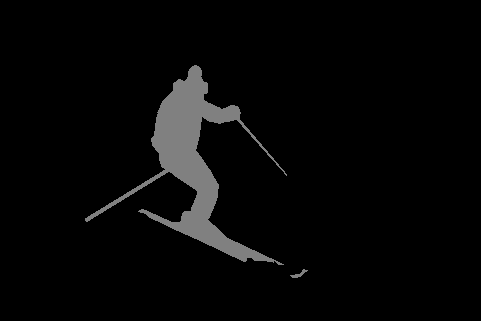} 
 \includegraphics[width = 1.6cm]{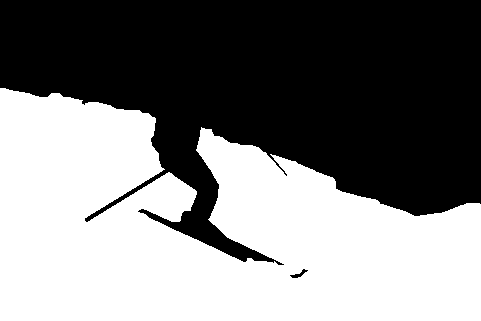} \\
 & & \cellcolor{gray!25}$\tilde{p}(I/H_1)=0.9665$ & & \cellcolor{gray!25}$\tilde{p}(I/H_2)=0.7964$ & & \cellcolor{gray!25}$\tilde{p}(I/H_3)=0.3679$\\
 \multicolumn{2}{c}{\scriptsize{Image 5}} & \cellcolor{gray!25}$\tilde{p}(H_1)=0.5375$ & & \cellcolor{gray!25}$\tilde{p}(H_2)=0.3679$ & & \cellcolor{gray!25}$\tilde{p}(H_3)=0.3679$\\
 & & \cellcolor{gray!25}$\mathbf{p(H_1/I)=0.5479}$ & & \cellcolor{gray!25}$p(H_2/I)=0.3090$ & & \cellcolor{gray!25}$p(H_3/I)=0.1432$\\ \\
 {\includegraphics[width = 1.6cm]{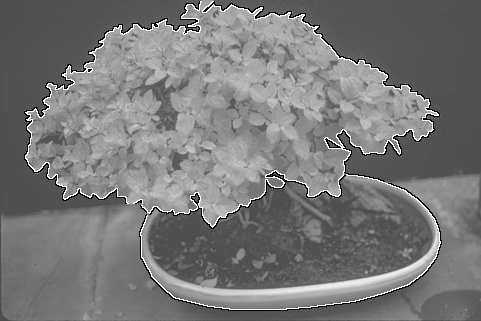}} & 
 {\includegraphics[width = 1.6cm]{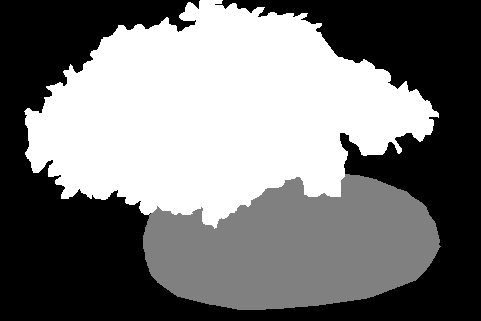}} & 
 \cellcolor{gray!25} \includegraphics[width = 1.6cm]{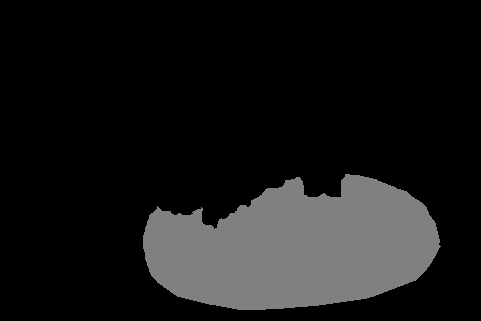} 
 \includegraphics[width = 1.6cm]{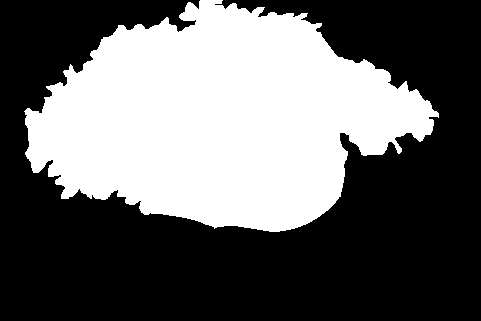} & &
 \cellcolor{gray!25} \includegraphics[width = 1.6cm]{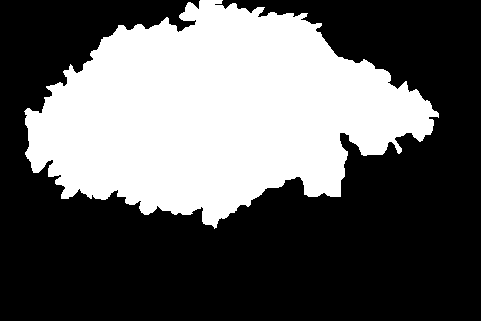} 
 \includegraphics[width = 1.6cm]{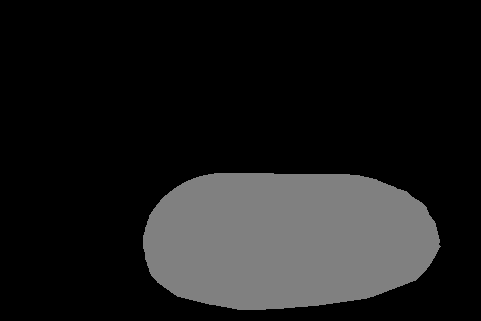} & &
 \cellcolor{gray!25} \includegraphics[width = 1.6cm]{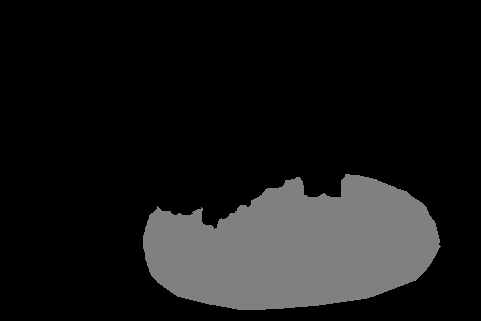} 
 \includegraphics[width = 1.6cm]{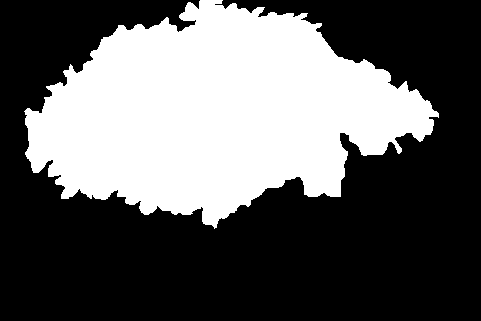} \\
 & & \cellcolor{gray!25}$\tilde{p}(I/H_1)=0.7035$ & & \cellcolor{gray!25}$\tilde{p}(I/H_2)=0.8029$ & & \cellcolor{gray!25}$\tilde{p}(I/H_3)=0.3679$\\
 \multicolumn{2}{c}{\scriptsize{Image 6}} & \cellcolor{gray!25}$\tilde{p}(H_1)=0.3696$ & & \cellcolor{gray!25}$\tilde{p}(H_2)=0.3882$ & & \cellcolor{gray!25}$\tilde{p}(H_3)=0.3679$\\
 & & \cellcolor{gray!25}$p(H_1/I)=0.3677$ & & \cellcolor{gray!25}$\mathbf{p(H_2/I)=0.4409}$ & & \cellcolor{gray!25}$p(H_3/I)=0.1914$\\ \\
 {\includegraphics[width = 1.6cm]{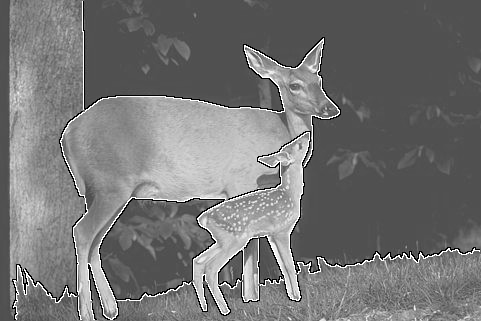}} & 
 {\includegraphics[width = 1.6cm]{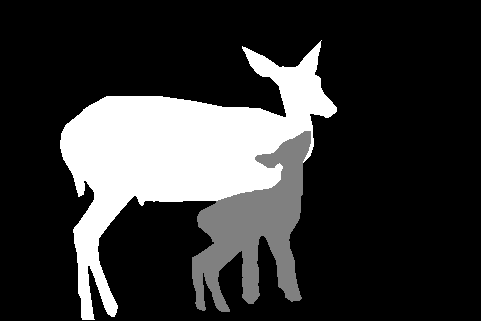}} & 
 \cellcolor{gray!25} \includegraphics[width = 1.6cm]{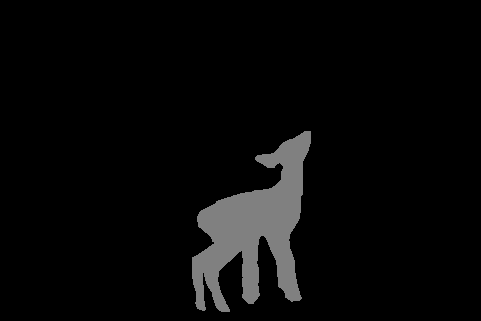} 
 \includegraphics[width = 1.6cm]{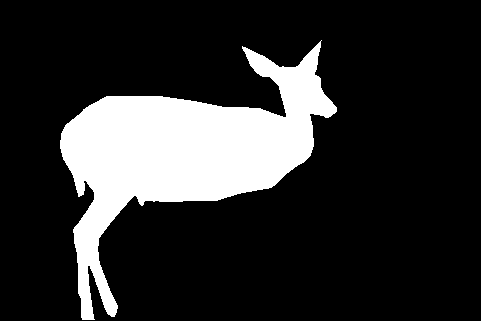} & &
 \cellcolor{gray!25} \includegraphics[width = 1.6cm]{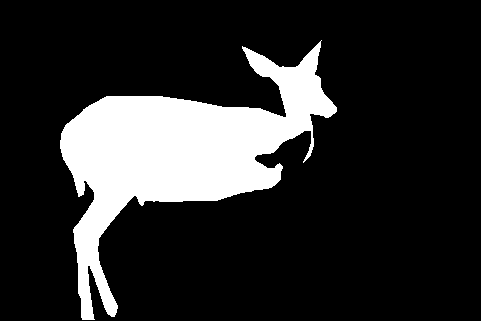} 
 \includegraphics[width = 1.6cm]{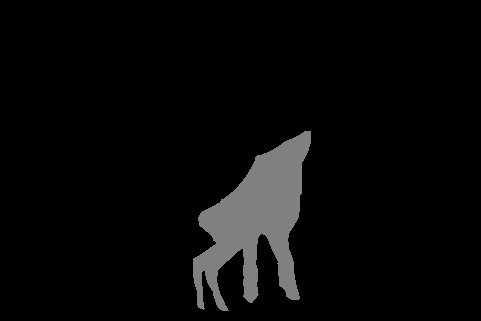} & &
 \cellcolor{gray!25} \includegraphics[width = 1.6cm]{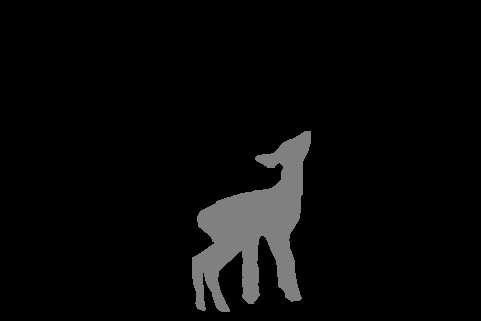} 
 \includegraphics[width = 1.6cm]{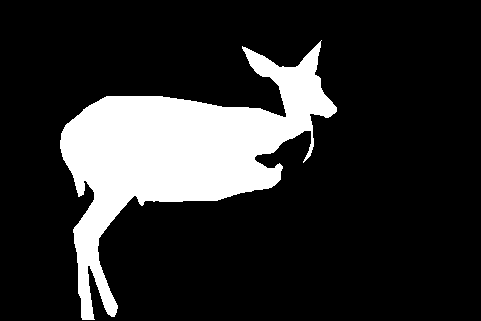} \\
 & & \cellcolor{gray!25}$\tilde{p}(I/H_1)=0.9112$ & & \cellcolor{gray!25}$\tilde{p}(I/H_2)=0.8257$ & & \cellcolor{gray!25}$\tilde{p}(I/H_3)=0.3679$\\
 \multicolumn{2}{c}{\scriptsize{Image 7}} & \cellcolor{gray!25}$\tilde{p}(H_1)=0.3727$ & & \cellcolor{gray!25}$\tilde{p}(H_2)=0.3679$ & & \cellcolor{gray!25}$\tilde{p}(H_3)=0.3706$\\
 & & \cellcolor{gray!25}$\mathbf{p(H_1/I)=0.4329}$ & & \cellcolor{gray!25}$p(H_2/I)=0.3914$ & & \cellcolor{gray!25}$p(H_3/I)=0.1757$\\ \\
 {\includegraphics[width = 1.6cm]{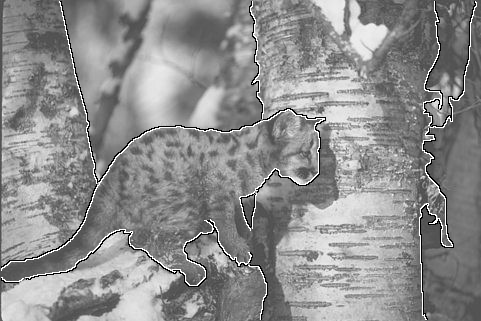}} & 
 {\includegraphics[width = 1.6cm]{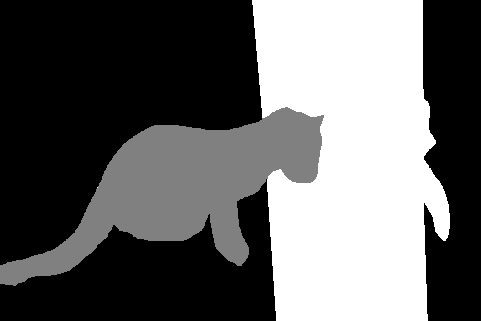}} & 
 \cellcolor{gray!25} \includegraphics[width = 1.6cm]{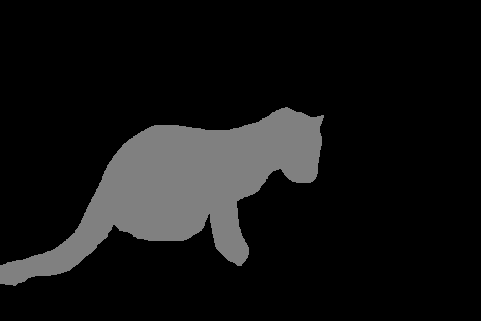} 
 \includegraphics[width = 1.6cm]{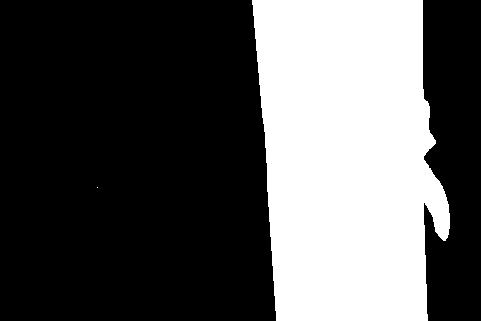} & &
 \cellcolor{gray!25} \includegraphics[width = 1.6cm]{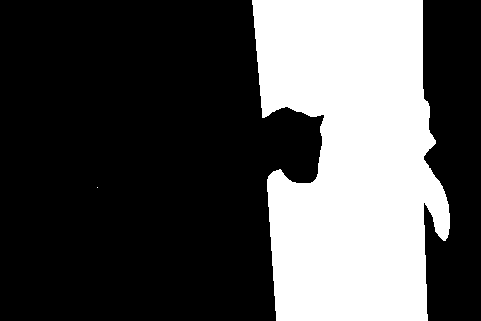} 
 \includegraphics[width = 1.6cm]{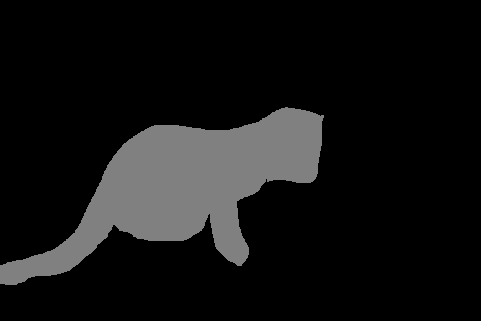} & &
 \cellcolor{gray!25} \includegraphics[width = 1.6cm]{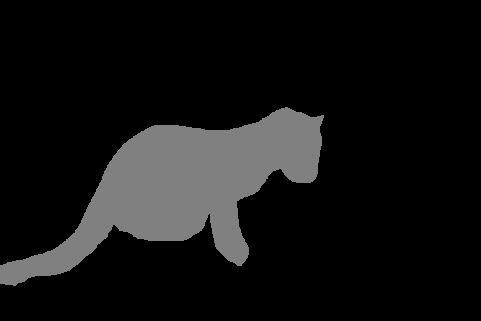} 
 \includegraphics[width = 1.6cm]{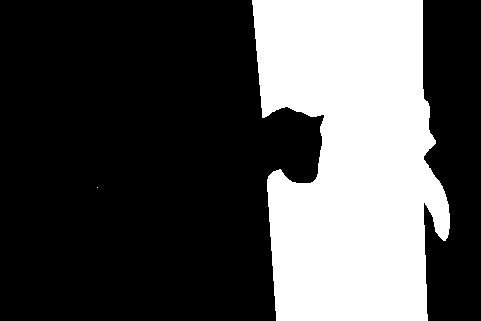} \\
 & & \cellcolor{gray!25}$\tilde{p}(I/H_1)=0.8831$ & & \cellcolor{gray!25}$\tilde{p}(I/H_2)=0.7771$ & & \cellcolor{gray!25}$\tilde{p}(I/H_3)=0.3679$\\
 \multicolumn{2}{c}{\scriptsize{Image 8}} & \cellcolor{gray!25}$\tilde{p}(H_1)=0.3821$ & & \cellcolor{gray!25}$\tilde{p}(H_2)=0.4018$ & & \cellcolor{gray!25}$\tilde{p}(H_3)=0.4001$\\
 & & \cellcolor{gray!25}$\mathbf{p(H_1/I)=0.4433}$ & & \cellcolor{gray!25}$p(H_2/I)=0.3789$ & & \cellcolor{gray!25}$p(H_3/I)=0.1778$\\ \\
 {\includegraphics[width = 1.6cm]{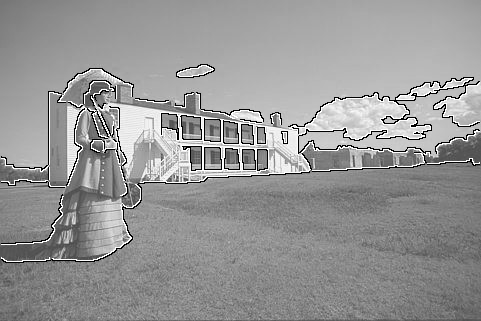}} & 
 {\includegraphics[width = 1.6cm]{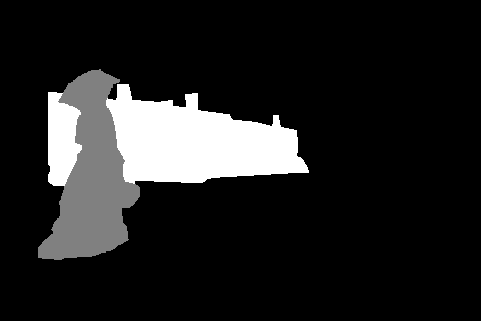}} & 
 \cellcolor{gray!25} \includegraphics[width = 1.6cm]{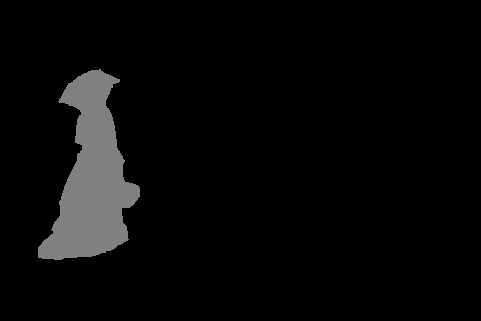} 
 \includegraphics[width = 1.6cm]{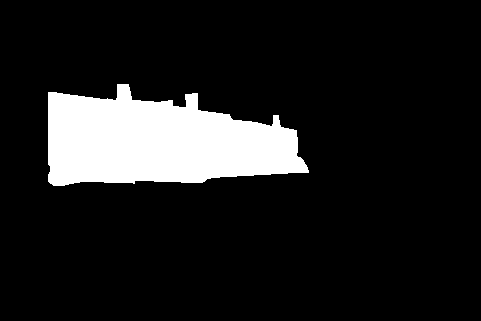} & &
 \cellcolor{gray!25} \includegraphics[width = 1.6cm]{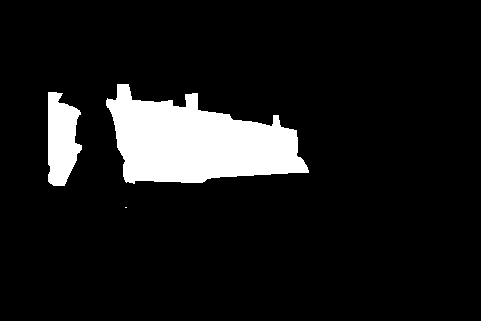} 
 \includegraphics[width = 1.6cm]{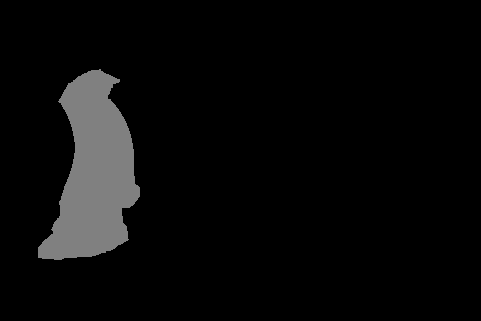} & &
 \cellcolor{gray!25} \includegraphics[width = 1.6cm]{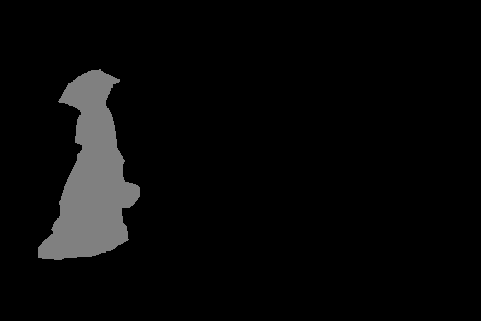} 
 \includegraphics[width = 1.6cm]{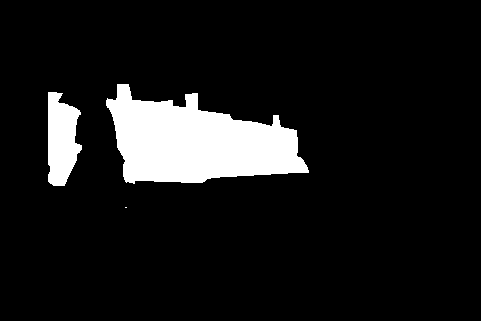} \\
 & & \cellcolor{gray!25}$\tilde{p}(I/H_1)=0.8338$ & & \cellcolor{gray!25}$\tilde{p}(I/H_2)=0.6028$ & & \cellcolor{gray!25}$\tilde{p}(I/H_3)=0.3679$\\
 \multicolumn{2}{c}{\scriptsize{Image 9}} & \cellcolor{gray!25}$\tilde{p}(H_1)=0.5462$ & & \cellcolor{gray!25}$\tilde{p}(H_2)=0.3700$ & & \cellcolor{gray!25}$\tilde{p}(H_3)=0.3679$\\
 & & \cellcolor{gray!25}$\mathbf{p(H_1/I)=0.5597}$ & & \cellcolor{gray!25}$p(H_2/I)=0.2740$ & & \cellcolor{gray!25}$p(H_3/I)=0.1663$\\ \\
 {\includegraphics[width = 1.6cm]{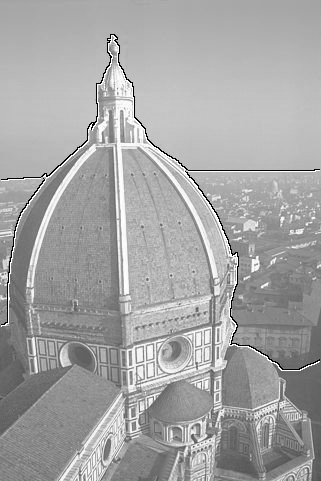}} & 
 {\includegraphics[width = 1.6cm]{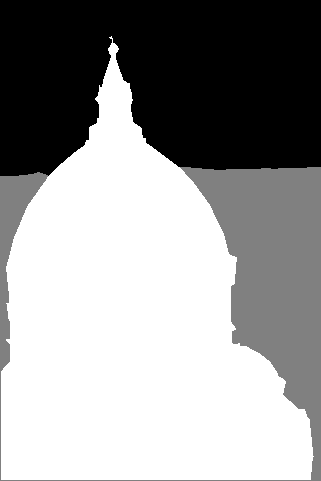}} & 
 \cellcolor{gray!25} \includegraphics[width = 1.6cm]{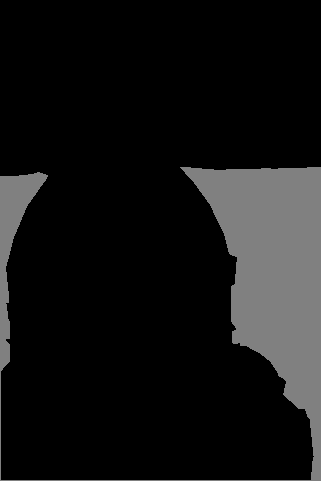} 
 \includegraphics[width = 1.6cm]{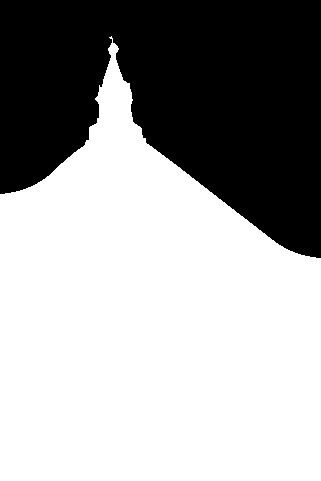} & &
 \cellcolor{gray!25} \includegraphics[width = 1.6cm]{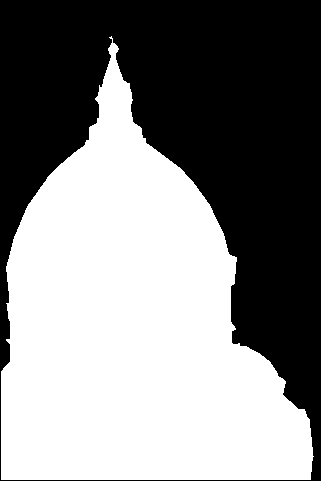} 
 \includegraphics[width = 1.6cm]{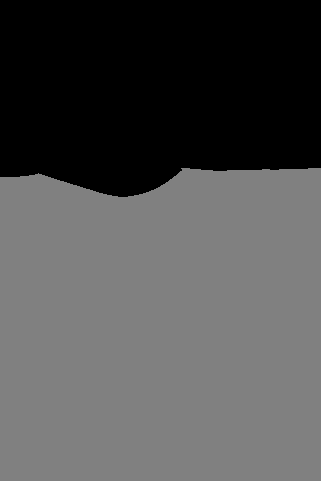} & &
 \cellcolor{gray!25} \includegraphics[width = 1.6cm]{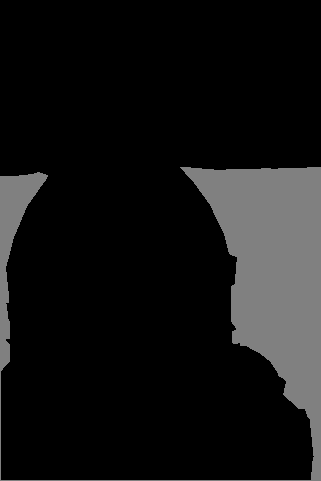} 
 \includegraphics[width = 1.6cm]{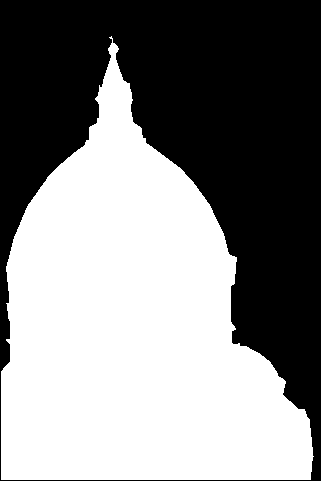} \\
 & & \cellcolor{gray!25}$\tilde{p}(I/H_1)=0.8935$ & & \cellcolor{gray!25}$\tilde{p}(I/H_2)=0.9504$ & & \cellcolor{gray!25}$\tilde{p}(I/H_3)=0.3679$\\
 \multicolumn{2}{c}{\scriptsize{Image 10}} & \cellcolor{gray!25}$\tilde{p}(H_1)=0.3679$ & & \cellcolor{gray!25}$\tilde{p}(H_2)=0.5218$ & & \cellcolor{gray!25}$\tilde{p}(H_3)=0.5219$\\
 & & \cellcolor{gray!25}$p(H_1/I)=0.3390$ & & \cellcolor{gray!25}$\mathbf{p(H_2/I)=0.5115}$ & & \cellcolor{gray!25}$p(H_3/I)=0.1495$\\ \\
 {\includegraphics[width = 1.6cm]{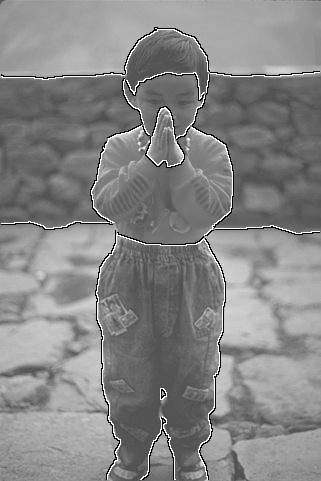}} & 
 {\includegraphics[width = 1.6cm]{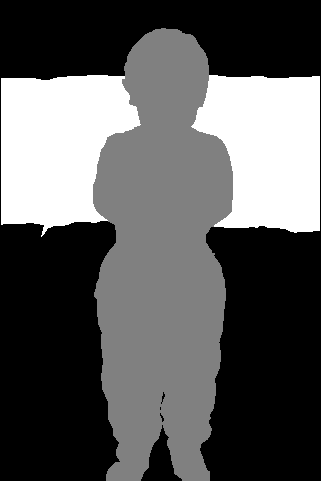}} & 
 \cellcolor{gray!25} \includegraphics[width = 1.6cm]{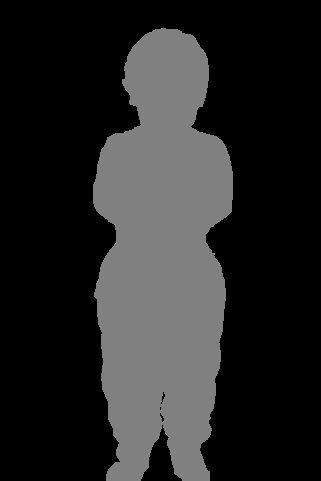} 
 \includegraphics[width = 1.6cm]{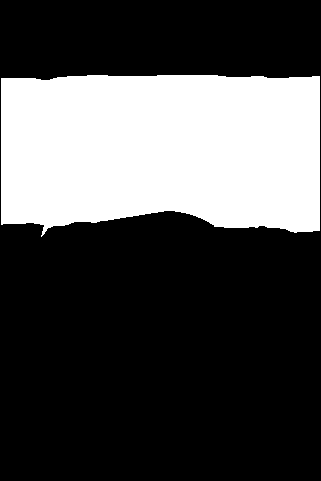} & &
 \cellcolor{gray!25} \includegraphics[width = 1.6cm]{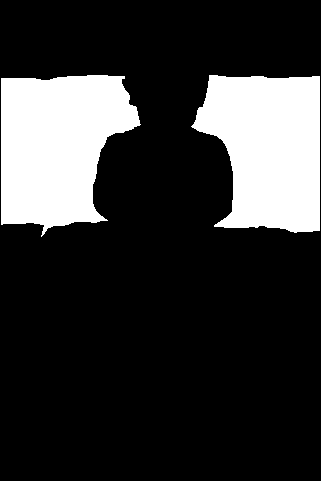} 
 \includegraphics[width = 1.6cm]{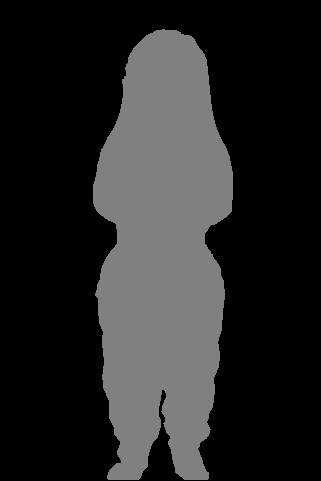} & &
 \cellcolor{gray!25} \includegraphics[width = 1.6cm]{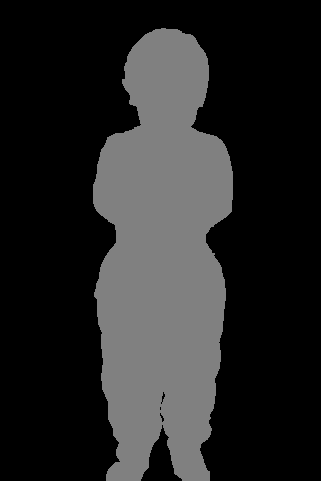} 
 \includegraphics[width = 1.6cm]{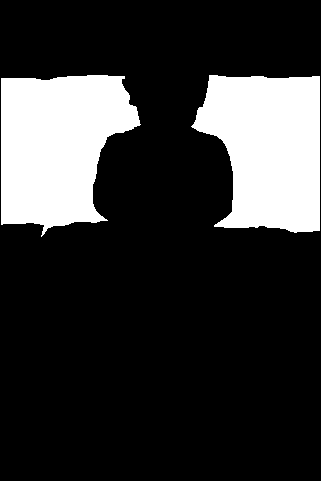} \\
 & & \cellcolor{gray!25}$\tilde{p}(I/H_1)=0.8059$ & & \cellcolor{gray!25}$\tilde{p}(I/H_2)=0.8029$ & & \cellcolor{gray!25}$\tilde{p}(I/H_3)=0.3679$\\
 \multicolumn{2}{c}{\scriptsize{Image 11}} & \cellcolor{gray!25}$\tilde{p}(H_1)=0.5367$ & & \cellcolor{gray!25}$\tilde{p}(H_2)=0.3696$ & & \cellcolor{gray!25}$\tilde{p}(H_3)=0.3679$\\
 & & \cellcolor{gray!25}$\mathbf{p(H_1/I)=0.5002}$ & & \cellcolor{gray!25}$p(H_2/I)=0.3432$ & & \cellcolor{gray!25}$p(H_3/I)=0.1565$\\ \\
 {\includegraphics[width = 1.6cm]{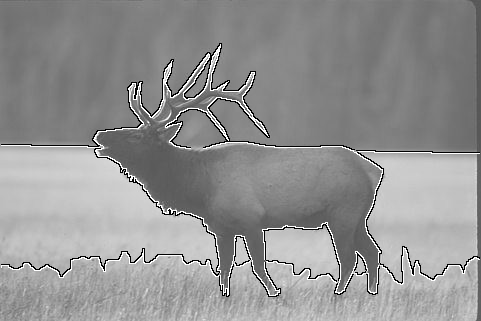}} & 
 {\includegraphics[width = 1.6cm]{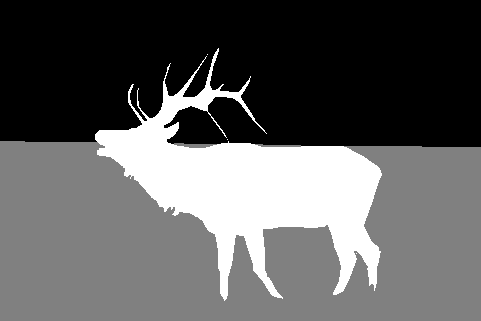}} & 
 \cellcolor{gray!25} \includegraphics[width = 1.6cm]{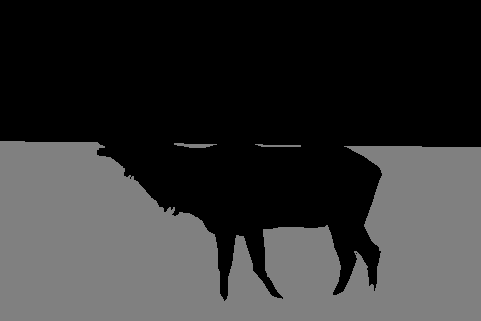} 
 \includegraphics[width = 1.6cm]{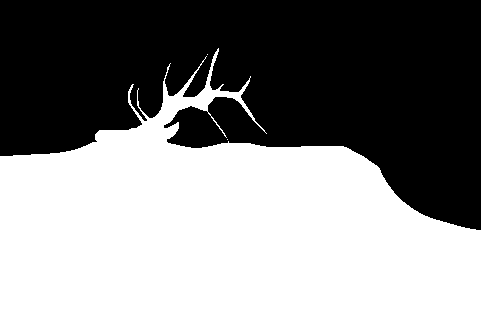} & &
 \cellcolor{gray!25} \includegraphics[width = 1.6cm]{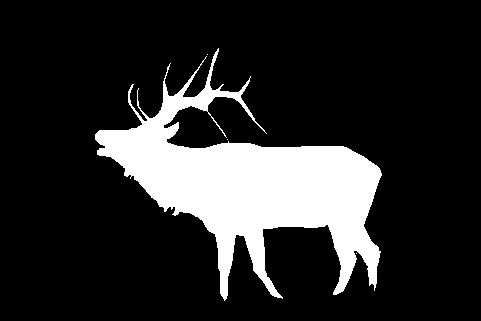} 
 \includegraphics[width = 1.6cm]{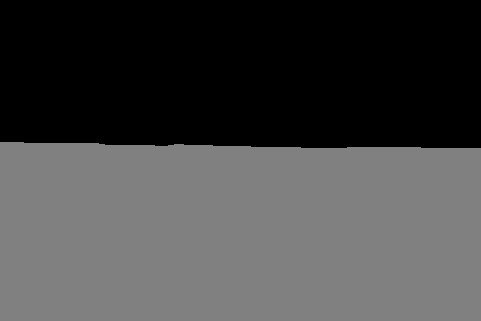} & &
 \cellcolor{gray!25} \includegraphics[width = 1.6cm]{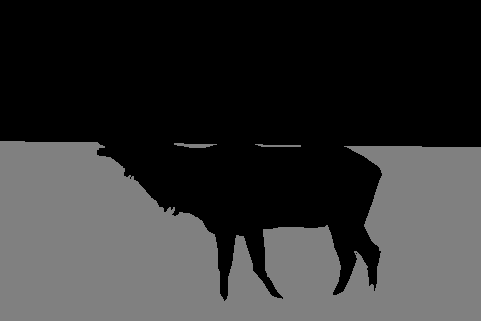} 
 \includegraphics[width = 1.6cm]{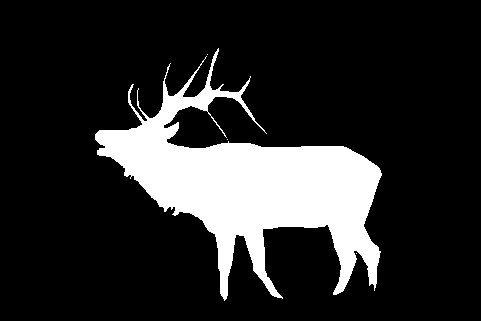} \\
 & & \cellcolor{gray!25}$\tilde{p}(I/H_1)=0.8940$ & & \cellcolor{gray!25}$\tilde{p}(I/H_2)=0.9192$ & & \cellcolor{gray!25}$\tilde{p}(I/H_3)=0.3679$\\
 \multicolumn{2}{c}{\scriptsize{Image 12}} & \cellcolor{gray!25}$\tilde{p}(H_1)=0.3764$ & & \cellcolor{gray!25}$\tilde{p}(H_2)=0.4979$ & & \cellcolor{gray!25}$\tilde{p}(H_3)=0.3679$\\
 & & \cellcolor{gray!25}$p(H_1/I)=0.3620$ & & \cellcolor{gray!25}$\mathbf{p(H_2/I)=0.4924}$ & & \cellcolor{gray!25}$p(H_3/I)=0.1456$\\ \\
 {\includegraphics[width = 1.6cm]{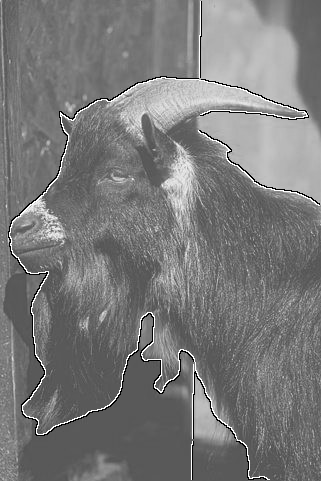}} & 
 {\includegraphics[width = 1.6cm]{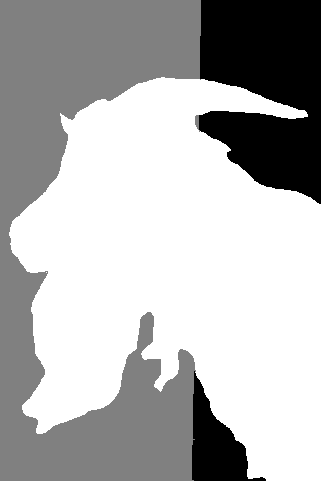}} & 
 \cellcolor{gray!25} \includegraphics[width = 1.6cm]{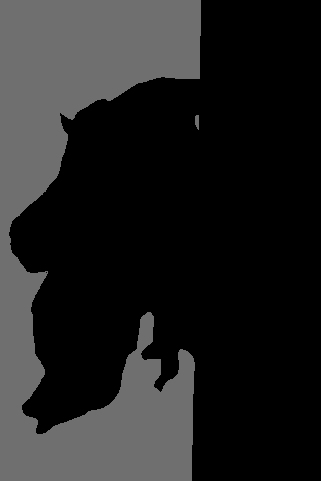} 
 \includegraphics[width = 1.6cm]{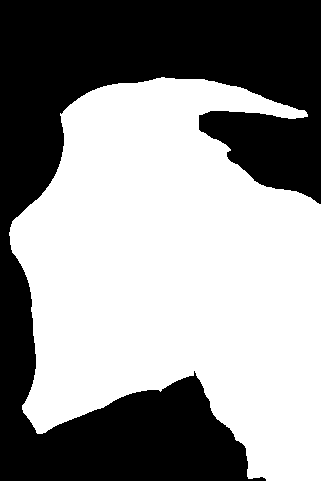} & &
 \cellcolor{gray!25} \includegraphics[width = 1.6cm]{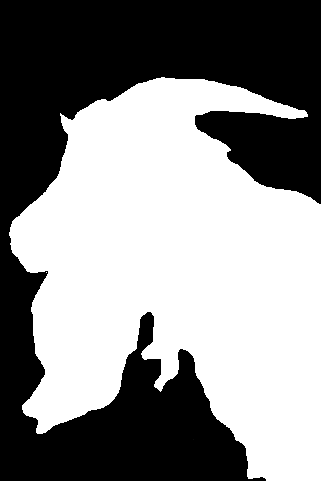} 
 \includegraphics[width = 1.6cm]{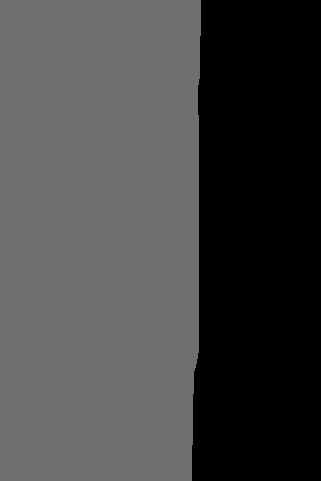} & &
 \cellcolor{gray!25} \includegraphics[width = 1.6cm]{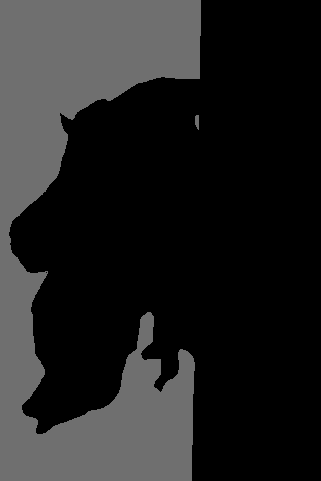} 
 \includegraphics[width = 1.6cm]{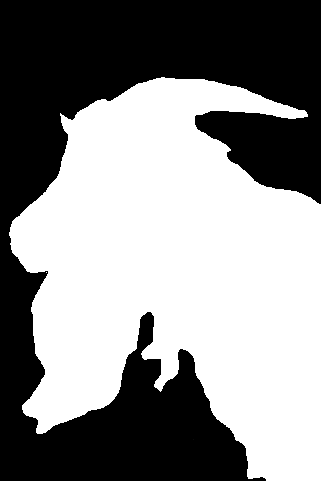} \\
 & & \cellcolor{gray!25}$\tilde{p}(I/H_1)=0.7809$ & & \cellcolor{gray!25}$\tilde{p}(I/H_2)=0.9201$ & & \cellcolor{gray!25}$\tilde{p}(I/H_3)=0.3679$\\
 \multicolumn{2}{c}{\scriptsize{Image 13}} & \cellcolor{gray!25}$\tilde{p}(H_1)=0.3712$ & & \cellcolor{gray!25}$\tilde{p}(H_2)=0.4947$ & & \cellcolor{gray!25}$\tilde{p}(H_3)=0.3679$\\
 & & \cellcolor{gray!25}$p(H_1/I)=0.3293$ & & \cellcolor{gray!25}$\mathbf{p(H_2/I)=0.5170
}$ & & \cellcolor{gray!25}$p(H_3/I)=0.1537$\\ \\
\caption{Experiments with real images from \cite{MartinFTM01}. Each row shows a different experiment: 
the two left-most images are, respectively, the original image and a segmentation of it. They are followed 
by the three different hypothesis (each one separated by a gray box). 
The lines superimposed in the original image indicate the figure/ground ground-truth labels (from \cite{figuregroundBerkeley})
for each boundary in the segmentation: the white boundary indicates the figure
side and the black one  the ground side. For the first two hypothesis, $H_i$ for $i=1,2$, we show: the object at depth 1 (left) and the disoccluded object at depth 2 (right). Notice that the object at depth 1 acts as a mask for disoccluding the object at depth 2. In the case of the third hypothesis, $H_3$, both objects are considered to be at the same depth and completely visible in the original image (no disocclusion is applied). 
  In the lower part of each hypothesis we show the values $\tilde{p}(I/H_i)$, $\tilde{p}(H_2)$ (proportional, respectively, to the likelihood and prior probabilities), and the posterior probability $p(H_i/I)$. The probability value of the preferred hypothesis $H_P$ is highlighted in boldface.}
\label{tab:alltogether7}
\end{longtable}

\setlength{\tabcolsep}{3pt}
\begin{longtable}{lllllll}
  {\includegraphics[width = 1.6cm]{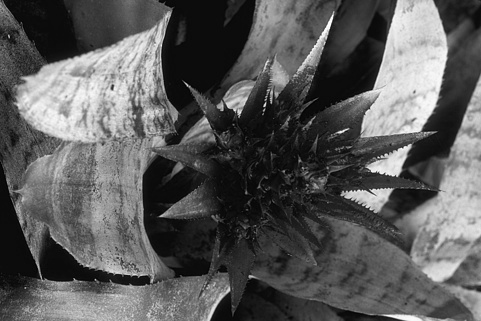}} & 
 {\includegraphics[width = 1.6cm]{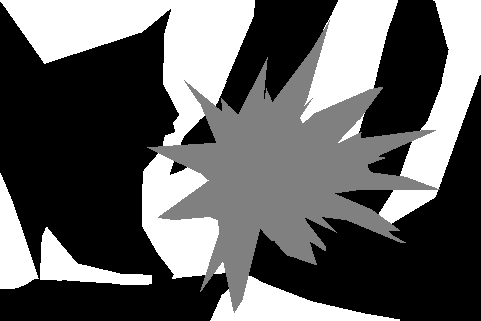}} & 
 \cellcolor{gray!25} \includegraphics[width = 1.6cm]{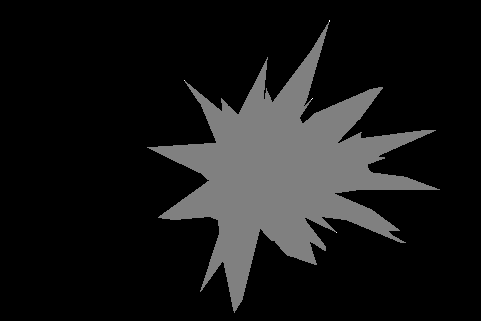} 
 \includegraphics[width = 1.6cm]{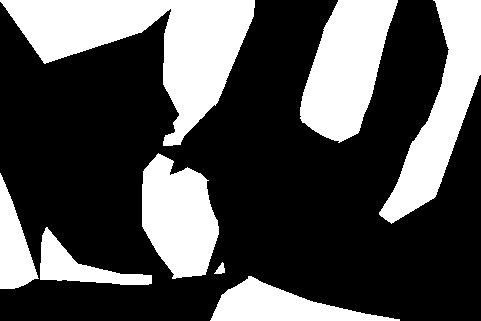} & &
 \cellcolor{gray!25} \includegraphics[width = 1.6cm]{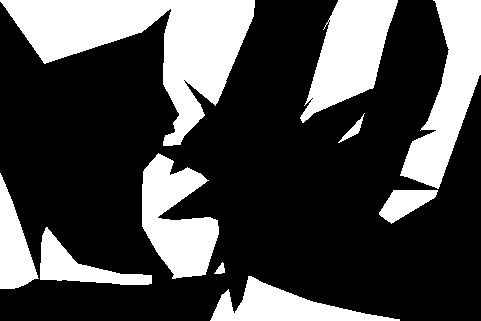} 
 \includegraphics[width = 1.6cm]{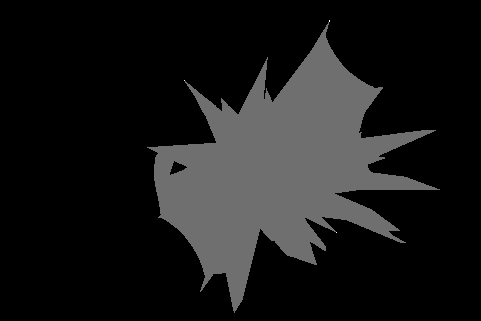} & &
 \cellcolor{gray!25} \includegraphics[width = 1.6cm]{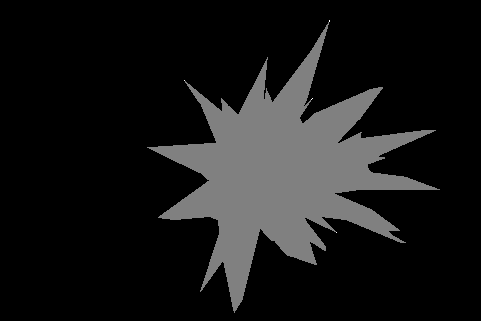} 
 \includegraphics[width = 1.6cm]{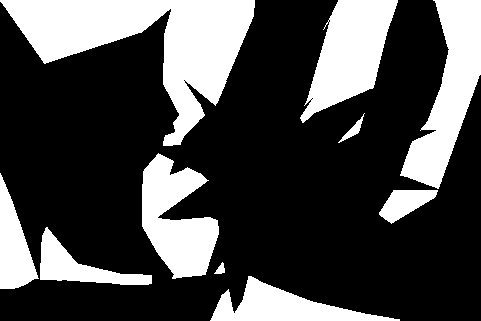} \\
 & & \cellcolor{gray!25}$\tilde{p}(I/H_1)=0.8526$ & & \cellcolor{gray!25}$\tilde{p}(I/H_2)=0.7416$ & & \cellcolor{gray!25}$\tilde{p}(I/H_3)=0.3679$\\
 \multicolumn{2}{c}{\scriptsize{Image 14}} & \cellcolor{gray!25}$\tilde{p}(H_1)=0.3701$ & & \cellcolor{gray!25}$\tilde{p}(H_2)=0.3688$ & & \cellcolor{gray!25}$\tilde{p}(H_3)=0.4472$\\
 & & \cellcolor{gray!25}$\mathbf{p(H_1/I)=0.4356}$ & & \cellcolor{gray!25}$p(H_2/I)=0.3776$ & & \cellcolor{gray!25}$p(H_3/I)=0.1868$\\ \\
  {\includegraphics[width = 1.6cm]{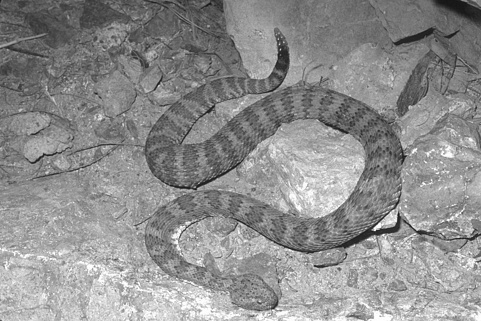}} & 
 {\includegraphics[width = 1.6cm]{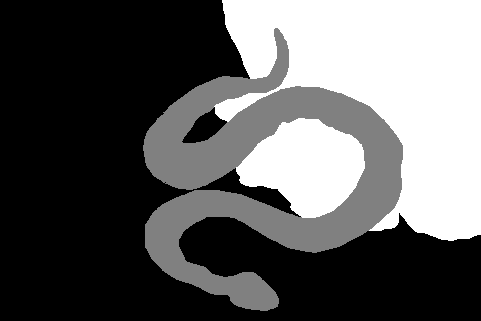}} & 
 \cellcolor{gray!25} \includegraphics[width = 1.6cm]{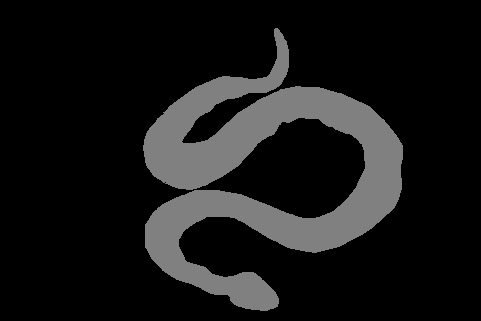} 
 \includegraphics[width = 1.6cm]{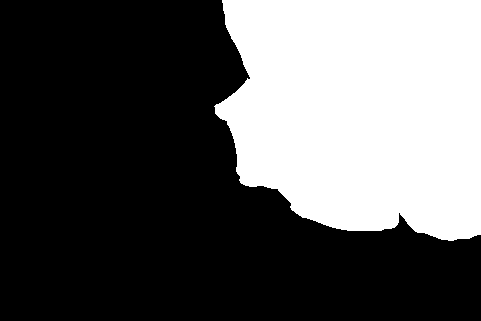} & &
 \cellcolor{gray!25} \includegraphics[width = 1.6cm]{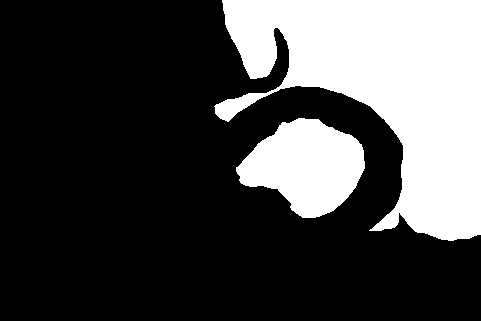} 
 \includegraphics[width = 1.6cm]{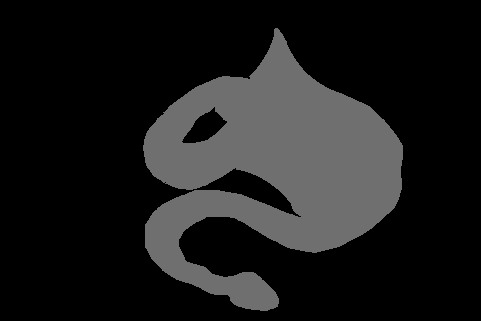} & &
 \cellcolor{gray!25} \includegraphics[width = 1.6cm]{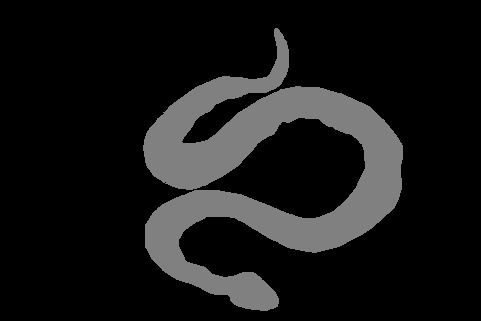} 
 \includegraphics[width = 1.6cm]{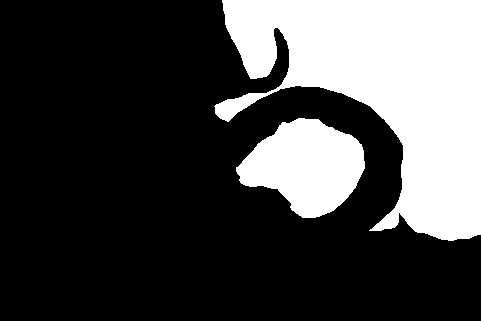} \\
 & & \cellcolor{gray!25}$\tilde{p}(I/H_1)=0.8796$ & & \cellcolor{gray!25}$\tilde{p}(I/H_2)=0.7074$ & & \cellcolor{gray!25}$\tilde{p}(I/H_3)=0.3679$\\
 \multicolumn{2}{c}{\scriptsize{Image 15}} & \cellcolor{gray!25}$\tilde{p}(H_1)=0.3750$ & & \cellcolor{gray!25}$\tilde{p}(H_2)=0.3680$ & & \cellcolor{gray!25}$\tilde{p}(H_3)=0.3729$\\
 & & \cellcolor{gray!25}$\mathbf{p(H_1/I)=0.4539}$ & & \cellcolor{gray!25}$p(H_2/I)=0.3582$ & & \cellcolor{gray!25}$p(H_3/I)=0.1879$\\ \\
  {\includegraphics[width = 1.6cm]{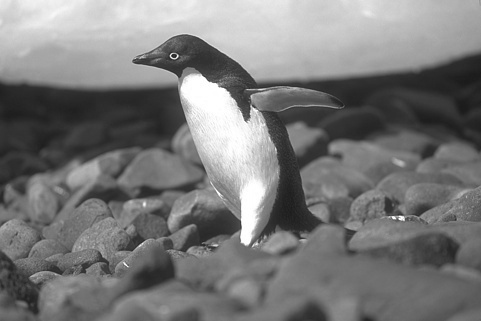}} & 
 {\includegraphics[width = 1.6cm]{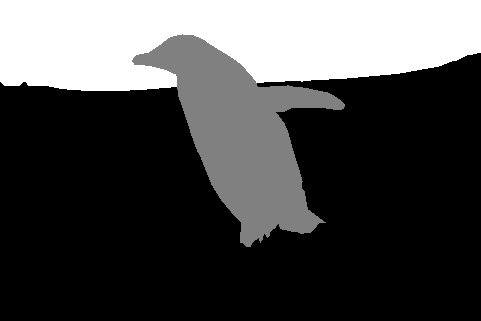}} & 
 \cellcolor{gray!25} \includegraphics[width = 1.6cm]{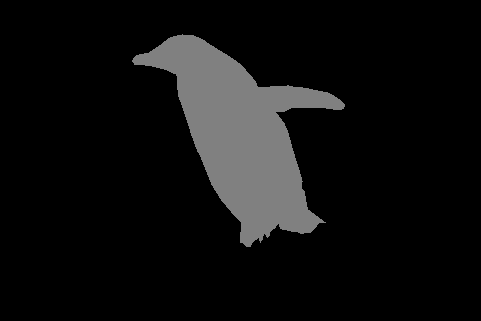} 
 \includegraphics[width = 1.6cm]{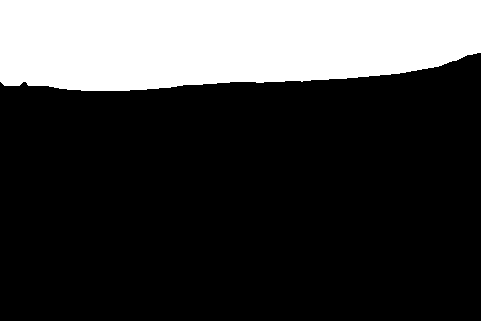} & &
 \cellcolor{gray!25} \includegraphics[width = 1.6cm]{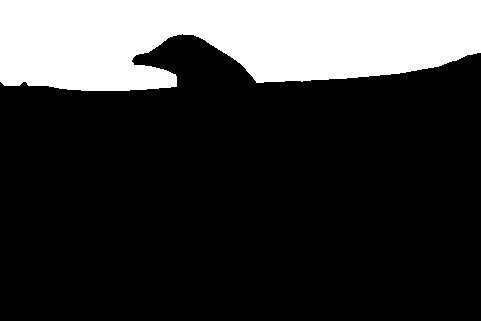} 
 \includegraphics[width = 1.6cm]{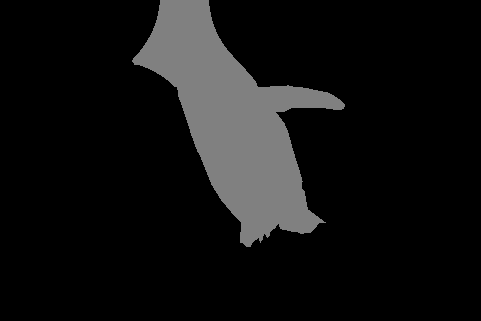} & &
 \cellcolor{gray!25} \includegraphics[width = 1.6cm]{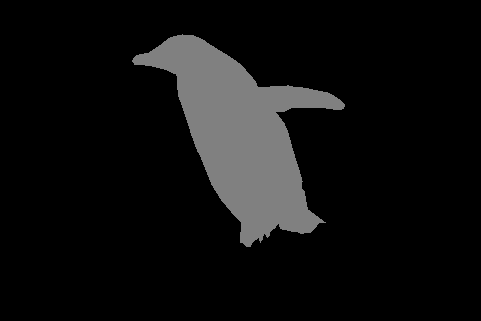} 
 \includegraphics[width = 1.6cm]{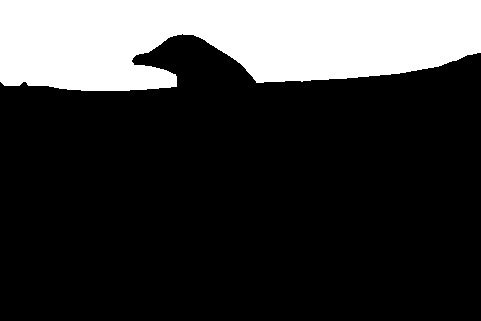} \\
 & & \cellcolor{gray!25}$\tilde{p}(I/H_1)=0.9112$ & & \cellcolor{gray!25}$\tilde{p}(I/H_2)=0.6035$ & & \cellcolor{gray!25}$\tilde{p}(I/H_3)=0.3679$\\
 \multicolumn{2}{c}{\scriptsize{Image 16}} & \cellcolor{gray!25}$\tilde{p}(H_1)=0.4034$ & & \cellcolor{gray!25}$\tilde{p}(H_2)=0.3679$ & & \cellcolor{gray!25}$\tilde{p}(H_3)=0.3679$\\
 & & \cellcolor{gray!25}$\mathbf{p(H_1/I)=0.5058}$ & & \cellcolor{gray!25}$p(H_2/I)=0.3055$ & & \cellcolor{gray!25}$p(H_3/I)=0.1887$\\ \\
  {\includegraphics[width = 1.6cm]{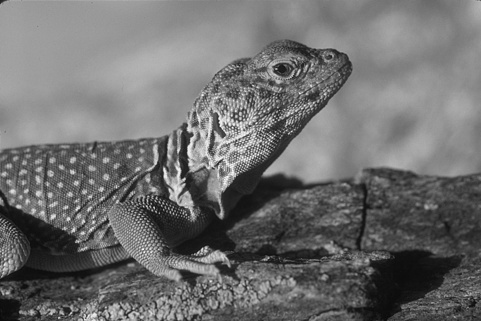}} & 
 {\includegraphics[width = 1.6cm]{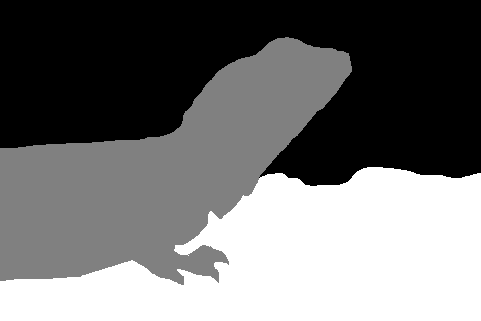}} & 
 \cellcolor{gray!25} \includegraphics[width = 1.6cm]{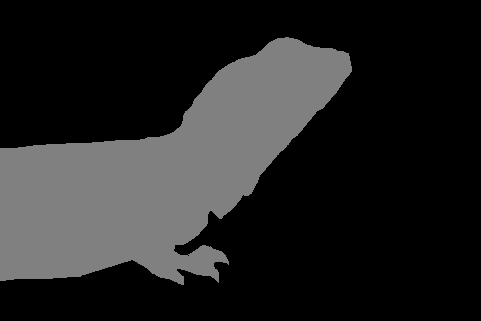} 
 \includegraphics[width = 1.6cm]{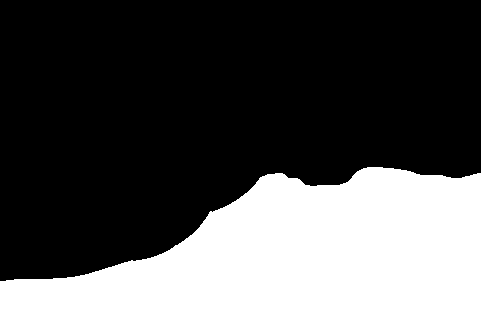} & &
 \cellcolor{gray!25} \includegraphics[width = 1.6cm]{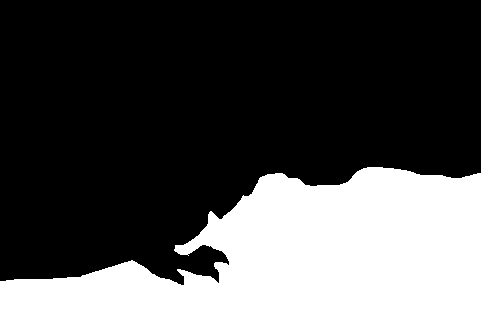} 
 \includegraphics[width = 1.6cm]{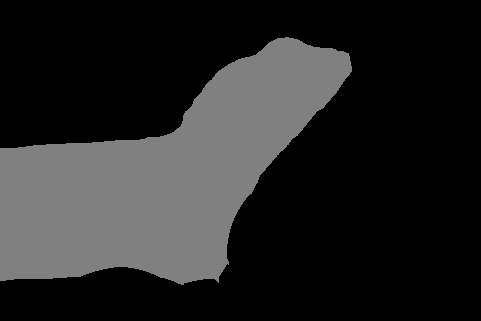} & &
 \cellcolor{gray!25} \includegraphics[width = 1.6cm]{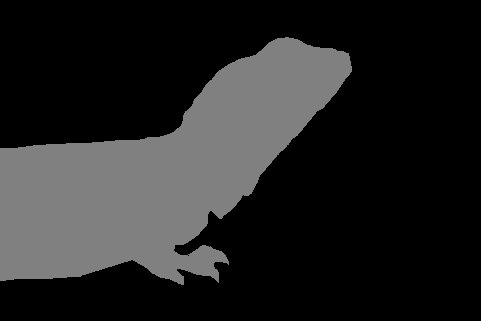} 
 \includegraphics[width = 1.6cm]{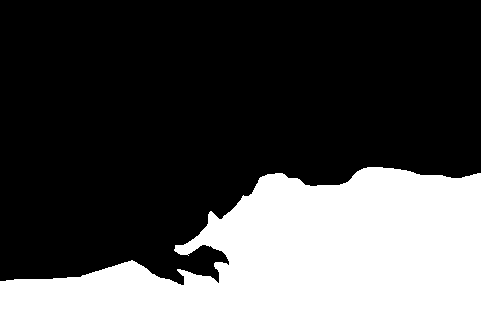} \\
 & & \cellcolor{gray!25}$\tilde{p}(I/H_1)=0.8833$ & & \cellcolor{gray!25}$\tilde{p}(I/H_2)=0.8279$ & & \cellcolor{gray!25}$\tilde{p}(I/H_3)=0.3679$\\
 \multicolumn{2}{c}{\scriptsize{Image 17}} & \cellcolor{gray!25}$\tilde{p}(H_1)=0.3718$ & & \cellcolor{gray!25}$\tilde{p}(H_2)=0.6354$ & & \cellcolor{gray!25}$\tilde{p}(H_3)=0.6334$\\
 & & \cellcolor{gray!25}$\mathbf{p(H_1/I)=0.4263}$ & & \cellcolor{gray!25}${p(H_2/I)=0.3980}$ & & \cellcolor{gray!25}$p(H_3/I)=0.1757$\\ \\
\caption{Experiments with real images from \cite{SemanticBenchmark}. Each row shows a different experiment: the two left-most images are, respectively, the original image and a segmentation of it, 
they are followed by the three different hypothesis (each one separated by a gray box). More details on the results shown for each hypothesis in Table \ref{tab:alltogether7}.}
\label{tab:alltogether5}
\end{longtable}

\setlength{\tabcolsep}{3pt}
\begin{longtable}{lllllll}
 {\includegraphics[width = 1.6cm]{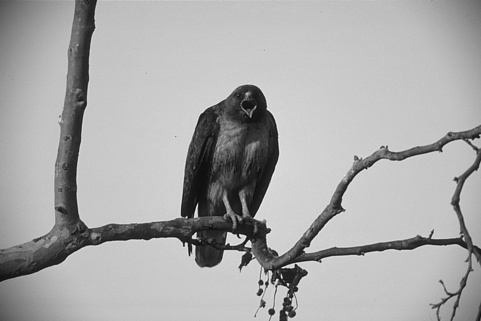}} & 
 {\includegraphics[width = 1.6cm]{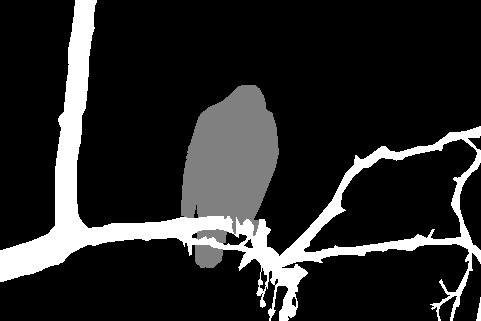}} & 
 \cellcolor{gray!25} \includegraphics[width = 1.6cm]{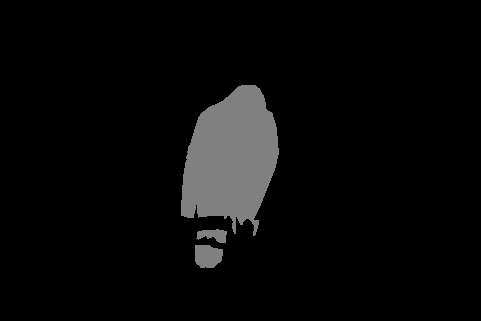} 
 \includegraphics[width = 1.6cm]{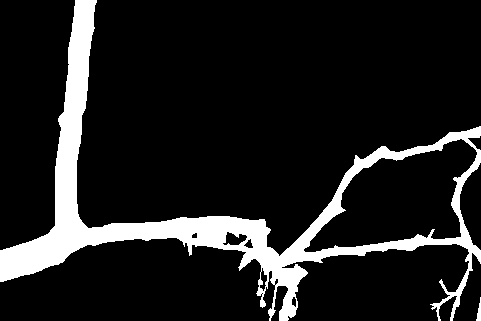} & &
 \cellcolor{gray!25} \includegraphics[width = 1.6cm]{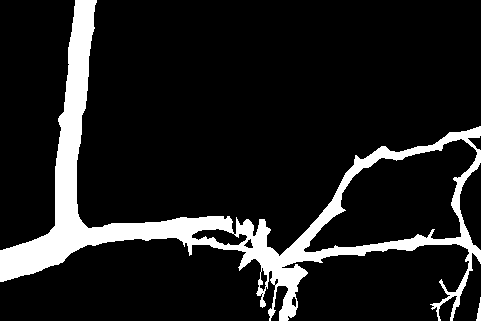} 
 \includegraphics[width = 1.6cm]{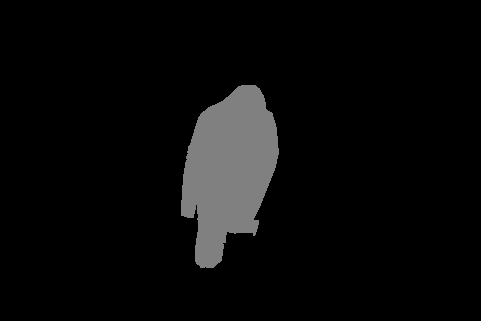} & &
 \cellcolor{gray!25} \includegraphics[width = 1.6cm]{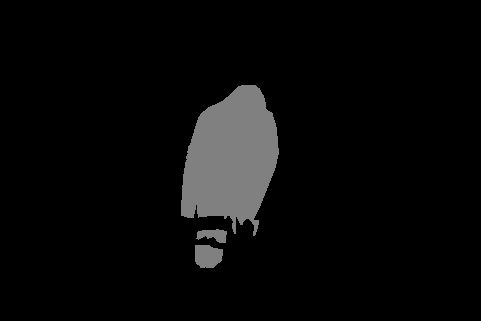} 
 \includegraphics[width = 1.6cm]{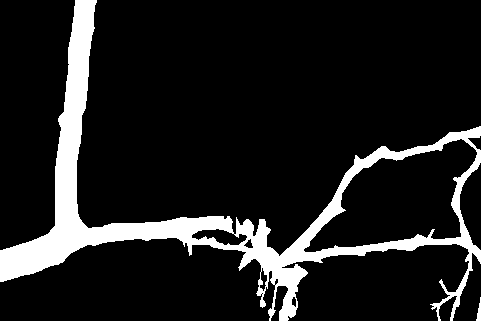} \\
 & & \cellcolor{gray!25}$\tilde{p}(I/H_1)=0.7913$ & & \cellcolor{gray!25}$\tilde{p}(I/H_2)=0.8845$ & & \cellcolor{gray!25}$\tilde{p}(I/H_3)=0.3679$\\
 \multicolumn{2}{c}{\scriptsize{Image 18}} & \cellcolor{gray!25}$\tilde{p}(H_1)=0.3685$ & & \cellcolor{gray!25}$\tilde{p}(H_2)=0.6163$ & & \cellcolor{gray!25}$\tilde{p}(H_3)=0.3679$\\
 & & \cellcolor{gray!25}$p(H_1/I)=0.2990$ & & \cellcolor{gray!25}$\mathbf{p(H_2/I)=0.5608}$ & & \cellcolor{gray!25}$p(H_3/I)=0.1392$\\ \\
 {\includegraphics[width = 1.6cm]{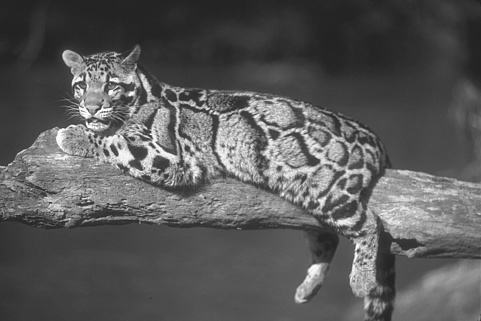}} & 
 {\includegraphics[width = 1.6cm]{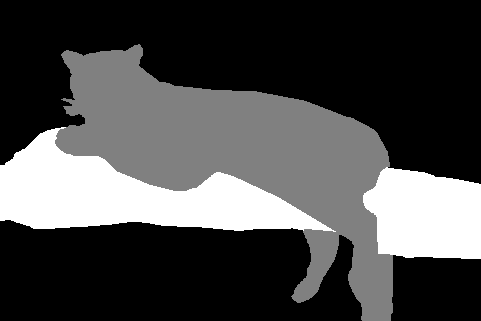}} & 
 \cellcolor{gray!25} \includegraphics[width = 1.6cm]{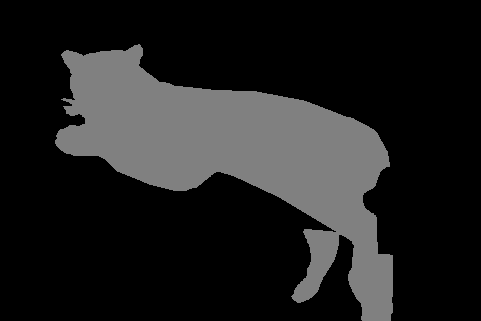} 
 \includegraphics[width = 1.6cm]{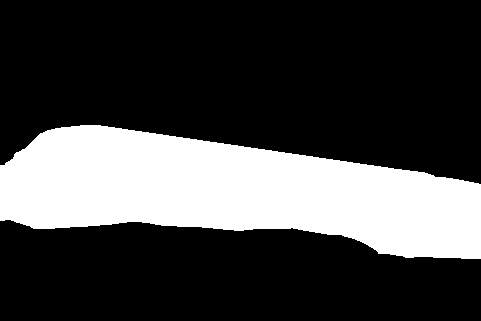} & &
 \cellcolor{gray!25} \includegraphics[width = 1.6cm]{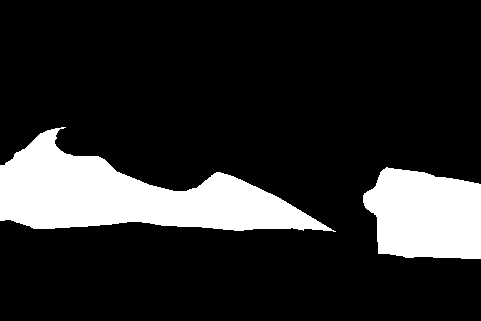} 
 \includegraphics[width = 1.6cm]{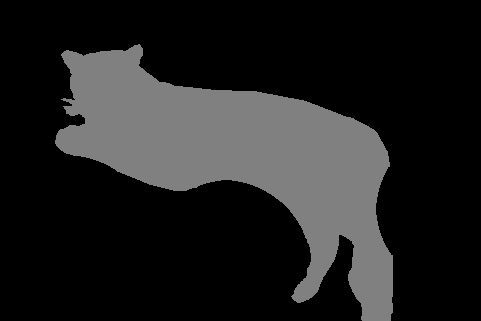} & &
 \cellcolor{gray!25} \includegraphics[width = 1.6cm]{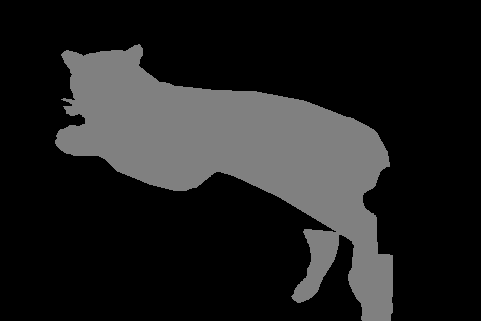} 
 \includegraphics[width = 1.6cm]{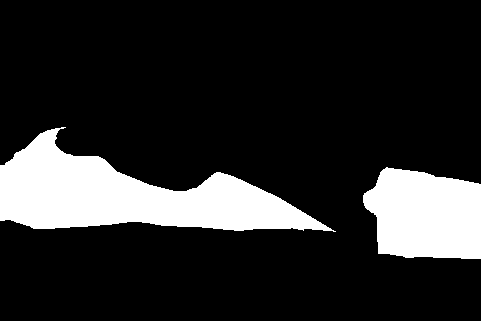} \\
 & & \cellcolor{gray!25}$\tilde{p}(I/H_1)=0.6789$ & & \cellcolor{gray!25}$\tilde{p}(I/H_2)=0.7031$ & & \cellcolor{gray!25}$\tilde{p}(I/H_3)=0.3679$\\
 \multicolumn{2}{c}{\scriptsize{Image 19}} & \cellcolor{gray!25}$\tilde{p}(H_1)=0.5380$ & & \cellcolor{gray!25}$\tilde{p}(H_2)=0.4302$ & & \cellcolor{gray!25}$\tilde{p}(H_3)=0.4313$\\
 & & \cellcolor{gray!25}$p(H_1/I)=\mathbf{0.4808}$ & & \cellcolor{gray!25}$p(H_2/I)=0.3405$ & & \cellcolor{gray!25}$p(H_3/I)=0.1787$\\ \\
 {\includegraphics[width = 1.6cm]{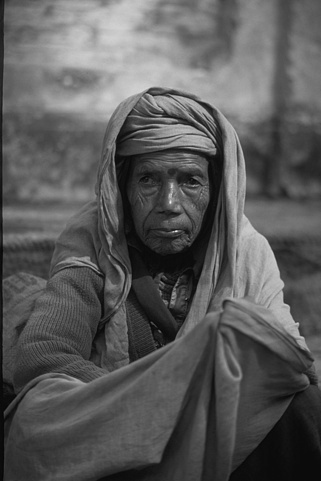}} & 
 {\includegraphics[width = 1.6cm]{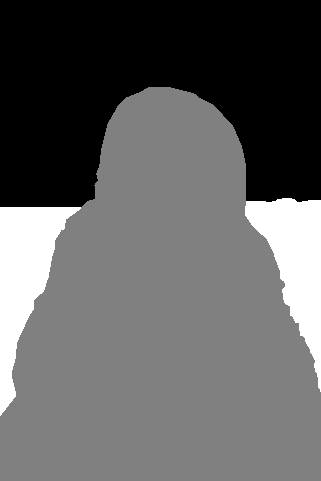}} & 
 \cellcolor{gray!25} \includegraphics[width = 1.6cm]{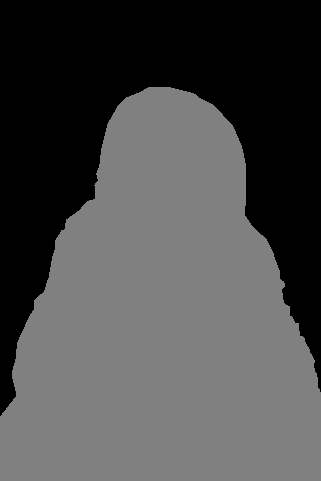} 
 \includegraphics[width = 1.6cm]{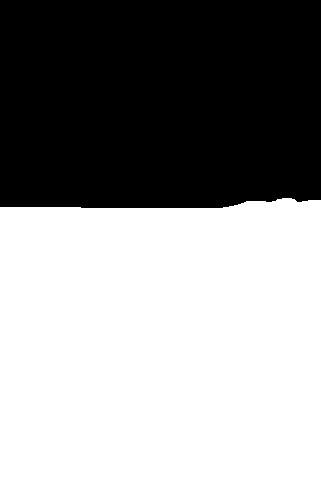} & &
 \cellcolor{gray!25} \includegraphics[width = 1.6cm]{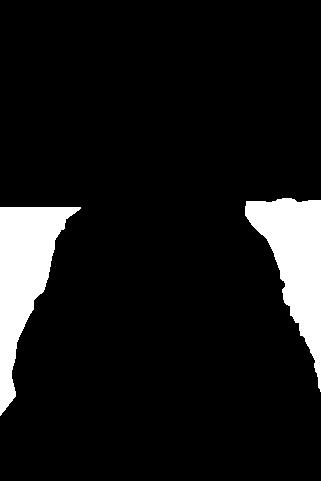} 
 \includegraphics[width = 1.6cm]{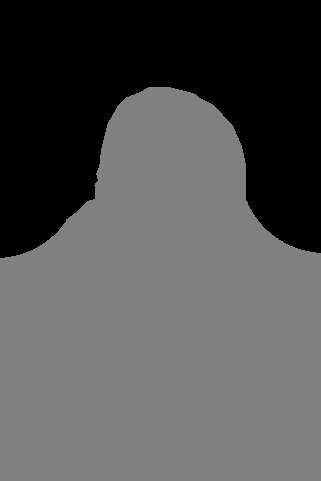} & &
 \cellcolor{gray!25} \includegraphics[width = 1.6cm]{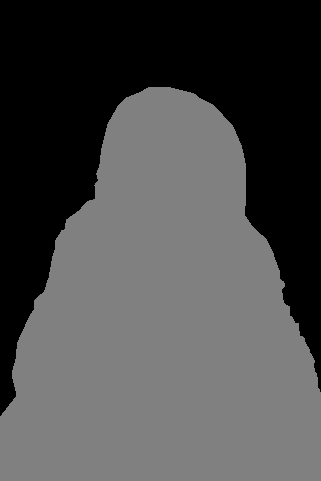} 
 \includegraphics[width = 1.6cm]{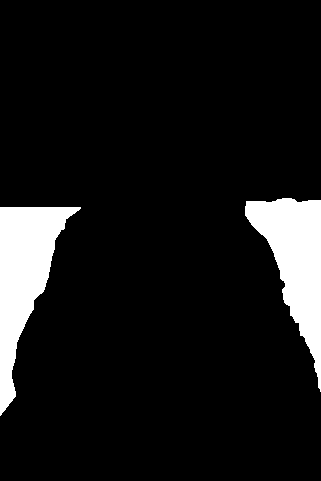} \\
 & & \cellcolor{gray!25}$\tilde{p}(I/H_1)=0.8933$ & & \cellcolor{gray!25}$\tilde{p}(I/H_2)=0.7496$ & & \cellcolor{gray!25}$\tilde{p}(I/H_3)=0.3679$\\
 \multicolumn{2}{c}{\scriptsize{Image 20}} & \cellcolor{gray!25}$\tilde{p}(H_1)=0.5970$ & & \cellcolor{gray!25}$\tilde{p}(H_2)=0.3715$ & & \cellcolor{gray!25}$\tilde{p}(H_3)=0.3679$\\
 & & \cellcolor{gray!25}$\mathbf{p(H_1/I)=0.5533}$ & & \cellcolor{gray!25}$p(H_2/I)=0.3063$ & & \cellcolor{gray!25}$p(H_3/I)=0.1404$\\ \\
 {\includegraphics[width = 1.6cm]{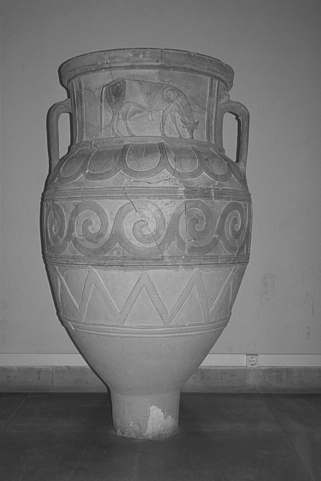}} & 
 {\includegraphics[width = 1.6cm]{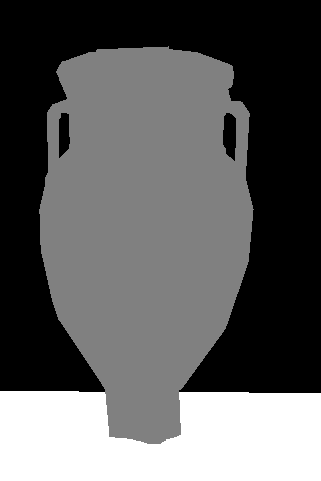}} & 
 \cellcolor{gray!25} \includegraphics[width = 1.6cm]{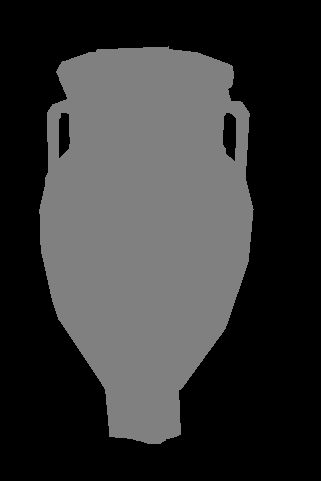} 
 \includegraphics[width = 1.6cm]{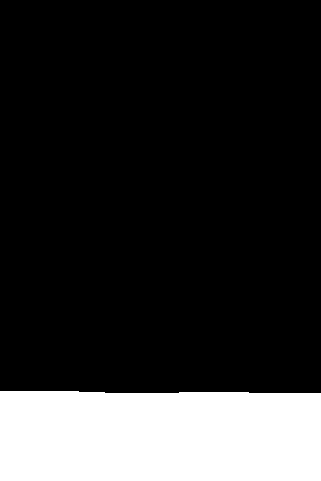} & &
 \cellcolor{gray!25} \includegraphics[width = 1.6cm]{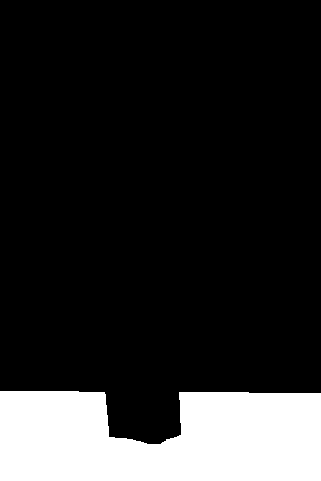} 
 \includegraphics[width = 1.6cm]{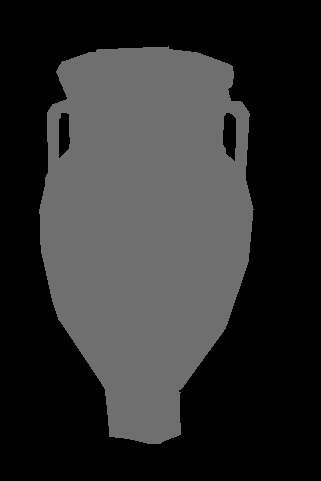} & &
 \cellcolor{gray!25} \includegraphics[width = 1.6cm]{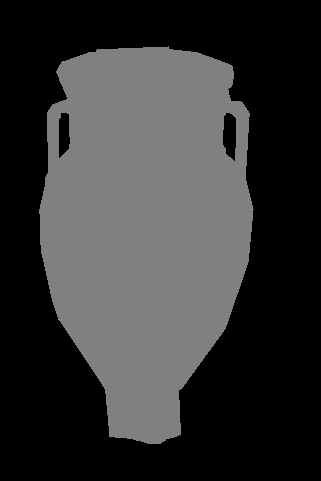} 
 \includegraphics[width = 1.6cm]{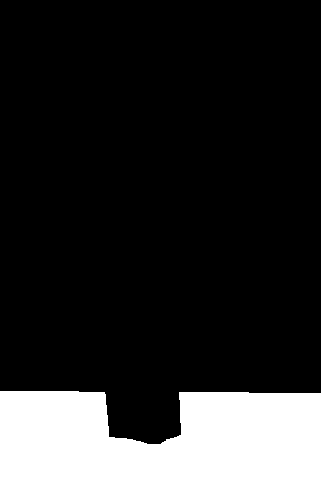} \\
 & & \cellcolor{gray!25}$\tilde{p}(I/H_1)=0.8579$ & & \cellcolor{gray!25}$\tilde{p}(I/H_2)=0.7293$ & & \cellcolor{gray!25}$\tilde{p}(I/H_3)=0.3679$\\
 \multicolumn{2}{c}{\scriptsize{Image 21}} & \cellcolor{gray!25}$\tilde{p}(H_1)=0.4983$ & & \cellcolor{gray!25}$\tilde{p}(H_2)=0.4976$ & & \cellcolor{gray!25}$\tilde{p}(H_3)=0.4976$\\
 & & \cellcolor{gray!25}$\mathbf{p(H_1/I)=0.4820}$ & & \cellcolor{gray!25}$p(H_2/I)=0.3441$ & & \cellcolor{gray!25}$p(H_3/I)=0.1739$\\ \\
 {\includegraphics[width = 1.6cm]{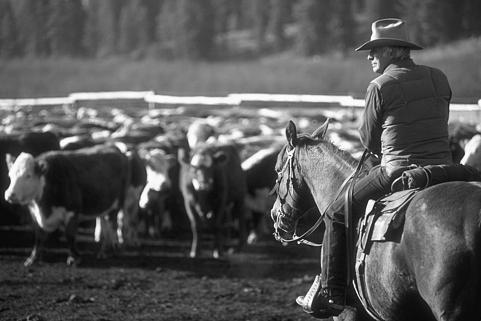}} & 
 {\includegraphics[width = 1.6cm]{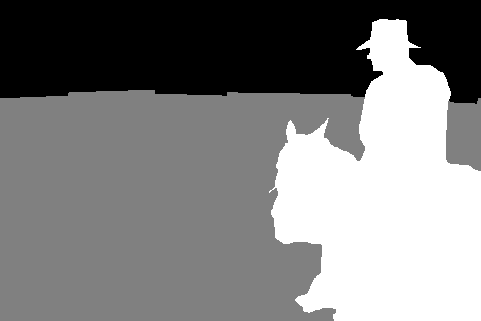}} & 
 \cellcolor{gray!25} \includegraphics[width = 1.6cm]{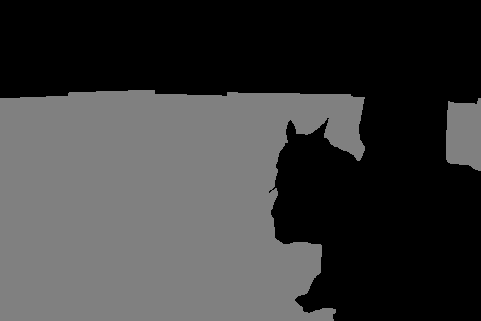} 
 \includegraphics[width = 1.6cm]{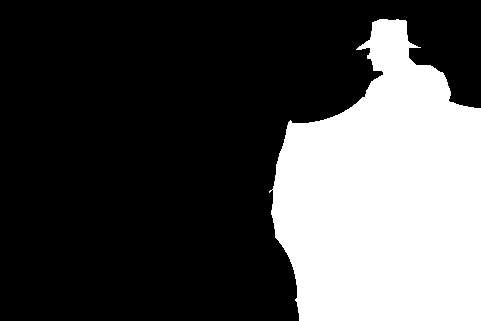} & &
 \cellcolor{gray!25} \includegraphics[width = 1.6cm]{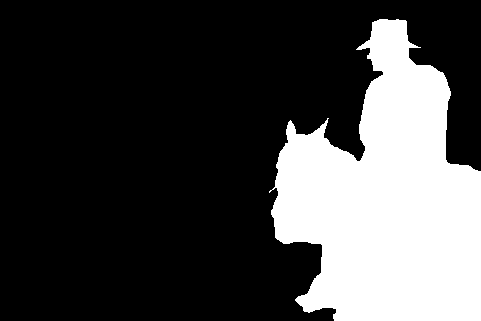} 
 \includegraphics[width = 1.6cm]{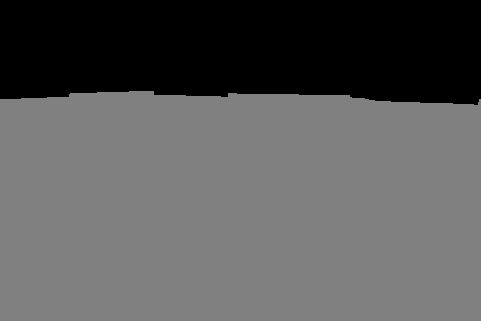} & &
 \cellcolor{gray!25} \includegraphics[width = 1.6cm]{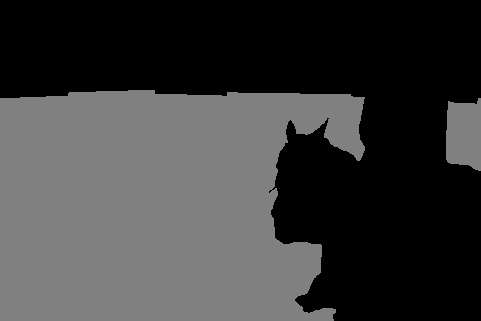} 
 \includegraphics[width = 1.6cm]{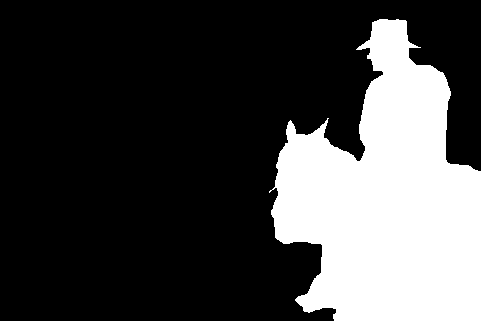} \\
 & & \cellcolor{gray!25}$\tilde{p}(I/H_1)=0.7574$ & & \cellcolor{gray!25}$\tilde{p}(I/H_2)=0.9523$ & & \cellcolor{gray!25}$\tilde{p}(I/H_3)=0.3679$\\
 \multicolumn{2}{c}{\scriptsize{Image 22}} & \cellcolor{gray!25}$\tilde{p}(H_1)=0.3679$ & & \cellcolor{gray!25}$\tilde{p}(H_2)=0.5717$ & & \cellcolor{gray!25}$\tilde{p}(H_3)=00.3709$\\
 & & \cellcolor{gray!25}$p(H_1/I)=0.2904$ & & \cellcolor{gray!25}$\mathbf{p(H_2/I)=0.5674}$ & & \cellcolor{gray!25}$p(H_3/I)=0.1422$\\ \\
 {\includegraphics[width = 1.6cm]{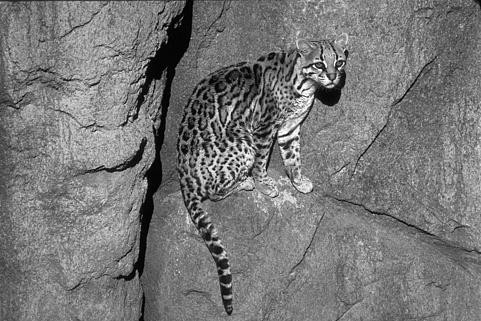}} & 
 {\includegraphics[width = 1.6cm]{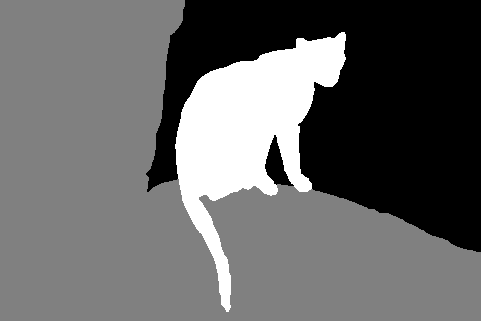}} & 
 \cellcolor{gray!25} \includegraphics[width = 1.6cm]{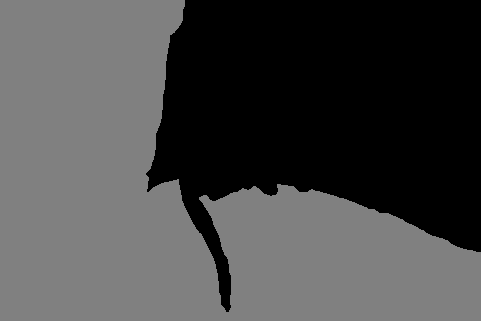} 
 \includegraphics[width = 1.6cm]{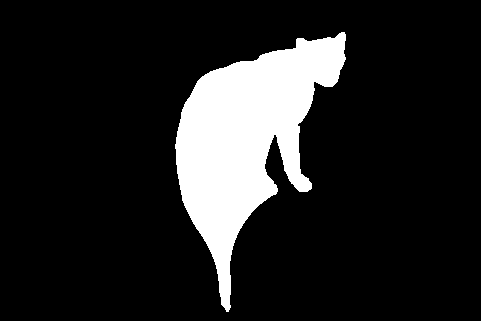} & &
 \cellcolor{gray!25} \includegraphics[width = 1.6cm]{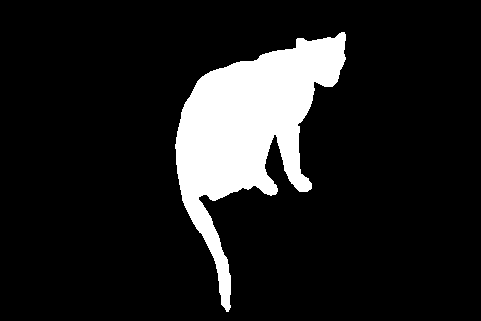} 
 \includegraphics[width = 1.6cm]{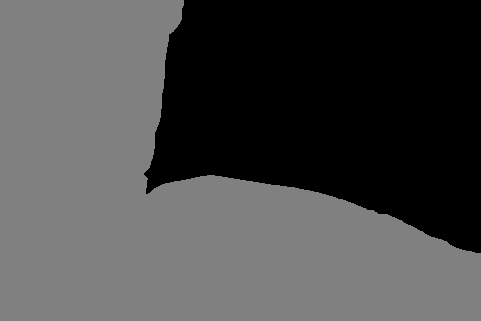} & &
 \cellcolor{gray!25} \includegraphics[width = 1.6cm]{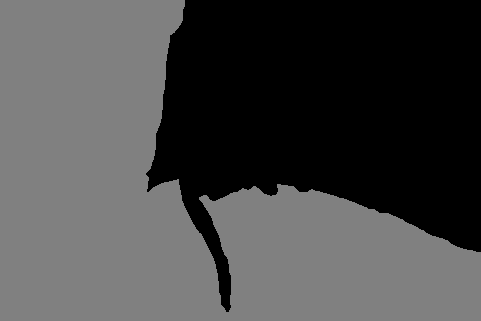} 
 \includegraphics[width = 1.6cm]{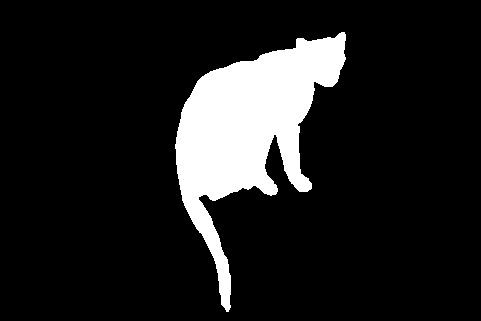} \\
 & & \cellcolor{gray!25}$\tilde{p}(I/H_1)=0.8382$ & & \cellcolor{gray!25}$\tilde{p}(I/H_2)=0.9034$ & & \cellcolor{gray!25}$\tilde{p}(I/H_3)=0.3679$\\
 \multicolumn{2}{c}{\scriptsize{Image 23}} & \cellcolor{gray!25}$\tilde{p}(H_1)=0.3722$ & & \cellcolor{gray!25}$\tilde{p}(H_2)=0.3877$ & & \cellcolor{gray!25}$\tilde{p}(H_3)=0.3679$\\
 & & \cellcolor{gray!25}$p(H_1/I)=0.3911$ & & \cellcolor{gray!25}$p\mathbf{(H_2/I)=0.4392}$ & & \cellcolor{gray!25}$p(H_3/I)=0.1697$\\ \\
 {\includegraphics[width = 1.6cm]{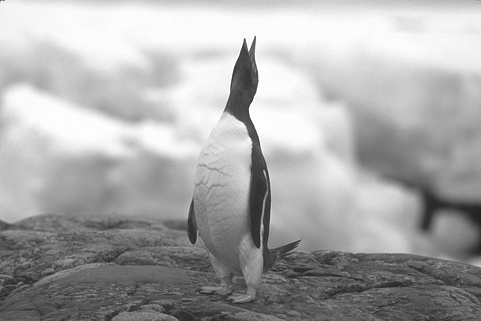}} & 
 {\includegraphics[width = 1.6cm]{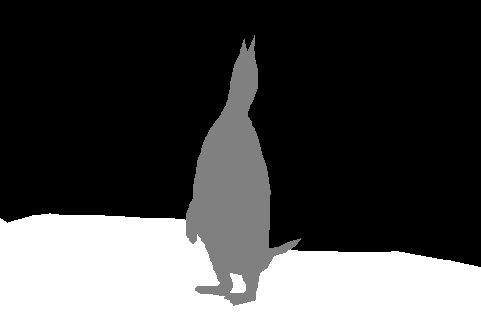}} & 
 \cellcolor{gray!25} \includegraphics[width = 1.6cm]{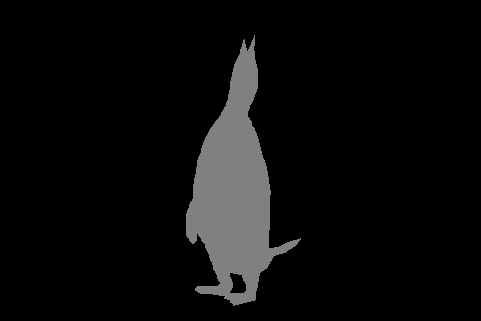} 
 \includegraphics[width = 1.6cm]{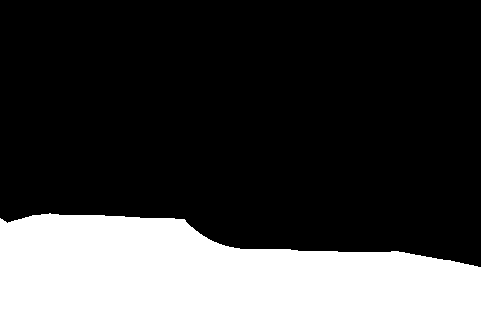} & &
 \cellcolor{gray!25} \includegraphics[width = 1.6cm]{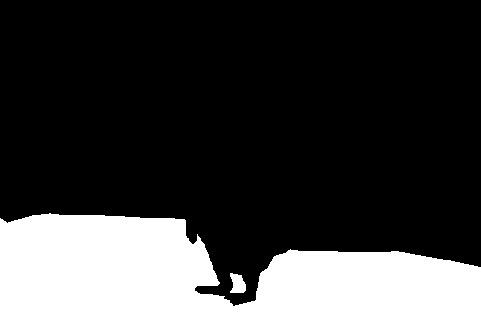} 
 \includegraphics[width = 1.6cm]{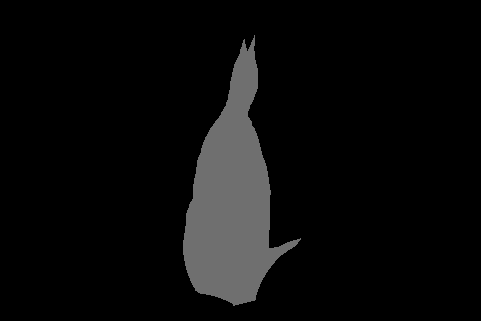} & &
 \cellcolor{gray!25} \includegraphics[width = 1.6cm]{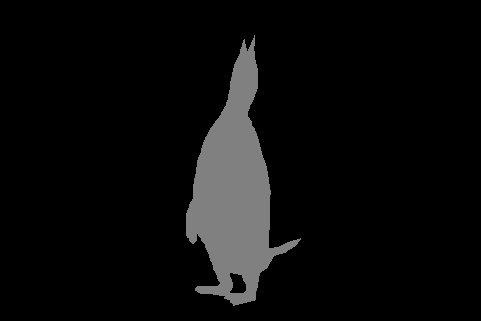} 
 \includegraphics[width = 1.6cm]{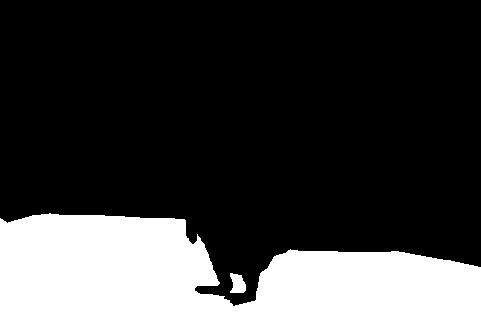} \\
 & & \cellcolor{gray!25}$\tilde{p}(I/H_1)=0.8392$ & & \cellcolor{gray!25}$\tilde{p}(I/H_2)=0.7695$ & & \cellcolor{gray!25}$\tilde{p}(I/H_3)=0.3679$\\
 \multicolumn{2}{c}{\scriptsize{Image 24}} & \cellcolor{gray!25}$\tilde{p}(H_1)=0.4028$ & & \cellcolor{gray!25}$\tilde{p}(H_2)=0.7417$ & & \cellcolor{gray!25}$\tilde{p}(H_3)=0.7414$\\
 & & \cellcolor{gray!25}$\mathbf{p(H_1/I)=0.4467}$ & & \cellcolor{gray!25}${p(H_2/I)=0.3742}$ & & \cellcolor{gray!25}$p(H_3/I)=0.1788$\\ \\
\caption{Experiments with real images from \cite{MartinFTM01}. Each row shows a different experiment: 
the two left-most images are, respectively, the original image and a segmentation of it, 
they are followed  by the three different hypothesis (each one separated by a gray box). 
More details on the results shown for each hypothesis in Table \ref{tab:alltogether7}.}
\label{tab:alltogether6}
\end{longtable}

We present in Table \ref{controvertides} two experiments with real images from \cite{MartinFTM01} where there is an ambiguity in the depth ordering (there are conflicting local depth cues). This situation can appear when the proximal image is made of objects that are not fronto-parallel to the camera or when their relative order changes due to, for example, mutual occlusions as in these examples. 
In other words, an object does not appear at a single depth layer. In this situation our algorithm chooses the object that is more occluded as being behind but let us remark how the posterior probabilities of the two first hypothesis are very close; in fact, these two hypothesis correspond to the two different depth orderings indicated by the local depth cues and figure/ground ground-truth labels superimposed on the original image.  
\setlength{\tabcolsep}{3pt}
\begin{longtable}{lllllll}
 {\includegraphics[width = 1.6cm]{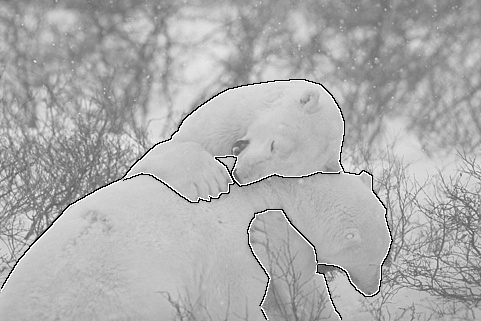}} & 
 {\includegraphics[width = 1.6cm]{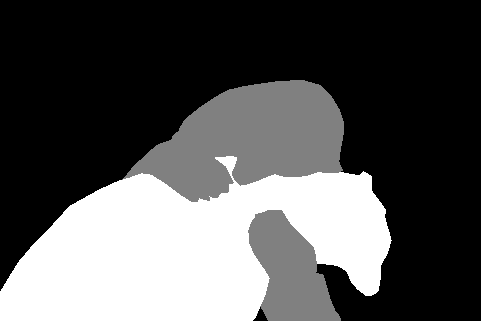}} & 
 \cellcolor{gray!25} \includegraphics[width = 1.6cm]{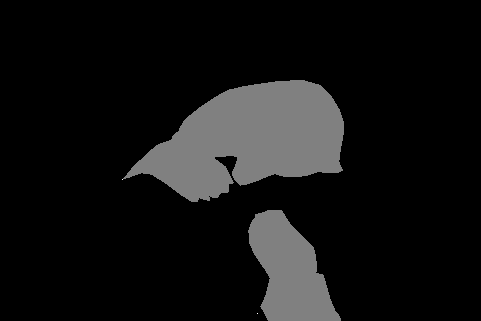} 
 \includegraphics[width = 1.6cm]{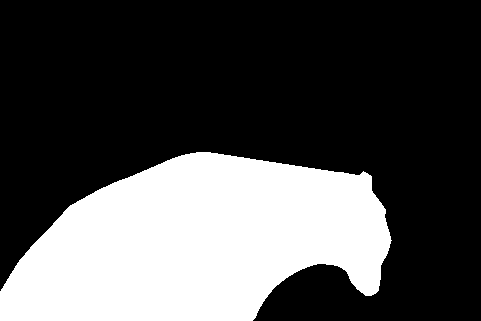} & &
 \cellcolor{gray!25} \includegraphics[width = 1.6cm]{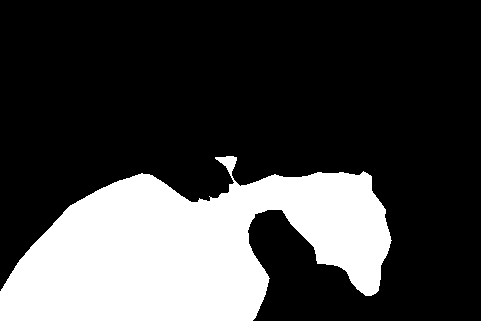} 
 \includegraphics[width = 1.6cm]{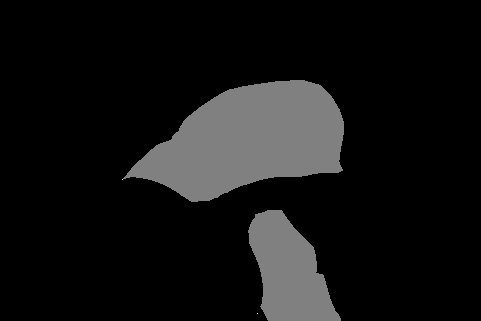} & &
 \cellcolor{gray!25} \includegraphics[width = 1.6cm]{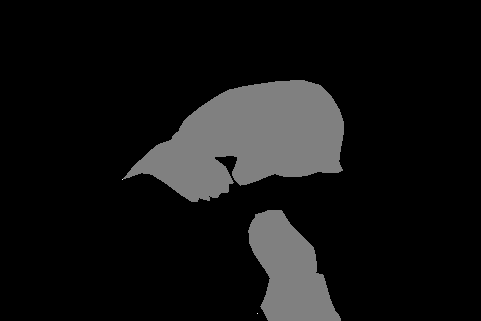} 
 \includegraphics[width = 1.6cm]{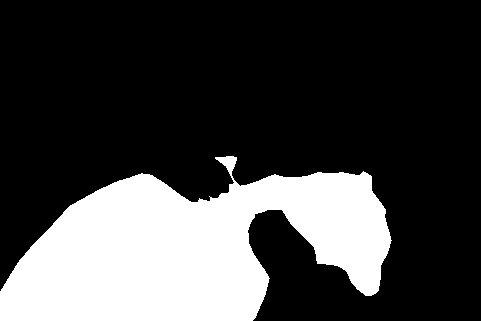} \\
 & & \cellcolor{gray!25}$\tilde{p}(I/H_1)=0.7757$ & & \cellcolor{gray!25}$\tilde{p}(I/H_2)=0.7942$ & & \cellcolor{gray!25}$\tilde{p}(I/H_3)=0.3679$\\
 \multicolumn{2}{c}{\scriptsize{Image 25}} & \cellcolor{gray!25}$\tilde{p}(H_1)=0.3709$ & & \cellcolor{gray!25}$\tilde{p}(H_2)=0.3763$ & & \cellcolor{gray!25}$\tilde{p}(H_3)=0.3679$\\
 & & \cellcolor{gray!25}$p(H_1/I)=0.3986$ & & \cellcolor{gray!25}$p(H_2/I)=0.5140$ & & \cellcolor{gray!25}$p(H_3/I)=0.1875$\\ \\
 {\includegraphics[width = 1.6cm]{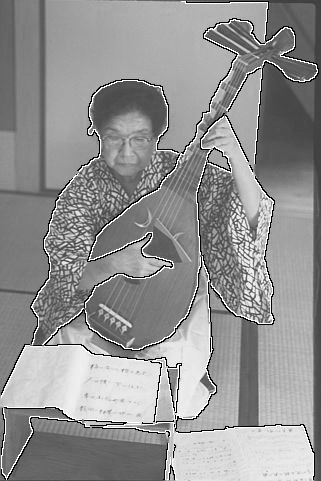}} & 
 {\includegraphics[width = 1.6cm]{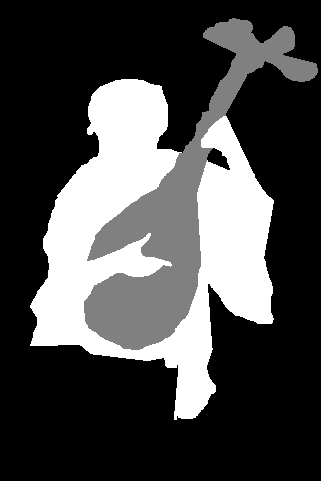}} & 
 \cellcolor{gray!25} \includegraphics[width = 1.6cm]{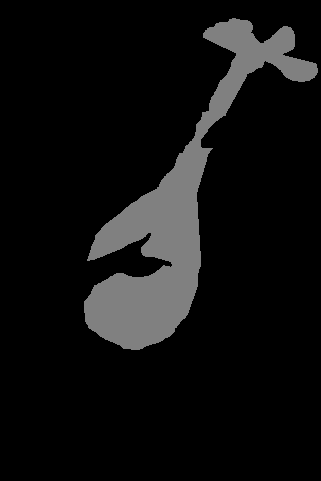} 
 \includegraphics[width = 1.6cm]{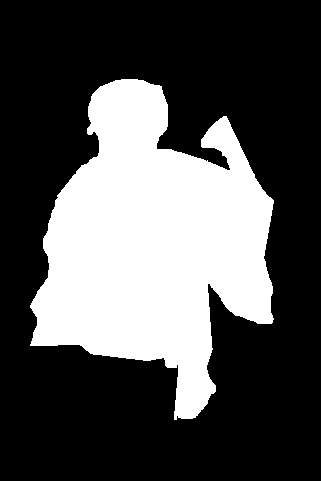} & &
 \cellcolor{gray!25} \includegraphics[width = 1.6cm]{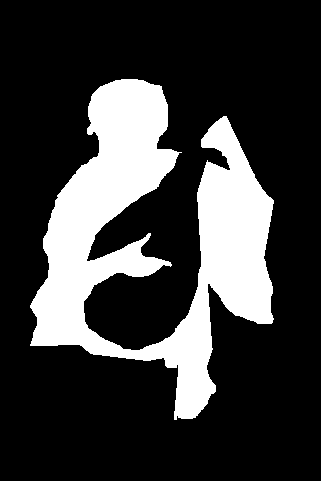} 
 \includegraphics[width = 1.6cm]{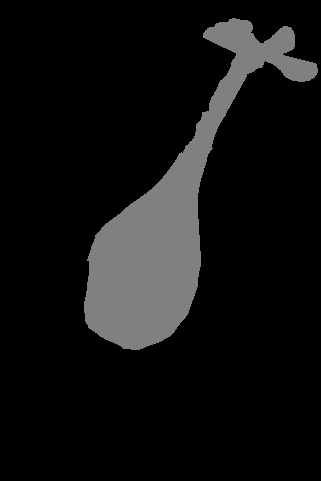} & &
 \cellcolor{gray!25} \includegraphics[width = 1.6cm]{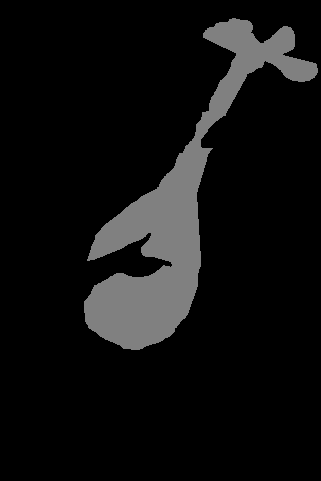} 
 \includegraphics[width = 1.6cm]{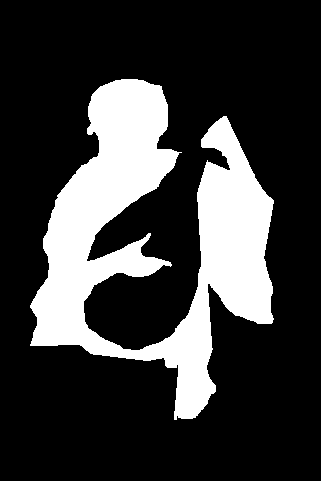} \\
 & & \cellcolor{gray!25}$\tilde{p}(I/H_1)=0.9787$ & & \cellcolor{gray!25}$\tilde{p}(I/H_2)=0.8768$ & & \cellcolor{gray!25}$\tilde{p}(I/H_3)=0.3679$\\
 \multicolumn{2}{c}{\scriptsize{Image 26}} & \cellcolor{gray!25}$\tilde{p}(H_1)=0.3689$ & & \cellcolor{gray!25}$\tilde{p}(H_2)=0.3695$ & & \cellcolor{gray!25}$\tilde{p}(H_3)=0.3679$\\
 & & \cellcolor{gray!25}$p(H_1/I)=0.4401$ & & \cellcolor{gray!25}$p(H_2/I)=0.3950$ & & \cellcolor{gray!25}$p(H_3/I)=0.1650$\\ \\
\caption{Two experiments with real images from \cite{MartinFTM01} where there is an ambiguity in the depth ordering (there are conflicting local depth cues). Notice how the posterior probabilities of the two first hypothesis are very close; in fact, these two hypothesis correspond to the two different depth orderings indicated by the local depth cues and figure/ground ground-truth labels superimposed on the original image.  
More details on the results shown in each row in Table \ref{tab:alltogether7}.}
\label{controvertides}
\end{longtable}

\subsection{Shapes in front of a background}
\label{subsec:fig-ground}
Finally, we present some results showing the ability of our method to also decide on (perceptually) fully visible objects over a background. Table~\ref{tab:alltogether8} displays several synthetic images of this type, where there are no T-junctions present 
and, according to human perception, the depth ordering is established by convexity cues~\cite{Kanizsa1991} (in the perception literature, they are also called figure-ground images).  For each row, the proximal stimulus is shown in the first column, and the following columns are the different hypothesis separated by a gray box. For hypothesis $H_1$ and $H_2$, the object which is supposed to be at depth 1, i.e, the occluding object, is displayed in blue. In $H_3$, where both objects are fitted together, the objects are presented with its original color. In these experiments, our method fails in Results 3, 4 and 5; in all of them the convexity cue is a stronger depth cue than symmetry, while the algorithm we are using for computing shape complexity  favors symmetries. Let us also remark that Result 9 allows both interpretations: black in front of white and white in front of black, as they form the same shape but with different orientation. In this case our algorithm prefers $H_1$ but with a small difference with respect to $H_2$. 

\setlength{\tabcolsep}{3pt}
\begin{longtable}{llllll}
 \includegraphics[width = 1.75cm]{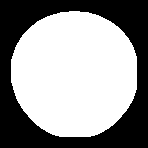} &
 \cellcolor{gray!25} \includegraphics[width = 1.75cm]{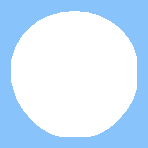} 
 \includegraphics[width = 1.75cm]{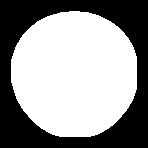} & &
 \cellcolor{gray!25} \includegraphics[width = 1.75cm]{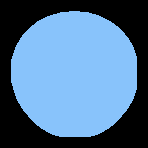} 
 \includegraphics[width = 1.75cm]{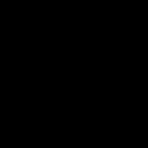} & &
 \cellcolor{gray!25} \includegraphics[width = 1.75cm]{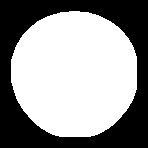} 
 \includegraphics[width = 1.75cm]{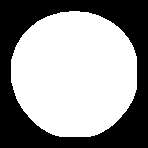} \\
 & \cellcolor{gray!25}$\tilde{p}(I/H_1)=0.3679$ & & \cellcolor{gray!25}$\tilde{p}(I/H_2)=1$ & & \cellcolor{gray!25}$\tilde{p}(I/H_3)=0.3679$\\
 Input 1 & \cellcolor{gray!25}$\tilde{p}(H_1)=0.3679$ & & \cellcolor{gray!25}$\tilde{p}(H_2)=0.6053$ & & \cellcolor{gray!25}$\tilde{p}(H_3)=0.3679$\\
 & \cellcolor{gray!25}$p(H_1/I)=0.1545$ & & \cellcolor{gray!25}$\mathbf{p(H_2/I)=0.6910}$ & & \cellcolor{gray!25}$p(H_3/I)=0.1545$\\ \\
 \includegraphics[width = 1.75cm]{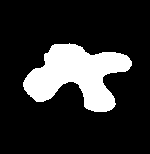} &
 \cellcolor{gray!25} \includegraphics[width = 1.75cm]{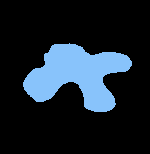} 
 \includegraphics[width = 1.75cm]{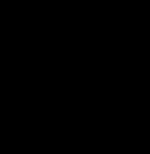} & &
 \cellcolor{gray!25} \includegraphics[width = 1.75cm]{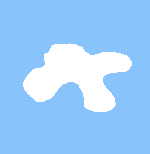} 
 \includegraphics[width = 1.75cm]{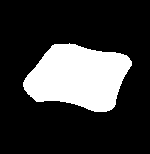} & &
 \cellcolor{gray!25} \includegraphics[width = 1.75cm]{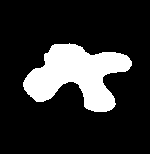} 
 \includegraphics[width = 1.75cm]{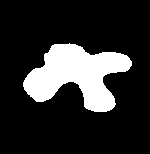} \\
 & \cellcolor{gray!25}$\tilde{p}(I/H_1)=1$ & & \cellcolor{gray!25}$\tilde{p}(I/H_2)=0.8243$ & & \cellcolor{gray!25}$\tilde{p}(I/H_3)=0.3679$\\
 Input 2 & \cellcolor{gray!25}$\tilde{p}(H_1)=0.6059$ & & \cellcolor{gray!25}$\tilde{p}(H_2)=0.3689$ & & \cellcolor{gray!25}$\tilde{p}(H_3)=0.3679$\\
 & \cellcolor{gray!25}$\mathbf{p(H_1/I)=0.5740}$ & & \cellcolor{gray!25}$p(H_2/I)=0.2909$ & & \cellcolor{gray!25}$p(H_3/I)=0.1295$\\ \\
 \includegraphics[width = 1.75cm]{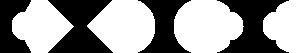} &
 \cellcolor{gray!25} \includegraphics[width = 1.75cm]{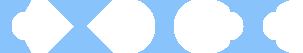} 
 \includegraphics[width = 1.75cm]{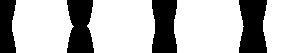} & &
 \cellcolor{gray!25} \includegraphics[width = 1.75cm]{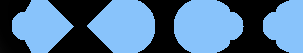} 
 \includegraphics[width = 1.75cm]{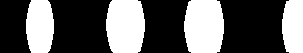} & &
 \cellcolor{gray!25} \includegraphics[width = 1.75cm]{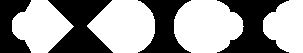} 
 \includegraphics[width = 1.75cm]{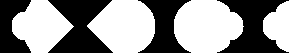} \\
 & \cellcolor{gray!25}$\tilde{p}(I/H_1)=0.7791$ & & \cellcolor{gray!25}$\tilde{p}(I/H_2)=0.7457$ & & \cellcolor{gray!25}$\tilde{p}(I/H_3)=0.3679$\\
 Input 3 & \cellcolor{gray!25}$\tilde{p}(H_1)=0.5273$ & & \cellcolor{gray!25}$\tilde{p}(H_2)=0.3679$ & & \cellcolor{gray!25}$\tilde{p}(H_3)=0.3843$\\
 & \cellcolor{gray!25}$\mathbf{p(H_1/I)=0.4971}$ & & \cellcolor{gray!25}$p(H_2/I)=0.3319$ & & \cellcolor{gray!25}$p(H_3/I)=0.1710$\\ \\
 \includegraphics[width = 1.75cm]{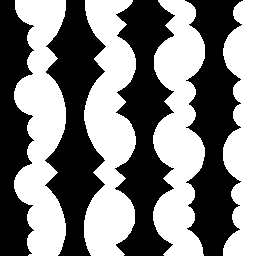} &
 \cellcolor{gray!25} \includegraphics[width = 1.75cm]{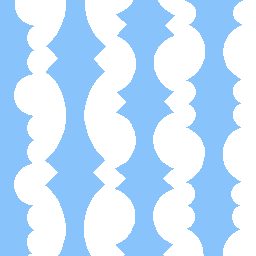} 
 \includegraphics[width = 1.75cm]{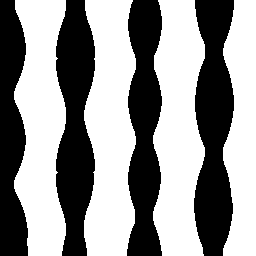} & &
 \cellcolor{gray!25} \includegraphics[width = 1.75cm]{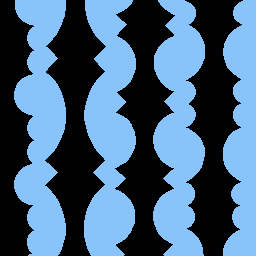} 
 \includegraphics[width = 1.75cm]{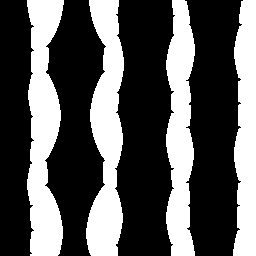} & &
 \cellcolor{gray!25} \includegraphics[width = 1.75cm]{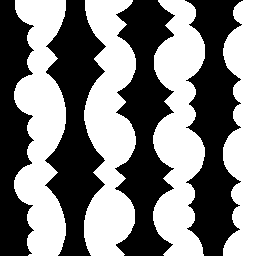} 
 \includegraphics[width = 1.75cm]{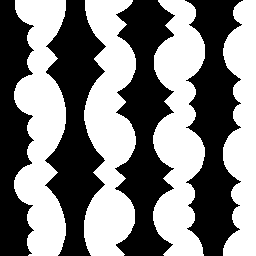} \\
 & \cellcolor{gray!25}$\tilde{p}(I/H_1)=0.7858$ & & \cellcolor{gray!25}$\tilde{p}(I/H_2)=0.7422$ & & \cellcolor{gray!25}$\tilde{p}(I/H_3)=0.3679$\\
 Input 4 & \cellcolor{gray!25}$\tilde{p}(H_1)=0.3891$ & & \cellcolor{gray!25}$\tilde{p}(H_2)=0.3862$ & & \cellcolor{gray!25}$\tilde{p}(H_3)=0.3679$\\
 & \cellcolor{gray!25}$\mathbf{p(H_1/I)=0.4204}$ & & \cellcolor{gray!25}$p(H_2/I)=0.3937$ & & \cellcolor{gray!25}$p(H_3/I)=0.1859$\\ \\
 \includegraphics[width = 1.75cm]{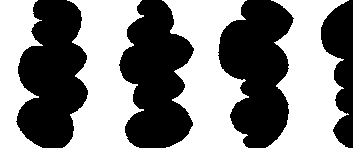} &
 \cellcolor{gray!25} \includegraphics[width = 1.75cm]{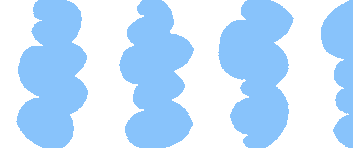} 
 \includegraphics[width = 1.75cm]{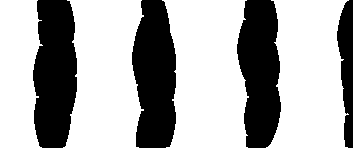} & &
 \cellcolor{gray!25} \includegraphics[width = 1.75cm]{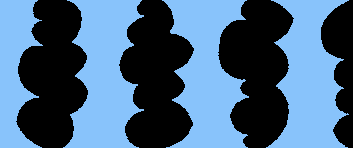} 
 \includegraphics[width = 1.75cm]{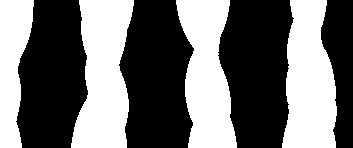} & &
 \cellcolor{gray!25} \includegraphics[width = 1.75cm]{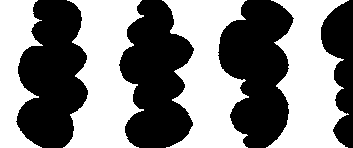} 
 \includegraphics[width = 1.75cm]{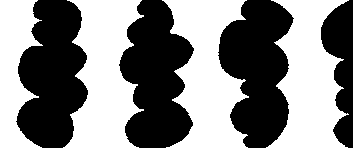} \\
 & \cellcolor{gray!25}$\tilde{p}(I/H_1)=0.7656$ & & \cellcolor{gray!25}$\tilde{p}(I/H_2)=0.7854$ & & \cellcolor{gray!25}$\tilde{p}(I/H_3)=0.3679$\\
  Input 5 & \cellcolor{gray!25}$\tilde{p}(H_1)=0.3924$ & & \cellcolor{gray!25}$\tilde{p}(H_2)=0.3926$ & & \cellcolor{gray!25}$\tilde{p}(H_3)=0.3679$\\
 & \cellcolor{gray!25}$p(H_1/I)=0.4037$ & & \cellcolor{gray!25}$\mathbf{p(H_2/I)=0.4144}$ & & \cellcolor{gray!25}$p(H_3/I)=0.1819$\\ \\
 \includegraphics[width = 1.75cm]{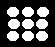} &
 \cellcolor{gray!25} \includegraphics[width = 1.75cm]{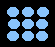} 
 \includegraphics[width = 1.75cm]{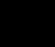} & &
 \cellcolor{gray!25} \includegraphics[width = 1.75cm]{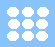} 
 \includegraphics[width = 1.75cm]{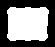} & &
 \cellcolor{gray!25} \includegraphics[width = 1.75cm]{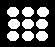} 
 \includegraphics[width = 1.75cm]{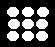} \\
 & \cellcolor{gray!25}$\tilde{p}(I/H_1)=1$ & & \cellcolor{gray!25}$\tilde{p}(I/H_2)=0.9508$ & & \cellcolor{gray!25}$\tilde{p}(I/H_3)=0.3679$\\
  Input 6 & \cellcolor{gray!25}$\tilde{p}(H_1)=0.5870$ & & \cellcolor{gray!25}$\tilde{p}(H_2)=0.4250$ & & \cellcolor{gray!25}$\tilde{p}(H_3)=0.3679$\\
 & \cellcolor{gray!25}$\mathbf{p(H_1/I)=0.5211}$ & & \cellcolor{gray!25}$p(H_2/I)=0.3588$ & & \cellcolor{gray!25}$p(H_3/I)=0.1201$\\ \\
 \includegraphics[width = 1.75cm]{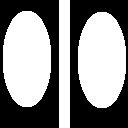} &
 \cellcolor{gray!25} \includegraphics[width = 1.75cm]{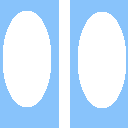} 
 \includegraphics[width = 1.75cm]{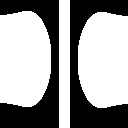} & &
 \cellcolor{gray!25} \includegraphics[width = 1.75cm]{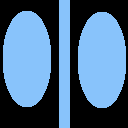} 
 \includegraphics[width = 1.75cm]{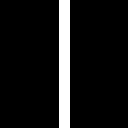} & &
 \cellcolor{gray!25} \includegraphics[width = 1.75cm]{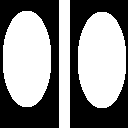} 
 \includegraphics[width = 1.75cm]{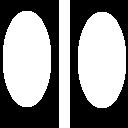} \\
 & \cellcolor{gray!25}$\tilde{p}(I/H_1)=0.9093$ & & \cellcolor{gray!25}$\tilde{p}(I/H_2)=0.7799$ & & \cellcolor{gray!25}$\tilde{p}(I/H_3)=0.3679$\\
 Input 7  & \cellcolor{gray!25}$\tilde{p}(H_1)=0.3746$ & & \cellcolor{gray!25}$\tilde{p}(H_2)=0.4755$ & & \cellcolor{gray!25}$\tilde{p}(H_3)=0.3679$\\
 & \cellcolor{gray!25}$p(H_1/I)=0.4023$ & & \cellcolor{gray!25}$\mathbf{p(H_2/I)=0.4379}$ & & \cellcolor{gray!25}$p(H_3/I)=0.1598$\\ \\
 \includegraphics[width = 1.75cm]{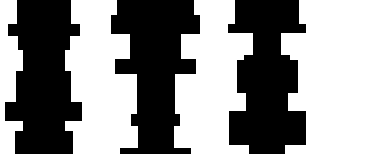} &
 \cellcolor{gray!25} \includegraphics[width = 1.75cm]{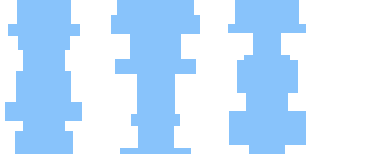} 
 \includegraphics[width = 1.75cm]{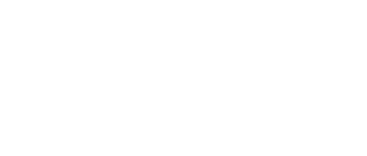} & &
 \cellcolor{gray!25} \includegraphics[width = 1.75cm]{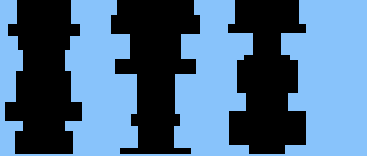} 
 \includegraphics[width = 1.75cm]{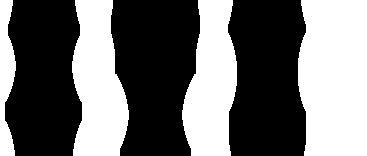} & &
 \cellcolor{gray!25} \includegraphics[width = 1.75cm]{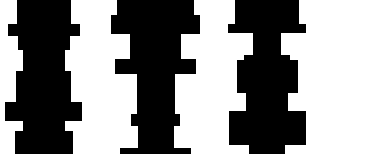} 
 \includegraphics[width = 1.75cm]{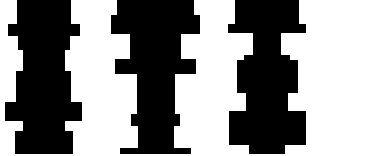} \\
 & \cellcolor{gray!25}$\tilde{p}(I/H_1)=1$ & & \cellcolor{gray!25}$\tilde{p}(I/H_2)=0.7191$ & & \cellcolor{gray!25}$\tilde{p}(I/H_3)=0.3679$\\
 Input 8 & \cellcolor{gray!25}$\tilde{p}(H_1)=0.6961$ & & \cellcolor{gray!25}$\tilde{p}(H_2)=0.3679$ & & \cellcolor{gray!25}$\tilde{p}(H_3)=0.4584$\\
 & \cellcolor{gray!25}$\mathbf{p(H_1/I)=0.6164}$ & & \cellcolor{gray!25}$p(H_2/I)=0.2343$ & & \cellcolor{gray!25}$p(H_3/I)=0.1493$\\ \\
 \includegraphics[width = 1.75cm]{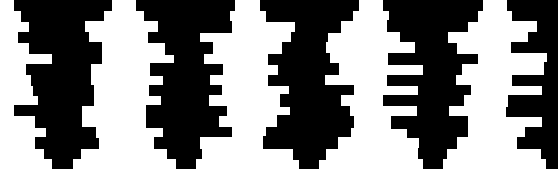} &
 \cellcolor{gray!25} \includegraphics[width = 1.75cm]{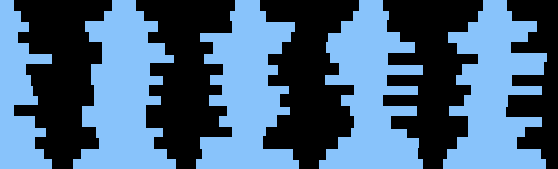} 
 \includegraphics[width = 1.75cm]{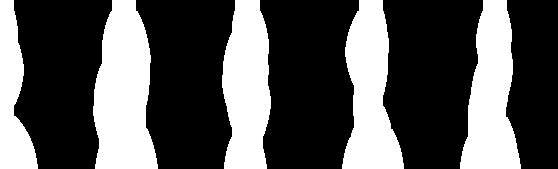} & &
 \cellcolor{gray!25} \includegraphics[width = 1.75cm]{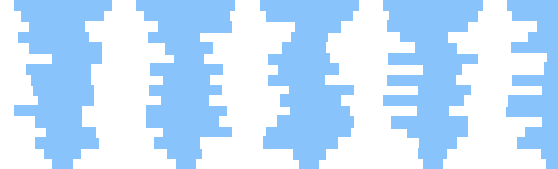} 
 \includegraphics[width = 1.75cm]{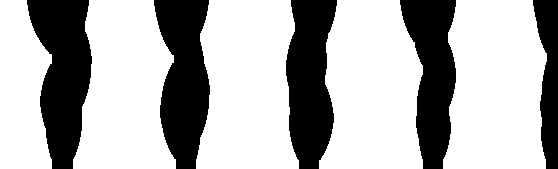} & &
 \cellcolor{gray!25} \includegraphics[width = 1.75cm]{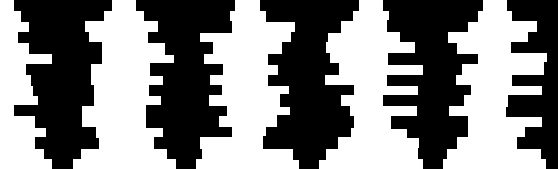} 
 \includegraphics[width = 1.75cm]{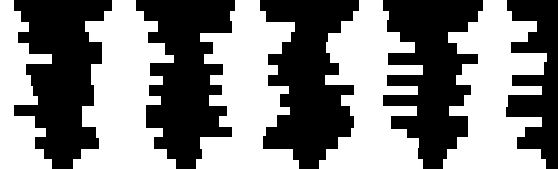} \\
 & \cellcolor{gray!25}$\tilde{p}(I/H_1)=0.7790$ & & \cellcolor{gray!25}$\tilde{p}(I/H_2)=0.7438$ & & \cellcolor{gray!25}$\tilde{p}(I/H_3)=0.3679$\\
  Input 9 & \cellcolor{gray!25}$\tilde{p}(H_1)=0.3812$ & & \cellcolor{gray!25}$\tilde{p}(H_2)=0.3679$ & & \cellcolor{gray!25}$\tilde{p}(H_3)=0.4757$\\
 & \cellcolor{gray!25}$\mathbf{p(H_1/I)=0.3812}$ & & \cellcolor{gray!25}$p(H_2/I)=0.3679$ & & \cellcolor{gray!25}$p(H_3/I)=0.2347$\\ \\
 \includegraphics[width = 1.75cm]{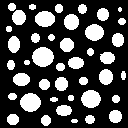} &
 \cellcolor{gray!25} \includegraphics[width = 1.75cm]{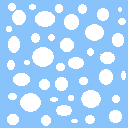} 
 \includegraphics[width = 1.75cm]{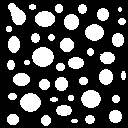} & &
 \cellcolor{gray!25} \includegraphics[width = 1.75cm]{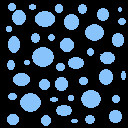} 
 \includegraphics[width = 1.75cm]{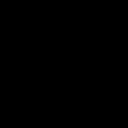} & &
 \cellcolor{gray!25} \includegraphics[width = 1.75cm]{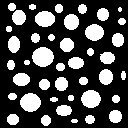} 
 \includegraphics[width = 1.75cm]{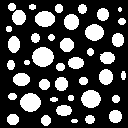} \\
 & \cellcolor{gray!25}$\tilde{p}(I/H_1)=0.9966$ & & \cellcolor{gray!25}$\tilde{p}(I/H_2)=1$ & & \cellcolor{gray!25}$\tilde{p}(I/H_3)=0.3679$\\
 Input 10 & \cellcolor{gray!25}$\tilde{p}(H_1)=0.3679$ & & \cellcolor{gray!25}$\tilde{p}(H_2)=0.3977$ & & \cellcolor{gray!25}$\tilde{p}(H_3)=0.3803$\\
 & \cellcolor{gray!25}$p(H_1/I)=0.4055$ & & \cellcolor{gray!25}$\mathbf{p(H_2/I)=0.4398}$ & & \cellcolor{gray!25}$p(H_3/I)=0.1547$\\ \\
\caption{Synthetic experiments with a shape in front of a background. Each row shows a different experiment: the original image (proximal stimulus) is shown on the left and it is followed by the three different hypothesis (each one separated by a gray box). For the first two hypothesis, $H_i$ for $i=1,2$, we show: the object at depth 1 in blue (left) and the disoccluded object at depth 2 (right). In the case of the third hypothesis, $H_3$, both objects are considered to be at the same depth and completely visible in the original image (no disocclusion is applied). In the lower part of each hypothesis we show the values $\tilde{p}(I/H_i)$, $\tilde{p}(H_2)$ (proportional, respectively, to the likelihood and prior probabilities), and the posterior probability $p(H_i/I)$. The probability value of the preferred hypothesis $H_P$ is highlighted in boldface.}
\label{tab:alltogether8}
\end{longtable}
\subsection{Discussion of failure cases}
\label{failures}

This section is devoted to present and discuss situations where our method can fail. 
Let us first comment on the experiments on synthetic images. Example 1 and Example 2 of Table~\ref{tab:alltogether1_fail} do not agree with human perception. As commented before,  in a T-junction, the occluder is the surface on the T-head side while the surfaces on the T-stem side continue behind the occluder. 
However, in these two results, the local occlusion signatures given by the T-junctions indicate that there is an occluded square which continues behind an incomplete disk (the occluder). Taking this into account, our method fails to give the hypothesis that agrees with the T-junction cues (which should be $H_2$). In Example 1, the likelihood of the hypothesis $H_1$ and $H_2$ are similar but the global complexity of the shapes in $H_1$ is smaller (thus higher prior) than the global complexity of the shape in $H_2$. In particular, any of the two shapes present in $H_1$ is simpler that any of the two shapes in $H_2$.
Regarding Example 2, the highly irregular contour of the shapes makes difficult a straightforward analysis and the final chosen hypothesis is due to a balance among the corresponding complexities and likelihoods. 
 In the examples of Table~\ref{tab:alltogether2_fail}, according to the local cues given by the T-junctions, the preferred option should be $H_1$. However, our method obtains $H_2$. In Example 3, although a higher prior due to a smaller complexity of the objects in $H_1$, the likelihood of $H_2$  is higher due to smaller elastica values in $H_2$. Example 4 is the opposite: $H_1$ presents a higher complexity (thus smaller prior) and a higher likelihood.

\setlength{\tabcolsep}{3pt}
\begin{longtable}{llllll}
 {\includegraphics[width = 1.75cm]{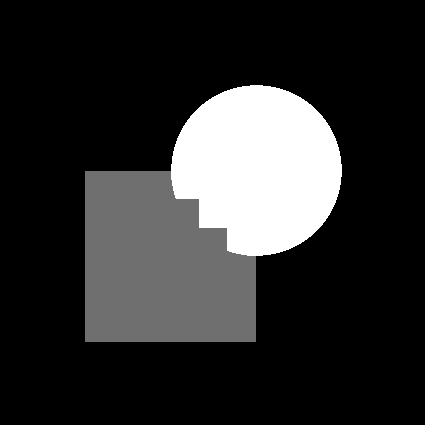}} & 
 \cellcolor{gray!25} \includegraphics[width = 1.75cm]{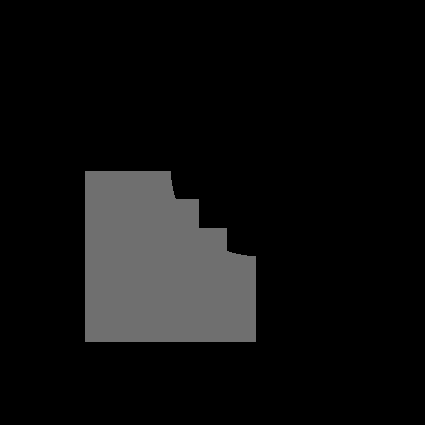} 
 \includegraphics[width = 1.75cm]{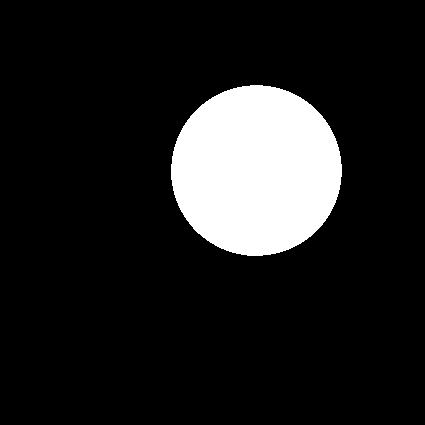} & &
 \cellcolor{gray!25} \includegraphics[width = 1.75cm]{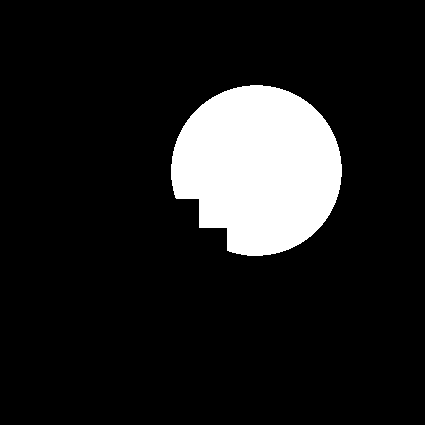} 
 \includegraphics[width = 1.75cm]{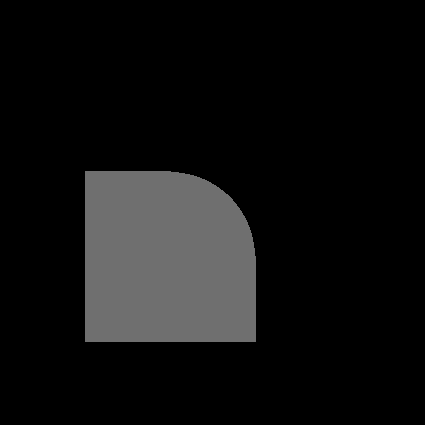} & &
 \cellcolor{gray!25} \includegraphics[width = 1.75cm]{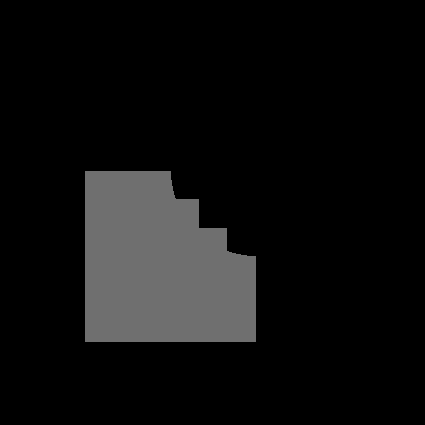} 
 \includegraphics[width = 1.75cm]{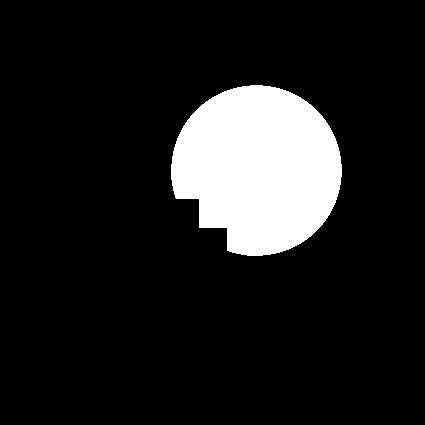} \\
 & \cellcolor{gray!25}$\tilde{p}(I/H_1)=0.6198$ & & \cellcolor{gray!25}$\tilde{p}(I/H_2)=0.4949$ & & \cellcolor{gray!25}$\tilde{p}(I/H_3)=0.3679$\\
 {\scriptsize{Example 1}} & \cellcolor{gray!25}$\tilde{p}(H_1)=0.6784$ & & \cellcolor{gray!25}$\tilde{p}(H_2)=0.3679$ & & \cellcolor{gray!25}$\tilde{p}(H_3)=0.4077$\\
 & \cellcolor{gray!25}$\mathbf{p(H_1/I)=0.5587}$ & & \cellcolor{gray!25}$p(H_2/I)=0.2420$ & & \cellcolor{gray!25}$p(H_3/I)=0.1993$\\ \\
 {\includegraphics[width = 1.75cm]{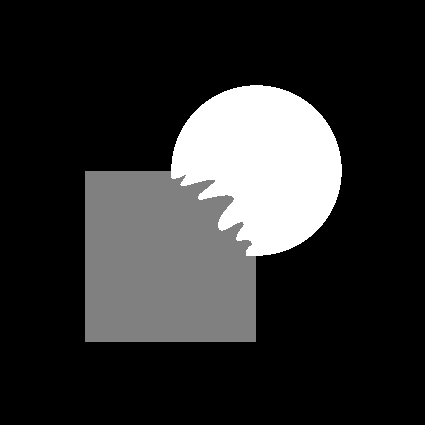}} & 
 \cellcolor{gray!25} \includegraphics[width = 1.75cm]{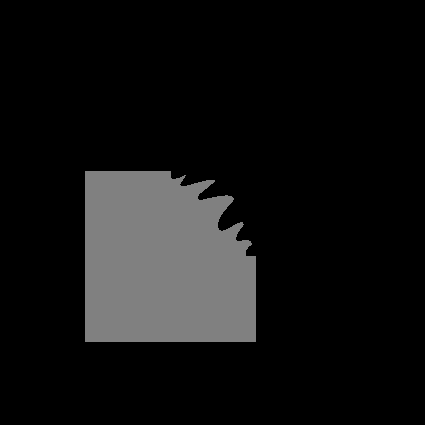} 
 \includegraphics[width = 1.75cm]{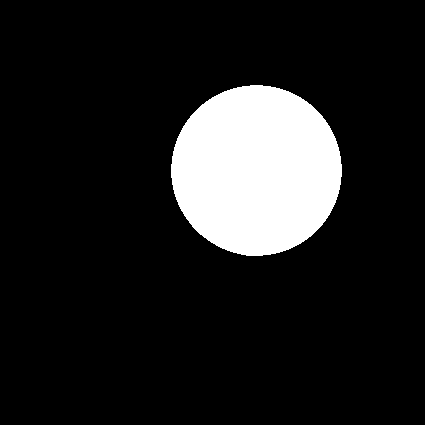} & &
 \cellcolor{gray!25} \includegraphics[width = 1.75cm]{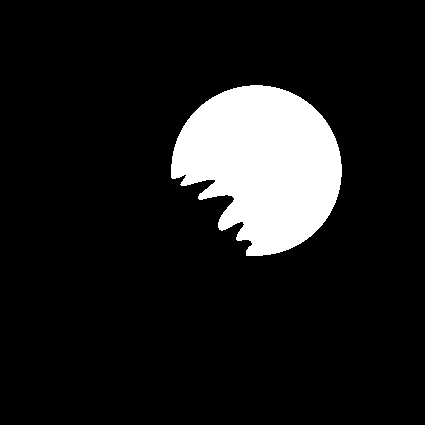} 
 \includegraphics[width = 1.75cm]{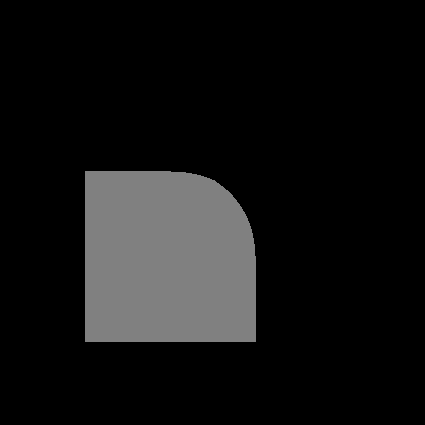} & &
 \cellcolor{gray!25} \includegraphics[width = 1.75cm]{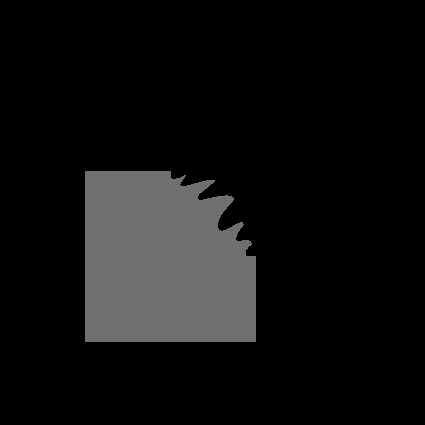} 
 \includegraphics[width = 1.75cm]{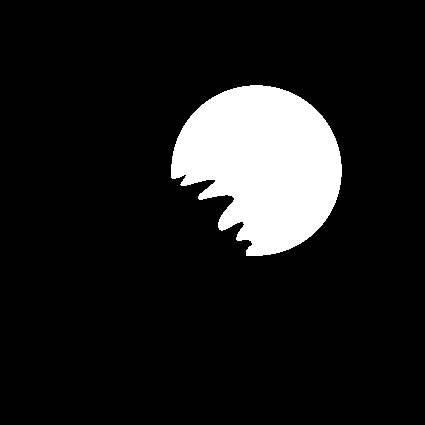} \\
 & \cellcolor{gray!25}$\tilde{p}(I/H_1)=0.6948$ & & \cellcolor{gray!25}$\tilde{p}(I/H_2)=0.6535$ & & \cellcolor{gray!25}$\tilde{p}(I/H_3)=0.3679$\\
  {\scriptsize{Example 2}}& \cellcolor{gray!25}$\tilde{p}(H_1)=0.3876$ & & \cellcolor{gray!25}$\tilde{p}(H_2)=0.4043$ & & \cellcolor{gray!25}$\tilde{p}(H_3)=0.3679$\\
 & \cellcolor{gray!25}$\mathbf{p(H_1/I)=0.4026}$ & & \cellcolor{gray!25}$p(H_2/I)=0.3951$ & & \cellcolor{gray!25}$p(H_3/I)=0.2023$\\ \\
  \caption{Synthetic experiments that fail. Each row shows a different experiment: the original image (proximal stimulus) is shown on the left and it is followed by the three different hypothesis (each one separated by a gray box). For the first two hypothesis, $H_i$ for $i=1,2$, we show: the object at depth 1 (left) and the disoccluded object at depth 2 (right). Notice that the object at depth 1 acts as a mask for disoccluding the object at depth 2. In the case of the third hypothesis, $H_3$, both objects are considered to be at the same depth and completely visible in the original image (no disocclusion is applied). 
  In the lower part of each hypothesis we show the values $\tilde{p}(I/H_i)$, $\tilde{p}(H_2)$ (proportional, respectively, to the likelihood and prior probabilities), and the posterior probability $p(H_i/I)$. The probability value of the preferred hypothesis $H_P$ is highlighted in boldface.}
\label{tab:alltogether1_fail}
\end{longtable}

\begin{longtable}{llll}
 {\includegraphics[width = 1.75cm]{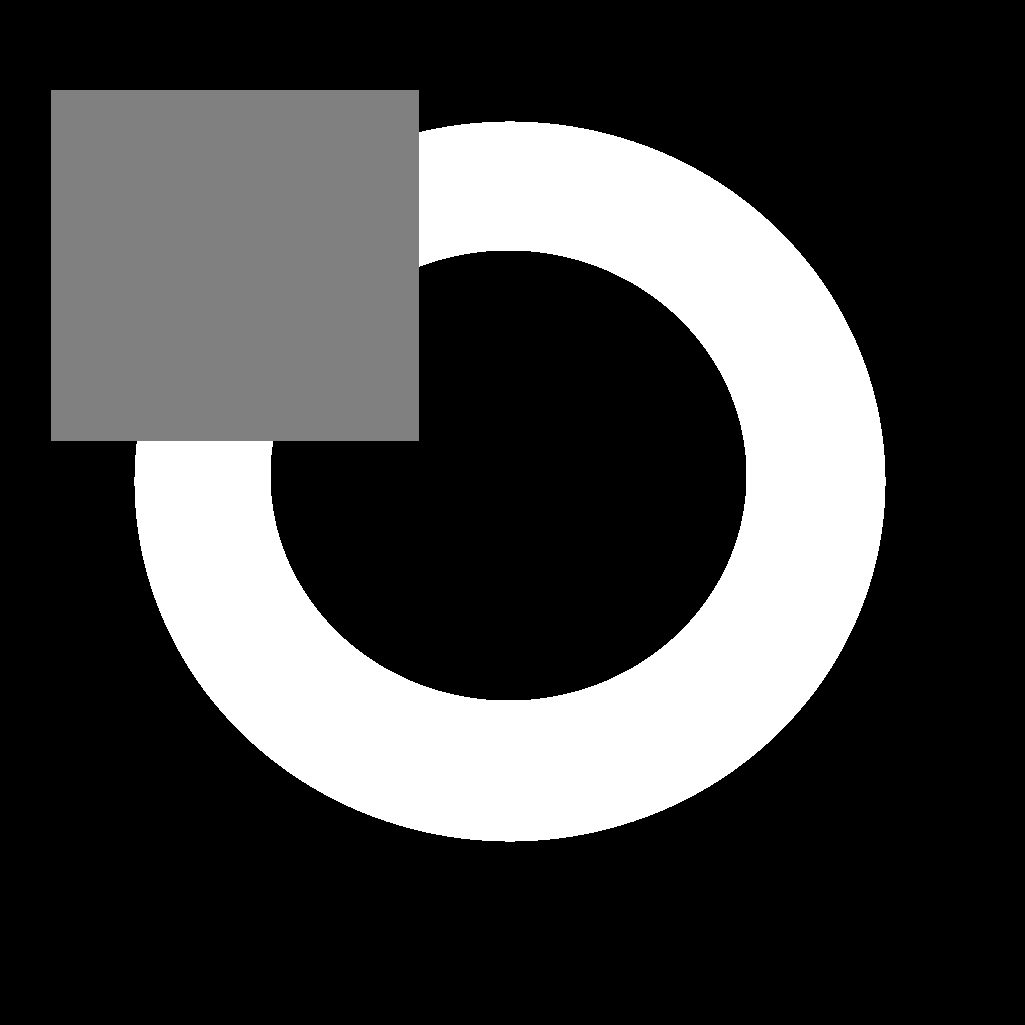}} & 
 \cellcolor{gray!25} \includegraphics[width = 1.75cm]{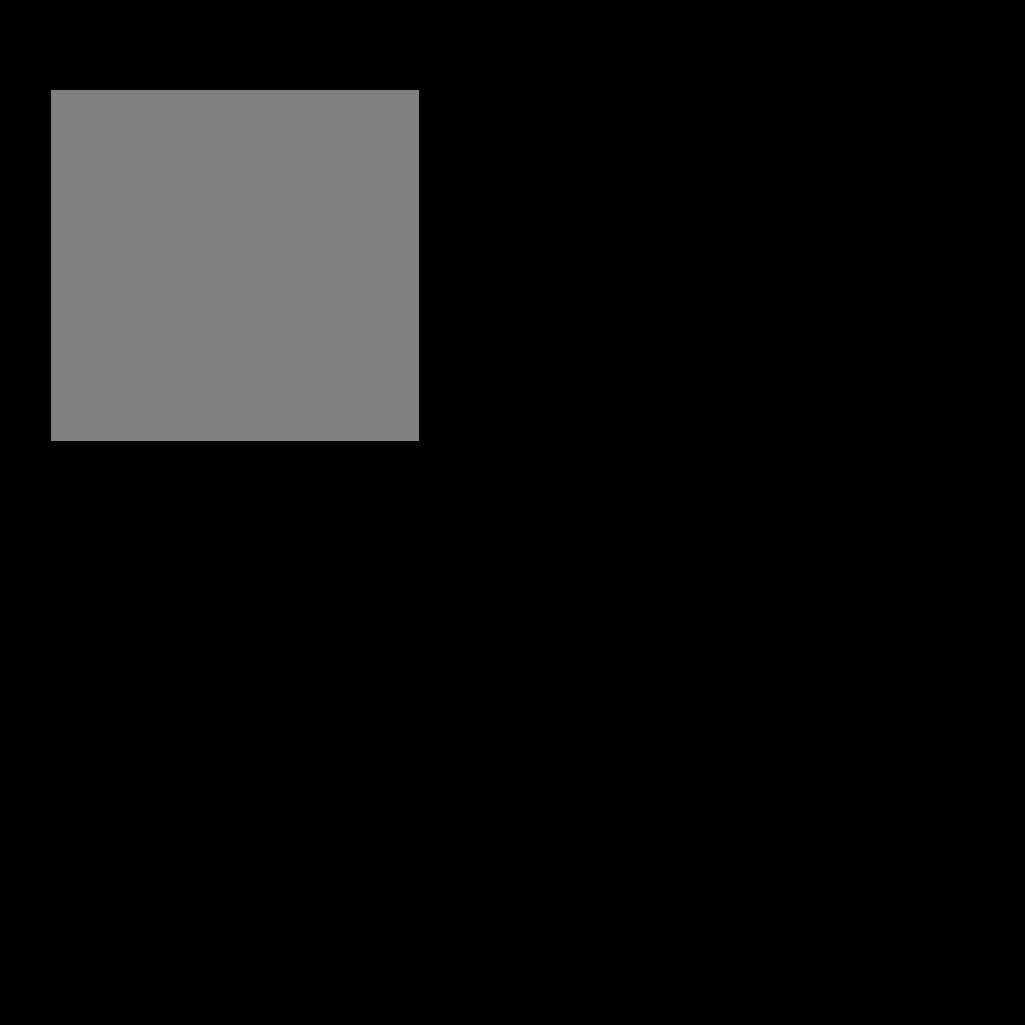} 
 \includegraphics[width = 1.75cm]{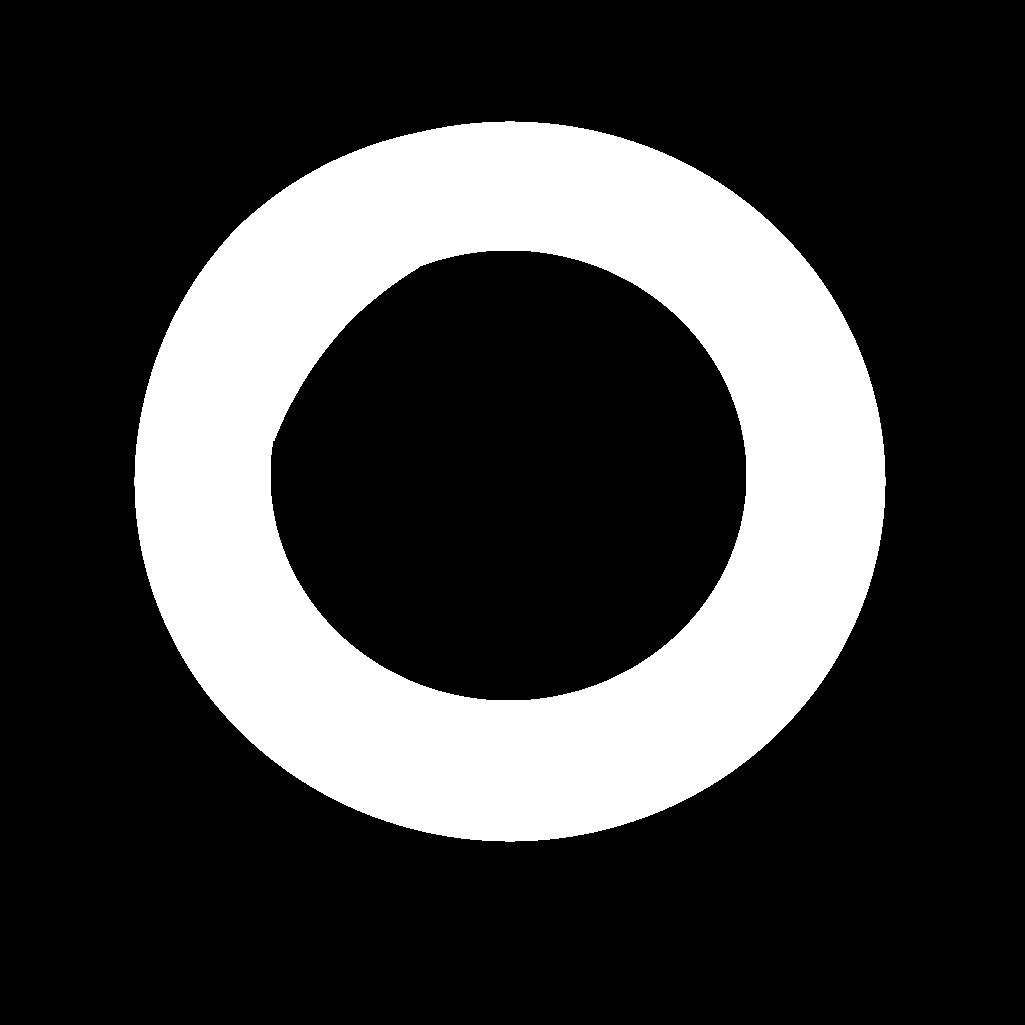} & &
 \cellcolor{gray!25} \includegraphics[width = 1.75cm]{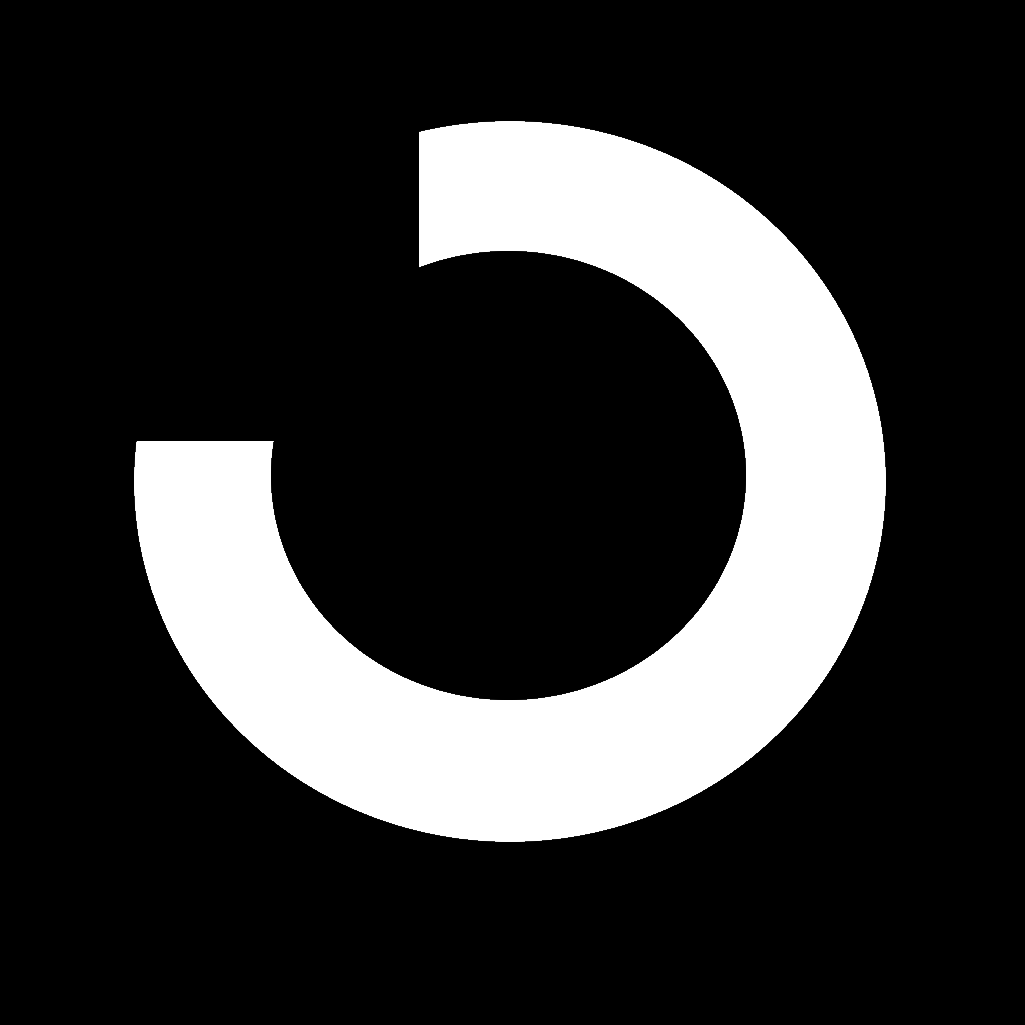} 
 \includegraphics[width = 1.75cm]{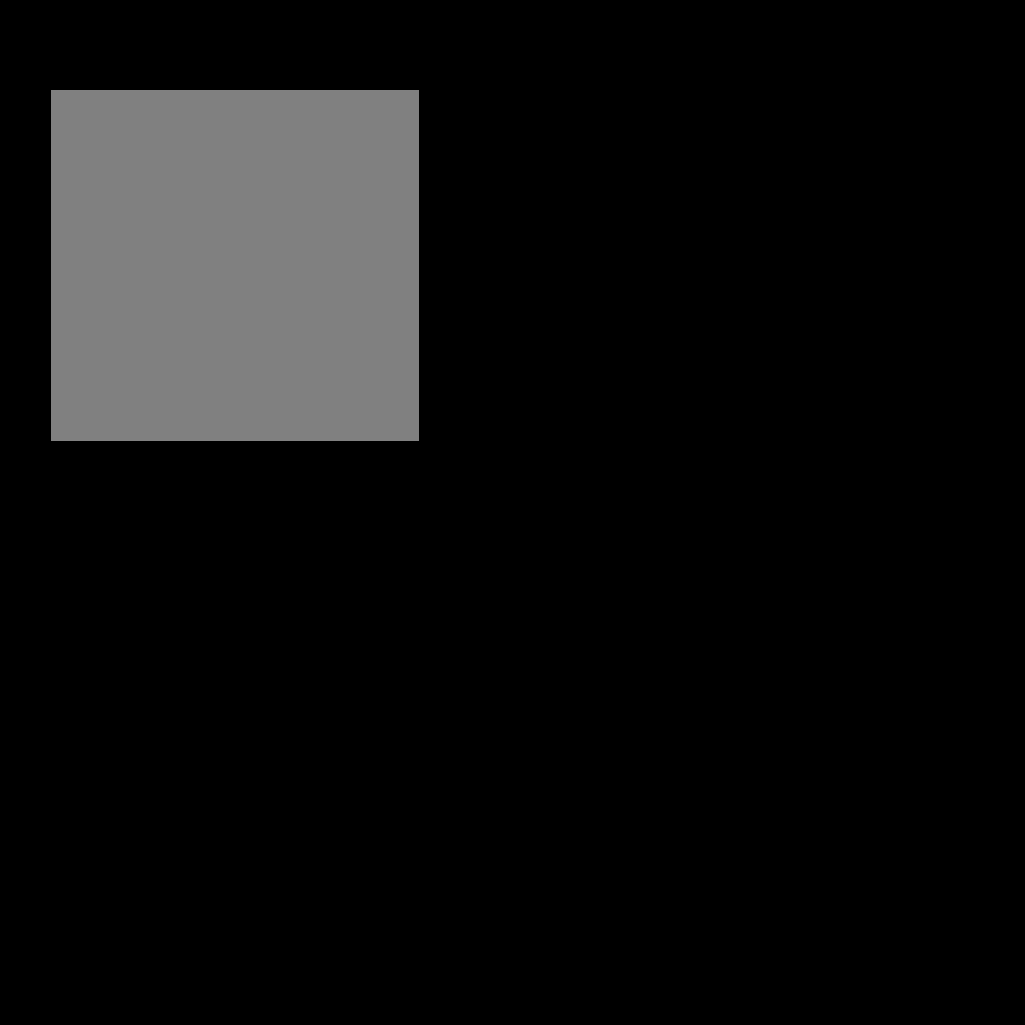} \\
 & \cellcolor{gray!25}$\tilde{p}(I/H_1)=0.3679$ & & \cellcolor{gray!25}$\tilde{p}(I/H_2)=0.4501$\\
  {\scriptsize{Example 3}}& \cellcolor{gray!25}$\tilde{p}(H_1)=0.3770$ & & \cellcolor{gray!25}$\tilde{p}(H_2)=0.3679$\\
 & \cellcolor{gray!25}$p(H_1/I)=0.2951$ & & \cellcolor{gray!25}$\mathbf{p(H_2/I)=0.3524}$\\ \\
 {\includegraphics[width = 1.75cm]{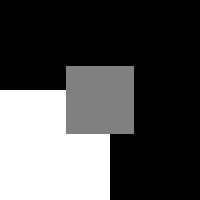}} & 
 \cellcolor{gray!25} \includegraphics[width = 1.75cm]{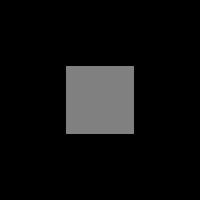} 
 \includegraphics[width = 1.75cm]{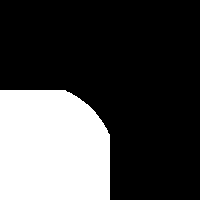} & &
 \cellcolor{gray!25} \includegraphics[width = 1.75cm]{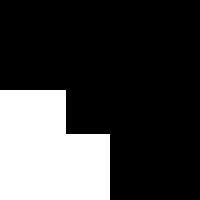} 
 \includegraphics[width = 1.75cm]{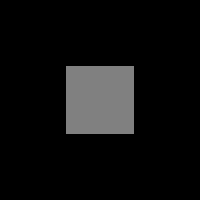} \\
 & \cellcolor{gray!25}$\tilde{p}(I/H_1)=0.4062$ & & \cellcolor{gray!25}$\tilde{p}(I/H_2)=0.3679$\\
  {\scriptsize{Example 4}}& \cellcolor{gray!25}$\tilde{p}(H_1)=0.3679$ & & \cellcolor{gray!25}$\tilde{p}(H_2)=0.5089$\\
 & \cellcolor{gray!25}${p(H_1/I)=0.2853}$ & & \cellcolor{gray!25}$p(H_2/I)=\mathbf{0.3574}$\\ \\
  \caption{Synthetic experiments that fail. Each row shows a different experiment: the original image (proximal stimulus) is shown on the left and it is followed by two different hypothesis (each one separated by a gray box). For each hypothesis $H_i$ we show: the object at depth 1 (left) and the disoccluded object at depth 2 (right). Notice that the object at depth 1 acts as a mask for disoccluding the object at depth 2. 
  In the lower part of each hypothesis we show the values $\tilde{p}(I/H_i)$, $\tilde{p}(H_2)$ (proportional, respectively, to the likelihood and prior probabilities), and the posterior probability $p(H_i/I)$. The probability value of the preferred hypothesis $H_P$ is highlighted in boldface. The third hypothesis, $H_3$, where both objects are considered to be at the same depth and completely visible in the original image (no disocclusion is applied) is not shown here because it coincides with $H_2$ (due to the fact that the disocclusion algorithm does not change the objects being disoccluded in $H_2$). More details are given in the text.}
\label{tab:alltogether2_fail}
\end{longtable}

\setlength{\tabcolsep}{3pt}
\begin{longtable}{lllllll}
 {\includegraphics[width = 1.6cm]{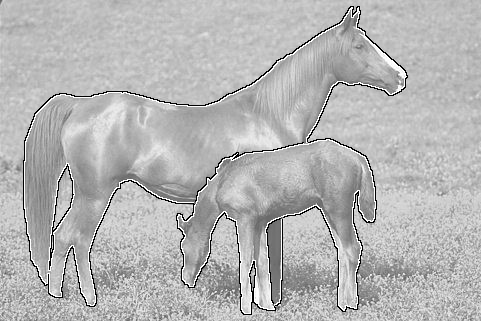}} & 
 {\includegraphics[width = 1.6cm]{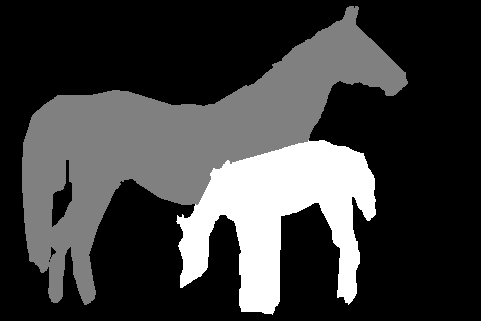}} & 
 \cellcolor{gray!25} \includegraphics[width = 1.6cm]{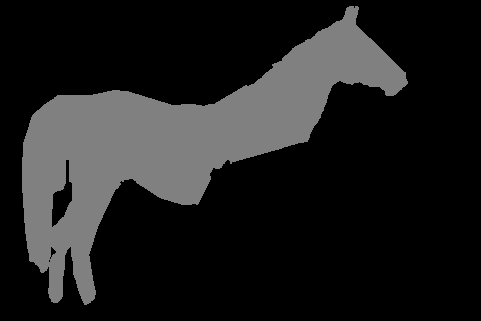} 
 \includegraphics[width = 1.6cm]{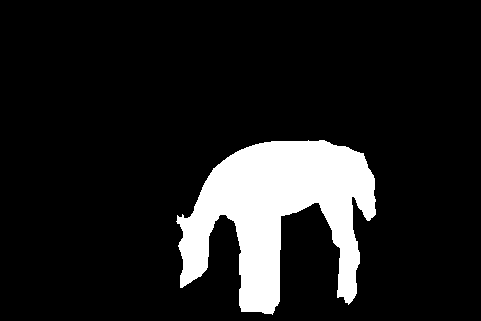} & &
 \cellcolor{gray!25} \includegraphics[width = 1.6cm]{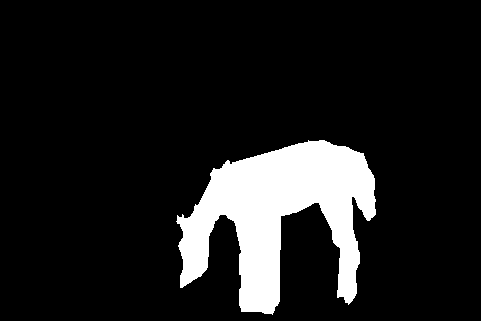} 
 \includegraphics[width = 1.6cm]{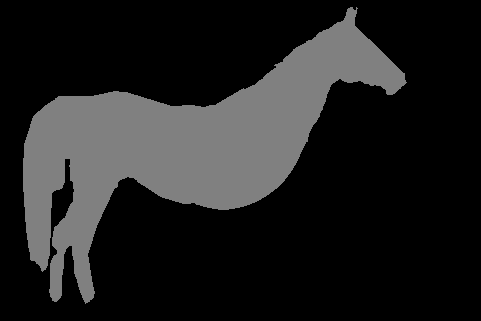} & &
 \cellcolor{gray!25} \includegraphics[width = 1.6cm]{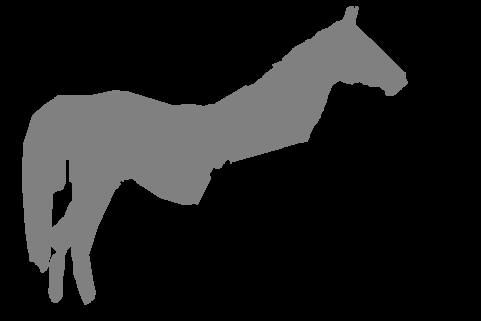} 
 \includegraphics[width = 1.6cm]{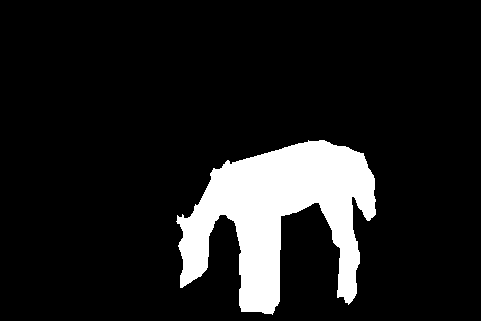} \\
 & & \cellcolor{gray!25}$\tilde{p}(I/H_1)=0.6858$ & & \cellcolor{gray!25}$\tilde{p}(I/H_2)=0.0.3726$ & & \cellcolor{gray!25}$\tilde{p}(I/H_3)=0.3679$\\
 \multicolumn{2}{c}{\scriptsize{Example 5}} & \cellcolor{gray!25}$\tilde{p}(H_1)=0.3726$ & & \cellcolor{gray!25}$\tilde{p}(H_2)=0.3704$ & & \cellcolor{gray!25}$\tilde{p}(H_3)=0.3679$\\
 & & \cellcolor{gray!25}$\mathbf{p(H_1/I)=0.3962}$ & & \cellcolor{gray!25}$p(H_2/I)=0.3939$ & & \cellcolor{gray!25}$p(H_3/I)=0.2099$\\ \\
 {\includegraphics[width = 1.6cm]{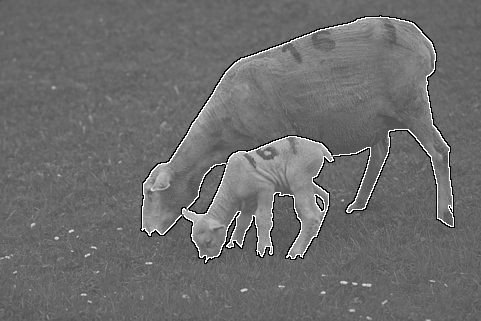}} & 
 {\includegraphics[width = 1.6cm]{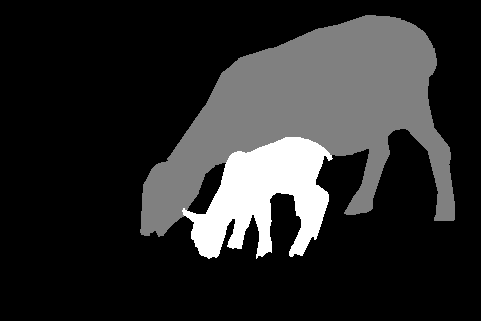}} & 
 \cellcolor{gray!25} \includegraphics[width = 1.6cm]{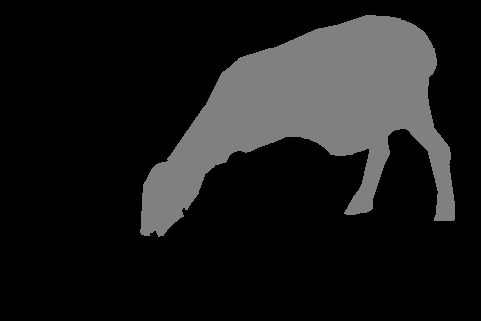} 
 \includegraphics[width = 1.6cm]{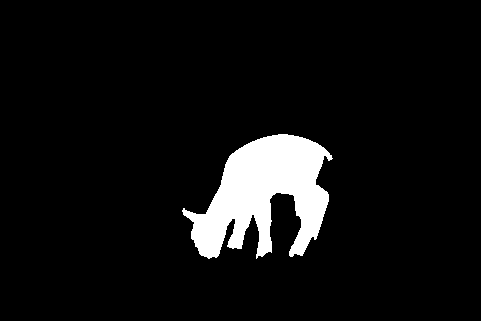} & &
 \cellcolor{gray!25} \includegraphics[width = 1.6cm]{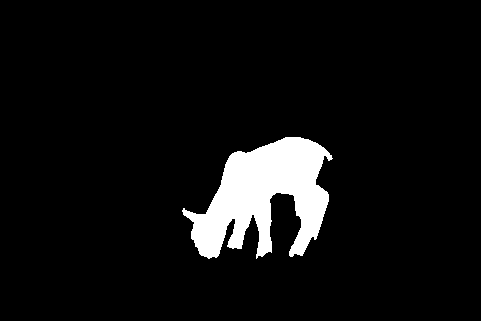} 
 \includegraphics[width = 1.6cm]{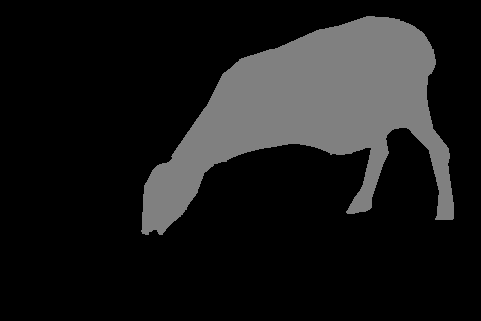} & &
 \cellcolor{gray!25} \includegraphics[width = 1.6cm]{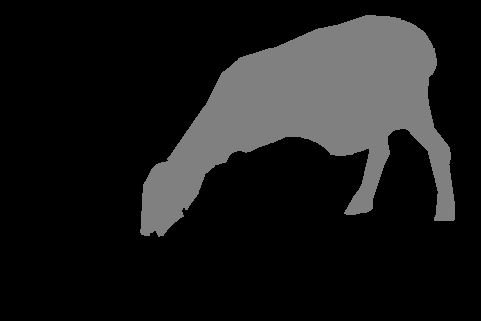} 
 \includegraphics[width = 1.6cm]{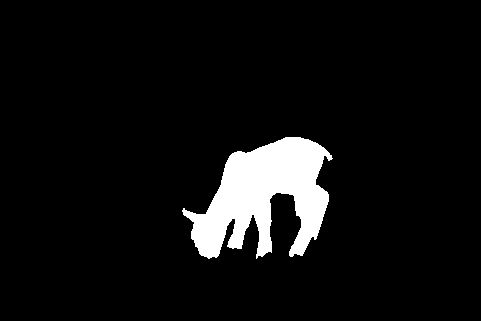} \\
 & & \cellcolor{gray!25}$\tilde{p}(I/H_1)=0.7469$ & & \cellcolor{gray!25}$\tilde{p}(I/H_2)=0.6515$ & & \cellcolor{gray!25}$\tilde{p}(I/H_3)=0.3679$\\
 \multicolumn{2}{c}{\scriptsize{Example 6}} & \cellcolor{gray!25}$\tilde{p}(H_1)=0.3686$ & & \cellcolor{gray!25}$\tilde{p}(H_2)=0.3695$ & & \cellcolor{gray!25}$\tilde{p}(H_3)=0.3679$\\
 & & \cellcolor{gray!25}$\mathbf{p(H_1/I)=0.4227}$ & & \cellcolor{gray!25}$p(H_2/I)=0.3696$ & & \cellcolor{gray!25}$p(H_3/I)=0.2078$\\ \\
 {\includegraphics[width = 1.6cm]{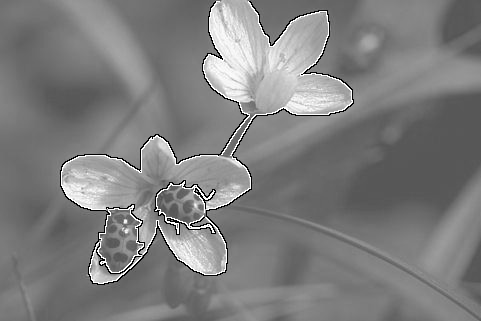}} & 
 {\includegraphics[width = 1.6cm]{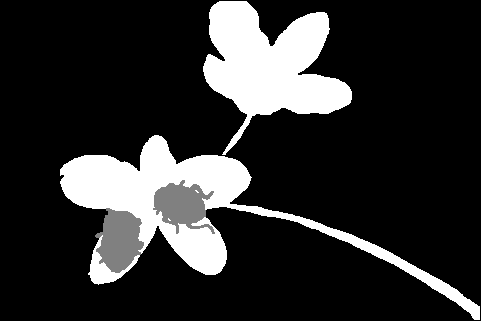}} & 
 \cellcolor{gray!25} \includegraphics[width = 1.6cm]{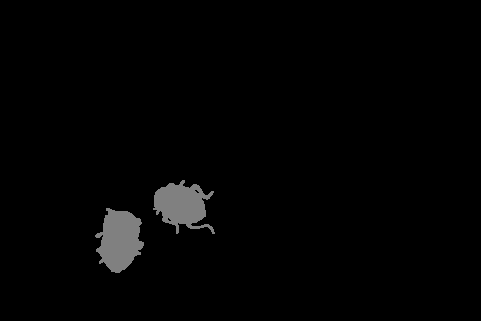} 
 \includegraphics[width = 1.6cm]{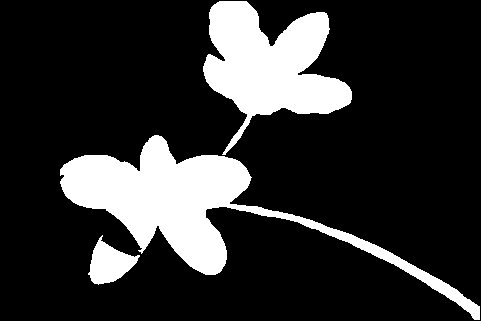} & &
 \cellcolor{gray!25} \includegraphics[width = 1.6cm]{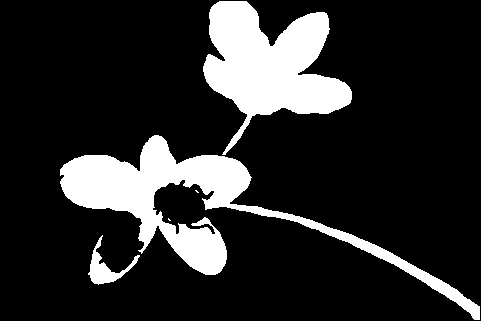} 
 \includegraphics[width = 1.6cm]{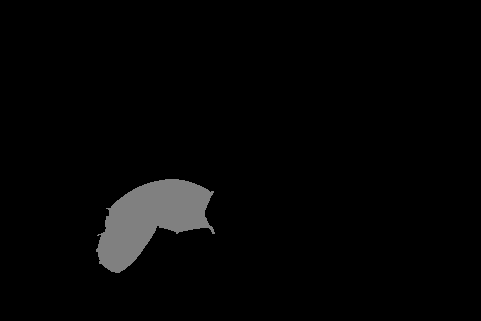} & &
 \cellcolor{gray!25} \includegraphics[width = 1.6cm]{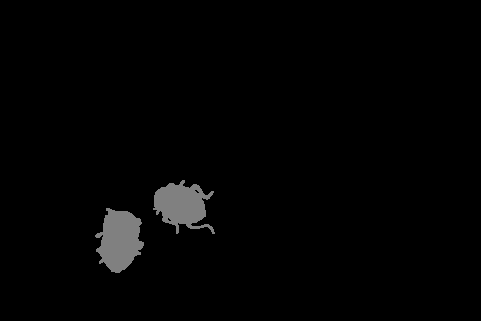} 
 \includegraphics[width = 1.6cm]{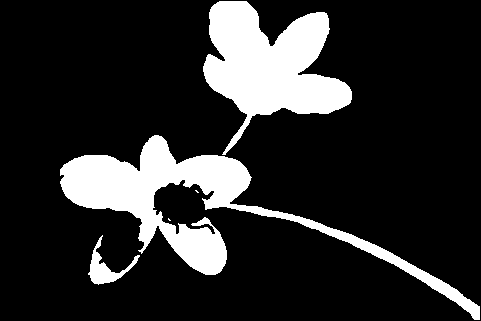} \\
 & & \cellcolor{gray!25}$\tilde{p}(I/H_1)=0.8575$ & & \cellcolor{gray!25}$\tilde{p}(I/H_2)=0.7297$ & & \cellcolor{gray!25}$\tilde{p}(I/H_3)=0.3679$\\
 \multicolumn{2}{c}{\scriptsize{Example 7}} & \cellcolor{gray!25}$\tilde{p}(H_1)=0.3709$ & & \cellcolor{gray!25}$\tilde{p}(H_2)=0.5332$ & & \cellcolor{gray!25}$\tilde{p}(H_3)=0.3679$\\
 & & \cellcolor{gray!25}$p(H_1/I)=0.3709$ & & \cellcolor{gray!25}$\mathbf{p(H_2/I)=0.5332}$ & & \cellcolor{gray!25}$p(H_3/I)=0.3679$\\ \\
 {\includegraphics[width = 1.6cm]{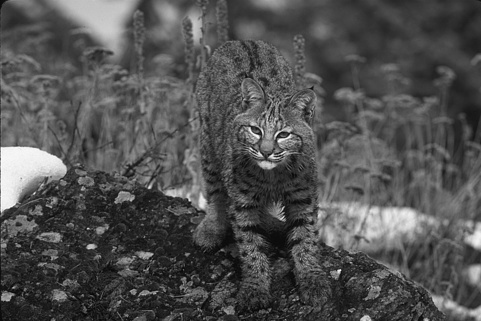}} & 
 {\includegraphics[width = 1.6cm]{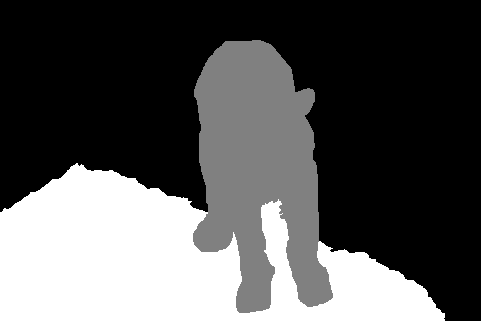}} & 
 \cellcolor{gray!25} \includegraphics[width = 1.6cm]{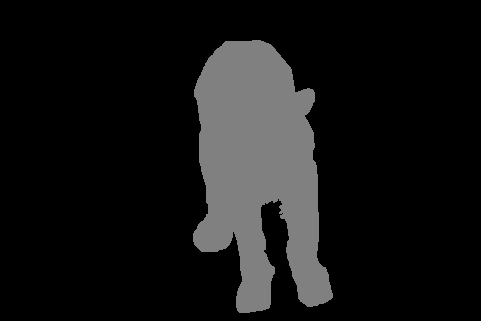} 
 \includegraphics[width = 1.6cm]{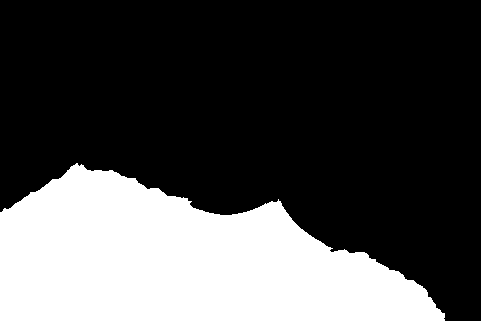} & &
 \cellcolor{gray!25} \includegraphics[width = 1.6cm]{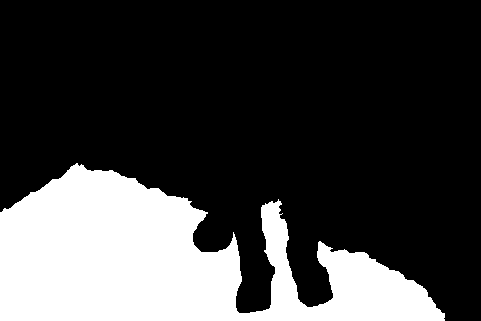} 
 \includegraphics[width = 1.6cm]{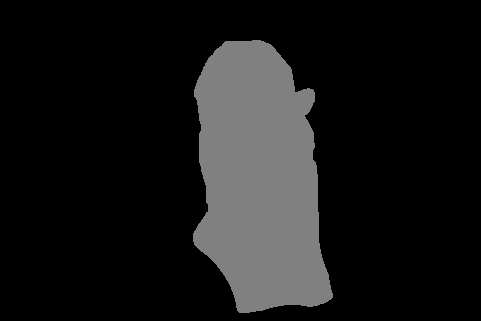} & &
 \cellcolor{gray!25} \includegraphics[width = 1.6cm]{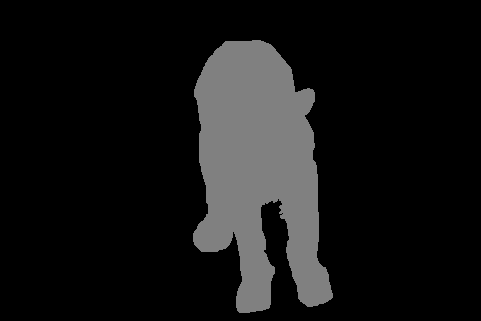} 
 \includegraphics[width = 1.6cm]{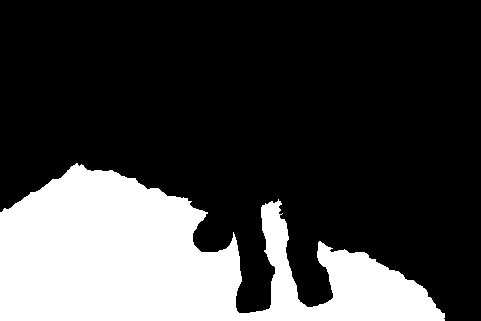} \\
 & & \cellcolor{gray!25}$\tilde{p}(I/H_1)=0.8732$ & & \cellcolor{gray!25}$\tilde{p}(I/H_2)=0.8325$ & & \cellcolor{gray!25}$\tilde{p}(I/H_3)=0.3679$\\
 \multicolumn{2}{c}{\scriptsize{Example 8}} & \cellcolor{gray!25}$\tilde{p}(H_1)=0.3679$ & & \cellcolor{gray!25}$\tilde{p}(H_2)=0.3792$ & & \cellcolor{gray!25}$\tilde{p}(H_3)=0.3679$\\
 & & \cellcolor{gray!25}${p(H_1/I)=0.4115}$ & & \cellcolor{gray!25}$\mathbf{p(H_2/I)=0.4118}$ & & \cellcolor{gray!25}$p(H_3/I)=0.1767$\\ \\
 {\includegraphics[width = 1.6cm]{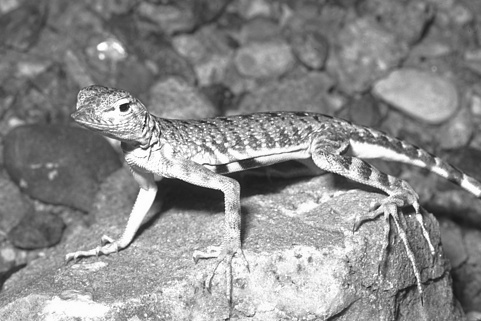}} & 
 {\includegraphics[width = 1.6cm]{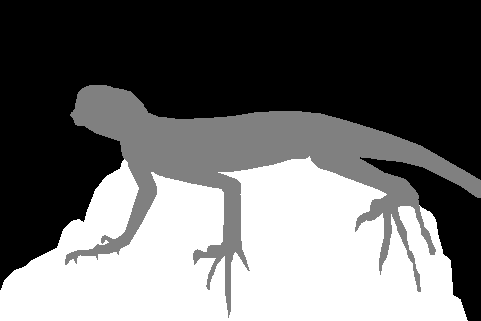}} & 
 \cellcolor{gray!25} \includegraphics[width = 1.6cm]{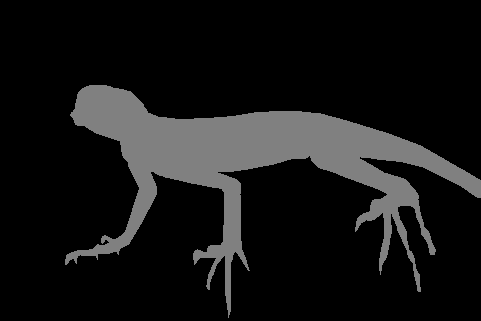} 
 \includegraphics[width = 1.6cm]{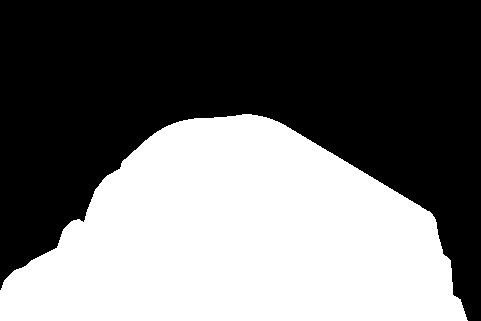} & &
 \cellcolor{gray!25} \includegraphics[width = 1.6cm]{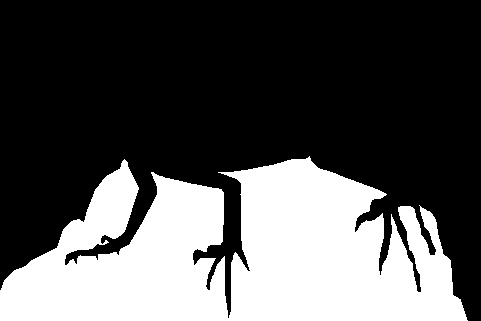} 
 \includegraphics[width = 1.6cm]{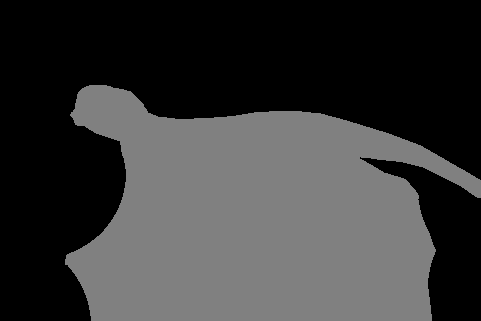} & &
 \cellcolor{gray!25} \includegraphics[width = 1.6cm]{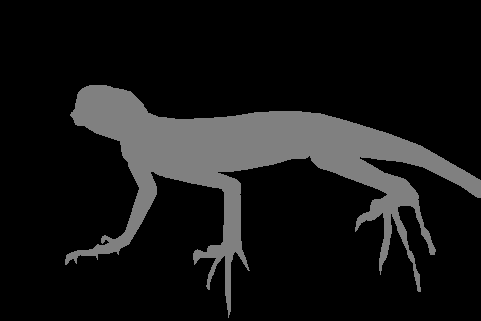} 
 \includegraphics[width = 1.6cm]{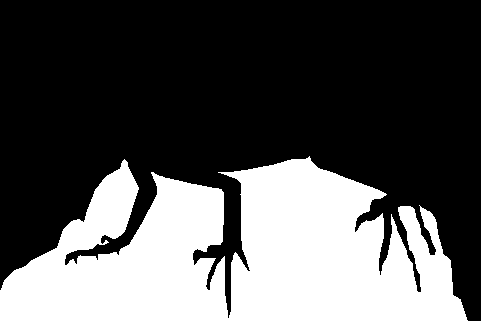} \\
 & & \cellcolor{gray!25}$\tilde{p}(I/H_1)=0.8744$ & & \cellcolor{gray!25}$\tilde{p}(I/H_2)=0.9118$ & & \cellcolor{gray!25}$\tilde{p}(I/H_3)=0.3679$\\
 \multicolumn{2}{c}{\scriptsize{Example 9}} & \cellcolor{gray!25}$\tilde{p}(H_1)=0.3763$ & & \cellcolor{gray!25}$\tilde{p}(H_2)=0.3747$ & & \cellcolor{gray!25}$\tilde{p}(H_3)=0.3679$\\
 & & \cellcolor{gray!25}$p(H_1/I)=0.4082$ & & \cellcolor{gray!25}$p\mathbf{(H_2/I)=0.4239}$ & & \cellcolor{gray!25}$p(H_3/I)=0.1679$\\ \\
 {\includegraphics[width = 1.6cm]{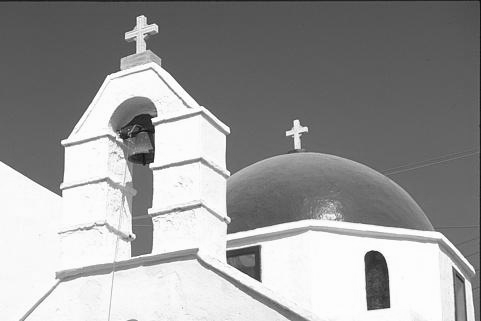}} & 
 {\includegraphics[width = 1.6cm]{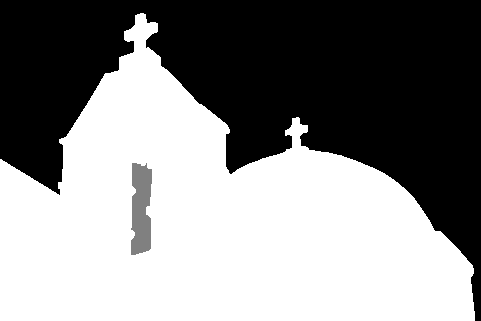}} & 
 \cellcolor{gray!25} \includegraphics[width = 1.6cm]{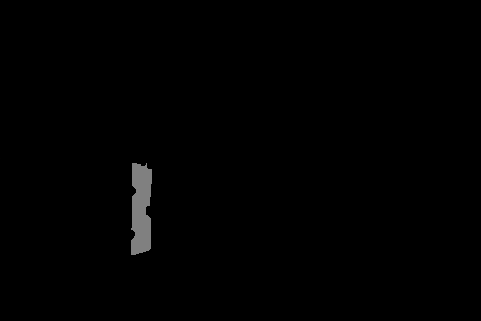} 
 \includegraphics[width = 1.6cm]{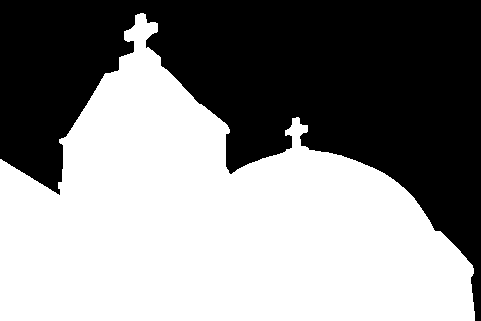} & &
 \cellcolor{gray!25} \includegraphics[width = 1.6cm]{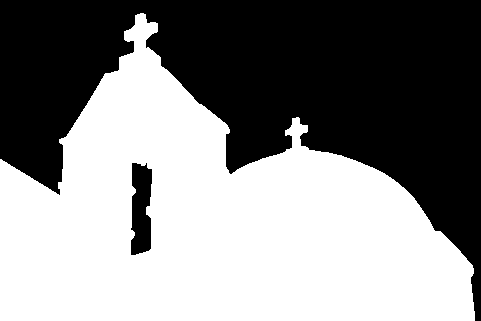} 
 \includegraphics[width = 1.6cm]{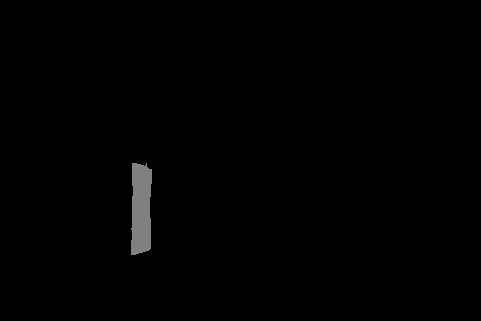} & &
 \cellcolor{gray!25} \includegraphics[width = 1.6cm]{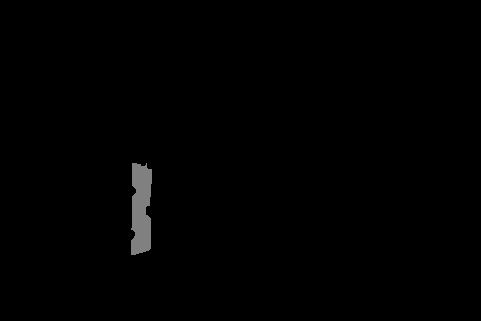} 
 \includegraphics[width = 1.6cm]{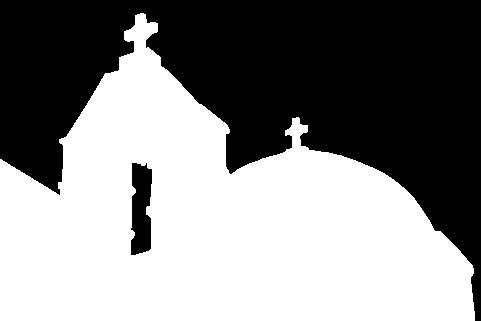} \\
 & & \cellcolor{gray!25}$\tilde{p}(I/H_1)=1$ & & \cellcolor{gray!25}$\tilde{p}(I/H_2)=0.8920$ & & \cellcolor{gray!25}$\tilde{p}(I/H_3)=0.3679$\\
 \multicolumn{2}{c}{\scriptsize{Example 10}} & \cellcolor{gray!25}$\tilde{p}(H_1)=0.3679$ & & \cellcolor{gray!25}$\tilde{p}(H_2)=0.4117$ & & \cellcolor{gray!25}$\tilde{p}(H_3)=0.3758$\\
 & & \cellcolor{gray!25}${\mathbf{p(H_1/I)=0.4213}}$ & & \cellcolor{gray!25}${p(H_2/I)=0.4205}$ & & \cellcolor{gray!25}$p(H_3/I)=0.1583$\\ \\
 {\includegraphics[width = 1.6cm]{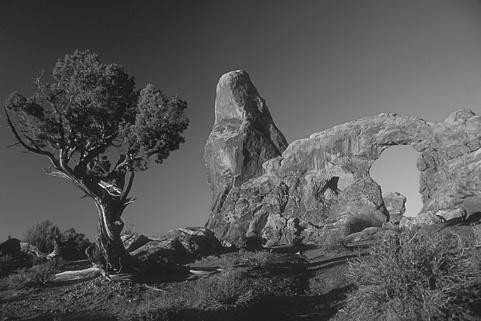}} & 
 {\includegraphics[width = 1.6cm]{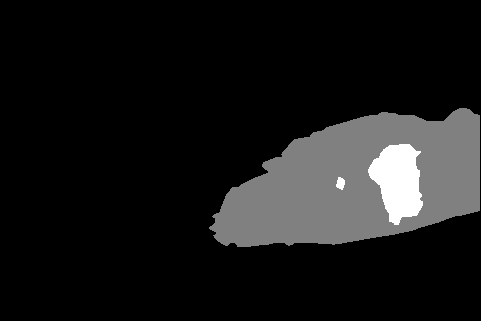}} & 
 \cellcolor{gray!25} \includegraphics[width = 1.6cm]{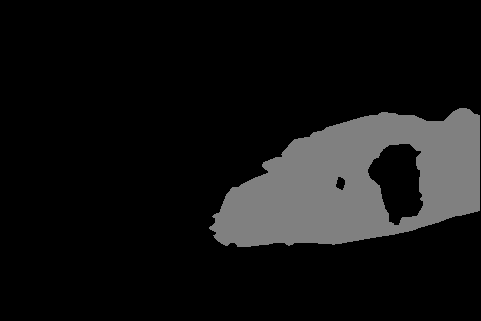} 
 \includegraphics[width = 1.6cm]{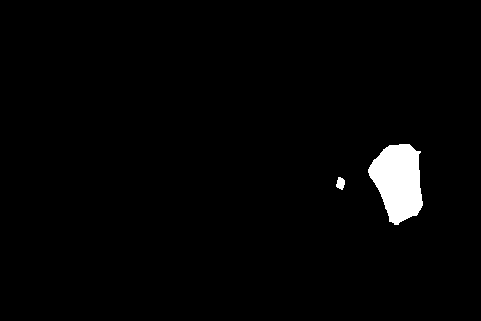} & &
 \cellcolor{gray!25} \includegraphics[width = 1.6cm]{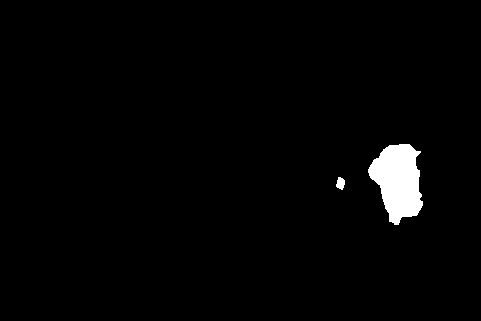} 
 \includegraphics[width = 1.6cm]{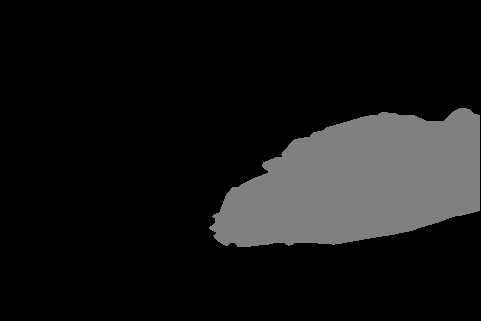} & &
 \cellcolor{gray!25} \includegraphics[width = 1.6cm]{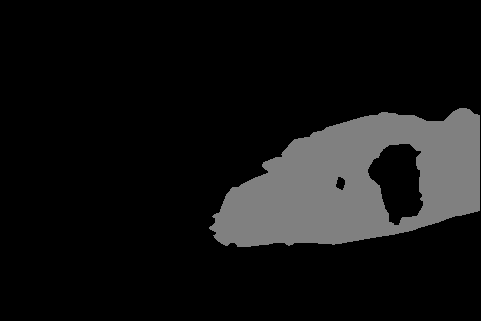} 
 \includegraphics[width = 1.6cm]{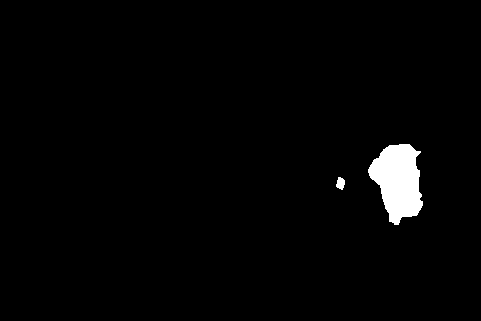} \\
 & & \cellcolor{gray!25}$\tilde{p}(I/H_1)=0.7878$ & & \cellcolor{gray!25}$\tilde{p}(I/H_2)=1$ & & \cellcolor{gray!25}$\tilde{p}(I/H_3)=0.3679$\\
 \multicolumn{2}{c}{\scriptsize{Example 11}} & \cellcolor{gray!25}$\tilde{p}(H_1)=0.3679$ & & \cellcolor{gray!25}$\tilde{p}(H_2)=0.3737$ & & \cellcolor{gray!25}$\tilde{p}(H_3)=0.3708$\\
 & & \cellcolor{gray!25}${p(H_1/I)=0.3623}$ & & \cellcolor{gray!25}$\mathbf{p(H_2/I)=0.4672}$ & & \cellcolor{gray!25}$p(H_3/I)=0.1705$\\ \\
\caption{Experiments with real images that fail  . Each row shows a different experiment: the two left-most images are, respectively, the original image and a segmentation of it, they are followed by the three different hypothesis (each one separated by a gray box). More details on the results shown for each hypothesis in Table \ref{tab:alltogether7}.}
\label{tab:alltogether5_fail}
\end{longtable}

Let us comment on the images of Table~\ref{tab:alltogether5_fail}.  Examples 5 and 6 reflect the same situation; the inpainting method is unable to recover the leg of the older horse or sheep. In any case, the difference among the posterior probabilities of the first two hypothesis is very small. On the other hand, in Examples 7 and 8, although according to the likelihood the preferred hypothesis is the correct one (e.g., two ladybugs in front of two flowers in Example 7), the complexity of the objects in the second hypothesis (flowers in front of ladybugs) is smaller (higher prior) because of the simplified completed object and this second hypothesis wins. 
In Example 9, the prior probabilities of $H_1$ and $H_2$ are similar but the likelihood of $H_2$ is slightly higher. Finally, Examples 10 and 11 show the same situation, where there appears a window showing the sky, which is behind. Our method fails in these cases, which are interpreted as small convex shapes over a biggest shape which is behind. Other example would be the arches of a bridge, which are further away compared to the bridge itself, which would be interpreted as closest by our method.

Finally, let us observe that the hypothesis of several objects fitting together (called $H_3$ in this work) implies an ambiguity in depth. When $H_3$ coincides with one of the other hypothesis, we assume that both objects appear at the same depth, but this is not always  true (although this proximal configuration would probably have smaller likelihood due to its oddity). 

\section{{Conclusions}}\label{sec:conclusions}
We have proposed a computational model of amodal completion that allows to compute the most preferred structure of a scene given a still image of it. 
As we are considering scenes where objects appear at two different depths, we  take into account the three different hypothesis: Object A occluding object B, B occluding A, or A and B fitting together forming a mosaic.  Our main contribution is  a Bayesian probabilistic model based on the elastica and the global complexity of the hypothesized objects in order to choose the most preferred explanation of the image. This explanation includes both the disoccluded objects that form the scene and their ordering according to depth.
Furthermore, we have proposed a disocclusion method, to compute the hypothesized objects, based on human visual completion, which is modeled by a binary inpainting method based on the Euler's elastica and that takes into account perceptual findings related to amodal completion, such as relatability, convexity, and good continuation.
Finally, we have shown the capability of our method with numerical experiments, both with real and synthetic images.

As future work, we plan to extend the approach to scenes with more than two depth layers. Furthermore, we plan to incorporate other disocclusion strategies (such as, e.g., exemplar-based methods~\cite{inpainting_aujol_2010,inpainting_arias_2011} 
or~\cite{HayashiSasaki}) allowing to model global completions taking into account properties such as symmetries or repetitions. Last but not least, 
we are also interested in the extension of the model to video sequences.

\paragraph{Acknowledgements.}
The first, second and fifth authors acknowledge partial support by MICINN project, reference  MTM2012-30772, and by GRC reference 2014 SGR 1301, Generalitat de Catalunya. The third author is supported by a Beatriu de Pin\'os fellowship (Marie-Curie COFUND action).
The final publication is available at Springer via \url{http://dx.doi.org/10.1007/s10851-016-0652-x}

\bibliographystyle{abbrv}
\bibliography{refs}{}  

\begin{thebibliography}{10}

\bibitem{inpainting_arias_2011}
P.~Arias, G.~Facciolo, V.~Caselles, and G.~Sapiro.
\newblock A variational framework for exemplar-based image inpainting.
\newblock {\em International Journal of Computer Vision}, 93:319--347, 2011.

\bibitem{inpainting_aujol_2010}
J.-F. Aujol, S.~Ladjal, and S.~Masnou.
\newblock Exemplar-based inpainting from a variational point of view.
\newblock {\em SIAM Journal on Mathematical Analysis}, 42(3):1246--1285, 2010.

\bibitem{BBCSV-01}
C.~Ballester, M.~Bertalm{\'{\i}}o, V.~Caselles, G.~Sapiro, and J.~Verdera.
\newblock Filling-in by joint interpolation of vector fields and gray levels.
\newblock {\em IEEE Trans. on IP}, 10(8):1200--1211, 2001.

\bibitem{BelletiniEtAl}
G.~Bellettini, G.~Dal~Maso, and M.~Paolini.
\newblock Semicontinuity and relaxation properties of a curvature depending
  functional in 2d.
\newblock {\em Annali della Scuola Normale Superiore di Pisa-Classe di
  Scienze}, 20(2):247--297, 1993.

\bibitem{burge2010natural}
J.~Burge, C.~Fowlkes, and M.~Banks.
\newblock Natural-scene statistics predict how the figure--ground cue of
  convexity affects human depth perception.
\newblock {\em The Journal of Neuroscience}, 30(21):7269--7280, 2010.

\bibitem{inpaintingCao2011}
F.~Cao, Y.~Gousseau, S.~Masnou, and P.~Pérez.
\newblock Geometrically guided exemplar-based inpainting.
\newblock {\em SIAM Journal on Imaging Sciences}, 4(4):1143--1179, 2011.

\bibitem{carrigan2015differentiating}
S.~Carrigan, E.~Palmer, and P.~Kellman.
\newblock Differentiating local and global processes in amodal completion
  through dot localization.
\newblock {\em Journal of vision}, 15(12):1123--1123, 2015.

\bibitem{inpaintingTVchan2001}
T.~Chan and J.~H. Shen.
\newblock Mathematical models for local nontexture inpaintings.
\newblock {\em SIAM J. App. Math.}, 62(3):1019--43, 2001.

\bibitem{ChanShen}
T.~Chan and J.~H. Shen.
\newblock Nontexture inpainting by curvature-driven diffusions.
\newblock {\em Journal of Visual Communication and Image Representation},
  12(4):436--449, 2001.

\bibitem{complexity}
Y.~Chen and H.~Sundaram.
\newblock Estimating complexity of 2d shapes.
\newblock In {\em IEEE 7th Workshop on Multimedia Signal Processing}, pages
  1--4, 2005.

\bibitem{CittiSarti2006}
G.~Citti and A.~Sarti.
\newblock A cortical based model of perceptual completion in the
  roto-translation space.
\newblock {\em Journal of Mathematical Imaging and Vision}, 24(3):307--326,
  2006.

\bibitem{inpainting_criminisi_2004}
A.~Criminisi, P.~P{\'e}rez, and K.~Toyama.
\newblock Region filling and object removal by exemplar-based inpainting.
\newblock {\em IEEE Trans. on IP}, 13(9):1200--1212, 2004.

\bibitem{inpainting_demanet_2003}
L.~Demanet, B.~Song, and T.~Chan.
\newblock Image inpainting by correspondence maps: a deterministic approach.
\newblock {\em Applied and Computational Mathematics}, 1100:217--50, 2003.

\bibitem{esedoglu2008threshold}
S.~Esedoglu, S.~J. Ruuth, and R.~Tsai.
\newblock Threshold dynamics for high order geometric motions.
\newblock {\em Interfaces and Free Boundaries}, 10(3):263--282, 2008.

\bibitem{figuregroundBerkeley}
C.~Fowlkes, D.~Martin, and J.~Malik.
\newblock Local figure/ground cues are valid for natural images.
\newblock {\em Journal of Vision}, 7(8):1--9, 2007.

\bibitem{froyen2015modal}
V.~Froyen, N.~Kogo, M.~Singh, and J.~Feldman.
\newblock Modal and amodal shape completion.
\newblock {\em Journal of vision}, 15(12):321--321, 2015.

\bibitem{gerbino1987effect}
W.~Gerbino and D.~Salmaso.
\newblock The effect of amodal completion on visual matching.
\newblock {\em Acta Psychologica}, 65(1):25--46, 1987.

\bibitem{von2009toward}
R.~Grompone~von Gioi.
\newblock Toward a computational theory of perception.
\newblock {\em apcap 2009}, page~49, 2009.

\bibitem{GH-03}
R.~Grzibovskis and A.~Heintz.
\newblock A convolution-thresholding scheme for the willmore flow.
\newblock Technical Report~34.

\bibitem{SemanticBenchmark}
T.~N. J.~Z. H.~Li, J.~Cai.
\newblock A benchmark for semantic image segmentation.
\newblock In {\em ICME}, 2013.

\bibitem{HayashiSasaki}
T.~Hayashi and M.~Sasaki.
\newblock Contour completion of partly occluded skew-symmetry objects.
\newblock In {\em Multimedia (ISM), 2014 IEEE International Symposium on},
  pages 90--93, 2014.

\bibitem{kang2014illusory}
S.~H. Kang, W.~Zhu, and J.~Shen.
\newblock Illusory shapes via corner fusion.
\newblock {\em SIAM Journal on Imaging Sciences}, 7(4):1907--1936, 2014.

\bibitem{kanizsa1972amodal}
G.~Kanizsa.
\newblock Amodal completion and shrinking of visual fields.
\newblock {\em Studia Psychologica}, 1972.

\bibitem{Kanizsa1979}
G.~Kanizsa.
\newblock {\em Organization in vision: essays on Gestalt perception}.
\newblock NY, Praeger, 1979.

\bibitem{kanizsa1985seeing}
G.~Kanizsa.
\newblock Seeing and thinking.
\newblock {\em Acta Psychologica}, 59(1):23--33, 1985.

\bibitem{Kanizsa1991}
G.~Kanizsa.
\newblock {\em Vedere et pensare}.
\newblock Il Mulino, Bologna, 1991.

\bibitem{inpainting_kawai_2009}
N.~Kawai, T.~Sato, and N.~Yokoya.
\newblock Image inpainting considering brightness change and spatial locality
  of textures and its evaluation.
\newblock In {\em Advances in Image and Video Technology}, pages 271--282.
  2009.

\bibitem{KellmanShipley}
P.~Kellman and T.~Shipley.
\newblock A theory of visual interpolation in object perception.
\newblock {\em Cognitive psychology}, 23(2):141--221, 1991.

\bibitem{knill1996}
D.~Knill, D.~Kersten, and A.~Yuille.
\newblock Introduction: A bayesian formulation of visual perception.
\newblock In D.~Knill and W.~Richards, editors, {\em Perception as Bayesian
  inference}. Cambridge University Press, 1996.

\bibitem{koffka1935}
K.~Koffka.
\newblock {\em Principles of Gestalt psychology}.
\newblock London: Routledge and Kegan Paul, 1935.

\bibitem{SIT}
E.~Leeuwenberg and P.~A. van~der Helm.
\newblock {\em Structural Information Theory. The Simplicity of Visual Form}.
\newblock Cambridge University Press, 2012.

\bibitem{LeungMalik}
T.~Leung and J.~Malik.
\newblock Contour continuity in region based image segmentation.
\newblock In {\em Computer Vision ECCV}, pages 544--559. Springer, 1998.

\bibitem{mansfield_2011}
A.~Mansfield, M.~Prasad, C.~Rother, T.~Sharp, P.~Kohli, and L.~van Gool.
\newblock Transforming image completion.
\newblock In {\em Proceedings of BMVC}, pages 121.1--121.11, 2011.

\bibitem{MartinFTM01}
D.~Martin, C.~Fowlkes, D.~Tal, and J.~Malik.
\newblock A database of human segmented natural images and its application to
  evaluating segmentation algorithms and measuring ecological statistics.
\newblock In {\em Proc. 8th ICCV}, volume~2, pages 416--423, July 2001.

\bibitem{masnou2002disocclusion}
S.~Masnou.
\newblock Disocclusion: a variational approach using level lines.
\newblock {\em IEEE Trans. on IP}, 11(2):68--76, 2002.

\bibitem{masnoumorel1998}
S.~Masnou and J.-M. Morel.
\newblock Level lines based disocclusion.
\newblock In {\em Proc. of IEEE ICIP}, 1998.

\bibitem{meijster2002general}
A.~Meijster, J.~B. Roerdink, and W.~H. Hesselink.
\newblock A general algorithm for computing distance transforms in linear time.
\newblock In {\em Mathematical Morphology and its applications to image and
  signal processing}, pages 331--340. Springer, 2002.

\bibitem{MBO}
B.~Merriman, J.~Bence, and S.~Osher.
\newblock Diffusion generated motion by mean curvature.
\newblock In {\em J.E. Taylor, editor, Computational Crystal Growers Workshop},
  pages 73--83. American Mathematical Society, Providence, Rhode Island, 1992.
\newblock Also available as UCLA CAM Report 92-18, April1992.

\bibitem{michotte1991amodal}
A.~Michotte, G.~Thines, and G.~Crabb{\'e}.
\newblock Amodal completion of perceptual structures.
\newblock {\em Michotte’s experimental phenomenology of perception}, pages
  140--167, 1991.

\bibitem{moravec86}
L.~Moravec and J.~Beck.
\newblock Amodal completion: Simplicity is not the explanation.
\newblock {\em Bulletin of the Psychonomic Society}, 24(4):269--272, 1986.

\bibitem{Mumford}
D.~Mumford.
\newblock Elastica and computer vision.
\newblock In {\em Algebraic Geometry and its Applications}, pages 491--506.
  Springer, New York, 1994.

\bibitem{murray2004setting}
M.~Murray, D.~Foxe, D.~Javitt, and J.~Foxe.
\newblock Setting boundaries: brain dynamics of modal and amodal illusory shape
  completion in humans.
\newblock {\em The Journal of neuroscience}, 24(31):6898--6903, 2004.

\bibitem{NitzbergMumfordShiota}
M.~Nitzberg, D.~Mumford, and T.~Shiota.
\newblock {\em Filtering, segmentation, and depth}, volume 662.
\newblock Lecture notes in computer science, Springer, 1993.

\bibitem{otero2016computing}
I.~R. Otero and M.~Delbracio.
\newblock Computing an {E}xact {G}aussian {S}cale-{S}pace.
\newblock {\em Image Processing On Line}, 6:8--26, 2016.

\bibitem{ringach1996spatial}
D.~Ringach and R.~Shapley.
\newblock Spatial and temporal properties of illusory contours and amodal
  boundary completion.
\newblock {\em Vision research}, 36(19):3037--3050, 1996.

\bibitem{rubin2001role}
N.~Rubin.
\newblock The role of junctions in surface completion and contour matching.
\newblock {\em Perception}, 30(3):339--366, 2001.

\bibitem{ERT-05}
S.~R. S.~Esedoglu and R.~Tsai.
\newblock Threshold dynamics for shape reconstruction and disocclusion.
\newblock In {\em Proc of IEEE ICIP}, volume~2, pages 502--505, 2005.

\bibitem{Sekuler}
A.~B. Sekuler.
\newblock Local and global minima in visual completion: effects of symmetry and
  orientation.
\newblock {\em PERCEPTION-LONDON}, 23:529--529, 1994.

\bibitem{sethian1985curvature}
J.~A. Sethian.
\newblock Curvature and the evolution of fronts.
\newblock {\em Communications in Mathematical Physics}, 101(4):487--499, 1985.

\bibitem{SinghHoffman}
M.~Singh and D.~Hoffman.
\newblock Completing visual contours: The relationship between relatability and
  minimizing inflections.
\newblock {\em Perception \& Psychophysics}, 61(5):943--951, 1999.

\bibitem{ThornberWilliams}
K.~K. Thornber and L.~R. Williams.
\newblock Characterizing the distribution of completion shapes with corners
  using a mixture of random processes.
\newblock {\em Pattern Recognition}, 33(4):543--553, 2000.

\bibitem{van2011bayesian}
P.~van~der Helm.
\newblock Bayesian confusions surrounding simplicity and likelihood in
  perceptual organization.
\newblock {\em Acta psychologica}, 138(3):337--346, 2011.

\bibitem{van1995multiple}
R.~Van~Lier, E.~Leeuwenberg, and P.~Van~der Helm.
\newblock Multiple completions primed by occlusion patterns.
\newblock {\em PERCEPTION-LONDON-}, 24:727--727, 1995.

\bibitem{VanDerHelm1}
R.~van Lier, P.~van~der Helm, and E.~Leeuwenberg.
\newblock Integrating global and local aspects of visual occlusion.
\newblock {\em Perception}, 23:883--903, 1994.

\bibitem{van1995competing}
R.~Van~Lier, P.~Van~der Helm, and E.~Leeuwenberg.
\newblock Competing global and local completions in visual occlusion.
\newblock {\em Journal of Experimental Psychology: Human Perception and
  Performance}, 21(3):571, 1995.

\bibitem{lsd}
R.~G. von Gioi, J.~Jakubowicz, J.-M. Morel, and G.~Randall.
\newblock Lsd: a line segment detector.
\newblock {\em Image Processing On Line}, 2:35--55, 2012.

\bibitem{Wertheimer1923}
M.~Wertheimer.
\newblock {\em Untersuchungen zur Lehre von der Gestalt, II [Investigations in
  Gestalt Theory, II: Laws of organization in perceptual forms]}, pages
  301--350.
\newblock Psychologische Forschung, 1923.

\bibitem{inpainting_wexler_irani_2007}
Y.~Wexler, E.~Shechtman, and M.~Irani.
\newblock Space-time completion of video.
\newblock {\em IEEE Transactions on PAMI}, 29(3):463--476, 2007.

\bibitem{WilliamsJacobs}
L.~R. Williams and D.~W. Jacobs.
\newblock Stochastic completion fields: A neural model of illusory contour
  shape and salience.
\newblock {\em Neural computation}, 9(4):837--858, 1997.

\end{thebibliography}


@InProceedings{Yashtini2015,
    author  = {M. Yashtini, and S. Kang},
    title   = {Alternating Direction Method of Multiplier for Euler's Elastica-based Denoising},
    booktitle = {Proc. of the Fifth Int. Conf. on 
Scale Space and Variational Methods in Computer Vision},
    year    = {2015},
    address = {L\`ege cap Ferret, France}
}

@Article{Ambrosioetal,
  title={Connected components of sets of finite perimeter and applications to image processing},
  author={Ambrosio, L. and Caselles, V. and Masnou, S. and Morel, J.M.},
  journal={Journal of the European Mathematical Society},
  volume={3},
  number={1},
  pages={39--92},
  year={2001},
  publisher={Springer}
}

@Article{BBCSV-01,
   author = {C. Ballester and M. Bertalm{\'{\i}}o and V. Caselles  and G. Sapiro and J. Verdera},
   title = {Filling-In by joint interpolation of vector fields and gray levels},
   journal = {IEEE Trans. on IP},
   volume = {10},
   number = {8},
   pages = {1200--1211},
   year = {2001}
}

@Article{BelletiniEtAl,
   author = {Bellettini, G., and Dal Maso, G., and Paolini, M.},
   title = {Semicontinuity and relaxation properties of a curvature depending functional in 2D},
   journal = {Annali della Scuola Normale Superiore di Pisa-Classe di Scienze},
   volume = {20},
   number = {2},
   pages = {247--297},
   year = {1993}
}

@Article{CasellesCollMorel,
   author = {Caselles, V., Coll, B., Morel, J. M.},
   title = {Topographic maps and local contrast changes in natural imagess},
   volume = {33},
   number = {1},
   pages = {5--27},
   year = {1999},
   journal = {International Journal of Computer Vision}
}
@Article{ChanShen,
   author = {T. Chan and J. H. Shen},
   title = {Nontexture inpainting by curvature-driven diffusions},
   volume = {12},
   number = {4},
   pages = {436--449},
   year = {2001},
   journal = {Journal of Visual Communication and Image Representation}
}
@InProceedings{EfrosLeung,
    author  = {A.A. Efros and T.K. Leung},
    title   = {Texture Synthesis by Non-parametric Sampling},
    booktitle = {Proc. of the IEEE Conf. on CVPR},
    year    = {1999},
    address = {Corfu, Greece},
    pages   = {1033--1038}
}

@article{sethian1985curvature,
  title={Curvature and the evolution of fronts},
  author={Sethian, James A},
  journal={Communications in Mathematical Physics},
  volume={101},
  number={4},
  pages={487--499},
  year={1985},
  publisher={Springer}
}

@article{chambolle2004algorithm,
  title={An algorithm for total variation minimization and applications},
  author={Chambolle, Antonin},
  journal={Journal of Mathematical imaging and vision},
  volume={20},
  number={1-2},
  pages={89--97},
  year={2004},
  publisher={Springer}
}


@incollection{meijster2002general,
  title={A general algorithm for computing distance transforms in linear time},
  author={Meijster, Arnold and Roerdink, Jos BTM and Hesselink, Wim H},
  booktitle={Mathematical Morphology and its applications to image and signal processing},
  pages={331--340},
  year={2002},
  publisher={Springer}
}


@InProceedings{ERT-05,
  author = {S. Esedoglu, S. Ruuth and R. Tsai},
  title = {Threshold dynamics for shape reconstruction and disocclusion},
  booktitle = {Proc of IEEE ICIP},
  year = {2005},
  volume = {2},
  pages = {502--505}
}

@article{esedoglu2008threshold,
  title={Threshold dynamics for high order geometric motions},
  author={Esedoglu, Selim and Ruuth, Steven J and Tsai, Richard},
  journal={Interfaces and Free Boundaries},
  volume={10},
  number={3},
  pages={263--282},
  year={2008}
}

@inproceedings{MBO,
  author = {B. Merriman and J.K. Bence and S.  Osher},
  title = {Diffusion generated motion by mean curvature.},
  booktitle = {J.E. Taylor, editor, Computational Crystal Growers Workshop},
  organization = {American Mathematical Society, Providence, Rhode Island},
  year = {1992},
  pages = {73--83},
  note = {Also available as UCLA CAM Report 92-18, April1992}
}
@techreport{GH-03,
  title={A convolution-thresholding scheme for the Willmore flow},
  author = {R. Grzibovskis and A. Heintz},
  number = {34}}
}

@article{froyen2015modal,
  title={Modal and amodal shape completion.},
  author={Froyen, V. and Kogo, N. and Singh, M. and Feldman, J.},
  journal={Journal of vision},
  volume={15},
  number={12},
  pages={321--321},
  year={2015}
}


@article{carrigan2015differentiating,
  title={Differentiating Local and Global Processes in Amodal Completion Through Dot Localization.},
  author={Carrigan, S. and Palmer, E. and Kellman, P.},
  journal={Journal of vision},
  volume={15},
  number={12},
  pages={1123--1123},
  year={2015}
}

@Book{Kanizsa1979,
  title={Organization in vision: essays on Gestalt perception},
  author={Kanizsa, G.},
  volume={},
  year={1979},
  publisher={NY, Praeger}
}
@Book{Kanizsa1991,
  title={Vedere et pensare},
  author={Kanizsa, G.},
  volume={},
  year={1991},
  publisher={Il Mulino, Bologna}
}

@article{CittiSarti2006,
  title={A cortical based model of perceptual completion in the roto-translation space},
  author={Citti, G., and Sarti, A.},
  journal={Journal of Mathematical Imaging and Vision},
  volume={24},
  number={3},
  pages={307--326},
  year={2006}
}

@article{burge2010natural,
  title={Natural-scene statistics predict how the figure--ground cue of convexity affects human depth perception},
  author={Burge, J. and Fowlkes, C.C. and Banks, M.S.},
  journal={The Journal of Neuroscience},
  volume={30},
  number={21},
  pages={7269--7280},
  year={2010},
  publisher={Soc Neuroscience}
}

@inBook{Wertheimer1923,
  title={Untersuchungen zur Lehre von der Gestalt, II
[Investigations in Gestalt Theory, II: Laws of organization in perceptual
forms]},
  author={Wertheimer, M.},
 pages={301-350},
  year={1923},
  publisher={Psychologische Forschung}
}

@Article{KellmanShipley,
  author =   {P.J. Kellman and T.F. Shipley},
  title =        {A theory of visual interpolation in object perception},
  journal = 	 {Cognitive psychology},
  year = 	 {1991},
  volume = 	 {23(2)},
  pages = 	 {141-221}
}

@article{inpaintingTVchan2001,
   author = {T. Chan and J. H. Shen},
   title = {Mathematical models for local nontexture inpaintings},
   volume = {62},
   number = {3},
   pages = {1019-43},
   year = {2001},
   journal = {SIAM J. App. Math.}
}

@article{inpaintingCao2011,
	author = {Cao, F. and Gousseau, Y. and Masnou, S. and Pérez, P.},
	title = {Geometrically Guided Exemplar-Based Inpainting},
	journal = {SIAM Journal on Imaging Sciences},
	volume = {4},
	number = {4},
	pages = {1143-1179},
	year = {2011}
}

@InProceedings{masnoumorel1998,
   author = {S. Masnou and J.-M. Morel},
   title = {Level lines based disocclusion},
   year = {1998},
   booktitle = {Proc. of IEEE ICIP}
}
@Article{masnou2002disocclusion,
   author = {S. Masnou},
   title = {Disocclusion: a variational approach using level lines},
   journal = {IEEE Trans. on IP},
   volume = {11},
   number = {2},
   year = {2002},
   pages = {68--76},
   publisher = {IEEE Computer Society},
   address = {Washington, DC, USA}
}
@Book{NitzbergMumfordShiota,
  title={Filtering, segmentation, and depth},
  author={Nitzberg, M. and Mumford, D. and Shiota, T.},
  publisher={Lecture notes in computer science, Springer},
  volume={662},
  number = {5},
  pages = {1--3},
  year={1993}
}
@Book{Serra,
  title={Image analysis and mathematical morphology},
  author={Serra, J.},
  year={1982},
  publisher={Academic press}
}
@InProceedings{SerraSalembier,
  author = 	 {Serra, J. C., Salembier, P.},
  title =        {Connected operators and pyramids},
  booktitle =        {SPIE's 1993 International Symposium on Optics, Imaging, and Instrumentation},
  organization = {International Society for Optics and Photonics},
  year = 	 {1993}
}

@article{van2011bayesian,
  title={Bayesian confusions surrounding simplicity and likelihood in perceptual organization},
  author={van der Helm, P.A.},
  journal={Acta psychologica},
  volume={138},
  number={3},
  pages={337--346},
  year={2011},
  publisher={Elsevier}
}

@article{murray2004setting,
  title={Setting boundaries: brain dynamics of modal and amodal illusory shape completion in humans},
  author={Murray, M.M. and Foxe, D.M. and Javitt, D.C. and Foxe, J.J.},
  journal={The Journal of neuroscience},
  volume={24},
  number={31},
  pages={6898--6903},
  year={2004},
  publisher={Soc Neuroscience}
}
@article{kanizsa1985seeing,
  title={Seeing and thinking},
  author={Kanizsa, G.},
  journal={Acta Psychologica},
  volume={59},
  number={1},
  pages={23--33},
  year={1985},
  publisher={Elsevier}
}

@article{ringach1996spatial,
  title={Spatial and temporal properties of illusory contours and amodal boundary completion},
  author={Ringach, D.L. and Shapley, R.},
  journal={Vision research},
  volume={36},
  number={19},
  pages={3037--3050},
  year={1996},
  publisher={Elsevier}
}
@article{gerbino1987effect,
  title={The effect of amodal completion on visual matching},
  author={Gerbino, W. and Salmaso, D.},
  journal={Acta Psychologica},
  volume={65},
  number={1},
  pages={25--46},
  year={1987},
  publisher={Elsevier}
}

@article{otero2016computing,
  title={Computing an {E}xact {G}aussian {S}cale-{S}pace},
  author={Otero, Ives Rey and Delbracio, Mauricio},
  journal={Image Processing On Line},
  volume={6},
  pages={8--26},
  year={2016}
}


@article{michotte1991amodal,
  title={Amodal completion of perceptual structures},
  author={Michotte, A. and Thines, G. and Crabb{\'e}, G.},
  journal={Michotte’s experimental phenomenology of perception},
  pages={140--167},
  year={1991},
  publisher={Hillsdale, NJ: Erlbaum}
}

@article{van1995multiple,
  title={Multiple completions primed by occlusion patterns},
  author={Van Lier, R.J. and Leeuwenberg, E.L.J. and Van der Helm, P.A.},
  journal={PERCEPTION-LONDON-},
  volume={24},
  pages={727--727},
  year={1995},
  publisher={Pion ltd}
}
@article{kanizsa1972amodal,
  title={Amodal completion and shrinking of visual fields.},
  author={Kanizsa, G.},
  journal={Studia Psychologica},
  year={1972},
  publisher={Slovak Academy of Sciences}
}

@Article{SinghHoffman,
  author =       {M. Singh and D.D. Hoffman},
  title =        {Completing visual contours: The relationship between relatability and minimizing inflections},
  journal = 	 {Perception \& Psychophysics},
  year = 	 {1999},
  volume = 	 {61(5)},
  pages = 	 {943-951}
}
@Article{VanDerHelm1,
  author =       {R.J. van Lier and P.A. van der Helm and E.L.J. Leeuwenberg},
  title =        {Integrating global and local aspects of visual occlusion},
  journal = 	 {Perception},
  year = 	 {1994},
  volume = 	 {23},
  pages = 	 {883-903}
}
@Article{VanDerHelm2,
  author =       {R.J. van Lier and P.A. van der Helm and E.L.J. Leeuwenberg},
  title =        {Competing global and local completions in visual occlusion},
  journal = 	 {Journal of Experimental Psychology: Human Perception and Performance},
  year = 	 {1995},
  volume = 	 {21},
  pages = 	 {571-583}
}
@Book{SIT,
  title={Structural Information Theory. The Simplicity of Visual Form},
  author={E.L.J. Leeuwenberg and P. A. van der Helm},
  year={2012},
  publisher={Cambridge University Press}
}
@Book{koffka1935,
  title={Principles of Gestalt psychology},
  author={Koffka, K.},
  year={1935},
  publisher={London: Routledge and Kegan Paul}
}


@InProceedings{VanDerHelm3,
   author = {van der Helm, P.A.},
   title = {Simplicity in perceptual organization},
   year = {2014},
   booktitle = {J. Wagemans (Ed.), Oxford Handbook of Perceptual Organization (in press). Oxford , U.K.: Oxford University Press}
}
@Article{Takeichietal,
  author =       {Takeichi, H., Nakazawa, H., Murakami, I., Shimojo, S.},
  title =        {The theory of the curvature-constraint line for amodal completion},
  journal = 	 {Perception-London-},
  year = 	 {1995},
  volume = 	 {24},
  pages = 	 {373-373}
}

@incollection{Mumford,
   author = {D. Mumford},
   title = {Elastica and Computer Vision},
   pages = {491--506},
   year = {1994},
   booktitle = {Algebraic Geometry and its Applications},
   publisher = {Springer, New York}
}

@Article{WilliamsJacobs,
  author =       {Williams, L. R., and Jacobs, D. W.},
  title =        {Stochastic completion fields: A neural model of illusory contour shape and salience},
  journal = 	 {Neural computation},
  year = 	 {1997},
  volume = 	 {9},
  number= {4},
  pages = 	 {837-858}
} 

@InProceedings{complexity,
   title = {Estimating Complexity of 2D Shapes},
   author = {Y. Chen and H. Sundaram},
   pages={1--4},
   year = {2005},
   booktitle = {IEEE 7th Workshop on Multimedia Signal Processing}
}


@Article{ThornberWilliams,
  author =       {Thornber, K. K. and Williams, L. R.},
  title =        {Characterizing the distribution of completion shapes with corners using a mixture of random processes},
  journal = 	 {Pattern Recognition},
  year = 	 {2000},
  volume = 	 {33(4)},
  pages = 	 {543-553}
}
@InProceedings{LeungMalik,
   author = {T. Leung and J. Malik},
   title = {Contour continuity in region based image segmentation},
   booktitle={Computer Vision ECCV},
   pages={544--559},
   year={1998},
   publisher={Springer}
}

@article{van1995competing,
  title={Competing global and local completions in visual occlusion.},
  author={Van Lier, R.J. and Van der Helm, P.A. and Leeuwenberg, E.L.J.},
  journal={Journal of Experimental Psychology: Human Perception and Performance},
  volume={21},
  number={3},
  pages={571},
  year={1995},
  publisher={American Psychological Association}
}

@InProceedings{SemanticBenchmark,
  author = {H. Li, J. Cai, T. Nguyen, J. Zheng},
  title = {A BENCHMARK FOR SEMANTIC IMAGE SEGMENTATION},
  booktitle = {ICME},
  year = {2013}
}

@inproceedings{MartinFTM01,
  author = {D. Martin and C. Fowlkes and D. Tal and J. Malik},
  title = {A Database of Human Segmented Natural Images and its Application to Evaluating Segmentation Algorithms and Measuring Ecological Statistics},
  booktitle = {Proc. 8th ICCV},
  volume = {2},
  pages = {416--423},
  month = {July},
  year = {2001}
}


@article{figuregroundBerkeley,
  title={Local Figure/Ground Cues are Valid for Natural Images},
  author={C. Fowlkes and D. Martin and J. Malik},
  journal={Journal of Vision},
  pages={1-9},
  volume={7},
  number={8},
  year={2007}
}

@article{von2009toward,
  title={Toward a Computational Theory of Perception},
  author={Grompone von Gioi, R. },
  journal={apcap 2009},
  pages={49},
  year={2009},
  publisher={Citeseer}
}


@Article{lsd,
  author =       "R. Grompone von Gioi and J. Jakubowicz and J.-M. Morel and G. Randall",
  title =        "LSD: a Line Segment Detector",
  journal = 	 "Image Processing On Line",
  year = 	 "2012",
  volume = 	 "2",
  pages = 	 "35--55"
}

@Article{rubin2001role,
    author = {Rubin, N.},
    title = {The role of junctions in surface completion and contour matching},
    journal = {Perception},
    number = {3},
    pages = {339--366},
    volume = {30},
    year = {2001}
}

@article{inpainting_demanet_2003,
  title={Image inpainting by correspondence maps: a deterministic approach},
  author={Demanet, L. and Song, B. and Chan, T.},
  journal={Applied and Computational Mathematics},
  volume={1100},
  pages={217--50},
  year={2003},
}
@incollection{inpainting_kawai_2009,
    author = {Kawai, N. and Sato, T. and Yokoya, N.},
    booktitle = {Advances in Image and Video Technology},
    keywords = {inpainting, nonlocal, variational},
    pages = {271--282},
    title = {Image Inpainting Considering Brightness Change and Spatial Locality of Textures and Its Evaluation},
    year = {2009}
}
@article{inpainting_criminisi_2004,
   author= "A. Criminisi and P. P{\'e}rez and K. Toyama",
   title= "Region filling and object removal by exemplar-based inpainting",
   journal= "IEEE Trans. on IP",
   pages= "1200-1212",
   volume="13",
   number="9",
   year= "2004",
}
@article{inpainting_wexler_irani_2007,
   author = {Y. Wexler and E. Shechtman and M. Irani},
   title = {Space-Time Completion of Video},
   journal = {IEEE Transactions on PAMI},
   volume = {29},
   number = {3},
   year = {2007},
   pages = {463--476},
   publisher = {IEEE Computer Society},
   address = {Washington, DC, USA},
}
@article{inpainting_aujol_2010,
  author = {J.-F. Aujol and S. Ladjal and S. Masnou},
  journal = {SIAM Journal on Mathematical Analysis},
  number = {3},
  pages = {1246-1285},
  title = {Exemplar-Based Inpainting from a Variational Point of View.},
  volume = {42},
  year = {2010}
}
@article{inpainting_arias_2011,
   author = {Arias, P. and Facciolo, G. and Caselles, V. and Sapiro, G.},
   affiliation = {Dept. of Information and Communication Technologies, Universitat Pompeu Fabra, Barcelona, Spain},
   title = {A Variational Framework for Exemplar-Based Image Inpainting},
   journal = {International Journal of Computer Vision},
   publisher = {Springer Netherlands},
   keyword = {Computer Science},
   pages = {319-347},
   volume = {93},
   issue = {3},
   year = {2011}
}
@inproceedings{mansfield_2011,
      title = {Transforming Image Completion},
      author = {A. Mansfield and  M. Prasad and C. Rother and  T. Sharp and  P. Kohli and  L. van Gool},
      year={2011},
      pages={121.1--121.11},
      booktitle={Proceedings of BMVC}
}

@article{kang2014illusory,
  title={Illusory shapes via corner fusion},
  author={Kang, S. H. and Zhu, W. and Shen, J.},
  journal={SIAM Journal on Imaging Sciences},
  volume={7},
  number={4},
  pages={1907--1936},
  year={2014}
}

@incollection{knill1996,
  author      = {Knill, D. and Kersten, D. and Yuille, A.},
  title       = {Introduction: A Bayesian formulation of visual perception},
  editor      = {D. Knill and W. Richards},
  booktitle   = {Perception as Bayesian inference},
  publisher   = {Cambridge University Press},
  year        = {1996}
}

@article{moravec86,
  title={Amodal completion: Simplicity is not the explanation},
  author={Moravec, L., and Beck, J.},
  journal={Bulletin of the Psychonomic Society},
  volume={24},
  number={4},
  pages={269--272},
  year={1986}
}

@inproceedings{HayashiSasaki,
      title = {Contour Completion of Partly Occluded Skew-Symmetry Objects},
      author = {T. Hayashi and M. Sasaki},
      year={2014},
      pages={90--93},
      booktitle={Multimedia (ISM), 2014 IEEE International Symposium on}
}

@article{Sekuler,
  title={Local and global minima in visual completion: effects of symmetry and orientation},
  author={Sekuler, A. B.},
  journal={PERCEPTION-LONDON},
  volume={23},
  number={},
  pages={529--529},
  year={1994}
}

\end{document}